\documentclass[authoryear,3p,11pt]{elsarticle}
\usepackage{titlesec}

\usepackage{geometry}
\usepackage{lscape}
\usepackage{setspace}
\newcommand{\w}{\omega}

\newdefinition{definition}{Definition}%[section]

\usepackage{bm}
\usepackage{dsfont}
\usepackage{amssymb}
\usepackage{amsmath}
\usepackage{url}
\usepackage{tablefootnote}
\usepackage{subcaption}

\usepackage{algorithm}
\usepackage{algpseudocode}
\usepackage{setspace}

\usepackage{enumitem}
\usepackage{xcolor}
\usepackage{relsize}
\definecolor{darkgreen}{rgb}{0.0, 0.0, 0.0}
\usepackage[colorlinks=true,linkcolor=blue,citecolor=blue,urlcolor=blue]{hyperref}

\journal{European Journal of Operational Research}

\begin{document}
\makeatletter
\def\ps@pprintTitle{%
     \let\@oddhead\@empty
     \let\@evenhead\@empty
     \let\@oddfoot\@empty
     \let\@evenfoot\@empty
}
\makeatother
\begin{frontmatter}

%% Title, authors and addresses

%% use the tnoteref command within \title for footnotes;
%% use the tnotetext command for theassociated footnote;
%% use the fnref command within \author or \affiliation for footnotes;
%% use the fntext command for theassociated footnote;
%% use the corref command within \author for corresponding author footnotes;
%% use the cortext command for theassociated footnote;
%% use the ead command for the email address,
%% and the form \ead[url] for the home page:
%% \title{Title\tnoteref{label1}}
%% \tnotetext[label1]{}
%% \author{Name\corref{cor1}\fnref{label2}}
%% \ead{email address}
%% \ead[url]{home page}
%% \fntext[label2]{}
%% \cortext[cor1]{}
%% \affiliation{organization={},
%%            addressline={}, 
%%            city={},
%%            postcode={}, 
%%            state={},
%%            country={}}
%% \fntext[label3]{}

\title{Soft decision trees for survival analysis} %% Article title

%% use optional labels to link authors explicitly to addresses:
%% \author[label1,label2]{}
%% \affiliation[label1]{organization={},
%%             addressline={},
%%             city={},
%%             postcode={},
%%             state={},
%%             country={}}
%%
%% \affiliation[label2]{organization={},
%%             addressline={},
%%             city={},
%%             postcode={},
%%             state={},
%%             country={}}
\author[mymainaddress]{Antonio Consolo\corref{mycorrespondingauthor}}
%\ead{antonio.consolo@polimi.it}
\author[mymainaddress]{Edoardo Amaldi}
%\ead{edoardo.amaldi@polimi.it}
\author[mysecondaryaddress]{Emilio Carrizosa}
%\ead{andrea.manno@univaq.it}

\cortext[mycorrespondingauthor]{Corresponding author  \\ \textcolor{white}{abc} Currently affiliation of Antonio Consolo: DISCo, Università di Milano-Bicocca, Milano, Italy}

\address[mymainaddress]{DEIB, Politecnico di Milano, Milano, Italy}
\address[mysecondaryaddress]{Instituto de Matematicas de la Universidad de Sevilla (IMUS), Sevilla, Spain}

%% Abstract
\begin{abstract}
Decision trees are popular in survival analysis for their interpretability and ability to model complex relationships. 
\textcolor{black}{Survival trees, which predict the timing of singular events using censored historical data, are typically built through heuristic approaches.} 
% Survival trees adapt decision trees to predict the timing of singular events using censored historical data. Survival tree algorithms typically use heuristic approaches, which can lead to sub-optimal results. 
Recently, there has been growing interest in globally optimized trees, where the overall tree is trained by minimizing the error function over all its parameters. %While traditional survival tree algorithms often rely on heuristic approaches, which may yield sub-optimal results, recent methods that globally optimized the overall tree parameters have been proposed. 
We propose a new soft survival tree model (SST), \textcolor{black}{with a soft splitting rule at each branch node, trained via} a nonlinear optimization formulation amenable to decomposition.
%We propose a new soft tree model for survival analysis, where a soft splitting rule is applied at each branch node and a nonlinear optimization formulation amenable to decomposition.
\textcolor{black}{Since SSTs provide for every input vector a specific survival function associated to a single leaf node, they satisfy the conditional computation property and inherit the related benefits.}
%E In SST, every input vector is associated with a potential survival function at all leaf nodes, but the final tree prediction is determined by the survival function of a particular leaf node, exhibiting the benefits of conditional computation. 
SST and the \textcolor{black}{training} formulation combine flexibility \textcolor{black}{with interpretability: any} smooth survival function (parametric, semiparametric, or nonparametric) estimated through maximum likelihood \textcolor{black}{can be used, and each leaf node of an SST yields a cluster of distinct survival functions which are associated to the data points routed to it.} 
%E allowing the use of any smooth survival function (whether parametric, semiparametric, or nonparametric) estimated through maximum likelihood,  as leaf nodes represent clusters of distinct survival curves corresponding to different data points. 
%E Moreover, to the best of our knowledge, SST is the first survival tree model which explicitly takes into account group fairness. 
Numerical experiments on 15 well-known datasets show that SSTs, \textcolor{black}{with} parametric and spline-based semiparametric survival functions, \textcolor{black}{trained using an} adaptation of the node-based decomposition algorithm proposed by Consolo et al.  \textcolor{black}{(2024)} \textcolor{black}{for soft regression trees}, outperform \textcolor{black}{three} benchmark survival trees in terms of \textcolor{black}{four} widely-used discrimination and calibration measures. \textcolor{black}{SSTs can also be extended to consider group fairness.}

\end{abstract}

%%Graphical abstract
%\begin{graphicalabstract}
%\includegraphics{grabs}
%\end{graphicalabstract}

%%Research highlights

%\begin{highlights}
%\item Research highlight 1
%\item Research highlight 2
%\end{highlights}

%% Keywords
\begin{keyword}
\textcolor{black}{Machine Learning \sep Survival Analysis} \sep \textcolor{black}{Soft decision trees}\sep \textcolor{black}{Decomposition algorithm} % Optimal Trees
%% keywords here, in the form: keyword \sep keyword

%% PACS codes here, in the form: \PACS code \sep code

%% MSC codes here, in the form: \MSC code \sep code
%% or \MSC[2008] code \sep code (2000 is the default)

\end{keyword}

\end{frontmatter}

%% Add \usepackage{lineno} before \begin{document} and uncomment 
%% following line to enable line numbers
%% \linenumbers

%% main text
%%
%\documentclass[../main.tex]{subfiles}

\section{Introduction}\label{sec:introduction}

In various disciplines and their practical applications, collecting and tracking observations over time is crucial. In areas like %medicine or insurance, 
\textcolor{black}{medicine, maintenance or insurance,} one of the main goals is to determine when a certain event of interest, such as %death or mechanical failure, occurs. 
\textcolor{black} {death, mechanical failure or claim, occurs.} One of the challenging aspects in addressing these problems is the fact that some data points are usually censored, that is, the event of interest is not observed either due to the limited duration of the observed time or to loss of contact during that period. Specifically, while for some data points it is known that \textcolor{black}{the} event is experienced at a certain given time $t$, for the remaining ones, we only have the information that the event did not occur during the observed time. In the latter case, when it is known that the event of interest happened after the observed time (e.g., a patient is confirmed to be alive until time $t$) the data point is said to be right-censored.

Survival analysis is an important subfield of Statistics that provides various methods to handle such censored data which are frequently encountered in disciplines such as healthcare, finance, and sociology \citep{flynn2012survival,eloranta2021cancer,lanczky2021web,gepp2008role,brockett2008survival,zhou2022recurrence,kramer2003survival,plank2008high}. The aim of survival models is to estimate for a given input vector $\mathbf{x}$ the relative survival function $S_\mathbf{x}(t)$, which represents the probability that the event of interest has not occurred to $\mathbf{x}$ at a certain time $t$. Early works in survival analysis either are based on a nonparametric approach, as in Kaplan-Meier curves \citep{kaplan1958nonparametric} where no features are considered (i.e., the specific $\mathbf{x}$ is not considered), or assume some particular parametric form based on linear combination of features, as seen in Cox proportional hazard models \citep{cox1972regression}. Later, several sophisticated statistical models have been proposed (see e.g., the semiparametric approach via splines in \cite{gray1992flexible,royston2002flexible,royston2011use,luo2016spline}), but they often require specific assumptions to be met in order to obtain accurate results. 

In addition to traditional statistical methods, various machine learning \textcolor{black}{(ML)} models have been developed over the years \citep{evers2008sparse,ishwaran2008random,van2011support,ranganath2016deep,wang2019machine}, taking advantage of advances in training and optimization techniques. The adaptation of ML methods to properly handle censored data and time estimation has led to \textcolor{black}{alternatives to statistical models} that can capture new non-linear relationships between features and survival times. Several ``black-box'' ML models have been proposed by adapting the original task to take into account survival data (see e.g., \cite{ripley1998neural,che2018recurrent,ching2018cox,giunchiglia2018rnn,katzman2018deepsurv,hu2021transformer}). Although such black-box models are able to achieve high level of accuracy, the \textcolor{black}{output} %survival curves 
they produce %are 
\textcolor{black}{is} often not interpretable, which makes them less practical for use in high-stake domain such as healthcare. Over the years, interpretable ML models have also been proposed and adapted to survival analysis.

Decision trees are widely used for classification and regression due to their inherent interpretability.
%Since data points are routed along the binary tree from the root following decisions based on splitting rules, the domain experts know the sequence of decisions that leads to the tree response for any data point.
\textcolor{black}{Since every input vector is routed along the binary tree from the root following a sequence of splitting rules at branch nodes, the} domain experts know the sequence of decisions \textcolor{black}{that leads to the tree response}. Hard or soft splitting rules \textcolor{black}{can be considered.}
 \textcolor{black}{In hard splits, the left branch is followed if a single feature (univariate) or a linear combination of the features (multivariate) exceeds a threshold value. In soft splits, the two branches are followed with complementary probabilities given by a sigmoid function of a linear combination of the features.}

%In the past several 
%E \textcolor{black}{Several} greedy approaches have been proposed to adapt decision tree to tackle survival task \citep{gordon1985tree,segal1988regression,ciampi1988recpam,davis1989exponential,therneau1990martingale,leblanc1992relative,leblanc1993survival,zhang1995splitting,kelecs2002residual,molinaro2004tree,jin2004alternative,hothorn2006unbiased}. These methods use a top-down approach to build tree partitions. At each branch node, a measure of survival dissimilarity (e.g., logrank statistic, survival time variance) identifies the best local split. This process is recursively applied to child nodes, with pruning heuristics often employed later to prevent overfitting. However, performance can suffer as incorrect splits are irreversible.

Early works on decision trees for classification, regression or survival analysis were based on greedy-like algorithms with often a pruning post-processing step. Due to the remarkable progresses in optimization methods and solvers, growing attention has been recently devoted to globally optimized decision trees, i.e., trees whose parameters are simultaneously tuned and which exhibit local or global optimality guarantees.
Examples of Mixed-Integer Linear \textcolor{darkgreen}{Programming (MILP)} approaches to design deterministic classification and regression trees can be found in \citep{bertsimas2019machine,aghaei2020learning,TU2024},
of nonlinear optimization approaches to train soft classification and regression trees in \citep{suarez1999globally,blanquero2020sparsity,blanquero2022sparse}, and of MILP ones to build deterministic survival trees in \citep{bertsimas2022optimal,zhang2024optimal}. \textcolor{darkgreen}{Recent works have also focused on methods such as dynamic programming, beam search techniques, and the use of different splitting rules, including those derived from linear programming (e.g., \cite{brita2025optimal,kiossou2025generic,lin2024improved,ENGUR2024127354}).}  See \citep{costa2022} for a survey on decision trees and \citep{CARRITOP} for a review on globally optimized classification and regression trees. %classification and regression trees. 
In some articles the aim is not only to enhance the predictive performance but also to improve interpretability and to consider fairness issues.

In this work, we propose a \textcolor{black}{soft multivariate survival tree model, which provides for each data point a survival function associated to a single leaf node, and a continuous nonlinear optimization formulation for training. These soft trees satisfy the conditional computational property, that is, each prediction depends on a subset of nodes (parameters), leading to computational and statistical benefits. Another important advantage of our approach is flexibility: domain experts can integrate prior knowledge by selecting an appropriate parametric or semiparametric survival function model, and it yields for each data point a distinct survival function.}

The remainder of the paper is organized as follows. \textcolor{black}{In Section \ref{sec:previous-work} we mention previous work on survival trees.} In Section \ref{sec:model}, we present the new soft survival tree model, the associated formulation for training, and describe the parametric and spline-based semiparametric \textcolor{black}{models considered in this work for estimating} the survival functions within the leaf nodes. 
We also show how our soft survival tree model can be adapted to take into account group fairness. %E Moreover, we provide a brief overview of the accuracy measures commonly used to evaluate performance in survival problems. 
\textcolor{black}{In Section \ref{sec:decomposition}, we extend to soft survival trees the node-based decomposition algorithm proposed in \citep{consolo2025} for training soft regression trees.} %, which includes an initialization procedure and an ad hoc reassignment heuristic.}
%E Section \ref{sec:decomposition} is devoted to \textcolor{black}{give} an overview of the node-based decomposition algorithm with an initialization procedure and a reassignment heuristic proposed in \citep{consolo2025}  and how we adapt this algorithm to survival task. 
In Section \ref{sec:experiments}, we provide an experimental comparison with \textcolor{black}{three} established survival tree models across 15 datasets from the literature. 
\textcolor{black}{Besides demonstrating how soft trees improve interpretability, we show on a job search dataset %\citep{romeo1999conducting} 
how group fairness can be taken into account.}
%E Additionally, we demonstrate how our soft survival tree improves interpretability and present results on a job search dataset \cite{romeo1999conducting}, showing its ability to address group fairness. 
Finally, Section \ref{sec:conclusions} \textcolor{black}{contains some concluding remarks and future research directions.}
%E presents the conclusions and outlines potential directions for future research. 
Details on the datasets 
\textcolor{black}{and additional material related to the comparative experiments and the benefits of interpretability are included in the Appendices.}
%E additional computational results from the comparative experiments, and further results on the interpretability benefits of our approach are included in the Appendices.
%\documentclass[../main.tex]{subfiles}

\section{\textcolor{black}{Previous work on survival trees}}\label{sec:previous-work}

\textcolor{black}{Since the mid 80s, several greedy approaches have been proposed to adapt decision trees to tackle survival tasks} \citep{gordon1985tree,segal1988regression,ciampi1988recpam,davis1989exponential,therneau1990martingale,leblanc1992relative,zhang1995splitting,kelecs2002residual,molinaro2004tree,jin2004alternative}. These methods use a top-down approach to build tree partitions. At each branch node, a measure of survival dissimilarity (e.g., logrank statistic, survival time variance) identifies the best local split. \textcolor{black}{This process is recursively applied to child nodes without reconsidering previous splits, and hence affecting performance. Pruning heuristics are often employed to prevent overfitting. Widely used greedy methods include the extension to survival analysis of Conditional Inference Trees \citep{hothorn2006unbiased}, CART \citep{breimanclassification}, and the survival trees in \citep{leblanc1993survival} where the branch node splits maximize the separation between survival functions using the logrank test.}

%\textcolor{black}{Among the greedy models that are still widely used, we mention the extension of Conditional Inference Trees \citep{hothorn2006unbiased} to survival analysis, CART \citep{breimanclassification} for survival data, and the survival trees proposed in \cite{leblanc1993survival}, where the branch node splits maximize the separation between survival functions using the logrank test.}

\textcolor{black}{Although building globally optimized trees for survival analysis is more challenging than for classical} classification or regression problems, in the last three years there have been some attempts to construct %survival deterministic trees.
\textcolor{black}{deterministic survival trees.}

%E To the best of our knowledge, the first effort to develop globally optimized survival trees was made in \cite{bertsimas2022optimal}. %E In their work, 
\textcolor{black}{In \cite{bertsimas2022optimal} the authors extend} the local search method introduced in \citep{dunn2018optimal} for univariate and multivariate classification and regression trees to the context of survival analysis. 
%E In particular, the local search considers 
\textcolor{black}{They only} consider univariate splitting derived from a proportional hazards model, following the likelihood-based approach proposed by \cite{leblanc1992relative}. At each leaf node the Nelson-Aalen estimator is used as baseline hazard function, and all individuals within the node are assumed to share the same survival function.

In \cite{huisman2024optimal}, the authors extend the dynamic programming approach proposed in \citep{Demirovic2020MurTreeOD} for classification trees to % the context of 
univariate survival trees. As done in \citep{bertsimas2022optimal}, the splitting rules follow the method of \cite{leblanc1992relative}, with the Nelson-Aalen estimator used as the \textcolor{black}{hazard} function at each leaf node. For the specific case of trees of depth $D=2$, the authors introduce a specialized algorithm to enhance scalability.

In \cite{zhang2024optimal}, a dynamic programming approach is proposed for building optimal univariate survival trees. The loss function employed for the survival task is the \textcolor{black}{so-called} Integrated Brier Score (IBS). To limit the exploration of the feasible solution space, bounds on the IBS loss function are introduced. At each leaf node, the Kaplan-Meier estimator is applied. %Experimental results demonstrate that the bounds effectively reduce computational complexity by pruning the search space.

It is worth pointing out that these three recent survival tree approaches make restrictive assumptions concerning different aspects of the survival tree model and of the training error function. In \citep{bertsimas2022optimal,huisman2024optimal}
the type of univariate splits at the branch nodes are based on proportional hazard, assuming that the ratio between the hazard functions of any two data points remains constant over time.
In \citep{zhang2024optimal} the survival functions at leaf nodes are estimated via Kaplan-Meier, while in \citep{bertsimas2022optimal,huisman2024optimal} via Nelson-Aalen estimators. \textcolor{darkgreen}{Therefore, all these models impose the same piecewise-constant survival function on all data points in a leaf node. It is important to note that when only a few data points are contained in the leaf node, this survival function has a small number of large jumps, which may result in an exaggerated drop in survival probability from $t$ to $t+\Delta$ without strong justification.}
Moreover, in \citep{zhang2024optimal} the error function in the training formulation only takes into account the IBS \textcolor{black}{calibration} measure.

\textcolor{black}{Finally, an evolutionary algorithm with specific mutation and crossover operators is introduced in \cite{kretowska2024global} for constructing multivariate deterministic survival trees, where Kaplan–Meier estimators are adopted. In the error function a term accounting for the complexity of the tree's topology is added to the IBS measure.}
\section{Soft survival trees}\label{sec:model}
In this section, we introduce the soft multivariate survival tree model and its formulation. %The 
\textcolor{black}{Then, we show how the} proposed method offers flexibility by allowing the use of parametric distributions within a parametric framework, as well as splines within a semiparametric framework, as possible choices for modeling survival functions at the leaf nodes.
\textcolor{black}{Moreover, we show how our framework allows one to consider two relevant practical aspects, namely,} %explainability 
\textcolor{black}{interpretability} \textcolor{black}{and group fairness. In particular, we}
discuss how soft survival trees with single leaf node prediction favor interpretability by considering the clusters of individuals survival functions associated to each leaf node \textcolor{black}{and the possibility for such trees of identifying relevant features through global sparsity.} Furthermore, we present a way to address group fairness in soft survival trees.
%\textcolor{black}{Subsequently, we highlight the interpretability of soft survival trees with single leaf node predictions by examining clusters of individual survival functions associated with each leaf node the possibility for such trees of identifying relevant features through global sparsity.}

%Finally, we briefly recall and describe the accuracy measures used to assess the performance of survival models in terms of discrimination and calibration.

\subsection{\textcolor{black}{Soft survival tree model}}

In survival analysis, the training set \textcolor{black}{is denoted by $I = \{ (\mathbf{x}_i, t_i, c_i) \}_{1 \leq i \leq N}$, where $N$ is the number of data points. For the $i$-th data point, $\mathbf{x}_i \in \mathbb{R}^p$ is the $p$-dimensional vector of input features, $t_i \in \mathbb{R}$ is the last observation time,} and $c_i \in \{0,1\}$ indicates the event status. In particular, $c_i = 1$ denotes that the event of interest occurred, while $c_i = 0$ indicates that the observation is censored. In this work, we consider the classical setting of right-censored data, meaning that the exact event time is not observed and is only known to exceed $t_i$.

As in 
\textcolor{black}{\citep{suarez1999globally,blanquero2022sparse}} for classification and regression tasks, soft survival trees are defined as maximal binary multivariate trees with a fixed depth $D$, where branch nodes have two children and the leaf nodes are positioned at the same depth.

\textcolor{darkgreen}{Let $\tau_{L}$ and $\tau_{B}$ \textcolor{darkgreen}{denote} the sets of leaf and branch nodes, respectively.}
%E At each branch node, the sigmoid function $F(v)=\frac{1}{1+e^{-v}}$ determines the probability of routing a data point left or right. For each $n \in \tau_{B}$, the decision variables $\w_{jn}\in \mathbb{R}$ and $\w_{0n}\in \mathbb{R}$ denote the coefficients and intercept of the hyperplane that serves as the input to the sigmoid function. The probability of taking the left (or right) branch of node $n$ is defined as follows:
\textcolor{darkgreen}{At each branch node, the probabilities that a data point is routed along the left or right branches are defined as follows.}

\vspace{-20pt}

\textcolor{darkgreen}{
\begin{definition}
%For each input vector $\mathbf{x}_i$, with $1 \leq i \leq N$, and branch node $n \in \tau_{B}$, the probability of being routed to the left branch is  
\textcolor{darkgreen}{For each branch node $n \in \tau_{B}$, the probability that an input vector $\mathbf{x}_i$, with $1 \leq i \leq N$, is routed along the left branch of $n$ is defined as}
\begin{equation}\label{eq:prob_left}
    p_{in}=F\left(\sum_{j=1}^{p}\w_{jn}x_{ij}-\w_{0n}\right) \hspace{-0.13cm},
\end{equation}
 \textcolor{darkgreen}{where $F(v)=\frac{1}{1+e^{-v}}$ is a sigmoid function, and the coefficients %the decision variables 
 $\w_{jn}\in \mathbb{R}$, with $1 \leq j \leq p$, and $\w_{0n}\in \mathbb{R}$ are the decision variables.}
 %are the coefficients and intercept of the hyperplane that serves as the input to the sigmoid function $F(v)=\frac{1}{1+e^{-v}}$}. 
 \textcolor{darkgreen}{Obviously, the probability of $\mathbf{x}_i$ being routed along the right branch of $n$ is $1-p_{in}$.}
\end{definition}
}

\textcolor{darkgreen}{Let $A_{L(n)}$ denote the set of ancestor nodes of a leaf \textcolor{darkgreen}{node} $n$ whose left branches are included in the \textcolor{darkgreen}{unique} path from the root to $n$, and $A_{R(n)}$ the set of ancestor nodes whose right branches belong to that path. %E Based on the 
\textcolor{darkgreen}{Due to the} soft splitting rule \eqref{eq:prob_left}, the probability that an input vector $\mathbf{x}_i$ %reaches 
\textcolor{darkgreen}{falls into} a given leaf node %$n \in \tau_L$ 
is defined as follows.
%E the product of the (independent) probabilities along the path from the root to $n$:
}
\vspace{-20pt}

\textcolor{darkgreen}{\begin{definition}
%E For each input vector $\mathbf{x}_i$, with $1 \leq i \leq N$, and leaf node $n \in \tau_{L}$, the probability of falling into $n$ is  
\textcolor{darkgreen}{For each leaf node $n \in \tau_{L}$, the probability that an input vector $\mathbf{x}_i$, with $1 \leq i \leq N$, falls into $n$ is} 
\begin{align}\label{eq:prob_tot}
P_{in} = 
\prod_{n_{l}\in A_{L(n)}}p_{in_{l}}\;\prod_{n_{r}\in A_{R(n)}}(1-p_{in_{r}}),
\end{align}
\textcolor{darkgreen}{namely, it amounts to the product of the (independent) probabilities of $\mathbf{x}_i$ being routed along the branches contained in the path from the root to $n$.}
\end{definition}}

%Due to the soft splitting rule induced by the sigmoid function at each branch node, every input vector $\mathbf{x}_i$ is associated with a probability of falling into each leaf node $n \in \tau_L$: 
%\begin{align*}
%P_{in} = 
%\prod_{n_{l}\in A_{L(n)}}p_{in_{l}}\;\prod_{n_{r}\in A_{R(n)}}(1-p_{in_{r}}),\label{eq:pr_it_lit}
%\end{align*}
%\noindent where $A_{L(n)}$ denotes the set of ancestor nodes of leaf node $n$ whose left branches belong to the path from the root to $n$, while $A_{R(n)}$ the set of ancestor nodes for the right branches. 

\begin{figure}[h]
\centering
\includegraphics[scale=0.34]{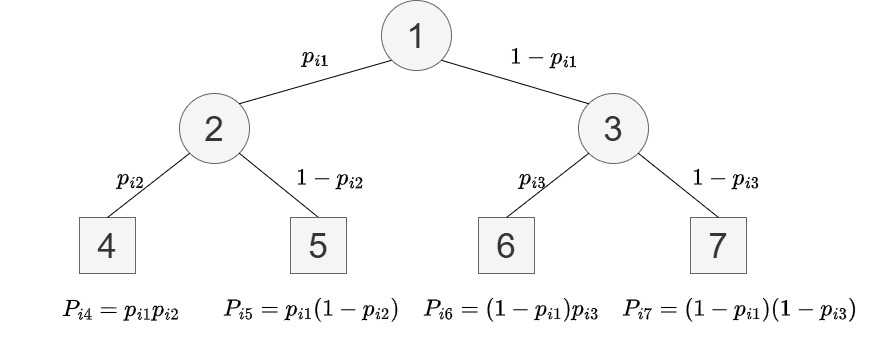}
\caption{\label{fig:fig1234}A soft tree of depth $D=2$.}
\end{figure}

\textcolor{darkgreen}{An illustrative example of a soft tree of depth $D=2$ is shown in Figure \ref{fig:fig1234}.}

For any input vector $\mathbf{x} \in \mathbb{R}^p$ and any leaf node $n \in \tau_{L}$, the output is defined as the \textcolor{black}{so-called } survival function $S_{\mathbf{x}}(t; \bm{\beta}_n)$, 
\textcolor{black}{which indicates the probability that the corresponding event does not occur before time $t$,}
where the vector $\bm{\beta}_n$ denotes the parameters of the survival function associated with the leaf node $n$. In this work, we consider both parametric and semiparametric approaches for estimating survival functions. \textcolor{black}{In SSTs with parametric models at leaf nodes,} the parameter vector $\bm{\beta}_n$ includes both the coefficients of the features and the parameters characterizing the mean, shape, variance, or higher moments of the survival functions. \textcolor{black}{In SSTs with semiparametric models at leaf nodes, $\bm{\beta}_n$ includes the coefficients related to the features and to the basis functions.}%In the spline-based semiparametric approach, $\bm{\beta}_n$ contains not only the coefficients of the features but also the coefficients of the considered basis functions.

\textcolor{black}{For any input vector $\mathbf{x}$ and any leaf node $n \in \tau_{L}$, we also have, besides the survival function, the related hazard function and the cumulative hazard function, denoted by $h_{\mathbf{x}}(t; \bm{\beta}_n)$ and $H_{\mathbf{x}}(t; \bm{\beta}_n)$, respectively. For the sake of completeness, we briefly recall the relationship between the above} %function. 
\textcolor{black}{functions, and we refer the reader to e.g. \citep{OAKES19833} for further details.}

\textcolor{black}{The hazard function $h_{\mathbf{x}}(\cdot;\bm{\beta}_n)$ is defined as follows: $h_{\mathbf{x}}(t;\bm{\beta}_n) \Delta$ 
is, for infinitesimal $\Delta >0,$ the conditional probability of failure before $t+\Delta$, given survival beyond $t$, of an individual with \textcolor{black}{input vector} $\mathbf{x}$.
} 
The cumulative hazard function $H_{\mathbf{x}}(t;\bm{\beta}_n)$  is defined as the integral of \textcolor{black}{$h_{\mathbf{x}}(t;\bm{\beta}_n)$}:
\begin{equation}\label{eq:hazard_def}
    H_{\mathbf{x}}(t;\bm{\beta}_n) = \int_{0}^{t} h_{\mathbf{x}}(s;\bm{\beta}_n)ds.
\end{equation}

\noindent The survival function is then expressed in terms of the cumulative hazard function \eqref{eq:hazard_def} as:
\begin{equation}\label{eq:survival_def}
S_{\mathbf{x}}(t;\bm{\beta}_n)= e^{-H_{\mathbf{x}}(t;\bm{\beta}_n)}.
\end{equation}
%
%The survival functions for each leaf node are estimated via maximum likelihood.
The estimation of the survival functions $S^n_{\mathbf{x}}(t;\bm{\beta}_n)$ %is 
\textcolor{black}{can be} performed via maximum likelihood. 

\begin{figure}[H]
    \centering
    \includegraphics[width = 0.7\textwidth]{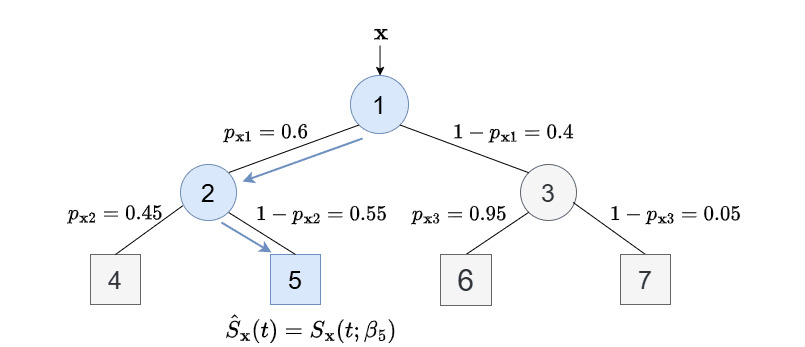}
    \caption{An example of soft survival tree with single leaf node prediction where the arrows indicate the HBP path for any input $\mathbf{x}$, and the corresponding predicted survival function is $\hat{S}_{\mathbf{x}}(t) = S_{\mathbf{x}}(t; \bm{\beta}_5)$.}
    \label{fig:percorso}
\end{figure}

\textcolor{black}{For any given values of the parameters $\bm{\omega}_n$ with $n \in \tau_B$ and $\bm{\beta}_n$ with $n \in \tau_L$, the soft survival tree provides, for any input vector $\mathbf{x}_i \in \mathbb{R}^{p}$, $\vert \tau_{L} \vert$ potential survival function predictions (one for each leaf node), along with the associated probabilities $P_{in}$ that $\mathbf{x}_i$ is assigned to each corresponding leaf node $n$. %E In line with
\textcolor{darkgreen}{Similarly to what was proposed in \citep{consolo2025} for soft regression trees}, for each input $\mathbf{x}$ %E the actual tree prediction is
\textcolor{darkgreen}{we define as the actual tree prediction} the survival function $\hat{S}_{\mathbf{x}}(t) = S_{\mathbf{x}}(t;\bm{\beta}_{n_{\mathbf{x}}})$ associated with a single leaf node $n_{\mathbf{x}}$. This leaf node is determined by routing $\mathbf{x}$ from the root along the branches with the highest probability, which correspond to the so-called Highest Branch Probability (HBP) path. Figure \ref{fig:percorso} illustrates how a tree of depth $D=2$  generates %E predictions 
\textcolor{darkgreen}{the prediction} for an input vector $\mathbf{x}$, where $p_{\textbf{x}n}$ represents the probability of $\textbf{x}$ being routed through the left branch of node $n$. The HBP path, highlighted in blue, leads to \textcolor{darkgreen}{the} leaf node $n_{\mathbf{x}}=5$, where the predicted survival function is $\hat{S}_{\mathbf{x}}(t) = S_{\mathbf{x}}(t; \bm{\beta}_5)$.
} %%% END CYAN

\textcolor{darkgreen}{
%E Formally, for any input vector $\mathbf{x}$, the survival function predicted by following the HBP is defined as:
The formal definition of the survival function single leaf node prediction based on the concept of HBP path is as follows.
}
\vspace{-20pt}
%Specifically, for any given input vector $\mathbf{x}$ the predicted survival function is defined as:

\textcolor{darkgreen}{\begin{definition}
For a given input vector $\mathbf{x} \in \mathbb{R}^p$, the predicted survival function is defined as  
\begin{equation}\label{eq:prediction_model}
\hat{S}_{\mathbf{x}}(t) = S_{\mathbf{x}}(t;\bm{\beta}_{n_{\mathbf{x}}}) = 
\sum_{n\in \tau _{L}}
\prod_{\ell\in A_{L(n)}}
\mathds{1}_{0.5}(p_{\mathbf{x}\ell}(\bm{\w}_{\ell})) \;
\prod_{r\in A_{R(n)}}
\mathds{1}_{0.5}(1-p_{\mathbf{x}r}(\bm{\w}_r))
S_{\mathbf{x}}(t;\bm{\beta}_{n}),
\end{equation}
where $\mathds{1}_{0.5}(v) = 1$ if $v \geq 0.5$ and $\mathds{1}_{0.5}(v) = 0$ otherwise.
\end{definition}}
The Soft Survival Tree model defined above will be referred to as SST.

\textcolor{black}{It is worth pointing out that the deterministic way the prediction is defined for every input vector based on a single leaf node guarantees the beneficial conditional computation property (see e.g., \cite{bengio2015conditional}), namely, for any input vector the output only depends on the parameters $\bm{\omega}_n$ and $\bm{\beta}_n$ for the nodes $n$ contained in the corresponding HBP path. The notable computational and statistical benefits of conditional computation include reducing parameter usage for faster inference and acting as a regularizer to enhance the statistical properties of the model \citep{breiman2001random,hastie2009elements,bengio2015conditional}. 
}  %%% END CYAN

\textcolor{black}{
\subsection{Training formulation and advantages of soft survival trees} \label{sec:formulation-advantages}
}  %%% END CYAN

\textcolor{black}{
Before presenting the nonlinear optimization formulation that we propose to train the above-mentioned SSTs, we need to specify the log-likelihood to be used in the estimation of the survival functions associated to the leaf nodes. 
}  %%% END CYAN

Assuming non-informative censoring, that is, assuming that the censoring times of individuals are statistically independent from the occurrence times of the considered event\footnote{In clinical settings, this assumption may fail when individuals at higher risk of drop out yield censored survival times, often due to worsening disease status. In extreme cases, dropout may occur shortly before death.} \textcolor{darkgreen}{(see, e.g., \cite{oquigley2008proportional})}, the negative log-likelihood associated with each leaf node $n \in \tau_L$ and each data point $(\mathbf{x}_i,t_i,c_i)$ is defined as:
\begin{equation}\label{eq:neg_log_lik}
    L^-_n({\mathbf{x}_i},t_i,c_i;\bm{\beta}_n) =\begin{cases}
	-log(h_{\mathbf{x}_i}(t_i;\bm{\beta}_n)) + H_{\mathbf{x}_i}(t_i;\bm{\beta}_n)\hspace{0.5cm} \text{if } c_i = 1, \\
	H_{\mathbf{x}_i}(t_i;\bm{\beta}_n)\hspace{3.8cm} \text{if } c_i = 0.
\end{cases}
\end{equation} 

  %%% END CYAN

%Once the contents of the leaf and branch nodes, along with their variables, have been defined, 

\textcolor{darkgreen}{Based on the definitions provided in \eqref{eq:prob_left}, \eqref{eq:prob_tot}, and \eqref{eq:neg_log_lik} for the leaf and branch nodes, we propose the following unconstrained nonlinear optimization problem to train SSTs:}

%In order to train SSTs, we propose to solve the following unconstrained nonlinear optimization problem:

\begin{equation}
\label{eq:formulation}
\begin{split}
\min_{\bm{\w},\bm{\beta}} E(\bm{\w,\beta})= \frac{1}{N}\sum_{i \in I}\sum_{k\in \tau _{L}}\prod_{\ell\in A_{L(n)}}p_{i\ell}(\bm{\w}_{\ell})\;\prod_{r\in A_{R(n)}}(1-p_{ir}(\bm{\w}_r))L^{-}_n({\mathbf{x}_i},t_i,c_i;\bm{\beta}_n)\\
= \frac{1}{N}\sum_{i \in I}\sum_{k\in \tau _{L}}P_{in}(\bm{\w})L^{-}_n({\mathbf{x}_i},t_i,c_i;\bm{\beta}_n), \end{split}
\end{equation}

\noindent where the $|\tau_{B}|(p + 1)$  vector  $\bm{\omega} = (\omega_{jn})_{j\in\{0,1,\ldots,p\},\, n \in \tau_{B}}$ contains all the parameters of the branch nodes and the $|\tau_{L}|(p + m + 1)$ vector  $\bm{\beta} = (\beta_{jn})_{j\in\{0,1,\ldots,p+m\},\, n \in \tau_{L}}$ contains all the parameters of the leaf nodes. In this context, the number $m$ \textcolor{black}{denotes the number of additional variables, beside from the feature variables,} required to define the survival functions. The value of $m$ depends on whether a parametric or \textcolor{black}{semiparametric approach is adopted to estimate} the survival functions. See the next section for more details on $m$.

\textcolor{black}{To conclude, we emphasize the main advantages of the above SST model and training formulation which are related to flexibility, computational efficiency and interpretability.}

\textcolor{black}{
From the modeling point of view, the only assumption concerning the survival functions is that they can be expressed as or approximated by smooth  (parametric, semiparametric, or nonparametric) functions.}
%E It is worth pointing out that the above formulation makes no specific assumptions on whether the estimation of the survival function is parametric, semiparametric, or nonparametric. The only requirement is that it can be expressed as a smooth function.% to preserve the continuity of formulation \eqref{eq:formulation}. 

\textcolor{black}{From the computational point view, the number of variables in the formulation for training SSTs
% soft survival trees 
%with soft splitting rules at branch nodes
only depends on the number of features $p$, $m$  %(plus $m$ and the additional variables associated with the structure of the survival functions) 
and the depth $D$ of the tree. Thus SSTs tend to scale better than deterministic ones whose number of variables in the formulation increases with the
%, possibly very large, 
number of data points $N$.}

\textcolor{black}{
As to interpretability, SSTs offer two key advantages. First, unlike alternative (deterministic) survival tree models that are constrained to assign a single survival function (e.g., through the Kaplan-Meier estimator) to each leaf node for all input vectors routed to it, the SST approach provides greater flexibility by considering distinct survival functions for each leaf node and input vector. \textcolor{darkgreen}{This makes it possible to define a more detailed, data point–specific survival function that relies on the local information associated with the features of each data point.} Second, the predictions associated with single leaf nodes can be interpreted as clusters, favoring the analysis of survival functions to extract valuable information about the phenomenon under study. In Section \ref{sec:interpretability}, we illustrate the valuable additional insights that can be gained using SSTs for three real-world datasets.
}

%\textcolor{darkgreen}{Finally, it is worth pointing out that if a domain expert has already determined the most appropriate (continuous) survival functions for the \textcolor{darkgreen}{application under consideration}, the flexibility of SST allows this prior knowledge to be easily incorporated into the model.

\textcolor{darkgreen}{Finally, it is worth pointing out that 
the flexibility of SSTs allows the domain experts to incorporate prior knowledge concerning the application under consideration when selecting an appropriate type of (continuous) survival functions.}

\subsection{\textcolor{black}{Parametric and spline-based semiparametric models for survival functions}}

In statistical modeling, several alternatives have been proposed for the estimation of the survival and hazard functions. Traditionally, medical researchers and biostatisticians have relied on Cox regression \citep{cox1972regression} to handle censored survival data. This model \textcolor{black}{assumes} that the hazard rate for an event %as 
\textcolor{black}{is a linear combination of the features}, with the hazard function for an input vector $\mathbf{x}$ expressed as:
$$h_{\mathbf{x}}(t;\bm{\beta}) = h_0(t)\text{exp}(\bm{\beta}^T\mathbf{x}),$$
where $h_0(t)$ \textcolor{black}{corresponds to the baseline hazard function.} 

In the Cox model, the choice of $h_0(t)$ is critical since its estimation \textcolor{black}{in high-dimension is prone to overfitting and may complicate} the validation \citep{gelfand2000proportional,royston2002flexible}.
%The choice of $h_0(t)$ is crucial, as it is often estimated in high-dimensional settings, leading to potential overfitting. This overfitting complicates validation on independent datasets, posing significant challenges \cite{}.
%As emphasized in \cite{gelfand2000proportional,royston2002flexible}, 
Considering more sophisticated hazard functions can lead to more accurate estimates and offer deeper insights into the phenomenon under investigation (e.g., capturing the progression of a disease over time in medical studies). Moreover, the classical proportional hazards assumption of the Cox model may not be suitable for all contexts\footnote{Although the Cox model allows for non-proportional hazard assumption, such as through time-varying regression coefficients, there is no widely accepted or intuitive method for implementing this adjustment.}.

To overcome these limitations, more flexible statistical approaches, \textcolor{black}{including parametric distribution and spline-based semiparametric ones, have been proposed. These approaches} are well-suited to capture complex patterns in the data, such as time-varying hazards, interval censoring, frailties, and multiple responses across diverse datasets or time scales \citep{aalen2008survival}.

\textcolor{black}{The soft survival tree model that we propose in this work is very flexible since it encompasses parametric and spline-based semiparametric estimation of the survival functions (see the next two subsections) and it can consider any survival function model which preserves the smoothness of formulation \eqref{eq:formulation}.} 

%It is important to note that, without loss of generality, alternative choices that preserve the smoothness of formulation \eqref{eq:formulation} are equally valid. 

\subsubsection{\textcolor{black}{Parametric distribution models}}

\textcolor{black}{In the case of parametric distributions}, the survival, hazard, and cumulative hazard functions depend on a finite, often small, number $R$ of parameters \textcolor{black}{$\alpha_1,\ldots,\alpha_R$. Typically $\alpha_1$ is the mean ($\mu$) or the main parameter (location) of the distribution, and the other parameters $(\alpha_2, \dots, \alpha_R)$, referred to as ‘ancillary’, are related to its shape, variance, or higher moments.}

All parameters may be linearly dependent on the features %(i.e., $\alpha_r(\mathbf{x})$)
through a transformation $g$ such that $g(\alpha_r(\mathbf{x})) = \gamma_0 + \bm{\gamma}^T \mathbf{x}$ for $r = 1, \dots, R$. \textcolor{black}{Typically, $g$ is the logarithm function for positive parameters, and the identity function for parameters which are unrestricted in sign. For a review of the parametric distributions, the reader is referred to \citep{lawless2014parametric,klein2014handbook}.}

In this work, we \textcolor{black}{assume that only} the primary parameter $\mu$ linearly depends on the features (i.e., $\text{log}(\mu(\mathbf{x})) = \gamma_0 + \bm{\gamma}^T \mathbf{x}$). Thus, for each leaf node $n \in \tau_L$, the vector $\bm{\beta}_n$ contains $p+1+m$ components. \textcolor{black}{The first $p+1$ components correspond to the number of variables related to $\mu$, and the $m = R-1$ remaining ones to the ancillary parameters.}

The parametric distributions considered in this work are the Exponential (Exp), Weibull ($W$), and Log-Logistic (Llog) ones. \textcolor{black}{Table \ref{tab:parametric} summarizes their main characteristics, namely, the survival and hazard functions, the parameters involved, and the hazard shape.} The Exp distribution assumes a constant hazard function and involves only the rate parameter $\mu$ ($m=0$). The W distribution \textcolor{black}{accounts for} both monotonically increasing and decreasing hazards, and includes a shape parameter $\alpha$ ($m=1$) in addition to the scale parameter $\mu$. The Llog distribution can model arc-shaped and monotonically decreasing hazards, considering both the scale parameter $\mu$ and the shape parameter $\alpha$ ($m=1$). 

\begin{table}[]
\centering
\resizebox{\textwidth}{!}{
\begin{tabular}{lllcll}
\hline
\textbf{Distribution} & \textbf{Abbreviation} & \textbf{Parameters}                                                             & \multicolumn{1}{l}{\textbf{Hazard} $\bm{h(t)}$}                          & \textbf{Survival} $\bm{S(t)}$                                     & \textbf{Hazard shape}                                                                             \\ \hline
Exponential & Exp & rate $\mu>0$                                                           & $\mu$                                                   & \multicolumn{1}{c}{$\text{exp}(-\mu t)$}         & constant                                                                                  \\ \hline
Weibull  & W    & \begin{tabular}[c]{@{}l@{}}scale $\mu>0$\\ shape $\alpha>0$\end{tabular} & {\Large $\frac{\alpha}{\mu}(\frac{t}{\mu})^{\alpha-1}$}          & $\text{exp}(-(t/\mu)^{\alpha})$                  & \begin{tabular}[c]{@{}l@{}}constant,\\ monotonically\\ increasing/decreasing\end{tabular} \\ \hline
Log-logistic & Llog & \begin{tabular}[c]{@{}l@{}}scale $\mu>0$\\ shape $\alpha>0$\end{tabular} & {\Large$\frac{(\alpha/\mu)(t)^{\alpha-1}}{1+(t/\mu)^{\alpha}}$} & \multicolumn{1}{c}{\Large $\frac{1}{1+(t/\mu)^\alpha}$} & \begin{tabular}[c]{@{}l@{}}arc-shaped,\\ monotonically\\ decreasing\end{tabular}          \\ \hline
\end{tabular}
}\caption{\textcolor{black}{Considered parametric distributions with the corresponding hazard and survival functions.}}\label{tab:parametric}
\end{table}

\textcolor{black}{It is worth emphasizing that in SSTs any other parametric distribution can be considered}.

\subsubsection{\textcolor{black}{Spline-based semiparametric models}}

\textcolor{black}{For the estimation of the semiparametric survival functions, we adopt the methodology proposed by} \cite{royston2002flexible}, which assumes flexible spline-based models 
%for the survival function 
to take into account proportional hazards and proportional odds. \textcolor{black}{Other options} (see, e.g., \cite{younes1997link,shen1998propotional,royston2011use,bremhorst2016flexible}) \textcolor{black}{that preserve the smoothness of formulation \eqref{eq:formulation} can also be considered.}

Given any input vector $\mathbf{x}$, \cite{royston2002flexible} assume that the transformation of the survival function $g(S_{\mathbf{x}}(t;\bm{\beta}))$ is equal to the transformation of a baseline survival function  $S_{0}(t)$ plus a linear combination of the features: 
\begin{equation}\label{eq:g_survival}
    g(S_{\mathbf{x}}(t;\bm{\beta})) = g(S_{0}(t)) + \bm{\gamma}^T\mathbf{x},
\end{equation}

\noindent where $\bm{\gamma}$ \textcolor{black}{is the parameters vector related to the features.} 

\textcolor{darkgreen}{
%E For the form of the $g$ function, 
\textcolor{darkgreen}{Concerning the transformation $g$, 
following \citep{aranda1981two},}
\cite{royston2002flexible} define:
$$g(x;\theta) = \text{log}\frac{x^{-\theta}-1}{\theta},$$
\noindent where the value of $\theta$ determines whether the proportional odds or the proportional hazards model is obtained. %Specifically, 
\textcolor{darkgreen}{When} $\theta=1$, \textcolor{darkgreen}{we have} the proportional odds model (i.e., the classical logistic model), %E arises,
while as %E $\theta \rightarrow 0$ 
\textcolor{darkgreen}{$\theta$ tends to $0$ we recover}
the proportional hazards model. % is recovered.
}

\textcolor{darkgreen}{Following \cite{royston2002flexible}, we estimate the baseline survival function $S_{0}(t)$ using spline-based models. Specifically, in the proportional odds case we model the logarithm of the baseline cumulative odds $\text{log}\left(\tfrac{1-S_0(t)}{S_0(t)}\right)$, while in the proportional hazards case we model the logarithm of the baseline hazard $\text{log}H_{0}(t)$, both as natural cubic splines of log-time $y=\text{log}(t)$.}
%\textcolor{blue}{As in \cite{royston2002flexible}, we estimate the baseline survival function $S_{0}(t)$ by modeling, in the proportional odds case, the logarithm of the baseline cumulative odds (i.e., $\text{log}\left(\tfrac{1-S_0(t)}{S_0(t)}\right)$), and in the proportional hazards case, the logarithm of the baseline hazard function (i.e., $\text{log}H_{0}(t)$), both as natural cubic splines of log-time $y=\text{log}(t)$.}

\textcolor{darkgreen}{From \eqref{eq:g_survival},} this leads to the Proportional Odds (PO) spline model:
\vspace{-5pt}
$$g(S_{\mathbf{x}}(t;\bm{\eta},\bm{\gamma}))=\text{log}(S_{\mathbf{x}}(t;\bm{\eta},\bm{\gamma})^{-1}-1) = \text{log}\left(\frac{1-S_{\mathbf{x}}(t;\bm{\eta},\bm{\gamma})}{S_{\mathbf{x}}(t;\bm{\eta},\bm{\gamma})}\right) = \text{log}\left(\frac{1-S_0(t)}{S_0(t)}\right) + \bm{\gamma}^T\mathbf{x} = s(y;\bm{\eta})+ \bm{\gamma}^T\mathbf{x}$$

\noindent and to Proportional Hazard (PH) spline model:

\vspace{-20pt}

$$g(S_{\mathbf{x}}(t;\bm{\eta},\bm{\gamma}))= \text{log}(-\text{log}S_{\mathbf{x}}(t;\bm{\eta},\bm{\gamma}))= \text{log}H_{\mathbf{x}}(t;\bm{\eta},\bm{\gamma}) = \text{log}H_{0}(t) + \bm{\gamma}^T\mathbf{x} = s(y;\bm{\eta })+ \bm{\gamma}^T\mathbf{x},$$

\noindent where $s(y;\bm{\eta })$ is a nonlinear function approximated via natural cubic splines.

A natural cubic spline is a special type of cubic spline constrained to behave linearly beyond the boundary knots 
$k_{\text{min}}$ and $k_{\text{max}}$. Boundary knots are typically, though not necessarily, positioned at the extreme observed \textcolor{black}{logarithmic time $y=\text{log}(t)$} values with $m$ distinct internal knots $k_1 , \dots , k_m$ such that $k_{\text{min}} < k_1$ and $ k_m < k_{\text{max}}$. Assuming $\lambda_j = \frac{k_{\text{max}}-k_j}{k_{\text{max}}-k_{\text{min}}}$ and $(x-a)_{+} = \text{max}(0,x-a)$, the natural cubic splines have the following form: 

\vspace{-5pt}
$$s(y;\bm{\eta}) = \eta_0 +\eta_1 y + \eta_2 v_1(y)+ \dots + \eta_{m+1}v_{m}(y),$$

\noindent where the $j$-th basis function, for $j=1, \dots, m$, is defined as:
\vspace{-5pt}
$$v_j(y) = (x-k_j)_{+}^3 - \lambda_j(x-k_{\text{min}})_{+}^3- (1-\lambda_j)(x-k_{\text{max}})_{+}^3.$$

\noindent The derivative of $s(y;\bm{\eta})$ with respect to $y$ is:

$$\frac{\text{d}s(y;\bm{\eta})}{\text{d}y} = \eta_1 + \sum^{m}_{j=2}\eta_j[3(y-k_j)^2_{+}-3\lambda_j(y-k_{\text{min}})_{+}^2 -3(1-\lambda_j)(y-k_{\text{max}})^2_{+})].$$

The complexity of such splines is determined by the number of degrees of freedom, which is equal to $m+1$. For the placement of internal knots, \textcolor{black}{\cite{royston2002flexible} propose to use} the percentiles of the distribution of uncensored log survival times. Further details can be found in Table 1 of \cite{royston2002flexible}.

\textcolor{darkgreen}{According to \eqref{eq:neg_log_lik},} for any leaf node $n \in \tau_L$ and any right-censored data point  $(\textbf{x},t,c)$, the corresponding negative log-likelihood $L^-_n$ 
\textcolor{darkgreen}{in the PO model} %E under the PO assumption 
is:

\vspace{-0.6cm}
\begingroup
\fontsize{10.46}{13.2}\selectfont
\begin{equation}
 L^-_n({\mathbf{x}},t,c;\bm{\beta}_n) =\begin{cases}
- log \left[\frac{1}{t} \frac{\text{d}s(y;\bm{\eta}_n)}{\text{d}y}\text{exp}( s(y;\bm{\eta}_n)+ \bm{\gamma}_{n}^T\mathbf{x})(1+\text{exp}( s(y;\bm{\eta}_n)+ \bm{\gamma}_{n}^T\mathbf{x}))^{-2}\right]\hspace{0.0cm} \text{if } c = 1 \\
	- log \left[(1+\text{exp}( s(y;\bm{\eta}_n)+ \bm{\gamma}_{n}^T\mathbf{x}))^{-1}\right]\hspace{4.5cm} \text{if } c = 0,
\end{cases}
\end{equation}
\endgroup
\noindent whereas 
%E under the PH assumption 
\textcolor{darkgreen}{in the PH model}
it is as follows:
\vspace{-0.2cm}

\begingroup
\fontsize{10.46}{13.2}\selectfont
\begin{equation} L^-_n({\mathbf{x}},t,c;\bm{\beta}_n) =\begin{cases}
- log \left[\frac{1}{t} \frac{\text{d}s(y;\bm{\eta}_n)}{\text{d}y}\text{exp}( s(y;\bm{\eta}_n)+ \bm{\gamma}_{n}^T\mathbf{x}- \text{exp}( s(y;\bm{\eta}_n)+ \bm{\gamma}_{n}^T\mathbf{x}) )\right] \hspace{0.95cm} \text{if } c = 1 \\
	- log \left[\text{exp}( s(y;\bm{\eta}_n)+ \bm{\gamma}_{n}^T\mathbf{x})\right] \hspace{5.9cm} \text{if } c = 0.
\end{cases}
\end{equation}
%\noindent where $\bm{\beta}_n$ the vector consists
\textcolor{black}{The vector $\bm{\beta}_n$  consists}
of $p+m+1$ \textcolor{black}{components, where the first $p$ ones corresponds to the variables $\bm{\gamma}_n$ associated with the linear combination of the features, and the last $m+1$ ones} to the variables $\bm{\eta}_n$ related to the cubic spline.

\subsection{\textcolor{black}{Promoting fairness in soft survival trees}}

A crucial aspect in ML research is \textcolor{black}{to deal with bias and fairness issues \citep{mehrabi2021survey}. Indeed, ML models may perform an unfair discrimination by penalizing groups of data points characterized by sensitive attributes (features), like} gender or ethnicity.
%\textcolor{black}{These concerns arise} from the risk of unfair discrimination by ML models, which \textcolor{black}{may penalize groups characterized by sensitive attributes (features)}, such as gender or race. 
Such discrimination often stems from inherent biases in the data, including the marginalization of historically disadvantaged groups \citep{pedreshi2008discrimination}. \textcolor{darkgreen}{Even if a sensitive \textcolor{darkgreen}{feature} %attribute 
is removed or ignored, individuals in the sensitive group may still be disadvantaged, since other features \textcolor{darkgreen}{may often be} strongly correlated with \textcolor{darkgreen}{the group membership. %E Hence, the bias stems not from the sensitive attribute itself but from the individual’s condition as reflected in their characteristics.
%Hence, the bias may stem from such a subset of other features.[DELETE LAST SENTENCE?]
}
}

\textcolor{darkgreen}{\textcolor{darkgreen}{During the past decade,}
%Recently, 
several attempts have been made to train decision
trees % accounting for
\textcolor{darkgreen}{while accounting for fairness issues.} Recent contributions \textcolor{darkgreen}{include MILP-based} methods to incorporate group fairness into decision trees, combining feature selection with fairness constraints \citep{aghaei2019learning,wang2022synthesizing,jo2023learning}, as well as a dynamic programming approach that builds classification trees satisfying group fairness requirements \citep{van2022fair}.
To the best of our knowledge, no prior work \textcolor{darkgreen}{takes into account} fairness in survival trees.}

\textcolor{black}{In this section, we show how our SSTs can be extended to promote group fairness by reducing the ``distances'' between the survival functions of a sensitive group of data points and of its complement, based on attributes such as gender or ethnicity.}  
%Group fairness aims to ensure equality between sensitive groups, such as those defined by gender or ethnicity.

%\textcolor{black}{In this section, we show how our SSTs can be extended to promote important fairness properties. In particular, we present a way to address group fairness by controlling the distances between the survival functions of a sensitive group of data points and its complement.}  Group fairness aims to ensure equality between sensitive groups, such as those defined by ethnicity or gender.

Let \textcolor{black}{$G \subset I$ denote the sensitive group of data points (individuals)} to be protected against discrimination, \textcolor{black}{and let $\bar{G} = I\setminus G$ denote its complementary group. To induce group fairness in SSTs, we penalize the $\ell_2$ norm of the difference between survival functions corresponding to data points belonging to $G$ and $\bar{G}$ by adding in the loss function of the training formulation \eqref{eq:formulation} the following penalty term:}

%To impose that the survival functions of individuals in $G$ does not differ much from the survival functions in the complementary group $\bar{G}$, we introduce in the loss function in \eqref{eq:formulation} the following penalty term:
\begin{equation}\label{eq:fair_term}
    \sum_{i \in G}\sum_{j \in \bar{G}} \int^{t_{\text{max}}}_{t_{\text{min}}}(S_{\textbf{x}_i}(t;\bm{\beta}_{n_{\mathbf{x}_i}})-S_{\textbf{x}_j}(t;\bm{\beta}_{n_{\mathbf{x}_j}}))^2 dt,
\end{equation}

\noindent where $t_{\text{min}} = \min_{1 \leq i \leq N}\{t_i\}$ and $t_{\text{max}} = \max_{1 \leq i \leq N}\{t_i\}$ and the parameter vector $\bm{\beta}_{n_{\mathbf{x}}}$ corresponds to the leaf node variables of the HBP path related to the input vector $\mathbf{x}$.

\textcolor{black}{Thus, the nonlinear continuous optimization formulation for training SSTs promoting group fairness is as follows:}

\vspace{-0.2cm}
\begin{equation}
\label{eq:formulation_fair}
\begin{split}
\min_{\bm{\w},\bm{\beta}}  \frac{1}{N}\sum_{i \in I}\sum_{k\in \tau _{L}}P_{in}(\bm{\w})L^-_n(\mathbf{x}_i,t_i,c_i;\bm{\beta}_n) + \rho \sum_{i \in G}\sum_{j \in \bar{G}} \int^{t_{\text{max}}}_{t_{\text{min}}}(S_{\textbf{x}_i}(t;\bm{\beta}_{n_{\mathbf{x}_i}})-S_{\textbf{x}_j}(t;\bm{\beta}_{n_{\mathbf{x}_j}}))^2 dt, \end{split}
\end{equation}

\noindent where $\rho \geq 0$ is a hyperparameter to be tuned.

\section{\textcolor{black}{Node-based decomposition algorithm for training soft survival trees}}\label{sec:decomposition}

\textcolor{black}{To train a soft survival tree we need to minimize the error function in \eqref{eq:formulation} with respect to the branch and leaf node parameters $\bm{\w}$ and $\bm{\beta}$. This is a challenging nonconvex problem whose computational load increases with the size of the data set ($p$ and $N$), the depth $D$ of the tree, and the number $m$ of survival function parameters.}

\textcolor{black}{Decomposition algorithms have been devised to address similar challenges for other ML models such as Support Vector Machines \citep{chang2011libsvm}, multilayer perceptrons \citep{grippo2015decomposition} and soft classification and regression trees \citep{amaldi2021multivariate,consolo2025}. The idea is to split the training problem into a sequence of smaller subproblems, where at each iteration the error function is optimized over a subset of variables (whose indices correspond to the so-called working set), while keeping the other variables fixed at their current values.}  

\textcolor{black}{In this section we summarize the adaptation of the NOde-based DEComposition algorithm with Data points Reassignment heuristic (NODEC-DR) proposed in \citep{consolo2025} for training soft regression trees to soft survival trees, which will be referred to as NODEC-DR-SST. See \ref{sec:dec_details} for a more detailed description.}  

\textcolor{black}{NODEC-DR employes a node-based working set selection procedure that preserves the separation between branch node variables and leaf node variables. Furthermore, it includes an ad hoc heuristic for reassigning the data points throughout the tree to achieve a more balanced routing along the root-to-leaf-node paths, and hence better exploit the representative power of the soft survival tree.}

\textcolor{black}{In NODEC-DR-SST we consider the following $\ell_2$-regularized version of the training formulation \eqref{eq:formulation}: 
\begin{equation}\label{eq:obj}
\min_{\bm{\w},\bm{\beta}} E(\bm{\w,\beta})= \frac{1}{N}\sum_{i \in I}\sum_{k\in \tau _{L}}P_{in}(\bm{\w})L^-_n({\mathbf{x}_i},t_i,c_i;\bm{\beta}_n)+  \frac{\lambda_{\bm{\beta}}}{2} \|\bm{\beta}\|^2,
\end{equation}
where $\lambda_{\bm{\beta}}\geq 0$ is the regularization hyperparameter.}

\textcolor{black}{Once an initial solution $(\bm{\w}^{0},\bm{\beta}^{0})$ has been determined at random or using an ad hoc initialization procedure (see \ref{sec:dec_details}), NODEC-DR-SST proceeds with a main external loop of macro iterations, indexed by $it$, which is repeated until a termination criterion (involving $\| \nabla E \|$ or a maximum number of iterations) is met. Each macro iteration $it$ includes an internal loop of inner iterations, indexed by $k$, in which the working set is identified as the set of indices corresponding to a subset of nodes. At each inner iteration $k$, one branch node $s \in \tau_B$ is selected and the corresponding working set is defined as the subset of indices of all the branch and leaf nodes belonging to the subtree rooted at $s$. We distinguish the branch node working set $W^k_B \subseteq \tau_B$ \textcolor{black}{consisting of all the indices of the descendant branch nodes of $s$, denoted by ${\cal D}_B(s)$, from the} leaf node working set $W^k_L \subseteq \tau_L$ \textcolor{black}{consisting of all the indices of the descendant leaf nodes of $s$, denoted by ${\cal D}_{L}(s)$}. See Figure \ref{fig:tree_dec1} for an example of subtree-based working set. Whenever a node index $n$ is included in $W^k_B \subseteq \tau_B$ (or $W^k_L \subseteq \tau_L$), the corresponding variables, $\bm{\w}_n$ (or $\bm{\beta}_n$), are jointly considered in the corresponding subproblem. During each macro iteration, all the branch nodes are considered in turn as root $s$ of a subtree, except the SST root which is considered by itself.}

\begin{figure}[H]
    \centering
    \includegraphics[width=0.5\textwidth]{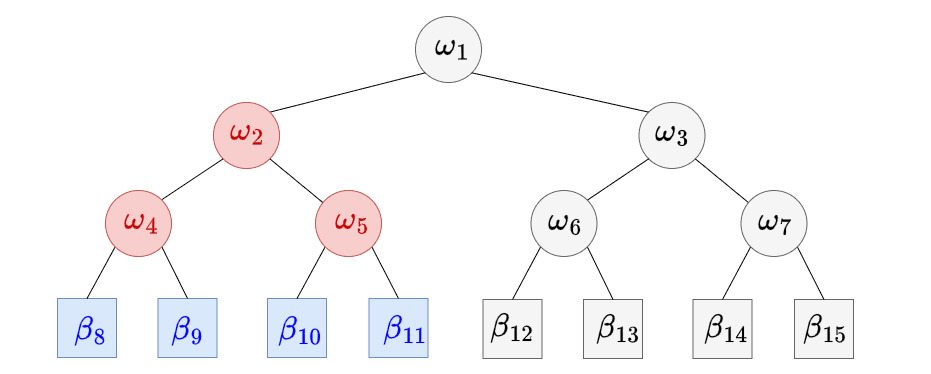}
    \caption{\textcolor{black}{Example of NODEC-DR-SST working set selection for a SST of depth $D=3$. The branch node $r_s=2$ is selected, along with the associated working sets $W_B=\{2,4,5\}$ (red) and $W_L = \{8,9,10,11\}$ (blue). The associated variable vectors $\bm{\w}$ and $\bm{\beta}$ are indicated inside each node.}}
    \label{fig:tree_dec1}
\end{figure}

\textcolor{black}{At each inner iteration $k$ of NODEC-DR-SST, \eqref{eq:obj} is first optimized with respect to the branch node variables whose node indices are in $W^k_B$ ({\bf BN Step}), and then with respect to the leaf node variables whose node indices are in $W^k_L$ ({\bf LN Step}). The two steps are as follows:
\begin{itemize}
 \item \textbf{BN Step}: starting from $(\bm{\w}^{k},\bm{\beta}^k)$ determine $\bm{\w}^{k+1}_n$ for $n\in W^k_B$ by minimizing $E(\bm{\w},\bm{\beta})$ with respect to
$\bm{\w}_n$ for $n\in W^k_B$ and return $(\bm\w^{k+1},\bm\beta^k)$ where $\bm{\w}_n^{k+1}=\bm{\w}_n^k$ for $ n\notin W^k_B$.
\vspace{-8pt}
\item \textbf{LN Step}: starting from $(\bm{\w}^{k+1},\bm{\beta}^k)$ determine $\bm{\beta}^{k+1}_n$ for $n\in W^k_L$ by minimizing $E(\bm{\w},\bm{\beta})$ with respect to $\bm{\beta}_n$ for $n\in W^k_L$ and return $(\bm\w^{k+1},\bm\beta^{k+1})$ where $\bm{\beta}_n^{k+1}=\bm{\beta}_n^k$ for $ n\notin W^k_L$.
%    \item \textbf{LN Step}: starting from $(\bm{\w}^{k+1},\bm{\beta}^k)$ minimize $E(\bm{\w},\bm{\beta})$  with respect to $\bm{\beta}_n$ for $n\in W^k_L$ determine $\bm{\beta}^*_n$ for $n\in W^k_L$, and update $(\bm\w^{k+1},\bm\beta^{k+1})$ by setting $\bm{\beta}_n^{k+1}=\bm{\beta}^*_n$ for $n\in W^k_L$ and $\bm{\beta}_n^{k+1}=\bm{\beta}_n^k$ for $ n\notin W^k_L$.
\end{itemize}
In the \textbf{BN Step}, we consider the restricted training set, denoted by $I_{s}$, containing only the data points which deterministically fall into the node $s$, and the restricted $E(\bm{\w},\bm{\beta})$ is minimized with respect to $\bm{\w}_n$ for $n\in W^k_B$ while all the other variables are kept fixed at their current values.}

\textcolor{black}{The pseudocode of NODEC-DR-SST is reported below.}

\renewcommand{\thealgorithm}{NODEC-DR-SST} %RIMUOVE IL NUMERO
\begin{algorithm}[H]
\floatname{algorithm}{} %PER RIMUOVERE ALGORITHM
\caption{- Decomposition training algorithm}
\label{alg:Decomposition1}
$\textbf{Input:}\mbox{ }\text{depth} \,D;\; \text{training set} \,I;\,\text{max}\,\text{number}\,\text{of}\,\text{iterations}\;M\_it;\; \varepsilon_{1}^0,\,\varepsilon_{2}^0,\,\varepsilon_{3}^0>0;\,\theta \in (0,1);\\(\bm{\w}^{0},\bm{\beta}^{0})\\\textbf{Output:}\, (\bm{\hat{\w}},\,\bm{\hat{\beta}})$
\vspace{-17pt}
  {\fontsize{9}{10}\selectfont
  \begin{algorithmic}[1]
  \Statex {} 
    %\State $(\bm{\w}^{0},\bm{\beta}^{0}) \gets\textsc{InitializationProcedure}(I,D,r)$
    \Procedure{NODEC-DR-SST}{$D,I,(\bm{\w^0},\bm{\beta^0}),\varepsilon_{1},\varepsilon_{2},\varepsilon_{3},\theta$}
      
      \State  $(\bm{\hat{\w}},\bm{\hat{\beta}}) \gets (\bm{\w}^{0},\bm{\beta}^{0}); \;\varepsilon_{1},\,\varepsilon_{2},\,\varepsilon_{3}\gets \varepsilon_{1}^0,\,\varepsilon_{2}^0,\,\varepsilon_{3}^0;\; k \gets 0;\, it \gets 1$
      %\State $\bullet$ $\varepsilon_{1},\,\varepsilon_{2},\,\varepsilon_{3}\gets \varepsilon_{1}^0,\,\varepsilon_{2}^0,\,\varepsilon_{3}^0$
      \State $error_{best} \gets E(\bm{\hat{\w}},\bm{\hat{\beta}})$
      %\State $\bullet$ $k \gets 0,\, it \gets 1$
      \While{$ it < M\_it\,and\,$ not {\it termination criterion}}
       
        \For{$ s \in \{1,...,2^{D}-1\} $}\Comment{loop over the branch nodes}
        \State \textbf{if} $s == 1\,\text{and}\, D > 1$ \textbf{then}\; $W^k_B \gets \{s\}, \, W^k_L \gets \emptyset$
        \State \textbf{else} \; $W^k_B \gets \{s,{\cal D}_{B}(s)\}, \, W^k_L \gets {\cal D}_{L}(s)$ \; \textbf{end if}
        %\If{$s == 1\,\text{and}\, D > 1$}
         %$W^k_B \gets \{s\}, \, W^k_L \gets \emptyset$
        %\Else \; $W^k_B \gets \{s,{\cal D}_{B}(s)\}, \, W^k_L \gets {\cal D}_{L}(s)$
        %\EndIf
        
        %\State $W^k_B \gets \{t\}, \, W^k_L \gets children_leaf(t)$
       %\Statex
        \Statex  {$\triangleright$ \, \textbf{BN Step}} \textit{(Optimization with respect to a subset of branch nodes parameters)}

        \State  $\bm{\w}_{W^k_B}^{k+1} \gets $\textsc{UpdateBranchNode}$(I,W^k_B,\bm{\w}^{k}_{W^k_B},\varepsilon_{1},\varepsilon_{2},\varepsilon_{3})$ \Comment{update determined using heuristic}    
        %\Statex
        \Statex  {$\triangleright$ \, \textbf{LN Step}} \textit{(Optimization with respect to a subset of leaf nodes parameters)}

            \State compute $\bm{\beta}_{W^k_L}^{k+1} $ by any method
        \State \textbf{if} $E(\bm{\w}^{k+1},\bm{\beta}^{k+1})< error_{best}$ \textbf{then} %\; $error_{best} \gets E(\bm{\w}^{k+1},\bm{\beta}^{k+1})$; \; ($\bm{\hat{\w}},\bm{\hat{\beta}}) \gets (\bm{\w}^{k+1},\bm{\beta}^{k+1})$ \; \textbf{end if}
        \State  \quad \; $error_{best} \gets E(\bm{\w}^{k+1},\bm{\beta}^{k+1})$
        \State  \quad \; ($\bm{\hat{\w}},\bm{\hat{\beta}}) \gets (\bm{\w}^{k+1},\bm{\beta}^{k+1})$ \; \textbf{end if}
        %\If{$E(\bm{\w}^{k+1},\bm{\beta}^{k+1})< error_{best}$}
        %\State  $error_{best} \gets E(\bm{\w}^{k+1},\bm{\beta}^{k+1})$
        %\State  ($\bm{\hat{\w}},\bm{\hat{\beta}}) \gets (\bm{\w}^{k+1},\bm{\beta}^{k+1})$
        %\EndIf
        %\State \texttt{left_{t} = descendant(t,right=\bm{True})}
        \State $k \gets k + 1$
        \EndFor
	  \State  $\varepsilon_{1},\,\varepsilon_{2},\,\varepsilon_{3} \gets \theta\,\varepsilon_{1},\,\theta\,\varepsilon_{2},\,\theta\,\varepsilon_{3};\; it \gets it + 1$
      \EndWhile\label{Decompositiondwhile}
      \State \textbf{return} $(\bm{\hat{\w}},\bm{\hat{\beta}})$
    \EndProcedure
  \end{algorithmic}
  }
\end{algorithm} 

\textcolor{black}{In the \textbf{BN Step}, the new update  
$\bm{\w}_{W_{B^k}}^{k+1}$ is determined using the \textsc{UpdateBranchNode} procedure, whose details are \textcolor{black}{described} in \ref{sec:dec_details}. Since the {\bf BN Step} problem is nonconvex, standard optimization methods often lead to poor-quality solutions such that the HBP paths of a significant portion of the data points end at a few leaf nodes. To address this imbalance issue, \textsc{UpdateBranchNode} includes a heuristic that reassigns data points by partially modifying their HBP paths, deterministically rerouting them across the tree to ensure a more balanced distribution along the root-to-leaf paths. When imbalance occurs, the goal is to adjust the routing at each branch node for a selected subset of data points to achieve a better balance between the number of data points following its two branches. Three positive thresholds, $\varepsilon_1$, $\varepsilon_2$ and $\varepsilon_3$, are introduced to detect the degree of imbalance across the tree and to enhance stability during the initial macro iterations of NODEC-DR-SST. The thresholds $\varepsilon_1$ and $\varepsilon_2$ (with $\varepsilon_1 > \varepsilon_2$) account for the low and high levels of imbalance, respectively, while $\varepsilon_3$ controls the fraction of data points to be routed along the other branch. After each macro iteration, the values of $\varepsilon_1$, $\varepsilon_2$, and $\varepsilon_3$ are decreased, allowing for progressively larger imbalance.}

\textcolor{black}{To determine the level of imbalance at a branch node $n$, we define $L_n$ as the ratio of the number of data points deterministically routed along the left branch of node $n$ to the cardinality of the restricted training set $I_n$. If the data point routing at branch node $n$ is sufficiently balanced, the update
$\bm{\w}_{W^k_{B}}^{k+1}$ is obtained by minimizing the error function in \eqref{eq:obj} associated with the restricted subtree rooted at $n$ and the restricted training set $I_n$, with respect to the variables $\bm{\w}_n$ for $n \in W_B^k$. If $L_n$ exceeds $\varepsilon_1$ 
but is below $\varepsilon_2$ (moderate imbalance), one tries to
improve the routing balance by solving a surrogate problem that updates the parameters of node $n$ while accounting for the relative weights of data points routed along its left and right branches. In this case,
$\bm{\w}_{W^k_{B}}^{k+1}$ is obtained by optimizing a \textcolor{black}{two-class} Weighted Logistic Regression (WLR) problem with respect to the variables $\bm{\w}_n$ and the restricted training set $I_n$, while keeping fixed all $\bm{\w}_{l}$ for $l \in \tau_B \setminus \{n\}$. Finally, if $L_n$ exceeds $\varepsilon_2$ (high level of imbalance), $\bm{\w}^{k+1}_{W_B^k}$ is obtained, as in the case of moderate imbalance, by minimizing a WLR function with respect to only the $\bm{\w}_n$ variables of branch node $n$. However, we try to reroute a fraction $\varepsilon_3$ of the data points in $I_n$, specifically those with a larger term in the negative log-likelihood, towards the other branch of node $n$ by swapping the class  (left branch versus right branch) in the corresponding WLR problem.}

\textcolor{black}{In the \textbf{LN Step}, the new update $\bm{\beta}_{W^k_{L}}^{k+1}$ is determined by minimizing a negative log-likelihood function using any appropriate unconstrained optimization solver. This is in line with the standard approach in parametric and semiparametric survival models, where the parameters of the survival functions are estimated via maximum likelihood.
% estimation. 
Similar to the \textbf{BN Step}, at each iteration $k$ the leaf node variables $\bm{\beta}$ are updated by considering the restricted training set $I_{s}$ of the branch node $s$, which is the root of the subtree defined by the working sets $W^k_{B}$ and $W^k_{L}$.  
% Similar to the \textbf{BN Step}, at each iteration $k$, the update of the variables $\bm{\beta}$ is performed using only the restricted training set $I_{s}$, which consists of the data points that deterministically fall into the branch node $s$ (i.e. the root node of the subtree defined by the working sets $W^k_{B}$ and $W^k_{L}$).  
It is worth pointing out that when using the objective function \eqref{eq:formulation_fair} to address fairness, the branch node variables $\bm{\w}$ remain fixed at their current values during the \textbf{LN Step}. Thus, for any input vector $\mathbf{x}$, the unique leaf node $n_{\mathbf{x}} \in \tau_L$ in the corresponding HBP path can be readily determined.}

 %This restricted training set is used to update $\bm{\beta}_n$ for all $n \in {\cal D}_L( s )$.
\section{Experimental results}\label{sec:experiments}

%This section aims to present the performance of our SSTs compared to standard tree-based benchmarking methods, while also emphasizing their ability to enhance interpretability and promote fairness. 

\textcolor{black}{In this work, we are interested not only in assessing the performance of SSTs and comparing them with three tree-based benchmarking methods, but also in demonstrating their considerable flexibility. In particular, we show the advantages of adopting different parametric or spline-based semiparametric survival models at leaf nodes, and we illustrate how the model interpretability can be substantially improved and one important type of fairness can be induced.} 
%In this section, we provide different types of numerical results to demonstrate the performance and the flexibility of our SSTs, which allow to consider different survival models (parametric or spline-based semiparametric) at leaf nodes, and to illustrate their ability to enhance interpretability and to promote fairness.

After describing the considered datasets and the experimental settings, \textcolor{darkgreen}{Section \ref{sec:perf_meas} provides a brief overview of the performance measures adopted in this work.} In Section \ref{sec:results_exp} we compare the performance of SSTs trained via our node-based decomposition training algorithm with three widely used methods, namely, the survival tree Scikit-Survival (SkSurv) in \citep{leblanc1993survival}, Conditional Inference Trees (CTree) in \citep{hothorn2006unbiased}, and the CART-like Recursive Partitioning and Regression Trees (RPART) \citep{breimanclassification}. 
In Section \ref{sec:interpretability}, we illustrate how our SSTs enhance interpretability by inducing a partition of all the data points into clusters associated to the different leaf nodes with distinct survival functions. In Section \ref{sec:fairness}, we show that the SST model and training formulation can be extended to promote group fairness in an unemployment dataset for a socioeconomic study \citep{romeo1999conducting}.

\subsection{Datasets and experimental setting}
\label{sec:dataset_exp} 

In the numerical experiments, we consider 15 real-world survival datasets, \textcolor{black}{all the
11 datasets from \cite{zhang2024optimal} and the first 4 listed in Table 1 of \cite{huisman2024optimal}}. Details about the type of application and the repositories containing the datasets are provided in \ref{sec:dataset_details}. Table \ref{tab:dataset} \textcolor{black}{lists the names and the main characteristics of the datasets, namely, the number of features $p$, the number of data points $N$, and the censoring level expressed as a percentage.} For all datasets, all input features are normalized to the $[0, 1]$ range, while missing values are handled by imputing the mean for continuous features and the mode for categorical ones.

\begin{table}[h]
\centering
\resizebox{0.42\textwidth}{!}{
\begin{tabular}{lccc}
\hline
\multicolumn{1}{c}{Dataset} & N     & p  & Censoring level \\ \hline
Aids                        & 1151  & 11 & 91.7 $\%$       \\
Aids\_death                 & 1151  & 11 & 91.7 $\%$       \\
Aids2                       & 2839  & 4  & 38.0 $\%$       \\
Churn                       & 2000  & 12 & 53.4 $\%$       \\
Credit                      & 1000  & 19 & 30.0 $\%$       \\
Dialysis                    & 6805  & 4  & 76.4 $\%$       \\
Employee                    & 11992\tablefootnote{The initial dataset contained 15000 data points, but we applied the preprocessing steps outlined in \cite{fotso2019pysurvival}, as described in \url{https://square.github.io/pysurvival/tutorials/employee_retention.html}.} & 7  & 77.2 $\%$       \\
Flchain                     & 7478  & 9  & 72.5 $\%$       \\
Framingham                  & 4658  & 7  & 68.5 $\%$       \\
Gbsg2                       & 686   & 8  & 56.4 $\%$       \\
Maintenance                 & 1000  & 5  & 60.3 $\%$       \\
Uissurv                     & 628    & 12 & 19.1 $\%$       \\
Unempdur                    & 3241  & 6  & 38.7 $\%$       \\
Veterans                    & 137   & 6  & 6.6 $\%$        \\
Whas500                     & 500   & 14 & 57.0 $\%$         \\
\hline
\end{tabular}
}
\caption{Summary of the 15 datasets selected for the comparative experiments}\label{tab:dataset}
\end{table}

The experiments are carried out on a PC with Intel(R) Core(TM) i7-11370H CPU @ 3.30GHz with 16 GB of RAM. \textcolor{black}{The SST model and the NODEC-DR-SST are implemented in \texttt{Python} 3.10.14.} As previously mentioned, the performance of our SSTs is compared with that of the following implementation of the above-mentioned three methods: SkSurv \citep{polsterl2020scikit}, CTree \citep{hothorn2015ctree}, and RPART \citep{therneau2021rpart}\footnote{The first from the Scikit-Survival package, and the last two from the CRAN package pec. Further information are reported in \ref{sec:soft_pack}.}.
Different measures are adopted to assess the  discrimination and calibration capabilities, namely, Harrell’s C-index ($C_H$), Uno’s C-index ($C_U$), and the Cumulative Dynamic Area Under the Curve (\text{CD-AUC}) as discrimination measures, and the Integrated Brier Score (\text{IBS}) as calibration one. \textcolor{darkgreen}{In Section \ref{sec:perf_meas}, we provide a concise overview of the four measures considered in this study (a detailed description and references can be found in \ref{sec:performance}), with particular emphasis on $C_H$ and IBS. These two measures, which assess discrimination and calibration respectively, are the ones whose results are reported in Section \ref{sec:results_exp} (results for $C_U$ and CD-AUC on the same experiments are provided in \ref{sec:tables_sd}).} %See \ref{sec:performance} for a detailed description and references.
Such performance measures are estimated by means of $k$-fold cross-validation, with $k=5.$
Due to the non convexity of the SST training formulation, 20 random initial solutions are considered for each fold. The final testing performance measures are then averaged over 5 folds and 20 initial solutions, that is, over 100 runs.
\textcolor{darkgreen}{Furthermore, to evaluate the statistical significance of the comparison between the best parametric and semiparametric SST and the best benchmark model, we perform statistical tests on the mean differences of the testing accuracies for each of the four considered measures. In particular, we employ the one-sided non-parametric Wilcoxon signed-rank test, which is preferred over the classical t-test to avoid the restrictive assumption of normally distributed differences.}

%\textcolor{black}{BN and LN steps are not yet defined. I guess they will be, in Section 2, in the subsection missing on decomposition and how to find the optimal parameters.}

\textcolor{black}{In the {\bf BN Step}, the SLSQP solver is used, except in the imbalanced cases where a Weighted Logistic Regression problem is solved via} the \texttt{LogisticRegression} function from the \texttt{sklearn} 1.5.2 package with its default settings. Concerning the {\bf LN Step}, the SLSQP solver is generally used for \textcolor{black}{SSTs with parametric and spline-based semiparametric survival function models.}%both parametric distributions and the spline-based semiparametric approach. 
\textcolor{black}{When numerical issues arise (mainly due to overflow), a possibly better initial solution for SLSQP is obtained with a \textcolor{black}{Nelder-Mead} solver. If SLSQP still does not provide a solution, only the  
\textcolor{black}{Nelder-Mead solver}
is applied (with at most 40,000 iterations).}

\textcolor{black}{As far as the initialization procedure for the Llog parametric distribution is concerned, we adopt} the same clustering-based method proposed in \citep{consolo2025} \textcolor{black}{for soft regression trees, based on the \texttt{Kmeans} function of the \texttt{sklearn} package}. \textcolor{black}{For SSTs with spline-based semiparametric survival functions,} \textcolor{black}{at each leaf node $n$ the initial solutions for the corresponding $\bm{\beta}_n$ vector \textcolor{black}{are} obtained using the method described in \citep{royston2002flexible}. In particular, least-squares regression is applied to the logarithms of the cumulative hazards values (related to PH) or the cumulative odds values (related to PO), which are estimated from a random subset of uncensored data points (a fraction $\frac{1}{2^D}\%$ sampled with replacement) identified by fitting a standard Cox model.} The COXHFitter function from the \texttt{lifelines} 0.29.0 package is used for fitting the Cox model. \textcolor{black}{In the few cases (only a few runs for two datasets) where the COXHFitter function faces numerical issues, random values are assigned to the leaf node $\beta$ variables}\footnote{\textcolor{black}{The values of the leaf node $\beta_n$ parameters } are selected to guarantee the positivity of the derivative $\frac{\text{d}s(y;\bm{\eta})}{\text{d}y}$, which is the argument of the logarithm in the negative log-likelihood.}. 

\textcolor{black}{Concerning the spline-based PH and PO survival functions at the SSTs leaf nodes, we consider $m=2$ internal knots, placed according to Table 1 in \citep{royston2002flexible}.}

As done in \citep{consolo2025}, we adopt an early-stopping strategy for the NODEC-DR algorithm, where $M_{it}=10$, $k_0 > M_{it}$, and the hyperparameters $\varepsilon_1=0.1$, $\varepsilon_2=0.3$, $\varepsilon_3=0.4$, and $\zeta=0.8$ are used for the data points reassignment heuristic. \textcolor{black}{To improve numerical stability, an $\ell_2$-penalty term is added to the error function \eqref{eq:formulation}. This term amounts to $\ell_2$ norm of the survival function variables $\bm{\beta}_{n}$ in the leaf nodes, multiplied by a constant value of $2$.}

\subsection{\textcolor{darkgreen}{Discrimination and calibration measures}}
\label{sec:perf_meas}
\textcolor{darkgreen}{Performance measures for survival models are typically concerned with two aspects: discrimination and calibration. Discrimination evaluates a model’s ability to separate \textcolor{darkgreen}{individuals} %subjects 
who experience the event from those who do not, by assigning higher predicted probabilities to the former \textcolor{darkgreen}{ones}. Calibration assesses the agreement between predicted probabilities and observed outcomes.}

\textcolor{darkgreen}{Among \textcolor{darkgreen}{the} discrimination measures, Harrell’s Concordance Statistic ($C_H$) is the most widely applied in survival analysis \citep{yan2008investigating}. $C_H$ is based on the principle that a reliable survival model should assign lower survival probabilities to \textcolor{darkgreen}{individuals} %subjects 
with shorter times-to-event. To define the concordance measure, it is necessary to introduce the notions of comparable and concordant pairs \textcolor{darkgreen}{of individuals}. A pair $(i,j)$ \textcolor{darkgreen}{of individuals} is \textit{comparable} if the order of \textcolor{darkgreen}{the} events can be determined and it is \textit{concordant} when the \textcolor{darkgreen}{individual} with the earlier event receives a lower predicted survival probability. Since survival predictions depend on time, concordance is assessed at $\min\{t_i,t_j\}$. Formally, Harrell’s C-index is defined as:}

\textcolor{darkgreen}{$$C_H = \frac{\sum\limits_{i} \sum\limits_{j} c_i \cdot \mathds{1}_{t_i < t_j} \cdot \left( \mathds{1}_{\hat{S}_{\mathbf{x}_i}(t_i) < \hat{S}_{\mathbf{x}_j}(t_i)} + 0.5 \cdot \mathds{1}_{\hat{S}_{\mathbf{x}_i}(t_i) = \hat{S}_{\mathbf{x}_j}(t_i)} \right)}{\sum\limits_{i} \sum\limits_{j} c_i \cdot \mathds{1}_{t_i < t_j}}.$$}
\textcolor{darkgreen}{$C_H$ ranges from 0 to 1 and higher values indicate better model discrimination.}

\textcolor{darkgreen}{Since $C_H$ may be biased \textcolor{darkgreen}{for} %in 
datasets with a high proportion of censored data points, \cite{uno2011c} introduced an extended version of the concordance index ($C_U$) based on Inverse Probability of Censoring Weights (see \ref{sec:performance} for details), which %enhances its robustness
\textcolor{darkgreen}{which is more robust} under extensive censoring. Like $C_H$, $C_U$ ranges from 0 to 1, with higher values indicating better discrimination.}

\textcolor{darkgreen}{%Among the discrimination measures used in survival analysis, there is also
\textcolor{darkgreen}{Another popular discrimination measure used in survival analysis is} the Cumulative Dynamic Area Under the Curve (CD-AUC). This measure extends the classical AUC \textcolor{darkgreen}{for}
%from 
binary classification to time-to-event data by using time-dependent %ROC 
\textcolor{darkgreen}{Receiver Operating Characteristic functions \citep{steyerberg2009clinical}}, where at each time $t$ true positives are \textcolor{darkgreen}{individuals} %subjects 
experiencing the event at or before $t$, and true negatives are those remaining event-free beyond $t$. The CD-AUC summarizes
 time-dependent AUC values over a chosen interval, yielding an overall measure of discrimination across the \textcolor{darkgreen}{the considered time interval}. %The time-dependent AUC evaluates how well a model discriminates events up to $t$, and the CD-AUC summarizes these values over a chosen time interval, providing an overall assessment of discrimination across the study period. 
Like the standard AUC, its values range from 0 to 1, with higher values indicating better discrimination. The reader is referred to \ref{sec:performance} for further details.}

\textcolor{darkgreen}{As to calibration, a widely used measure is the Brier Score (BS), originally developed for probabilistic classification and later adapted to censored survival data \citep{graf1999assessment}. In survival analysis, $BS(t)$ at time $t$ is the mean squared error between \textcolor{darkgreen}{the} observed outcomes and \textcolor{darkgreen}{the} predicted survival probabilities $\hat{S}_{\mathbf{x}_i}(t)$, adjusted through inverse probability of censoring weights:}

$$
\textcolor{darkgreen}{BS(t) = \frac{1}{N} \sum_{i=1}^N \left( 
\frac{(\hat{S}_{\mathbf{x}_i}(t) - 0)^2}{\hat{G}(t_i)} \cdot \mathds{1}_{t_i \leq t, c_i = 1} 
+ \frac{(\hat{S}_{\mathbf{x}_i}(t) - 1)^2}{\hat{G}(t)} \cdot \mathds{1}_{t_i > t} 
\right),} $$
\noindent \textcolor{darkgreen}{where $\hat{G}$ represents the Kaplan–Meier estimate of the censoring distribution $\mathbf{c}$.}

\textcolor{darkgreen}{
Similar to CD-AUC, the Integrated Brier Score (IBS) 
\textcolor{darkgreen}{defined as
\begin{equation*}
   \textcolor{darkgreen}{\text{IBS} = \frac{1}{t_{\text{max}}} \int_{0}^{t_{\text{max}}} BS(t) \, dt }
\end{equation*}
where $t_{\text{max}} = \max_{i=1}^{N} t_i$ being the latest observed time point, provides a global assessment of the odel performance over the time interval $[0, t_{\text{max}}]$.
BS and IBS range from 0 to 1, with lower values corresponding to better calibration, see \ref{sec:performance} for further details.
%[REFERENCES + pointer to the Appendix?]
%The reader is referred to \cite{...} and \ref{sec:performance} for further details.
}
}
%\begin{equation*}
%    \textcolor{darkgreen}{\text{IBS} = \frac{1}{t_{\text{max}}} \int_{0}^{t_{\text{max}}} BS(t) \, dt }
%\end{equation*}
%\textcolor{darkgreen}{The IBS is non-negative, and lower values indicate better model performance.}
% Similar to CD-AUC, the Integrated Brier Score (IBS) provides a global assessment over a time horizon $[0, t_{\text{max}}]$, with $t_{\text{max}} = \max_{i} t_i$ being the latest observed time point:}
%\begin{equation*}
%    \textcolor{darkgreen}{\text{IBS} = \frac{1}{t_{\text{max}}} \int_{0}^{t_{\text{max}}} BS(t) \, dt }
%\end{equation*}
%\textcolor{darkgreen}{The IBS is non-negative, and lower values indicate better model performance.}

\subsection{Comparison with three benchmark survival tree methods}
\label{sec:results_exp}

\textcolor{black}{In the numerical experiments, we consider SSTs with parametric or spline-based semiparametric models for the survival functions.} SSTs involving the parametric Exponential, Weibull, and LogLogistic distributions are denoted as Exp, W, and Llog, respectively. For the semiparametric \textcolor{black}{survival function models}, SSTs 
\textcolor{darkgreen}{with the PO and PH models}
%based on the PO and PH assumptions 
are referred to as PO and PH, respectively. To show the impact of an appropriate initial solution \textcolor{black}{on the calibration and discrimination measures, we also report the results for the Llog parametric distribution when NODEC-DR-SST is combined} with the K-means-based initialization procedure proposed in \citep{consolo2025}, referred to as Llog-init. %As to the benchmark survival tree models, in this work we consider SkSurv, CTree, and RPART. 

\textcolor{black}{For SSTs we consider depths $D = 1$ and $D = 2$, and for the three benchmarking survival tree methods depths ranging from $D=2$ to $D=5$.} %\textcolor{black}{For the sake of space, this section reports results for two performance measures, namely, $C_H$ for discrimination and IBS for calibration. The results include SSTs with depth $D = 2$ and are compared with those obtained using SkSurv, CTree, and RPART, with trees of depth $D = 5$. The complete results for the other two performance measures ($C_U$ and CD-AUC) and for the other depths can be found in \ref{sec:tables_sd}.} 
\textcolor{black}{For the sake of space, in this section we only report the results in terms of two performance measures, $C_H$ and IBS, obtained for SSTs of depth $D = 2$ and for SkSurv, CTree, and RPART, with trees of depth $D=5$. The complete results for the other two performance measures ($C_U$ and CD-AUC) and for the other depths can be found in \ref{sec:tables_sd}.} 
%It is worth pointing out that the choice to show only the depth $D=5$ for the benchmark models is only due to space limitations. 
\textcolor{black}{It is worth pointing out that for all the three benchmarking methods and for all the performance measures the best average results are consistently obtained with the trees of depth $D=5$, which are compared with our very shallow trees with depth $D=2$.}

\textcolor{black}{Tables \ref{tab:ch_test} and \ref{tab:ibs_test_sec} report the average testing $C_H$ and IBS for each method.} The arithmetic averages over all datasets appear at the bottom of each table. \textcolor{darkgreen}{Within each group of approaches, namely parametric SST, semiparametric spline-based SST, and benchmark models for each dataset, the best model is highlighted in bold.}

Concerning the $C_H$ discrimination measure in Table \ref{tab:ch_test}, for which \textcolor{black}{higher values correspond to better solutions}, 
\textcolor{black}{Llog and Llog-init (with the clustering-based initialization)} achieve the best average testing accuracy among the SSTs with parametric or semiparametric models at leaf nodes, and perform better than the three benchmarking methods. In particular, the average testing values for Llog, CTree (the best benchmarking method), and both the comparable semiparametric models PO and PH are, $C_H=0.744$, $C_H=0.711$, and $C_H=0.735$ and $C_H=0.736$. Overall, Llog outperforms CTree on 14 out of the 15 datasets, while PO and PH on 13 of them. \textcolor{darkgreen}{By applying the Wilcoxon test between Llog and CTree, a p-value of $6.6e^{-4}$ is obtained. When comparing PO and PH with CTree, the resulting p-values are $7.3e^{-3}$ and $5.7e^{-3}$, respectively. These results confirm, at the $1\%$ significance level, the improvement of SSTs over the best benchmark model.}  %In particular, the average testing values for Llog, PH (the best SSTs with semiparametric models), and CTree (the best benchmarking method) are, respectively, $C_H=0.744$, $C_H=0.736$ and $C_H=0.711$.} Overall, Llog outperforms CTree on 14 out of the 15 datasets \textcolor{black}{, while PH on 13 of them.}
The Employee dataset is the only one for which SkSurv, CTree, and RPART yield better $C_H$ values than the SST alternatives for all the four performance measures. Note that Llog-init, which  includes an appropriate initialization procedure, leads to a one percent improvement with respect to Llog for the three discrimination measures, see \ref{sec:performance}. \textcolor{darkgreen}{The superiority of Llog-init in terms of accuracy is further supported by the p-values below $1\%$ obtained in the comparison with CTree for $C_H$, $C_U$, and $\text{CD-AUC}$, which are $3.79e^{-3}$, $5.3e^{-3}$, and $5.29e^{-1}$, respectively. }

\begin{table}[ht]
\centering
\resizebox{\textwidth}{!}{
\renewcommand{\arraystretch}{0.85}
\begin{tabular}{lccccccccc}
\cline{2-10}
             & \multicolumn{9}{c}{Testing $C_H$} \\
\cline{2-10}
             & \multicolumn{4}{c}{D=2} & \multicolumn{3}{c}{D=5} & \multicolumn{2}{c}{D=2} \\\hline
Dataset      & \textbf{Llog} & \textbf{Llog-init} & \textbf{Exp} & \textbf{W} & \multicolumn{1}{|c}{\textbf{SkSurv}} & \textbf{RPART} & \textbf{CTree} & \multicolumn{1}{|c}{\textbf{PO}} & \textbf{PH} \\\hline
Aids         & 0.741 & \textbf{0.754} & 0.748 & 0.743 & \multicolumn{1}{|c}{0.621} & \textbf{0.727} & 0.706 & \multicolumn{1}{|c}{0.709} & \textbf{0.715} \\
Aids\_death  & 0.753 & \textbf{0.810} & 0.727 & 0.750 & \multicolumn{1}{|c}{0.545} & 0.624 & \textbf{0.645} & \multicolumn{1}{|c}{\textbf{0.703}} & 0.692 \\
Aids2        & 0.552 & \textbf{0.557} & \textbf{0.557} & 0.554 & \multicolumn{1}{|c}{\textbf{0.540}} & 0.518 & 0.537 & \multicolumn{1}{|c}{\textbf{0.560}} & 0.548 \\
Churn        & \textbf{0.838} & \textbf{0.838} & 0.830 & 0.832 & \multicolumn{1}{|c}{0.778} & 0.779 & \textbf{0.781} & \multicolumn{1}{|c}{\textbf{0.837}} & 0.834 \\
Credit risk  & \textbf{0.751} & 0.747 & 0.708 & 0.740 & \multicolumn{1}{|c}{0.718} & 0.723 & \textbf{0.726} & \multicolumn{1}{|c}{\textbf{0.745}} & 0.741 \\
Dialysis     & \textbf{0.735} & \textbf{0.735} & 0.733 & \textbf{0.735} & \multicolumn{1}{|c}{\textbf{0.642}} & 0.609 & \textbf{0.642} & \multicolumn{1}{|c}{0.720} & \textbf{0.728} \\
Employee     & 0.827 & \textbf{0.882} & 0.795 & 0.817 & \multicolumn{1}{|c}{\textbf{0.912}} & \textbf{0.912} & 0.903 & \multicolumn{1}{|c}{0.830} & \textbf{0.862} \\
Flchain      & 0.933 & \textbf{0.934} & 0.932 & 0.933 & \multicolumn{1}{|c}{0.848} & 0.876 & \textbf{0.880} & \multicolumn{1}{|c}{\textbf{0.873}} & 0.858 \\
Framingham   & \textbf{0.709} & \textbf{0.709} & 0.708 & 0.707 & \multicolumn{1}{|c}{0.666} & 0.658 & \textbf{0.676} & \multicolumn{1}{|c}{0.698} & \textbf{0.705} \\
Gbsg2        & \textbf{0.660} & \textbf{0.660} & 0.644 & 0.650 & \multicolumn{1}{|c}{0.626} & \textbf{0.643} & 0.633 & \multicolumn{1}{|c}{\textbf{0.661}} & 0.655 \\
Maintenance  & 0.942 & 0.940 & 0.870 & \textbf{0.944} & \multicolumn{1}{|c}{\textbf{0.934}} & 0.922 & 0.901 & \multicolumn{1}{|c}{\textbf{0.944}} & \textbf{0.944} \\
Uissurv      & \textbf{0.736} & 0.725 & 0.710 & 0.712 & \multicolumn{1}{|c}{0.717} & \textbf{0.721} & 0.700 & \multicolumn{1}{|c}{\textbf{0.735}} & 0.724 \\
Unempdur     & 0.681 & \textbf{0.684} & 0.681 & 0.682 & \multicolumn{1}{|c}{0.667} & 0.659 & \textbf{0.676} & \multicolumn{1}{|c}{\textbf{0.680}} & \textbf{0.680} \\
Veterans     & 0.593 & 0.600 & \textbf{0.608} & 0.588 & \multicolumn{1}{|c}{0.549} & \textbf{0.575} & \textbf{0.575} & \multicolumn{1}{|c}{\textbf{0.634}} & 0.633 \\
Whas500      & 0.713 & \textbf{0.730} & 0.717 & 0.711 & \multicolumn{1}{|c}{0.674} & \textbf{0.705} & 0.688 & \multicolumn{1}{|c}{0.701} & 0.718 \\\hline
Average      & 0.744 & \textbf{0.754} & 0.731 & 0.740 & 0.696 & 0.710 & \textbf{0.711} & 0.735 & \textbf{0.736} \\\hline
\end{tabular}
}
\caption{Testing accuracy in terms of $C_H$ discriminant measure (the higher the better). The comparison includes SSTs of depth $D=2$ with parametric distributions (Exp, W, Llog) and spline-based semiparametric models (PO and PH), as well as the three benchmarking survival tree methods  (SkSurv, CTree, and RPART) with depth $D=5$. Llog-init refers to Llog with the clustering-based initialization procedure. \textcolor{darkgreen}{Boldface indicates the best model within each group.}}
\label{tab:ch_test}
\end{table}

As to the calibration testing measure $\text{IBS}$ in Table \ref{tab:ibs_test_sec}, for which \textcolor{black}{lower values correspond to better solutions}, the spline-based semiparametric SSTs, PO and PH, outperform the \textcolor{black}{benchmarking methods} SkSurv, CTree, and RPART, \textcolor{black}{as well as} the parametric SSTs. Among the benchmarking methods, CTree achieves the lowest $\text{IBS}$ ($0.11$), comparable to Llog ($0.111$), but it turns out to be less accurate than PO and PH, which yield $\text{IBS}=0.074$ and $\text{IBS}=0.075$, respectively. \textcolor{darkgreen}{Indeed, the Wilcoxon test comparing PO with CTree yields a p-value of $6.22e^{-3}$, which is below the $1\%$ significance level.} Notice that the K-means-based initialization \textcolor{black}{procedure} further enhances the performance of Llog, achieving \textcolor{black}{an average $\text{IBS}$ of $0.107$ compared to $0.111$ with no initialization.}

\begin{table}[ht]
\centering
\resizebox{\textwidth}{!}{
\renewcommand{\arraystretch}{0.85}
\begin{tabular}{lccccccccc}
\cline{2-10}
             & \multicolumn{9}{c}{Testing IBS} \\
\cline{2-10}
             & \multicolumn{4}{c}{D=2} & \multicolumn{3}{c}{D=5} & \multicolumn{2}{c}{D=2} \\\hline
Dataset      & \textbf{Llog} & \textbf{Llog-init} & \textbf{Exp} & \textbf{W} & \multicolumn{1}{|c}{\textbf{SkSurv}} & \textbf{RPART} & \textbf{CTree} & \multicolumn{1}{|c}{\textbf{PO}} & \textbf{PH} \\\hline
Aids         & \textbf{0.057} & \textbf{0.057} & \textbf{0.057} & \textbf{0.057} & \multicolumn{1}{|c}{0.069} & 0.068 & \textbf{0.059} & \multicolumn{1}{|c}{0.033} & \textbf{0.030} \\
Aids\_death  & \textbf{0.016} & \textbf{0.016} & \textbf{0.016} & \textbf{0.016} & \multicolumn{1}{|c}{0.020} & 0.020 & \textbf{0.016} & \multicolumn{1}{|c}{0.010} & \textbf{0.008} \\
Aids2        & 0.157 & 0.158 & \textbf{0.141} & 0.144 & \multicolumn{1}{|c}{0.141} & 0.141 & \textbf{0.140} & \multicolumn{1}{|c}{\textbf{0.033}} & 0.034 \\
Churn        & 0.071 & \textbf{0.058} & 0.092 & 0.082 & \multicolumn{1}{|c}{0.103} & \textbf{0.101} & \textbf{0.101} & \multicolumn{1}{|c}{\textbf{0.008}} & \textbf{0.008} \\
Credit\_risk & \textbf{0.110} & 0.111 & 0.130 & 0.109 & \multicolumn{1}{|c}{0.122} & \textbf{0.113} & 0.114 & \multicolumn{1}{|c}{\textbf{0.109}} & 0.110 \\
Dialysis     & 0.145 & 0.146 & 0.145 & \textbf{0.144} & \multicolumn{1}{|c}{\textbf{0.175}} & 0.183 & 0.176 & \multicolumn{1}{|c}{0.112} & \textbf{0.110} \\
Employee     & 0.154 & \textbf{0.104} & 0.218 & 0.154 & \multicolumn{1}{|c}{\textbf{0.055}} & 0.057 & 0.065 & \multicolumn{1}{|c}{0.135} & \textbf{0.094} \\
Flchain      & 0.044 & 0.044 & 0.046 & \textbf{0.043} & \multicolumn{1}{|c}{0.062} & 0.064 & \textbf{0.060} & \multicolumn{1}{|c}{\textbf{0.025}} & 0.144 \\
Framingham   & \textbf{0.111} & \textbf{0.111} & 0.114 & \textbf{0.111} & \multicolumn{1}{|c}{0.119} & 0.118 & \textbf{0.117} & \multicolumn{1}{|c}{\textbf{0.046}} & \textbf{0.046} \\
Gbsg2        & \textbf{0.172} & \textbf{0.172} & 0.176 & 0.175 & \multicolumn{1}{|c}{0.189} & 0.180 & \textbf{0.173} & \multicolumn{1}{|c}{\textbf{0.087}} & \textbf{0.087} \\
Maintenance  & \textbf{0.019} & 0.021 & 0.098 & 0.020 & \multicolumn{1}{|c}{\textbf{0.001}} & 0.003 & 0.007 & \multicolumn{1}{|c}{\textbf{0.001}} & \textbf{0.001} \\
Uissurv      & \textbf{0.143} & 0.147 & 0.148 & 0.147 & \multicolumn{1}{|c}{0.146} & \textbf{0.144} & 0.146 & \multicolumn{1}{|c}{\textbf{0.095}} & 0.102 \\
Unempdur     & \textbf{0.163} & \textbf{0.163} & 0.165 & 0.164 & \multicolumn{1}{|c}{0.170} & 0.164 & \textbf{0.161} & \multicolumn{1}{|c}{\textbf{0.168}} & 0.169 \\
Veterans     & \textbf{0.124} & 0.126 & 0.126 & 0.131 & \multicolumn{1}{|c}{0.160} & \textbf{0.134} & \textbf{0.134} & \multicolumn{1}{|c}{\textbf{0.133}} & 0.134 \\
Whas500      & 0.178 & \textbf{0.170} & 0.180 & 0.181 & \multicolumn{1}{|c}{0.206} & 0.190 & \textbf{0.177} & \multicolumn{1}{|c}{0.118} & \textbf{0.112} \\\hline
Average      & 0.111 & \textbf{0.107} & 0.123 & 0.112 & 0.116 & 0.112 & \textbf{0.110} & \textbf{0.074} & 0.079 \\\hline
\end{tabular}
}
\caption{\textcolor{black}{Testing accuracy in terms of} calibration measure IBS (the lower the better). The comparison includes \textcolor{black}{SSTs} %models 
of depth $D=2$ with parametric distributions (Exp, W, Llog) and spline-based semiparametric \textcolor{black}{models} % approaches 
(PO and PH), as well as the three \textcolor{black}{benchmarking survival tree methods}   %models 
(SkSurv, CTree, and RPART) with depth $D=5$. Llog-init refers to Llog with the clustering-based initialization procedure. \textcolor{darkgreen}{Boldface indicates the best model within each group.}}
\label{tab:ibs_test_sec}
\end{table}

\textcolor{black}{To illustrate the variability in \textcolor{black}{testing} accuracy measures due to the changes in folds and initial solutions, Figures \ref{fig:box_plt_C} and \ref{fig:box_plt_IBS} present the boxplots of the testing $C_H$ and IBS obtained  \textcolor{black}{for} the 15 datasets \textcolor{black}{and} \textcolor{black}{the three different approaches:} LLog for SST with the parametric \textcolor{black}{model}, CTree as the benchmarking tree \textcolor{black}{method}, and PO  for SST with the semiparametric \textcolor{black}{model}. The boxplots for the remaining performance measures and models are reported in \ref{sec:tables_sd}.} 

%\textcolor{black}{To illustrate the variability in \textcolor{black}{testing} accuracy measures due to the changes in folds and initial solutions, Figures \ref{fig:box_plt_C} and \ref{fig:box_plt_IBS} present the boxplots of the testing $C_H$ and IBS obtained  
%\textcolor{black}{for} the 15 datasets \textcolor{black}{and} the three best-performing \textcolor{black}{ approaches} LLog for SST with the parametric \textcolor{black}{model}, CTree as the benchmarking tree \textcolor{black}{method}, and PO \textcolor{black}{PH???} for SST with the semiparametric \textcolor{black}{model}. The boxplots for the remaining performance measures and models are reported in \ref{sec:tables_sd}.}

\begin{figure}[h]
    \centering
    \includegraphics[width=1.\linewidth]{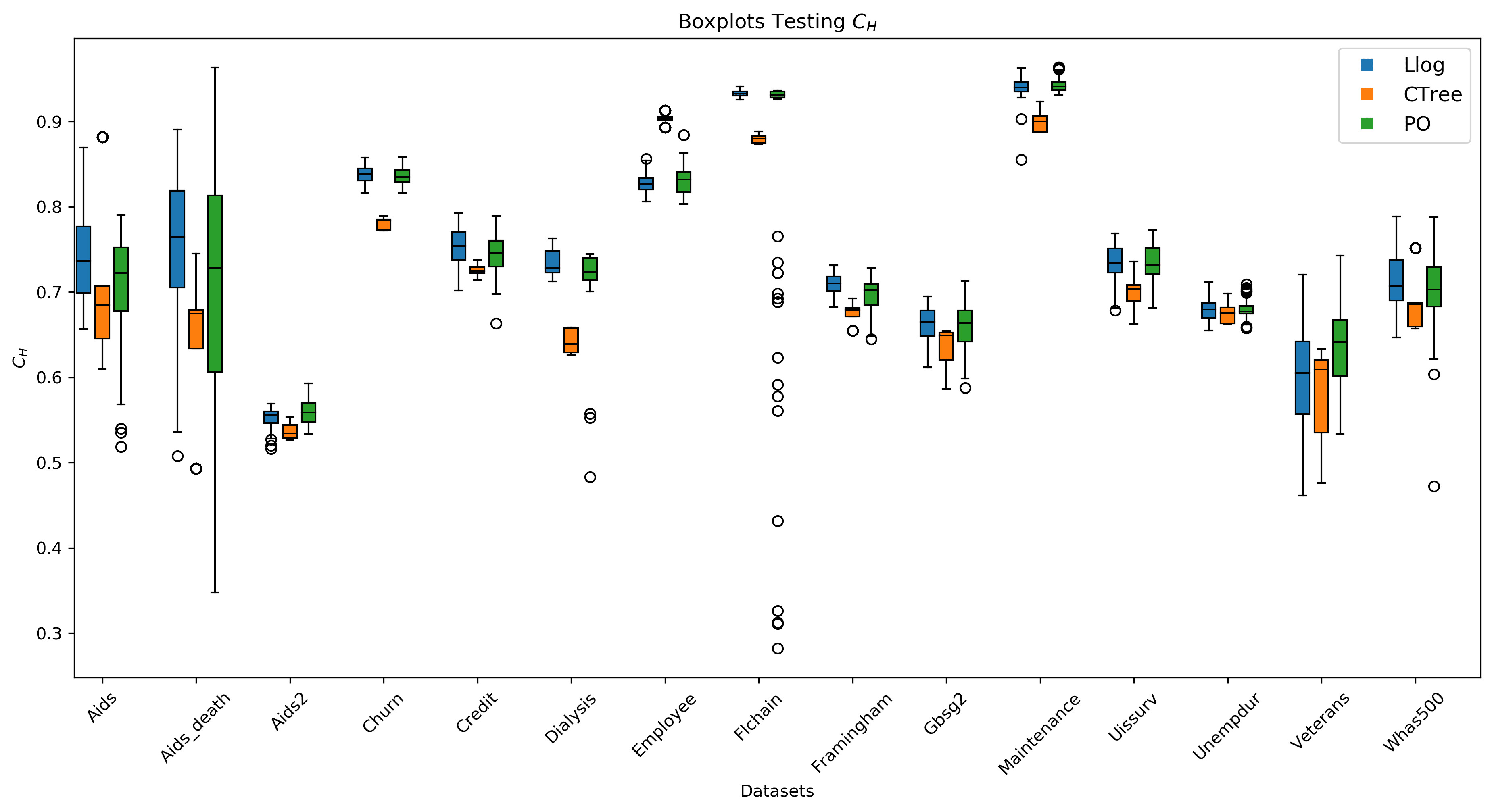}
    \caption{Boxplots of the testing $C_U$ \textcolor{black}{measure (the higher the better)} %results
    for the 15 datasets %showing the 
    \textcolor{black}{and for the}
    three best-performing \textcolor{black}{approaches, namely,} %models 
    LLog in blue, CTree in orange, and PO in green.}
    \label{fig:box_plt_C}
\end{figure}

\begin{figure}[H]
    \centering
    \includegraphics[width=1.\linewidth]{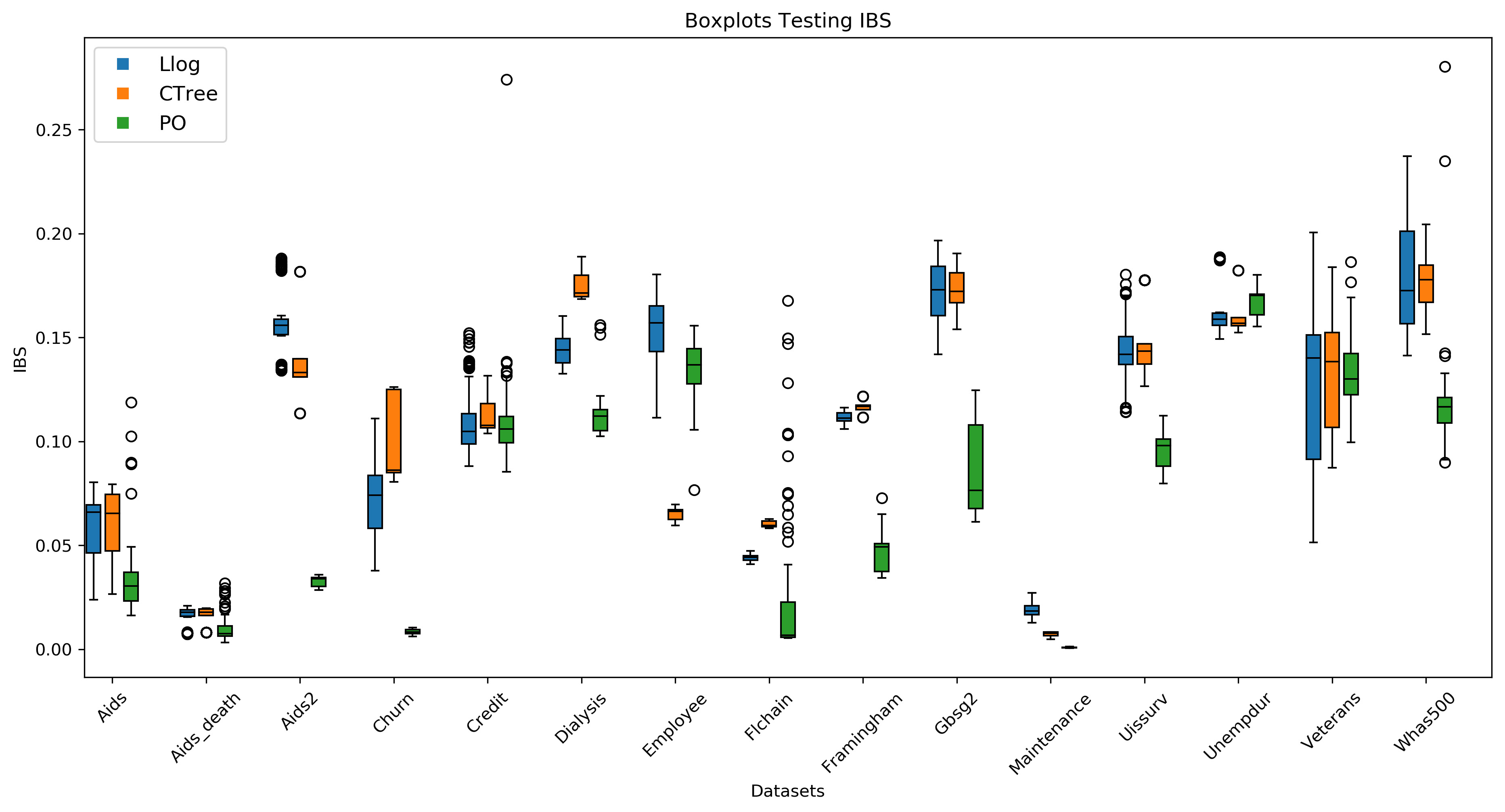}
    \caption{Boxplots of the testing IBS \textcolor{black}{measure} (the lower the better) %results 
    for the 15 datasets %showing 
    \textcolor{black}{and for the}
    the three best-performing 
    \textcolor{black}{approaches, namely,} %models   
    LLog in blue, CTree in orange, and PO in green.}
    \label{fig:box_plt_IBS}
\end{figure}

The sensitivity of SST to the initial solution due to the challenging nonconvex error function in \eqref{eq:formulation}, leads sometimes to longer whiskers and the presence of outliers for both LLog and PO \textcolor{black}{in terms of} both measures. \textcolor{black}{Nevertheless, it is worth noting} that LLog and PO often achieve significantly better results than CTree in specific runs. For the \textcolor{black}{testing} $C_U$, the box and top whisker associated with LLog are consistently higher than those of CTree for most datasets (12 out of 15). Similarly, for the IBS calibration measure %test, 
where lower values indicate better performance, the box-plot for PO is almost entirely below the box-plot for CTree in most datasets (11 out of 15). 

%\textcolor{black}{In general, the sensitivity of SST to the initial solution due to the challenging nonconvex error function in \eqref{eq:formulation}, leads sometimes to longer whiskers and the presence of outliers for both LLog and PO across both measures. Despite this, it is notable that LLog and PO often achieve significantly better results than CTree in specific runs. For the \textcolor{black}{testing} $C_U$ , the box and top whisker associated with LLog are consistently higher than those of CTree for most datasets (12 out of 15). Similarly, for the IBS test, where lower values indicate better performance, the box-plot for PO is almost entirely below the box-plot for CTree in most datasets (11 out of 15). }

To summarize, these numerical results show that SST, %whether employing 
\textcolor{black}{with} parametric or semiparametric survival functions at the leaf nodes, consistently outperforms the standard survival tree \textcolor{black}{benchmarking methods} SkSurv, CTree, and RPART in terms of calibration and discrimination testing accuracy.

\textcolor{darkgreen}{Finally, to assess the computational efficiency of the NODEC-DR-SST algorithm, we evaluate the CPU times of \textcolor{darkgreen}{SSTs with both parametric and non-parametric models for 15 datasets.} 
%both parametric and non-parametric SSTs across the 15 datasets. 
Table \ref{tab:tempi} reports the mean computational time in seconds for each dataset over 100 runs of every SST model, with the corresponding median shown in brackets. The last row %summarises
\textcolor{darkgreen}{indicates the averages} of the mean and \textcolor{darkgreen}{of} the median %across 
\textcolor{darkgreen}{over} all datasets. As expected,
\textcolor{darkgreen}{SSTs with} parametric models require less computational time than \textcolor{darkgreen}{SSTs with spline-based semiparametric ones.} %semiparametric SSTs. 
For all parametric models, computational times %remain 
\textcolor{darkgreen}{are} below 88 seconds, except for the Dialysis and Employee datasets, %where the
\textcolor{darkgreen}{whose} mean and median reach up to 150 and 140 seconds, respectively. Specifically, %in 
\textcolor{darkgreen}{for} 11 datasets, both the mean and median computational times are consistently below 42 seconds. Overall, when averaged across all datasets, \textcolor{darkgreen}{SSTs with} parametric models do not \textcolor{darkgreen}{require more than} %exceed 
35 seconds. For \textcolor{darkgreen}{SSTs with} spline-based models, 
%the need to handle a 
\textcolor{darkgreen}{the larger} number of variables results in longer computational times. With the exception of the Aids2, Dialysis, Flchain, and Framingham datasets, %where computational times exceed 200 seconds, 
\textcolor{darkgreen}{which require computational times above 200 seconds}, 
all other datasets \textcolor{darkgreen}{need} less than 57 seconds. 
For the Aids2 and Flchain datasets, %in both PO and PH models
\textcolor{darkgreen}{SSTs with} PO and PH models \textcolor{darkgreen}{yield a median which} is substantially lower than the mean. For instance, in the Aids2 dataset, PO achieves a median of $25$ seconds compared to an average of $234$ seconds, suggesting that only a small number of starting \textcolor{darkgreen}{solutions} %points 
result in longer computational times. The dataset with the highest computational burden is Flchain, %where the two models 
\textcolor{darkgreen}{for which PO and PH} require on average $4844.13$ and $9148.55$ seconds, respectively. However, the median values are lower than the mean \textcolor{darkgreen}{ones}, and in the case of PO the median is more than an order of magnitude smaller. This indicates that the choice of the starting %point 
\textcolor{darkgreen}{solution}
has a strong impact on the computation time. On average, 
%the PO model is faster than the PH model
\textcolor{darkgreen}{PO requires smaller computational times than PH}, with \textcolor{darkgreen}{average means} %mean times 
of $392.54$ and $711.01$ seconds, respectively, across all datasets. %However, it should be emphasized that these
\textcolor{darkgreen}{Notice that the} averages are substantially affected by the \textcolor{darkgreen}{computational} times %observed in
\textcolor{darkgreen}{for} Flchain. When Flchain is excluded, the average %times 
\textcolor{darkgreen}{means}
for PO and PH are $74.57$ and $108.33$ seconds, respectively.}

\begin{table}[htb!]
\centering
\textcolor{darkgreen}{
\resizebox{\textwidth}{!}{
\begin{tabular}{lcccccc}
\hline
Dataset     & \textbf{Llog} & \textbf{Llog-init} & \textbf{Exp} & \textbf{W} & \textbf{PO} & \textbf{PH} \\ \hline
Aids & 4.86 (4.13) & 14.78 (14.09) & 4.20 (4.12) & 6.29 (5.57) & 23.92 (24.30) & 34.82 (31.47) \\
Aids\_death & 5.25 (4.02) & 12.46 (11.51) & 4.72 (4.32) & 5.43 (4.77) & 28.82 (23.38) & 29.69 (20.42) \\
Aids2 & 10.56 (8.52) & 15.93 (16.11) & 12.83 (11.53) & 12.41 (10.97) & 234.03 (24.97) & 223.30 (18.90) \\
Churn & 17.08 (18.42) & 27.12 (26.94) & 22.57 (24.31) & 24.39 (24.39) & 22.32 (23.06) & 56.72 (52.33) \\
Credit & 20.42 (21.22) & 26.54 (26.99) & 36.31 (36.39) & 24.20 (24.65) & 25.95 (25.66) & 34.52 (33.03) \\
Dialysis & 84.21 (49.43) & 149.29 (166.67) & 138.93 (126.24) & 83.91 (52.65) & 402.18 (359.56) & 725.97 (699.34) \\
Employee & 46.19 (43.56) & 108.16 (117.98) & 49.12 (45.85) & 53.29 (49.32) & 74.78 (70.86) & 144.92 (139.37) \\
Flchain & 68.87 (71.38) & 87.80 (86.05) & 87.84 (85.29) & 76.21 (80.27) & 4844.13 (209.06) & 9148.55 (8688.07) \\
Framingham & 18.80 (14.71) & 32.78 (29.56) & 42.64 (42.81) & 24.01 (20.85) & 192.26 (190.47) & 212.38 (222.06) \\
Gbsg2 & 3.22 (3.14) & 5.36 (5.64) & 3.28 (3.41) & 3.68 (3.56) & 6.38 (6.04) & 9.31 (8.68) \\
Maintenance & 6.70 (6.97) & 7.46 (7.57) & 2.80 (2.66) & 6.56 (6.55) & 6.60 (6.58) & 9.51 (9.48) \\
Uissurv & 5.23 (5.47) & 4.69 (4.28) & 4.88 (4.89) & 5.06 (5.07) & 4.94 (4.77) & 7.74 (7.74) \\
Unempdur & 9.32 (8.80) & 9.47 (8.99) & 9.53 (8.57) & 9.60 (8.92) & 15.15 (15.53) & 18.42 (19.12) \\
Veterans & 1.01 (1.00) & 2.61 (2.65) & 0.98 (0.99) & 1.04 (1.02) & 1.29 (1.17) & 2.28 (1.34) \\
Whas500 & 3.27 (3.26) & 5.85 (5.95) & 3.91 (3.91) & 3.76 (3.78) & 5.32 (4.87) & 6.99 (6.81) \\ \hline
\textcolor{darkgreen}{Average} & 20.33 (17.60) & 34.02 (35.40) & 28.30 (27.02) & 22.66 (20.16) & 392.54 (66.02) & 711.01 (663.88) \\ \hline
\end{tabular}
}
}
\caption{For each dataset, the mean computational time in seconds obtained by the SSTs \textcolor{darkgreen}{with the specified survival functions model} over 100 runs is reported, while the median is shown in brackets. The last row reports the average of the mean and \textcolor{darkgreen}{of} the median across the 15 datasets.}\label{tab:tempi}
\end{table}

\subsection{Enhancing interpretability}
\label{sec:interpretability}

The aim of this section is to demonstrate how the SST model enhances interpretability by analyzing the clusters of survival functions corresponding to each leaf node. Unlike other single-tree methods, such as greedy algorithms \citep{hothorn2015ctree} or more recent approaches employing advanced optimization techniques \citep{bertsimas2022optimal,huisman2024optimal,zhang2024optimal}, the SST is sufficiently flexible to define distinct survival functions for each data point within a leaf node. This approach enables the exploration of differences both across and within leaf node clusters. For the sake of conciseness, the analysis focuses on the $9$ dataset, with results for other $6$ datasets presented in the \ref{sec:additional_explain}.

Figure \ref{fig:explain_maintenance} \textcolor{black}{depicts the survival functions obtained with a single run of NODEC-DR-SST} (a single fold and \textcolor{black}{one} initial solution) on the Maintenance\textcolor{black}{, Veterans and Churn} datasets for SSTs of depth $D=1$ and $D=2$. \textcolor{black}{Each color corresponds to a specific leaf node, with varying shades of each color to distinguish survival functions of the data points falling into the same leaf node.} \textcolor{black}{For all the datasets and for SSTs of both depths}, survival functions are easily distinguishable not only across leaf nodes but also within each leaf node, where distinct patterns are observable. \textcolor{black}{Results for SSTs of both depths highlight how the leaf nodes induce a partition of the survival functions accounting for distinct shapes and scales.} 

\textcolor{black}{For the Maintenance and Veterans datasets and for SSTs of depth $D=2$, pronounced differences in terms of local curvature and of global trend are evident between survival functions associated to leaf nodes 1 (blue) and 2 (red) compared with those associated to leaf nodes 3 (green) and 4 (purple). For the Maintenance dataset and SSTs of depth $D=2$, it is interesting to observe the presence of distinct bundles of survival functions associated to data points falling to the same leaf nodes (clusters). This additional information allows the domain expert to investigate not only the differences between survival functions of data points falling into distinct clusters defined by the SST splits, but also the finer-grained differences within the same clusters due to possibly more subtle feature characteristics. For the Veterans dataset with SSTs of depth $D=2$, it is worth pointing out that the survival functions associated with leaf nodes 3 (green) and 4 (purple) decrease at a slower rate than those associated with leaf nodes 1 (blue) and 2 (red). Note also that the purple survival functions are markedly concave for small values of $t$, while this is not the case for the red and blue survival functions associated to leaf nodes 1 and 2, which are very steep for small values of $t$.}

\begin{figure}[h]
    \centering
    \begin{minipage}[b]{0.48\textwidth}
        \centering
        \includegraphics[width=\textwidth]{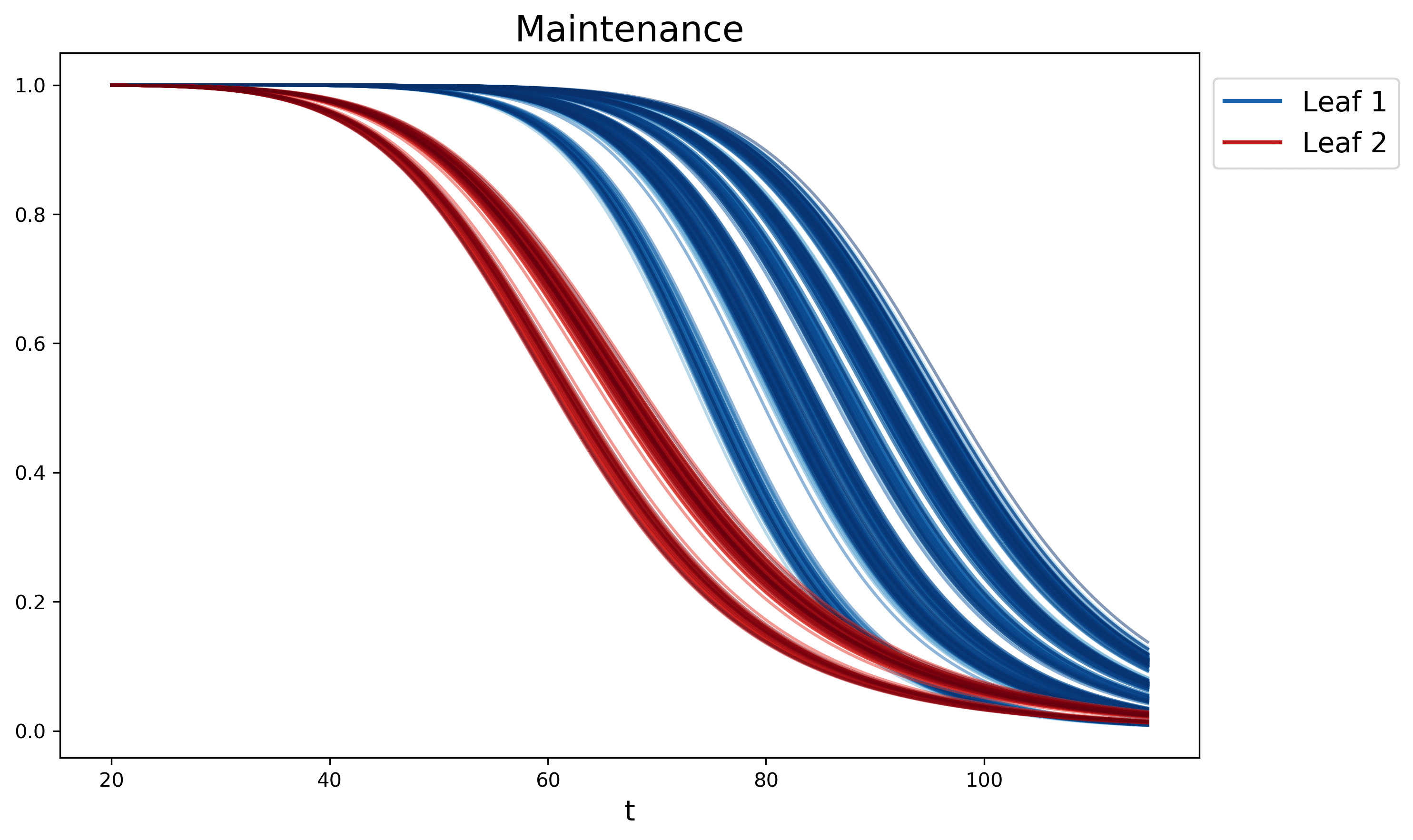}
    \end{minipage}
    \hfill
    \begin{minipage}[b]{0.48\textwidth}
        \centering
        \includegraphics[width=\textwidth]{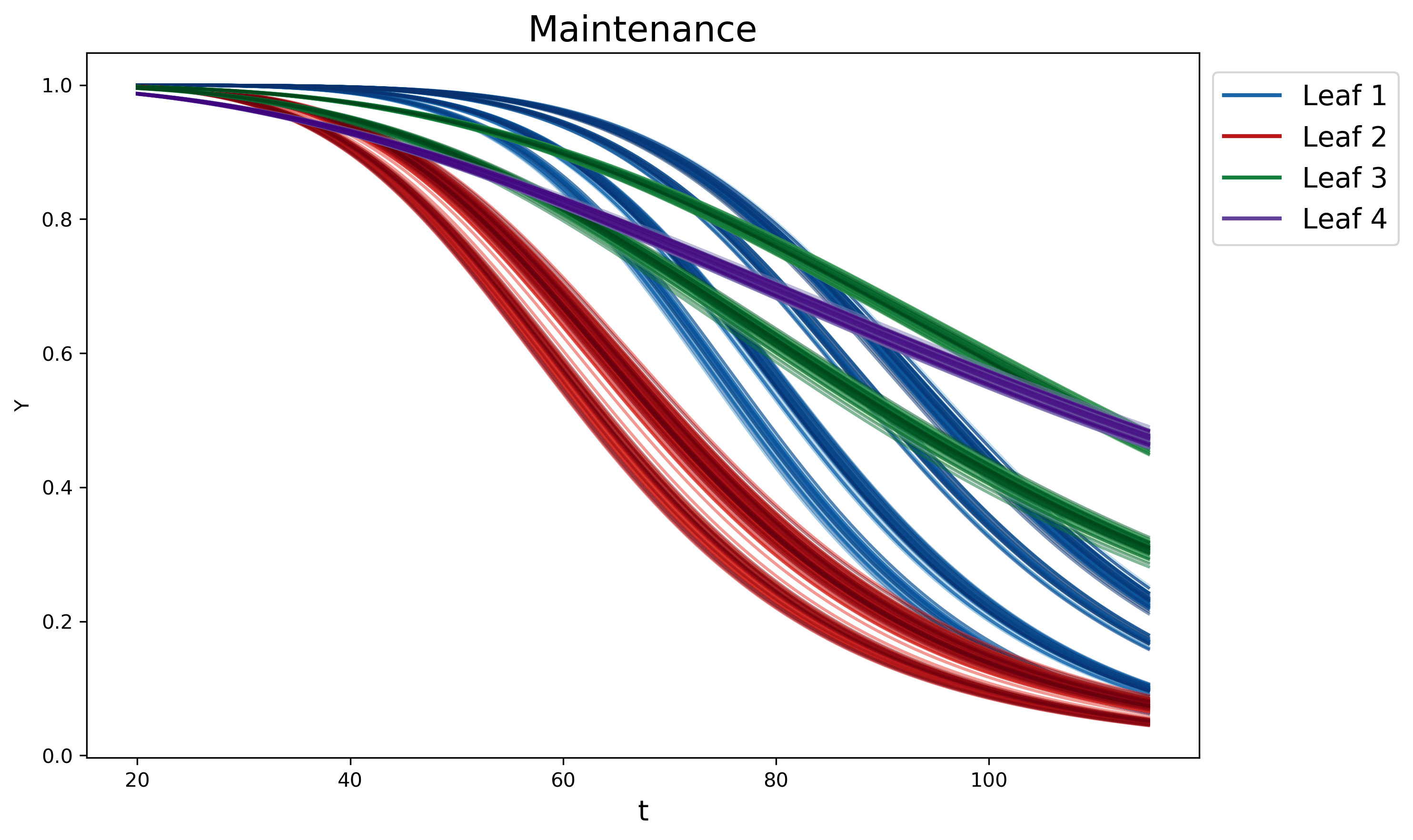}
    \end{minipage}
    \vspace{1em}
    \begin{minipage}[b]{0.48\textwidth}
        \centering
        \includegraphics[width=\textwidth]{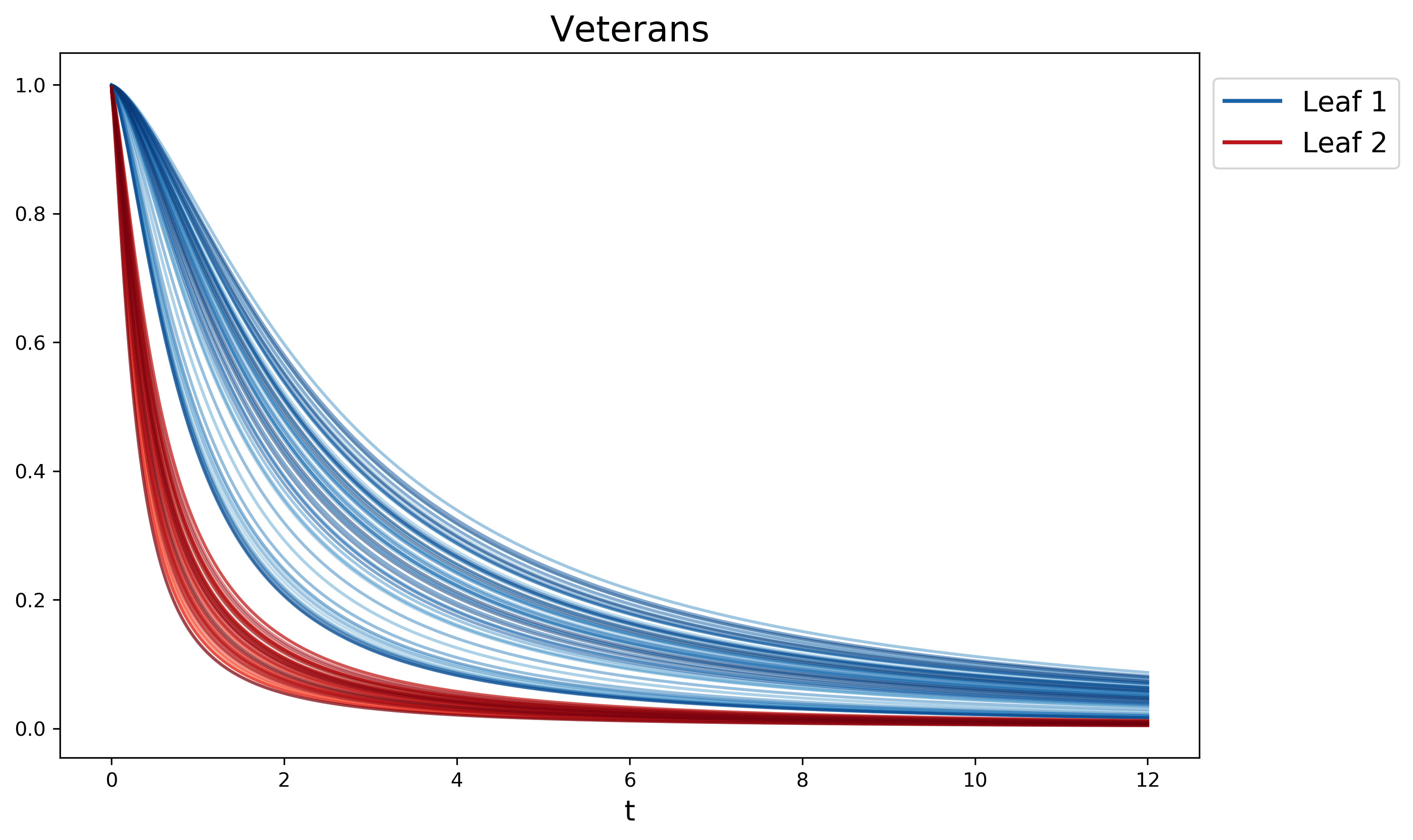}
    \end{minipage}
    \hfill
    \begin{minipage}[b]{0.48\textwidth}
        \centering
        \includegraphics[width=\textwidth]{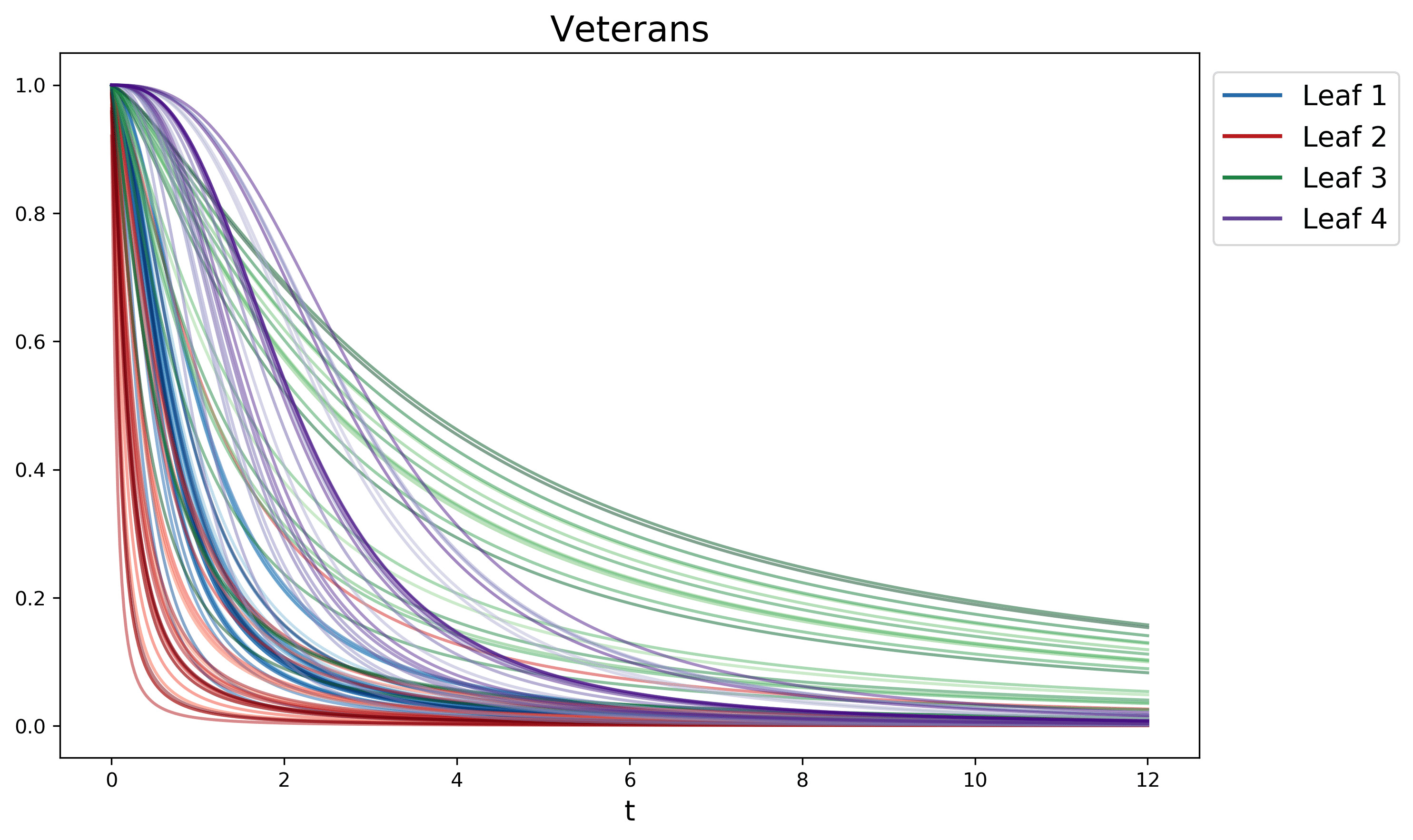}
    \end{minipage}
    \vspace{1em}
    \begin{minipage}[b]{0.48\textwidth}
        \centering
        \includegraphics[width=\textwidth]{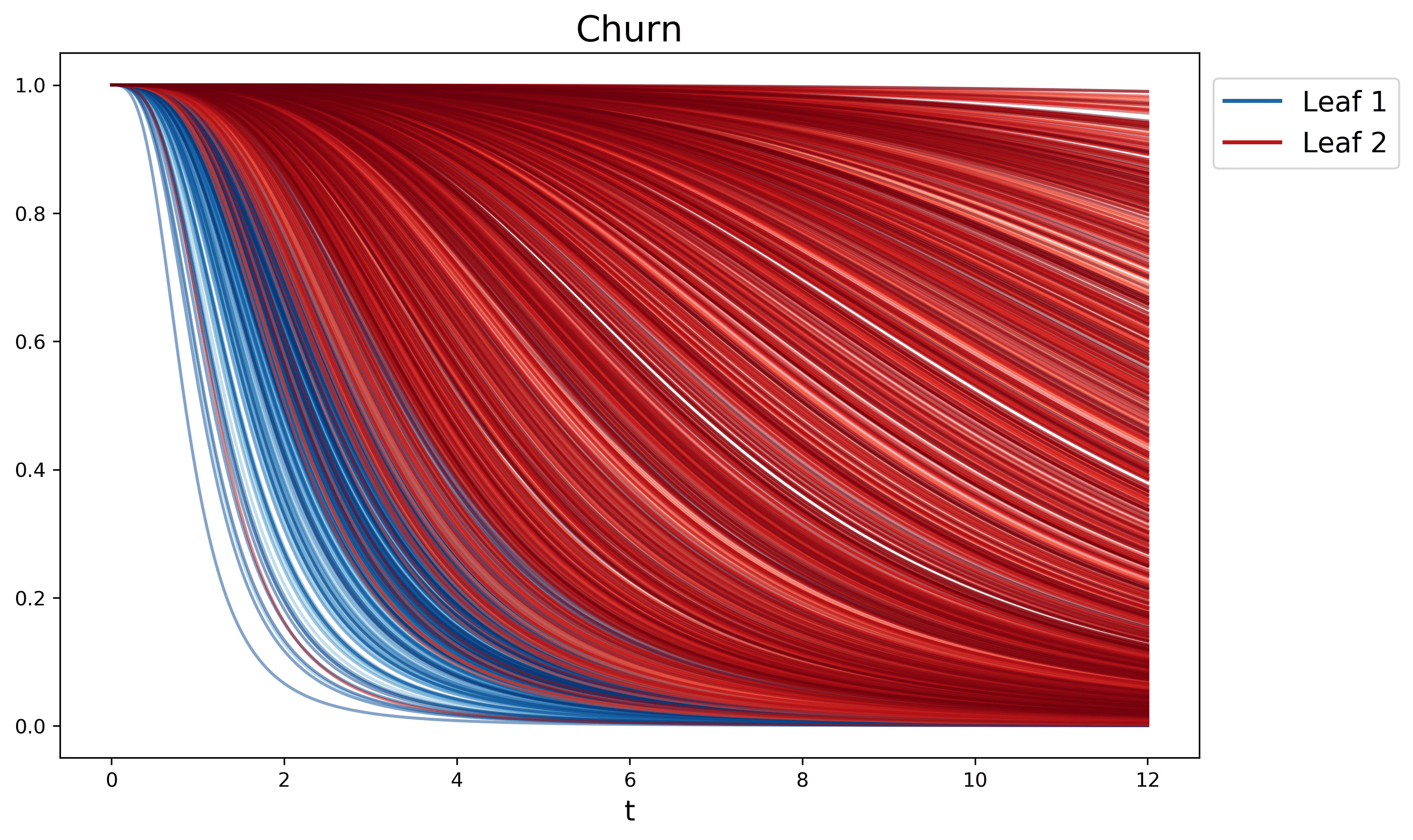}
    \end{minipage}
    \hfill
    \begin{minipage}[b]{0.48\textwidth}
        \centering
        \includegraphics[width=\textwidth]{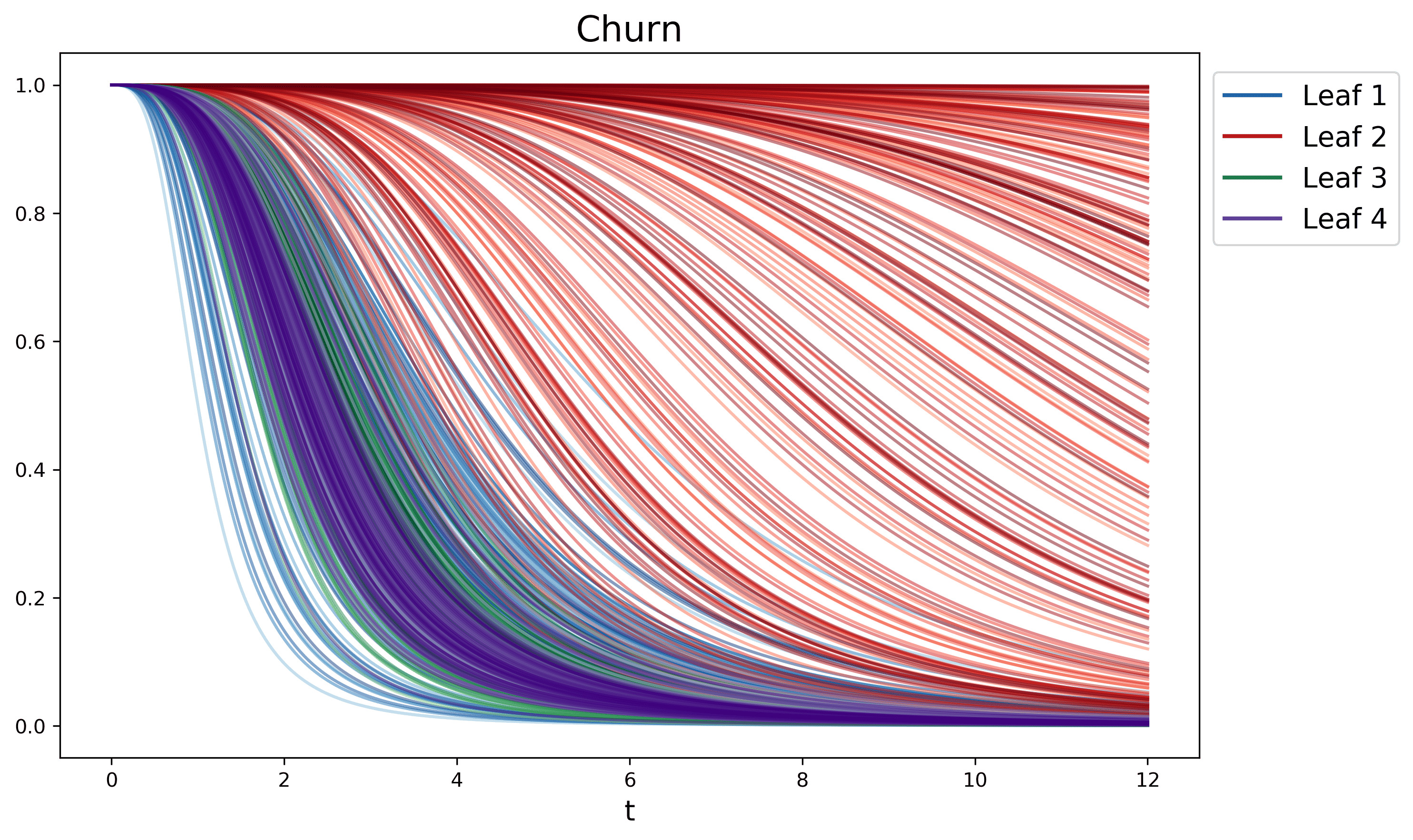}
    \end{minipage}
    
    \caption{Survival function distribution from a single run of the Maintenance, Veterans and Churn datasets, shown for depths $D=1$ (left) and $D=2$ (right). Each color represents a specific leaf node, with varying shades within each node highlighting differences among the survival functions of individual data points.} \label{fig:explain_maintenance}
\end{figure}

\textcolor{black}{For the Churn dataset, we observe a wide spectrum of survival functions which, for SSTs of depth $D=1$, is clearly split into two subsets. For SSTs of depth $D=2$, the blue and red survival functions associated with leaf nodes 1 and 2 (whose data points are identically routed at the root branch node) are clearly separated, while the bundle of purple survival functions (leaf node 4) is surrounded on both sides by two groups of green ones (leaf node 3).} 

In some cases shown in \ref{sec:additional_explain}, the separation of survival functions between and within leaf nodes is less pronounced. Nevertheless, clear differences between clusters are consistently evident, providing meaningful insights into the specific dataset.

\subsection{Fairness}
\label{sec:fairness}

In this work, we propose an \textcolor{black}{extension of SST training formulation} that takes into account group fairness into survival tasks by ensuring parity between sensitive groups, such as those based on ethnicity or gender, through controlling the distances between their survival functions.

\textcolor{black}{Formulation \eqref{eq:formulation_fair} is optimized using the node-based decomposition training algorithm on a socio-economic dataset where a sensitive group is identifiable.} Since the datasets discussed in Section \ref{sec:dataset_exp} involve applications where it is not %possible 
\textcolor{black}{natural} to identify a sensitive group, this section focuses on the Unemployment dataset from \cite{romeo1999conducting}. The Unemployment dataset consists of a four-month panel of revised Current Population Survey data from September 1993, where each observation corresponds to an unemployed individual searching for a job. Each row includes binary variables indicating the reason for unemployment (e.g., whether the individual is a job loser, job leaver, or labor force reentrant), the search method employed, and their demographic status. The dataset consists of $452$ data points, $7$ features, with a censoring level of $56.6\%$. Gender is selected as the sensitive feature for group $S$\footnote{This feature was not considered for SST training \textcolor{black}{and hence, it can be used to predict survival functions of forthcoming individuals without even knowing this feature}}.

\textcolor{black}{We aim to examine how the} fairness penalty term affects the distribution of individuals from the sensitive and non-sensitive groups across the leaf node clusters. For this purpose, we analyzed the effect on the Gini impurity measure, which reflects the distribution of males and females within the leaf nodes. The Gini impurity measure is a metric that quantifies inequality by evaluating the distribution of a variable, often to assess the balance between two groups. In this context, for the two classes of males and females, it measures the uniformity of gender distribution across the leaf node clusters, with 1 representing perfectly equal distribution and 0 indicating maximum imbalance.

For the sake of illustration, we consider SST with the Llog parametric distribution and a single fold of the 5-fold cross-validation. We also perform a single run without applying the clustering-based initialization procedure. It is important to highlight that any alternative parametric or semiparametric choice of survival function for the leaf nodes is feasible. The decision to present a single run is motivated solely by the aim of clearly illustrating the distribution of clusters across the 4 leaf nodes and its trend as the hyperparameter $\rho$, associated with the fairness penalty, increases. The $\rho$ hyperparameter starts from $0$ and reaches $\frac{20}{N_M N_F}$ in five steps, defined by $\frac{1}{N_M N_F} [0,1,5,10,15,20]$, where $N_M$ and $N_F$ correspond to the number of males and females in the training dataset.

Figure \ref{fig:gini_fair} displays the Gini index trends for the training and testing sets and the corresponding fairness penalty term values as $\rho$ grows. In particular, the figure presents two different Gini index measures: the blue line represents the simple average across the leaf nodes, while the green line corresponds to a weighted average, where each leaf node's Gini index is weighted by the number of data points it contains. The trend of both Gini index curves clearly shows that as the penalty parameter $\rho$ increases, the Gini index with respect to gender rises within the leaf nodes, indicating that the distribution of males and females becomes progressively more even.

\begin{figure}[htbp]
    \centering
    \begin{minipage}[b]{0.48\textwidth}
        \centering
        \includegraphics[width=\textwidth]{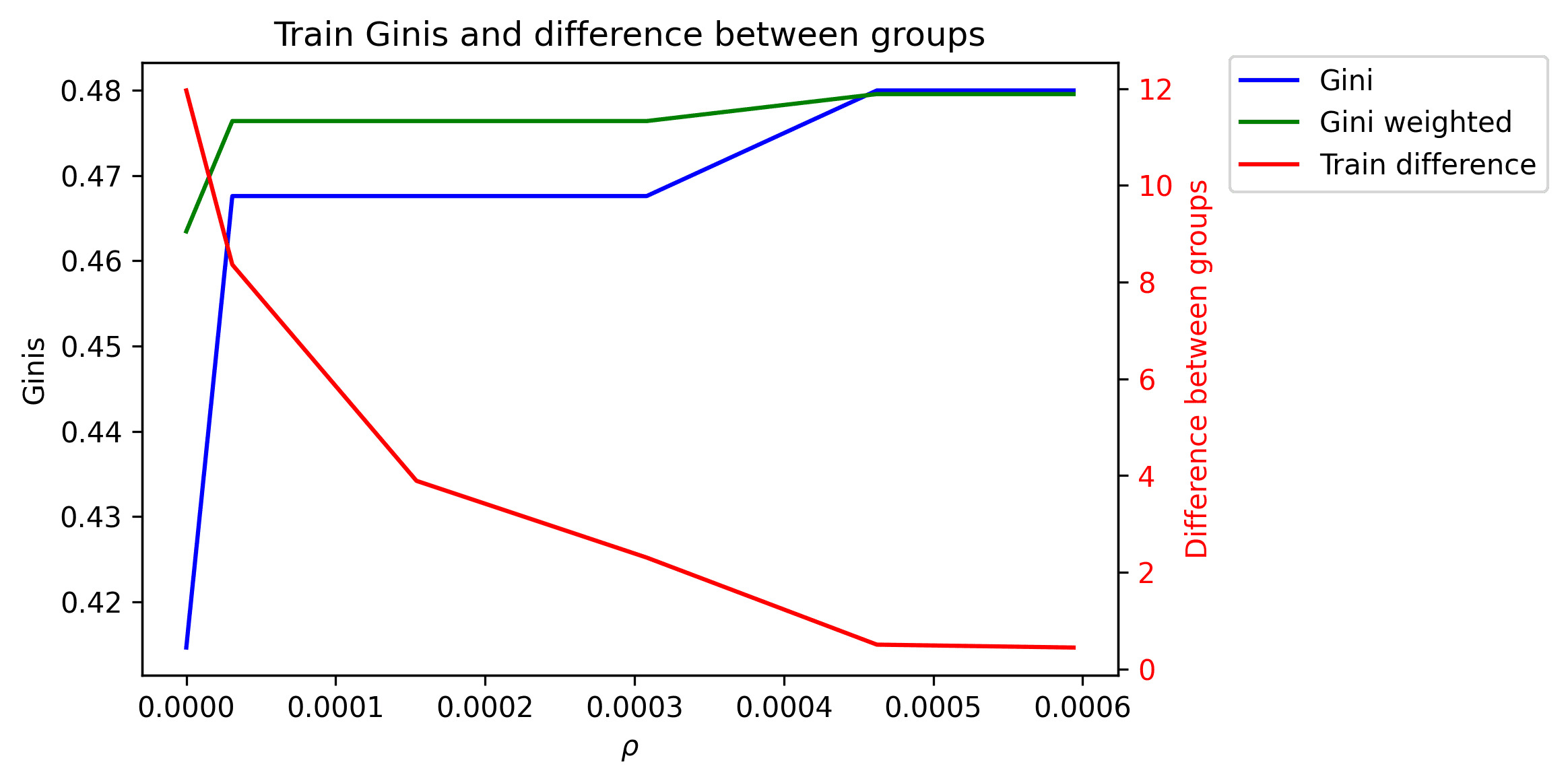}
    \end{minipage}
    \hfill
    \begin{minipage}[b]{0.48\textwidth}
        \centering
        \includegraphics[width=\textwidth]{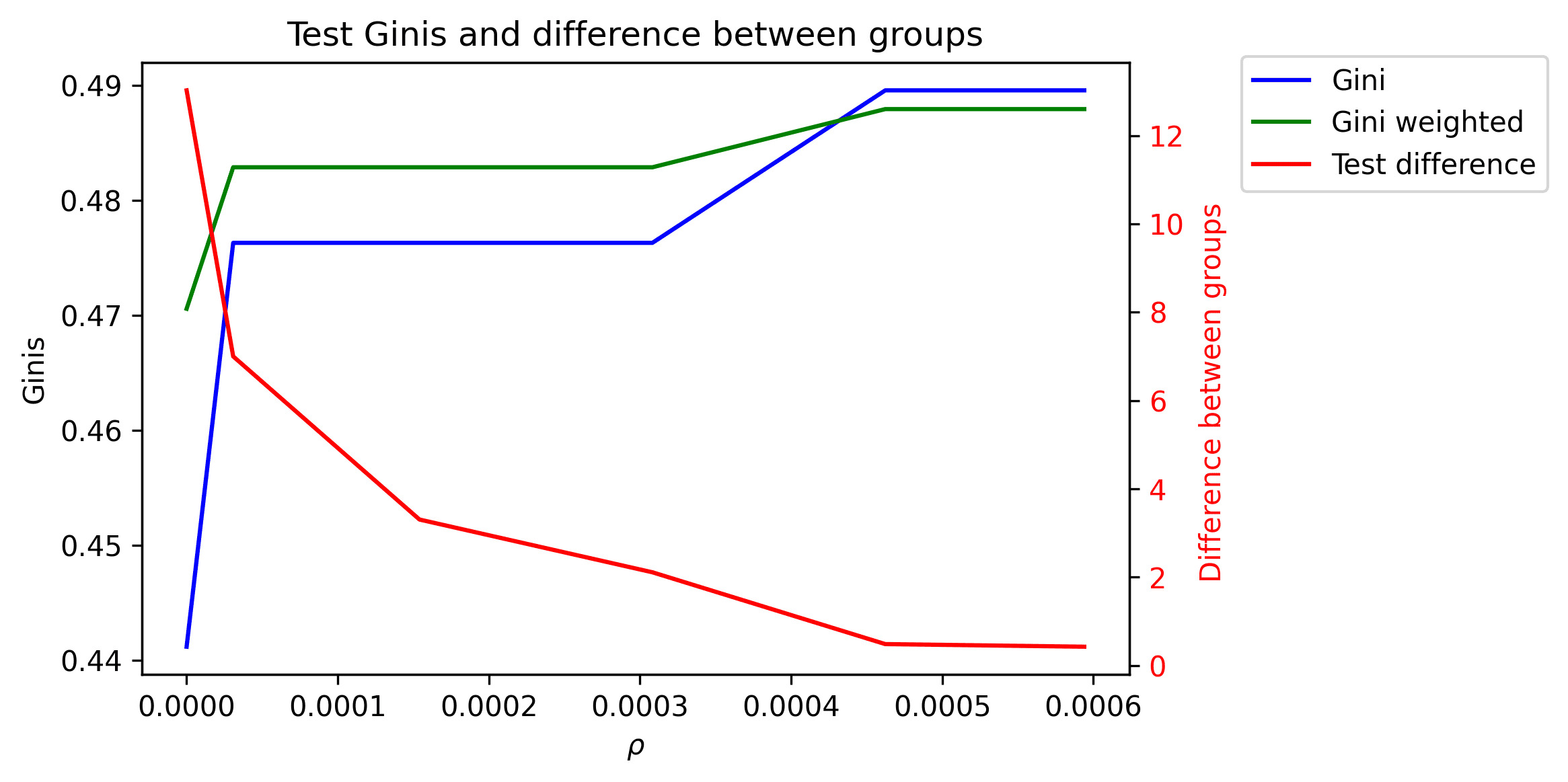}
    \end{minipage}
    \caption{Trends on the training set (left) and on the testing set (right) for the Gini index simple average (blue) and weighted average (green) over the leaf nodes of the distribution of males and females as the fairness penalty $\rho$ increases. The red curve represents the difference between the survival functions for the two groups \eqref{eq:fair_term}.} \label{fig:gini_fair}
\end{figure}

To explore the impact of the penalty term \eqref{eq:fair_term}, we \textcolor{black}{also investigate how the survival functions} change on the training and testing sets as $\rho$ grows. Figure \ref{fig:fair_rho} illustrates the survival functions for individual data points, with different colors representing distinct leaf nodes. The left column corresponds to the training set, and the right column to the testing set. The plots, arranged from top to bottom, show the evolution of survival functions for increasing values of $\rho$. For simplicity, the figure includes only three values of $\rho$, while \ref{sec:additional_fair} provides a comprehensive view for all $\rho$ values. In the first row of Figure \ref{fig:fair_rho}, which corresponds to the survival functions for $\rho=0$, it is clear that the functions for the second leaf node (in red) show a much lower probability of finding a job within the given time (indicating a higher survival probability). Further investigation of the composition of the cluster for leaf node 2 shows that it is largely composed of women ($2/13$ in the training set and $1/4$ in the testing set \textcolor{black}{are men}). As $\rho$ increases, the functions and their distribution across the nodes become more similar, demonstrating how the proposed approach reduces the effect of belonging to a sensitive group.

\begin{figure}[!ht]
    \centering
    % Prima riga
    \begin{subfigure}{0.45\textwidth}
        \centering
        \includegraphics[width=\linewidth]{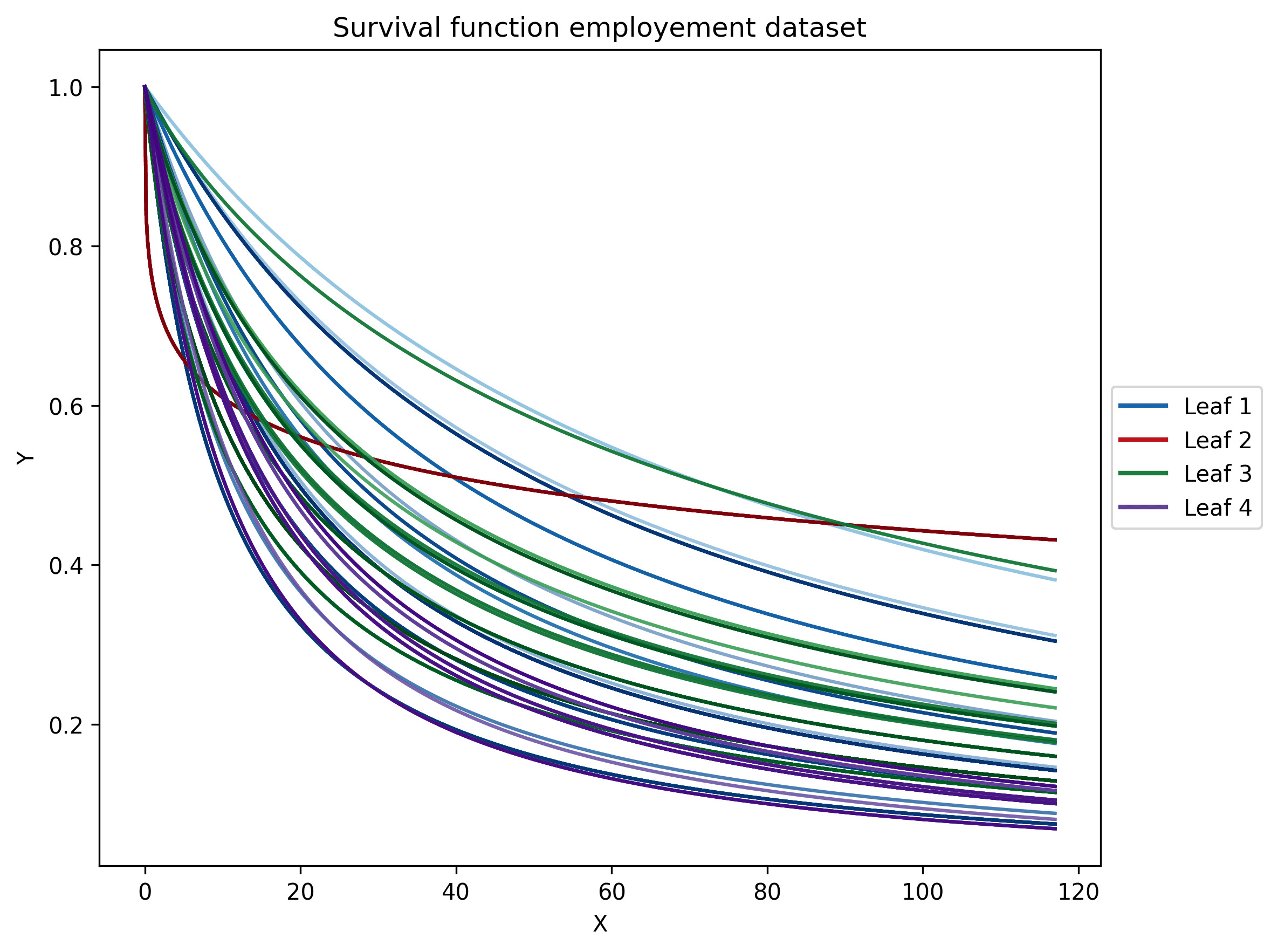}
    \end{subfigure}
    \begin{subfigure}{0.45\textwidth}
        \centering
        \includegraphics[width=\linewidth]{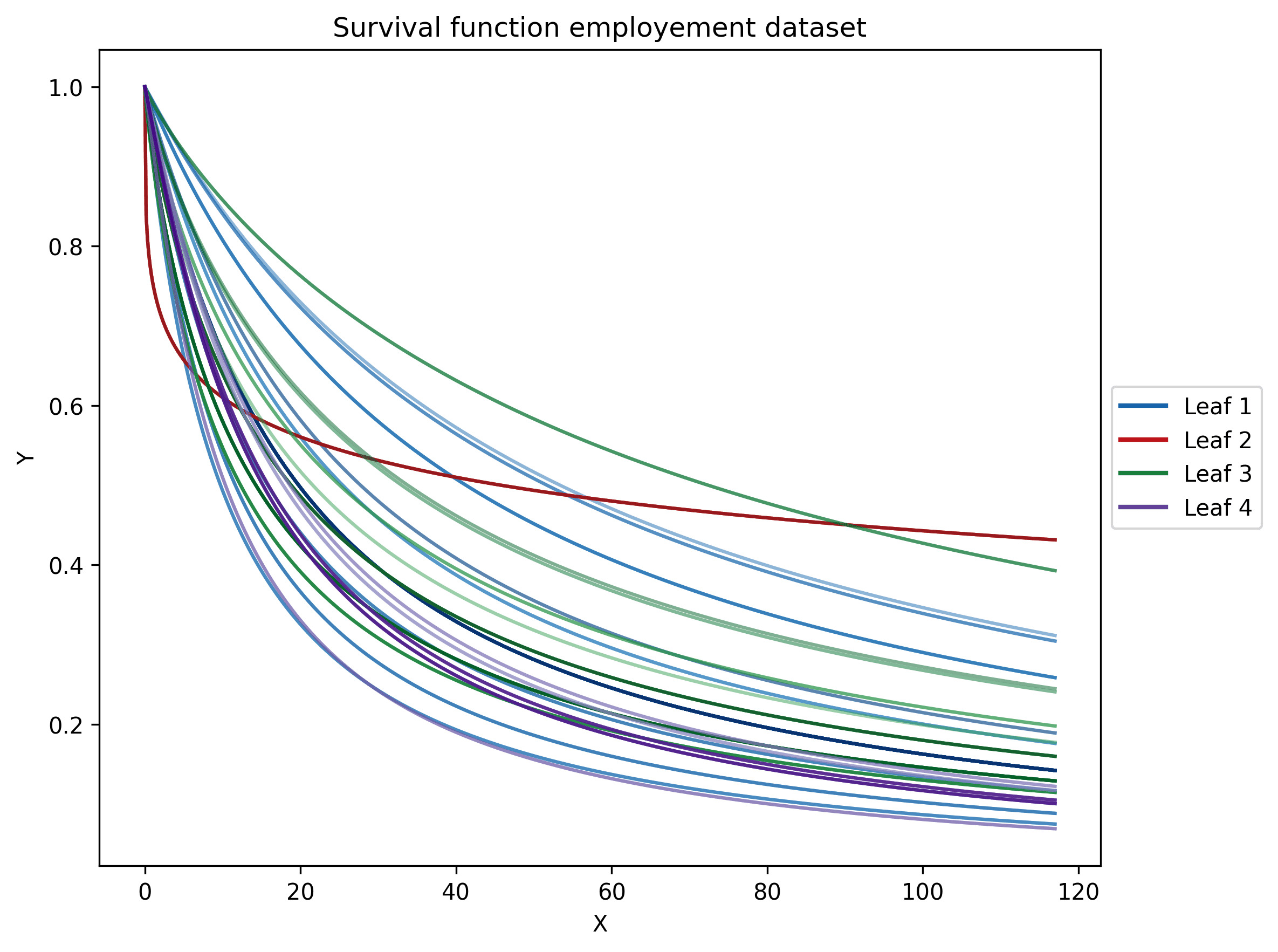}
    \end{subfigure}
    \ContinuedFloat % Mantiene numerazione della figura
\end{figure}

\clearpage % Salto pagina

% SECONDA PARTE CON LA CAPTION UNICA ALLA FINE
\begin{figure}[!ht]
    \ContinuedFloat
    \centering
    % Seconda riga
    \begin{subfigure}{0.45\textwidth}
        \centering
        \includegraphics[width=\linewidth]{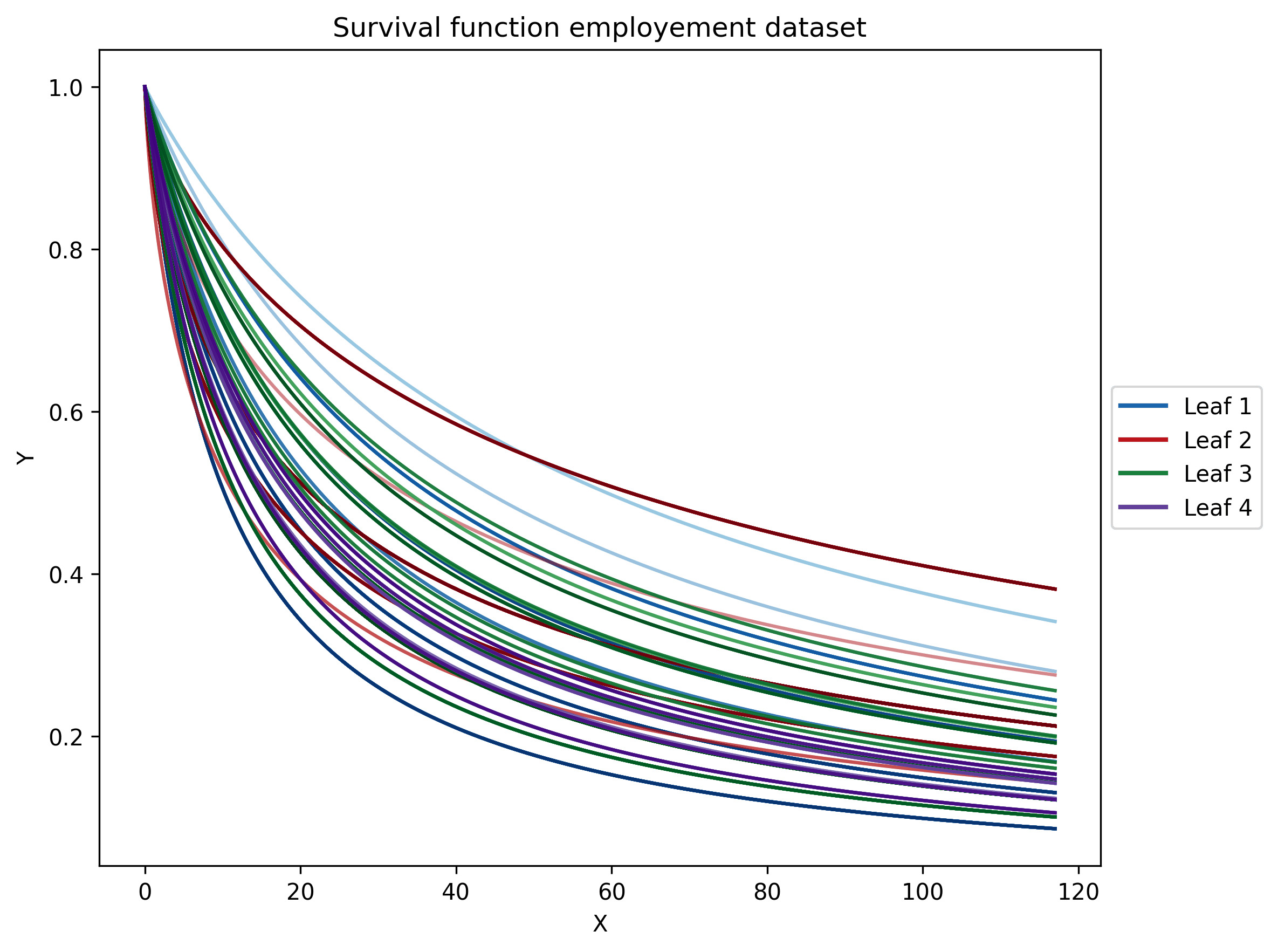}
    \end{subfigure}
    \begin{subfigure}{0.45\textwidth}
        \centering
        \includegraphics[width=\linewidth]{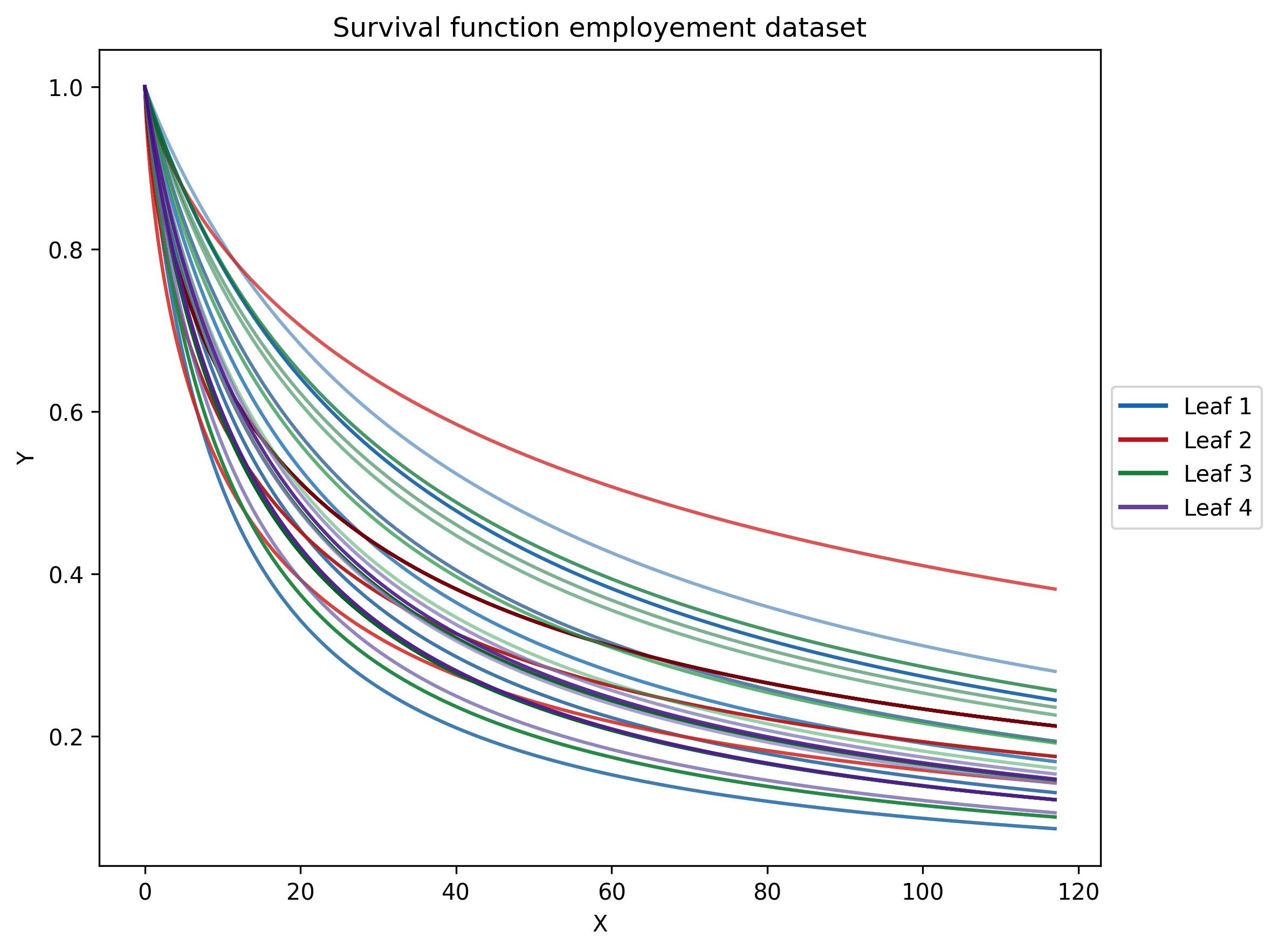}
    \end{subfigure}
    
    % Quarta riga
    \begin{subfigure}{0.45\textwidth}
        \centering
        \includegraphics[width=\linewidth]{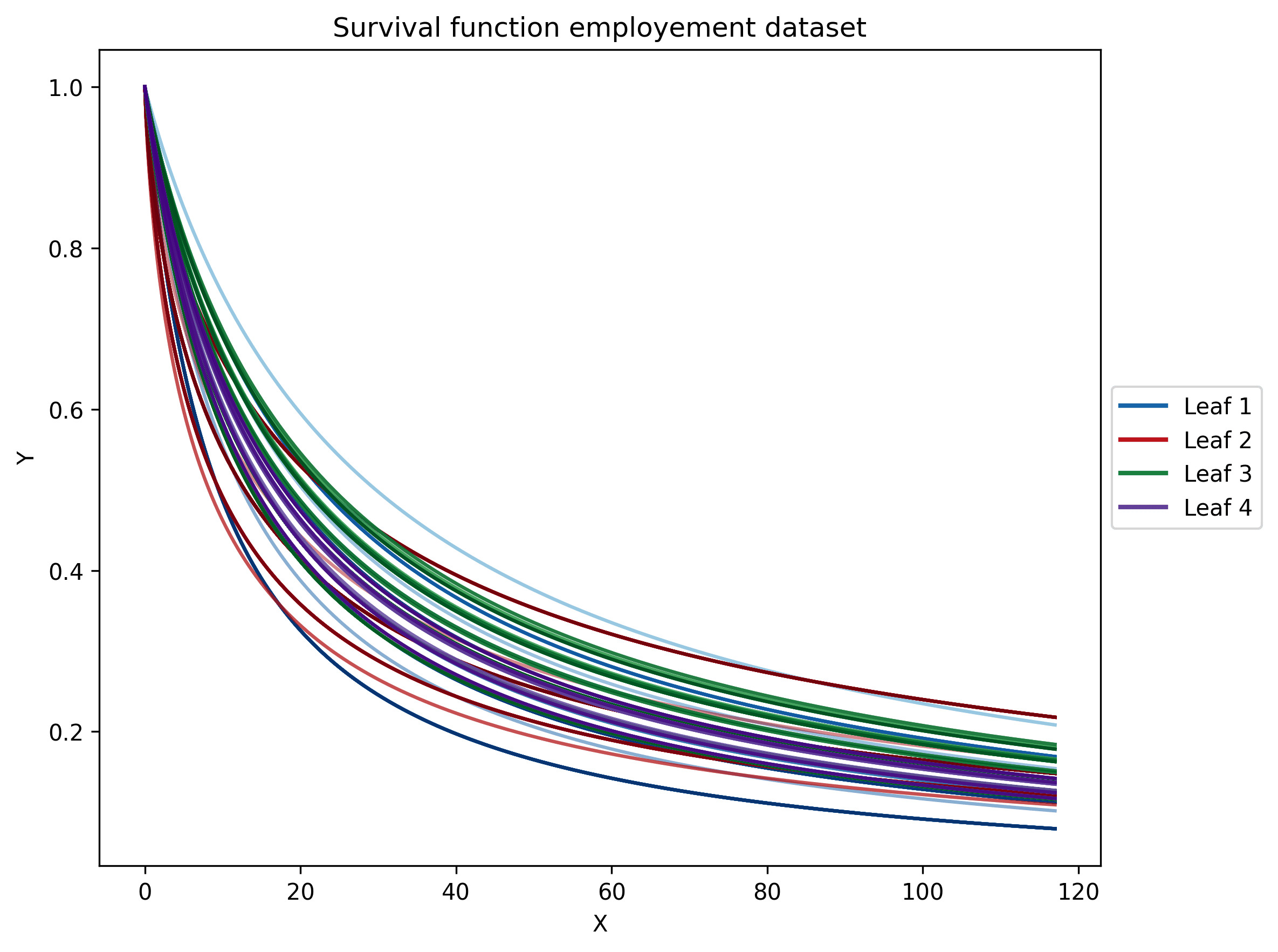}
    \end{subfigure}
    \begin{subfigure}{0.45\textwidth}
        \centering
        \includegraphics[width=\linewidth]{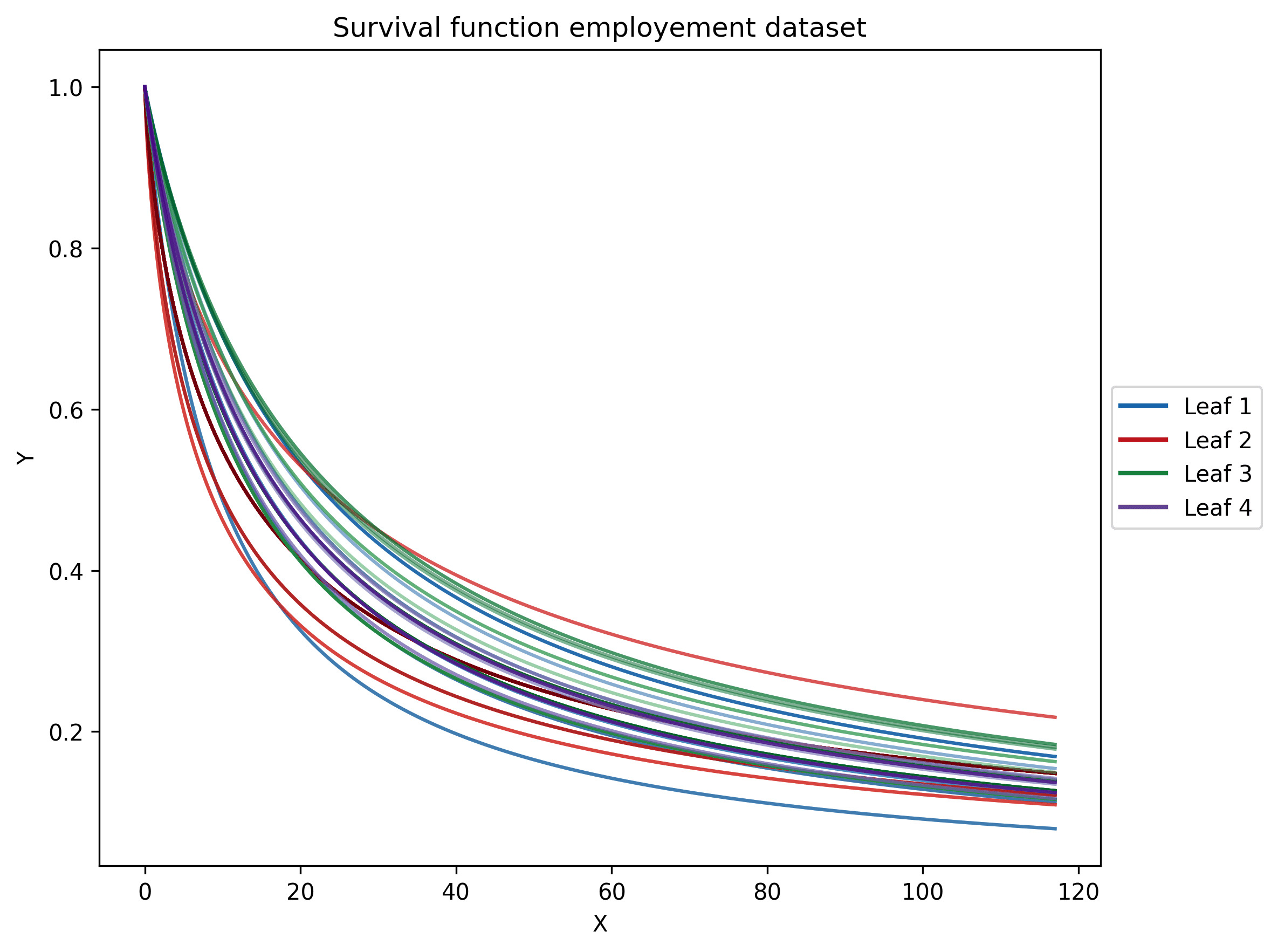}
    \end{subfigure}

    % La caption va qui
    \caption{Plot of survival functions where each row represents the results for a different value of fairness penalty. The red functions of the the first row (i.e. penalty equal to 0) represent a group composed mainly by women. Note well that only in the first line the red represents that group of women, after that they redistribute among the leaf nodes (the colors represent only the leaf nodes).}
    \label{fig:fair_rho}
\end{figure}

\textcolor{black}{In \ref{sec:additional_fair}, we present further results on calibration and discrimination measures, as well as the distance between survival functions as the penalty term $\rho$ increases. In general, the inclusion of the penalty term leads to improvements in group fairness for both the training and testing sets, as shown by the reduced distance between the survival functions of the two groups.} 
%The reader is referred to the \ref{sec:additional_fair} for more details.

\section{Concluding remarks}\label{sec:conclusions}

\textcolor{black}{In this work we proposed a new soft tree model for survival analysis where the actual prediction for any input vector  $\mathbf{x}$ is the corresponding survival function at the single leaf node obtained by routing the input vector from the root along the branches of higher probability.
Such single leaf node predictions, which guarantee the conditional computation property, bring important benefits such as computational efficiency, modeling flexibility and interpretability. The proposed nonlinear optimization formulation is amenable to decomposition, and the node-based decomposition training algorithm with reassignment heuristic, originally proposed in \citep{consolo2025} for regression task, was adapted to the survival problem.} 
% \textcolor{black}{In this work we proposed a new soft tree model for survival analysis. In our model, each input vector is associated with a potential survival function at every leaf node, but the final prediction corresponds to the survival function of a specific leaf node. This enables conditional computation, enhancing efficiency and flexibility. A nonlinear optimization formulation was designed to enable decomposition, and the node-based decomposition algorithm with reassignment heuristic, originally proposed in \citep{consolo2025} for regression task, was adapted to the survival problem.} 

%We showed that our SST and its formulation offer both flexibility, enabling the use of any smooth survival function estimated through maximum likelihood, and interpretability, allowing leaf nodes to represent clusters of distinct survival curves corresponding to different data points. Moreover,\textcolor{black}{as illustrated on a real-world data set, our} SST is capable of directly addressing fairness.

The numerical experiments conducted on 15 well-known datasets demonstrate that our SSTs, using parametric and spline-based semiparametric survival functions and trained with the node-based decomposition \textsc{Nodec-DR-SST} algorithm, outperform the benchmark survival trees in terms of discrimination measures ($C_H,\,C_U,\,\text{CD-AUC}$) as well as calibration (IBS) one. Moreover, as illustrated on a real-world dataset related to unemployment, SSTs can be extended to take into account group fairness.

\textcolor{black}{The above-mentioned results show that our SST model, whose leaf nodes correspond to clusters of distinct survival functions associated to different data points, the training formulation \eqref{eq:formulation} and our \textsc{NODEC-DR-SST} algorithm offer a viable flexible method for survival analysis allowing both flexibility in the choice of survival functions and interpretability. SSTs, which consider multivariate soft splits, allow the use of any smooth survival function estimated via maximum likelihood. This is in contrast with the three recent survival tree approaches, based on \textcolor{darkgreen}{MILP} or dynamic programming, which only consider deterministic univariate splits (with further strong assumptions in \cite{bertsimas2022optimal,huisman2024optimal}) and survival functions that are restricted to} \textcolor{black}{specific non-parametric estimator (e.g., Kaplan-Meier).}

%While the three recent survival tree approaches (based on MIO or dynamic programming) focus on deterministic trees with only univariate splits (in \cite{bertsimas2022optimal,huisman2024optimal} also rely on restrictive assumptions) SSTs consider multivariate soft splits and allow the use of any smooth survival function estimated via maximum likelihood. 

\textcolor{black}{As future work, it would be interesting to enhance the interpretability of SSTs by inducing sparsity to identify the most relevant features for prediction, as done in soft classification and regression trees (see, e.g., \cite{blanquero2020sparsity, amaldi2021multivariate, blanquero2022sparse}) via regularization or via a two-stage procedure (e.g., \cite{CARRIZOSA2023106180}). Furthermore, since SSTs trained via the proposed decomposition algorithm partitions the dataset into clusters, and each one of these clusters is associated to a single leaf node (with the same type of survival model), sparsity could also help to identify the relevant features that determine these partitions.}

\textcolor{black}{Another interesting direction would be to extend SSTs to handle more complex data, such as functional features, previously addressed for soft regression trees in \cite{BLANQUERO2023106152}, spatial data \cite{ripley2005spatial}, or unstructured information such as text, thus enhancing their applicability.}

\textcolor{darkgreen}{\section*{Acknowledgment}}

\textcolor{darkgreen}{This work was partially supported by project PID2022-137818OB-I00 (Ministerio de Ciencia e Innovación, Spain), with European Regional Development Fund (ERDF).}

\vspace{20pt}

\bibliographystyle{elsarticle-harv} 
\bibliography{mybibliography.bib}

@book{ripley2005spatial,
  title={Spatial statistics},
  author={Ripley, Brian D},
  year={2005},
  publisher={John Wiley \& Sons}
}

@inproceedings{aghaei2019learning,
  title={Learning optimal and fair decision trees for non-discriminative decision-making},
  author={Aghaei, Sina and Azizi, Mohammad Javad and Vayanos, Phebe},
  booktitle={Proceedings of the AAAI conference on artificial intelligence},
  volume={33},
  number={01},
  pages={1418--1426},
  year={2019}
}

@inproceedings{wang2022synthesizing,
  title={Synthesizing fair decision trees via iterative constraint solving},
  author={Wang, Jingbo and Li, Yannan and Wang, Chao},
  booktitle={International Conference on Computer Aided Verification},
  pages={364--385},
  year={2022},
  organization={Springer}
}

@inproceedings{jo2023learning,
  title={Learning optimal fair decision trees: Trade-offs between interpretability, fairness, and accuracy},
  author={Jo, Nathanael and Aghaei, Sina and Benson, Jack and Gomez, Andres and Vayanos, Phebe},
  booktitle={Proceedings of the 2023 AAAI/ACM Conference on AI, Ethics, and Society},
  pages={181--192},
  year={2023}
}

@book{oquigley2008proportional,
  title={Proportional hazards regression},
  author={O'Quigley, John},
  year={2008},
  publisher={Springer}
}

@article{van2022fair,
  title={Fair and optimal decision trees: A dynamic programming approach},
  author={van der Linden, Jacobus and de Weerdt, Mathijs and Demirovi{\'c}, Emir},
  journal={Advances in Neural Information Processing Systems},
  volume={35},
  pages={38899--38911},
  year={2022}
}

@article{BLANQUERO2023106152,
	author = {Rafael Blanquero and Emilio Carrizosa and Cristina Molero-R{\'\i}o and Dolores {Romero Morales}},
	journal = {Computers \& Operations Research},
	pages = {106152},
	title = {On optimal regression trees to detect critical intervals for multivariate functional data},
	volume = {152},
	year = {2023}
}

@article{CARRIZOSA2023106180,
title = {On clustering and interpreting with rules by means of mathematical optimization},
journal = {Computers \& Operations Research},
volume = {154},
pages = {106180},
year = {2023},
author = {Emilio Carrizosa and Kseniia Kurishchenko and Alfredo Mar\'{\i}n and Dolores {Romero Morales}}
}

@article{CARRITOP,
	author = {Carrizosa, Emilio and Molero-R{\'\i}o, Cristina and {Romero Morales}, Dolores},
	journal = {TOP},
	number = {1},
	pages = {5--33},
	title = {Mathematical optimization in classification and regression trees},
	volume = {29},
	year = {2021}}

@article{TU2024,
	author = {Jiancheng Tu and Zhibin Wu},
	journal = {European Journal of Operational Research},
	title = {Inherently interpretable machine learning for credit scoring: Optimal classification tree with hyperplane splits},
	year = {2024}
}

@article{OAKES19833,
	author = {David Oakes},
	journal = {European Journal of Operational Research},
	number = {1},
	pages = {3-14},
	title = {Survival analysis},
	volume = {12},
	year = {1983}
}

@article{aghaei2020learning,
  title={Strong optimal classification trees},
  author={Aghaei, Sina and G{\'o}mez, Andr{\'e}s and Vayanos, Phebe},
  journal={Operations Research},
  year={2024},
  publisher={INFORMS},
  volume = {0},
  pages = {0}
}

@book{bertsimas2019machine,
  title={Machine learning under a modern optimization lens},
  author={Bertsimas, Dimitris and Dunn, Jack},
  year={2019},
  publisher={Dynamic Ideas LLC}
}

@article{bertsimas2022optimal,
  title={Optimal survival trees},
  author={Bertsimas, Dimitris and Dunn, Jack and Gibson, Emma and Orfanoudaki, Agni},
  journal={Machine learning},
  volume={111},
  number={8},
  pages={2951--3023},
  year={2022},
  publisher={Springer}
}

@article{blanquero2020sparsity,
  title={Sparsity in optimal randomized classification trees},
  author={Blanquero, Rafael and Carrizosa, Emilio and Molero-R{\'\i}o, Cristina and Romero Morales, Dolores},
  journal={European Journal of Operational Research},
  volume={284},
  number={1},
  pages={255--272},
  year={2020},
  publisher={Elsevier}
}

@article{blanquero2022sparse,
title = {On sparse optimal regression trees},
author={Blanquero, Rafael and Carrizosa, Emilio and Molero-R{\'\i}o, Cristina and {Romero Morales}, Dolores},
journal = {European Journal of Operational Research},
volume = {299},
number = {3},
pages = {1045-1054},
year = {2022}
}

@article{Demirovic2020MurTreeOD,
  title={MurTree: Optimal Decision Trees via Dynamic Programming and Search},
  author={Emir Demirovi\'c and Anna Lukina and Emmanuel H{\'e}brard and Jeffrey Chan and James Bailey and Christopher Leckie and Kotagiri Ramamohanarao and Peter James Stuckey},
  journal={J. Mach. Learn. Res.},
  year={2020},
  volume={23},
  pages={26:1-26:47}
}

@phdthesis{dunn2018optimal,
  title={Optimal trees for prediction and prescription},
  author={Dunn, Jack},
  year={2018},
  school={Massachusetts Institute of Technology}
}

@inproceedings{huisman2024optimal,
  title={Optimal Survival Trees: A Dynamic Programming Approach},
  author={Huisman, Tim and van der Linden, Jacobus GM and Demirovi{\'c}, Emir},
  booktitle={Proceedings of the AAAI Conference on Artificial Intelligence},
  volume={38},
  number={11},
  pages={12680--12688},
  year={2024}
}

@inproceedings{kretowska2024global,
  title={Global induction of oblique survival trees},
  author={Kretowska, Malgorzata and Kretowski, Marek},
  booktitle={International Conference on Computational Science},
  pages={379--386},
  year={2024},
  organization={Springer}
}

@article{leblanc1992relative,
  title={Relative risk trees for censored survival data},
  author={LeBlanc, Michael and Crowley, John},
  journal={Biometrics},
  pages={411--425},
  year={1992},
  publisher={JSTOR}
}

@article{wang2019machine,
  title={Machine learning for survival analysis: A survey},
  author={Wang, Ping and Li, Yan and Reddy, Chandan K},
  journal={ACM Computing Surveys (CSUR)},
  volume={51},
  number={6},
  pages={1--36},
  year={2019},
  publisher={ACM New York, NY, USA}
}

@InProceedings{zhang2024optimal,
  title = 	 {Optimal Sparse Survival Trees},
  author =       {Zhang, Rui and Xin, Rui and Seltzer, Margo and Rudin, Cynthia},
  booktitle = 	 {Proceedings of The 27th International Conference on Artificial Intelligence and Statistics},
  pages = 	 {352--360},
  year = 	 {2024},
  volume = 	 {238},
  series = 	 {Proceedings of Machine Learning Research},
}

@article{kaplan1958nonparametric,
  title={Nonparametric estimation from incomplete observations},
  author={Kaplan, Edward L and Meier, Paul},
  journal={Journal of the American Statistical Association},
  volume={53},
  number={282},
  pages={457--481},
  year={1958},
  publisher={Taylor \& Francis}
}

@article{cox1972regression,
  title={Regression models and life-tables},
  author={Cox, David R},
  journal={Journal of the Royal Statistical Society: Series B (Methodological)},
  volume={34},
  number={2},
  pages={187--202},
  year={1972},
  publisher={Wiley Online Library}
}

@article{gelfand2000proportional,
  title={Proportional hazards models: a latent competing risk approach},
  author={Gelfand, Alan E and Ghosh, Sujit K and Christiansen, Cindy and Soumerai, Stephen B and McLaughlin, Thomas J},
  journal={Journal of the Royal Statistical Society: Series C (Applied Statistics)},
  volume={49},
  number={3},
  pages={385--397},
  year={2000},
  publisher={Wiley Online Library}
}

@article{gray1992flexible,
  title={Flexible methods for analyzing survival data using splines, with applications to breast cancer prognosis},
  author={Gray, Robert J},
  journal={Journal of the American Statistical Association},
  volume={87},
  number={420},
  pages={942--951},
  year={1992},
  publisher={Taylor \& Francis}
}

@article{royston2002flexible,
  title={Flexible parametric proportional-hazards and proportional-odds models for censored survival data, with application to prognostic modelling and estimation of treatment effects},
  author={Royston, Patrick and Parmar, Mahesh KB},
  journal={Statistics in medicine},
  volume={21},
  number={15},
  pages={2175--2197},
  year={2002},
  publisher={Wiley Online Library}
}

@article{royston2011use,
  title={The use of restricted mean survival time to estimate the treatment effect in randomized clinical trials when the proportional hazards assumption is in doubt},
  author={Royston, Patrick and Parmar, Mahesh KB},
  journal={Statistics in medicine},
  volume={30},
  number={19},
  pages={2409--2421},
  year={2011},
  publisher={Wiley Online Library}
}

@article{luo2016spline,
  title={Spline based survival model for credit risk modeling},
  author={Luo, Sirong and Kong, Xiao and Nie, Tingting},
  journal={European Journal of Operational Research},
  volume={253},
  number={3},
  pages={869--879},
  year={2016},
  publisher={Elsevier}
}

@article{bremhorst2016flexible,
  title={Flexible estimation in cure survival models using Bayesian P-splines},
  author={Bremhorst, Vincent and Lambert, Philippe},
  journal={Computational Statistics \& Data Analysis},
  volume={93},
  pages={270--284},
  year={2016},
  publisher={Elsevier}
}

@article{younes1997link,
  author  = {Younes, Naji and Lachin, John M.},
  title   = {Link-Based Models for Survival Data with Interval and Continuous Time Censoring},
  journal = {Biometrics},
  volume  = {53},
  number  = {4},
  pages   = {1199--1211},
  year    = {1997}
}

@article{shen1998propotional,
  title={Propotional odds regression and sieve maximum likelihood estimation},
  author={Shen, Xiaotong},
  journal={Biometrika},
  volume={85},
  number={1},
  pages={165--177},
  year={1998},
  publisher={Oxford University Press}
}

@incollection{ripley1998neural,
  author    = {Brian D. Ripley and Ruth M. Ripley},
  title     = {Neural Networks as Statistical Methods in Survival Analysis},
  booktitle = {Artificial Neural Networks: Prospects for Medicine},
  publisher = {Landes Bioscience},
  year      = {1998},
  pages     = {1--13},
}

@article{katzman2018deepsurv,
  title={DeepSurv: personalized treatment recommender system using a Cox proportional hazards deep neural network},
  author={Katzman, Jared L and Shaham, Uri and Cloninger, Alexander and Bates, Jonathan and Jiang, Tingting and Kluger, Yuval},
  journal={BMC Medical Research Methodology},
  volume={18},
  pages={1--12},
  year={2018},
  publisher={Springer}
}

@inproceedings{giunchiglia2018rnn,
  title={Rnn-surv: A deep recurrent model for survival analysis},
  author={Giunchiglia, Eleonora and Nemchenko, Anton and van der Schaar, Mihaela},
  booktitle={Artificial Neural Networks and Machine Learning--ICANN 2018: 27th International Conference on Artificial Neural Networks, Rhodes, Greece, October 4-7, 2018, Proceedings, Part III 27},
  pages={23--32},
  year={2018},
  organization={Springer}
}

@article{ching2018cox,
  title={Cox-nnet: an artificial neural network method for prognosis prediction of high-throughput omics data},
  author={Ching, Travers and Zhu, Xun and Garmire, Lana X},
  journal={PLoS computational biology},
  volume={14},
  number={4},
  pages={e1006076},
  year={2018},
  publisher={Public Library of Science San Francisco, CA USA}
}

@article{che2018recurrent,
  title={Recurrent neural networks for multivariate time series with missing values},
  author={Che, Zhengping and Purushotham, Sanjay and Cho, Kyunghyun and Sontag, David and Liu, Yan},
  journal={Scientific reports},
  volume={8},
  number={1},
  pages={6085},
  year={2018},
  publisher={Nature Publishing Group UK London}
}

@inproceedings{hu2021transformer,
  title={Transformer-based deep survival analysis},
  author={Hu, Shi and Fridgeirsson, Egill and van Wingen, Guido and Welling, Max},
  booktitle={Survival Prediction-Algorithms, Challenges and Applications},
  pages={132--148},
  year={2021},
  organization={PMLR}
}

@article{lanczky2021web,
  title={Web-based survival analysis tool tailored for medical research (KMplot): development and implementation},
  author={L{\'a}nczky, Andr{\'a}s and Gy{\H{o}}rffy, Bal{\'a}zs},
  journal={Journal of medical Internet research},
  volume={23},
  number={7},
  pages={e27633},
  year={2021},
  publisher={JMIR Publications Toronto, Canada}
}

@article{eloranta2021cancer,
  title={Cancer survival statistics for patients and healthcare professionals--a tutorial of real-world data analysis},
  author={Eloranta, S and Smedby, KE and Dickman, PW and Andersson, TM},
  journal={Journal of internal medicine},
  volume={289},
  number={1},
  pages={12--28},
  year={2021},
  publisher={Wiley Online Library}
}

@article{flynn2012survival,
  title={Survival analysis},
  author={Flynn, Robert},
  journal={Journal of Clinical Nursing},
  volume={21},
  number={19pt20},
  pages={2789--2797},
  year={2012},
  publisher={Wiley Online Library}
}

@article{gepp2008role,
  title={The role of survival analysis in financial distress prediction},
  author={Gepp, Adrian and Kumar, Kuldeep},
  journal={International research journal of finance and economics},
  volume={16},
  number={16},
  pages={13--34},
  year={2008},
  publisher={EuroJournals}
}

@article{zhou2022recurrence,
  title={The recurrence of financial distress: A survival analysis},
  author={Zhou, Fanyin and Fu, Lijun and Li, Zhiyong and Xu, Jiawei},
  journal={International Journal of Forecasting},
  volume={38},
  number={3},
  pages={1100--1115},
  year={2022},
  publisher={Elsevier}
}

@article{brockett2008survival,
  title={Survival analysis of a household portfolio of insurance policies: how much time do you have to stop total customer defection?},
  author={Brockett, Patrick L and Golden, Linda L and Guillen, Montserrat and Nielsen, Jens Perch and Parner, Jan and Perez-Marin, Ana Maria},
  journal={Journal of Risk and Insurance},
  volume={75},
  number={3},
  pages={713--737},
  year={2008},
  publisher={Wiley Online Library}
}

@inproceedings{ranganath2016deep,
  title={Deep survival analysis},
  author={Ranganath, Rajesh and Perotte, Adler and Elhadad, No{\'e}mie and Blei, David},
  booktitle={Machine Learning for Healthcare Conference},
  pages={101--114},
  year={2016},
  organization={PMLR}
}

@inproceedings{dudek2020silhouette,
  title={Silhouette index as clustering evaluation tool},
  author={Dudek, Andrzej},
  booktitle={Classification and Data Analysis: Theory and Applications 28},
  pages={19--33},
  year={2020},
  organization={Springer}
}

@article{grippo2015decomposition,
  title={Decomposition techniques for multilayer perceptron training},
  author={Grippo, Luigi and Manno, Andrea and Sciandrone, Marco},
  journal={IEEE Transactions on Neural Networks and Learning Systems},
  volume={27},
  number={11},
  pages={2146--2159},
  year={2015},
}

@inproceedings{bengio2015conditional,
  title={Conditional Computation in Neural Networks for Faster Models},
  author={Bengio, Emmanuel and Bacon, Pierre-Luc and Pineau, Joelle and Precup, Doina},
  booktitle={International Conference on Learning Representations (ICLR) Workshop Track},
  year={2015}
}

@article{breiman2001random,
  title={Random forests},
  author={Breiman, Leo},
  journal={Machine learning},
  volume={45},
  pages={5--32},
  year={2001},
  publisher={Springer}
}

@book{breimanclassification,
  title={Classification and regression trees},
  author={Breiman, Leo and Friedman, Jerome and Stone, Charles J and Olshen, Richard A},
  year={1984},
  publisher={CRC press}
}

@article{costa2022,
  title={Recent advances in decision trees: An updated survey},
  author={Costa, Vin{\'\i}cius G and Pedreira, Carlos E},
  journal={Artificial Intelligence Review},
  volume={56},
  number={5},
  pages={4765--4800},
  year={2023},
  publisher={Springer}
}

@article{suarez1999globally,
  title={Globally optimal fuzzy decision trees for classification and regression},
  author={Su{\'a}rez, Alberto and Lutsko, James F},
  journal={IEEE Transactions on Pattern Analysis and Machine Intelligence},
  volume={21},
  number={12},
  pages={1297--1311},
  year={1999},
  publisher={IEEE}
}

@book{hastie2009elements,
  title     = {The Elements of Statistical Learning: Data Mining, Inference, and Prediction},
  author    = {Hastie, Trevor and Tibshirani, Robert and Friedman, Jerome},
  year      = {2009},
  publisher = {Springer},
  edition   = {2}
}

@article{chang2011libsvm,
  title={{LIBSVM}: A library for support vector machines},
  author={Chang, Chih-Chung and Lin, Chih-Jen},
  journal={ACM Transactions on Intelligent Systems and Technology (TIST)},
  volume={2},
  number={3},
  pages={1--27},
  year={2011},
  publisher={Acm New York, NY, USA}
}

@article{amaldi2021multivariate,
  title={On multivariate randomized classification trees: l0-based sparsity, VC dimension and decomposition methods},
  author={Amaldi, Edoardo and Consolo, Antonio and Manno, Andrea},
  journal={Computers \& Operations Research},
  volume={151},
  pages={106058},
  year={2023},
  publisher={Elsevier}
}

@article{mehrabi2021survey,
  title={A survey on bias and fairness in machine learning},
  author={Mehrabi, Ninareh and Morstatter, Fred and Saxena, Nripsuta and Lerman, Kristina and Galstyan, Aram},
  journal={ACM computing surveys (CSUR)},
  volume={54},
  number={6},
  pages={1--35},
  year={2021},
  publisher={ACM New York, NY, USA}
}

@inproceedings{pedreshi2008discrimination,
  title={Discrimination-aware data mining},
  author={Pedreshi, Dino and Ruggieri, Salvatore and Turini, Franco},
  booktitle={Proceedings of the 14th ACM SIGKDD international conference on Knowledge discovery and data mining},
  pages={560--568},
  year={2008}
}

@article{davies1979cluster,
  title={A cluster separation measure},
  author={Davies, David L and Bouldin, Donald W},
  journal={IEEE transactions on pattern analysis and machine intelligence},
  number={2},
  pages={224--227},
  year={1979},
  publisher={IEEE}
}

@article{ishwaran2008random,
  title={Random survival forests},
  author={Ishwaran, Hemant and Kogalur, Udaya B and Blackstone, Eugene H and Lauer, Michael S},
journal = {The Annals of Applied Statistics},
 volume={2},
  number={3},
  pages={841--860},
  year={2008}
}

@article{evers2008sparse,
  title={Sparse kernel methods for high-dimensional survival data},
  author={Evers, Ludger and Messow, Claudia-Martina},
  journal={Bioinformatics},
  volume={24},
  number={14},
  pages={1632--1638},
  year={2008},
  publisher={Oxford University Press}
}

@article{van2011support,
  title={Support vector methods for survival analysis: a comparison between ranking and regression approaches},
  author={Van Belle, Vanya and Pelckmans, Kristiaan and Van Huffel, Sabine and Suykens, Johan AK},
  journal={Artificial intelligence in medicine},
  volume={53},
  number={2},
  pages={107--118},
  year={2011},
  publisher={Elsevier}
}

@book{aalen2008survival,
  title={Survival and Event History Analysis: A Process Point of View},
  author={Aalen, Odd O},
  year={2008},
  publisher={Springer-Verlag}
}

@article{lawless2014parametric,
  title={Parametric models in survival analysis},
  author={Lawless, Jerry F},
  journal={Wiley StatsRef: statistics reference online},
  year={2014},
  publisher={Wiley Online Library}
}

@book{klein2014handbook,
  title={Handbook of survival analysis},
  author={Klein, John P and Van Houwelingen, Hans C and Ibrahim, Joseph George and Scheike, Thomas H},
  year={2014},
  publisher={CRC Press Boca Raton, FL:}
}

@article{aranda1981two,
  title={On two families of transformations to additivity for binary response data},
  author={Aranda-Ordaz, Francisco J},
  journal={Biometrika},
  volume={68},
  number={2},
  pages={357--363},
  year={1981},
  publisher={Oxford University Press}
}

@article{ciampi1988recpam,
  title={RECPAM: a computer program for recursive partition and amalgamation for censored survival data and other situations frequently occurring in biostatistics. I. Methods and program features},
  author={Ciampi, Antonio and Hogg, Sheilah A and McKinney, Steve and Thiffault, Johanne},
  journal={Computer methods and programs in biomedicine},
  volume={26},
  number={3},
  pages={239--256},
  year={1988},
  publisher={Elsevier}
}

@article{davis1989exponential,
  title={Exponential survival trees},
  author={Davis, Roger B and Anderson, James R},
  journal={Statistics in medicine},
  volume={8},
  number={8},
  pages={947--961},
  year={1989},
  publisher={Wiley Online Library}
}

@article{gordon1985tree,
  title={Tree-structured survival analysis.},
  author={Gordon, Louis and Olshen, Richard A},
  journal={Cancer treatment reports},
  volume={69},
  number={10},
  pages={1065--1069},
  year={1985}
}

@article{hothorn2006unbiased,
  title={Unbiased recursive partitioning: A conditional inference framework},
  author={Hothorn, Torsten and Hornik, Kurt and Zeileis, Achim},
  journal={Journal of Computational and Graphical statistics},
  volume={15},
  number={3},
  pages={651--674},
  year={2006},
  publisher={Taylor \& Francis}
}

@article{jin2004alternative,
  title={Alternative tree-structured survival analysis based on variance of survival time},
  author={Jin, Hua and Lu, Ying and Stone, Kaite and Black, Dennis M},
  journal={Medical Decision Making},
  volume={24},
  number={6},
  pages={670--680},
  year={2004},
  publisher={Sage Publications Sage CA: Thousand Oaks, CA}
}

@article{kelecs2002residual,
  title={Residual-based tree-structured survival analysis},
  author={Kele{\c{s}}, S{\"u}nd{\"u}z and Segal, Mark R},
  journal={Statistics in Medicine},
  volume={21},
  number={2},
  pages={313--326},
  year={2002},
  publisher={Wiley Online Library}
}

@article{leblanc1993survival,
  title={Survival trees by goodness of split},
  author={LeBlanc, Michael and Crowley, John},
  journal={Journal of the American Statistical Association},
  volume={88},
  number={422},
  pages={457--467},
  year={1993},
  publisher={Taylor \& Francis}
}

@article{molinaro2004tree,
  title={Tree-based multivariate regression and density estimation with right-censored data},
  author={Molinaro, Annette M and Dudoit, Sandrine and Van der Laan, Mark J},
  journal={Journal of Multivariate Analysis},
  volume={90},
  number={1},
  pages={154--177},
  year={2004},
  publisher={Elsevier}
}

@article{segal1988regression,
  author  = {Segal, Mark R.},
  title   = {Regression Trees for Censored Data},
  journal = {Biometrics},
  volume  = {44},
  number  = {1},
  pages   = {35--47},
  year    = {1988}
}

@article{therneau1990martingale,
  title={Martingale-based residuals for survival models},
  author={Therneau, Terry M and Grambsch, Patricia M and Fleming, Thomas R},
  journal={Biometrika},
  volume={77},
  number={1},
  pages={147--160},
  year={1990},
  publisher={Oxford University Press}
}

@inproceedings{zhang1995splitting,
  title={Splitting criteria in survival trees},
  author={Zhang, Heping},
  booktitle={Statistical Modelling: Proceedings of the 10th International Workshop on Statistical Modelling Innsbruck, Austria, 10--14 July, 1995},
  pages={305--313},
  year={1995},
  organization={Springer}
}

@article{plank2008high,
  title={High school dropout and the role of career and technical education: A survival analysis of surviving high school},
  author={Plank, Stephen B and DeLuca, Stefanie and Estacion, Angela},
  journal={Sociology of Education},
  volume={81},
  number={4},
  pages={345--370},
  year={2008},
  publisher={SAGE Publications Sage CA: Los Angeles, CA}
}

@article{kramer2003survival,
  title={A survival analysis of timing of entry into prostitution: The differential impact of race, educational level, and childhood/adolescent risk factors},
  author={Kramer, Lisa A and Berg, Ellen C},
  journal={Sociological Inquiry},
  volume={73},
  number={4},
  pages={511--528},
  year={2003},
  publisher={Wiley Online Library}
}

@article{polsterl2020scikit,
  title={scikit-survival: A Library for Time-to-Event Analysis Built on Top of scikit-learn},
  author={P{\"o}lsterl, Sebastian},
  journal={Journal of Machine Learning Research},
  volume={21},
  number={212},
  pages={1--6},
  year={2020}
}

@misc{hothorn2015ctree,
  author       = {Hothorn, Torsten and Hornik, Kurt and Zeileis, Achim},
  title        = {ctree: Conditional Inference Trees},
  year         = {2015},
  note         = {The Comprehensive R Archive Network (CRAN) vignette}
}

@misc{therneau2021rpart,
  title={rpart: Recursive Partitioning and Regression Trees. {R} package version 4.1--15.},
  author={Therneau, Terry and Atkinson, Beth},
  year={2019},
}

@article{romeo1999conducting,
  title={Conducting inference in semiparametric duration models under inequality restrictions on the shape of the hazard implied by job search theory},
  author={Romeo, Charles J},
  journal={Journal of Applied Econometrics},
  volume={14},
  number={6},
  pages={587--605},
  year={1999},
  publisher={Wiley Online Library}
}

@inbook{steyerberg2009clinical,
  author   = {Ewout W. Steyerberg},
  title    = {Clinical {P}rediction {M}odels},
  chapter  = {15},
  pages    = {255-280},
  year     = 2009,
  publisher= {Springer},
  address  = {New York},
}

@book{lemeshow2008applied,
  title={Applied survival analysis: regression modeling of time-to-event data},
  author={Lemeshow, Stanley and May, Susanne},
  year={2008},
  publisher={Wiley}
}

@article{fotso2019pysurvival,
  title={PySurvival: open source package for survival analysis modeling},
  author={Fotso, Stephane and others},
  journal={URL: https://www. pysurvival.io},
  year={2019}
}

@book{kalbfleisch2002statistical,
  title={The statistical analysis of failure time data},
  author={Kalbfleisch, John D and Prentice, Ross L},
  year={2002},
  publisher={John Wiley \& Sons}
}

@inproceedings{dispenzieri2012use,
  title={Use of nonclonal serum immunoglobulin free light chains to predict overall survival in the general population},
  author={Dispenzieri, Angela and Katzmann, Jerry A and Kyle, Robert A and Larson, Dirk R and Therneau, Terry M and Colby, Colin L and Clark, Raynell J and Mead, Graham P and Kumar, Shaji and Melton III, L Joseph and others},
  booktitle={Mayo Clinic Proceedings},
  volume={87},
  number={6},
  pages={517--523},
  year={2012},
  organization={Elsevier}
}

@article{schumacher1994randomized,
  title={Randomized 2 x 2 trial evaluating hormonal treatment and the duration of chemotherapy in node-positive breast cancer patients. German Breast Cancer Study Group.},
  author={Schumacher, M and Bastert, G and Bojar, H and H{\"u}bner, K and Olschewski, M and Sauerbrei, W and Schmoor, C and Beyerle, C and Neumann, RL and Rauschecker, HF},
  journal={Journal of Clinical Oncology},
  volume={12},
  number={10},
  pages={2086--2093},
  year={1994}
}

@article{mccall1996unemployment,
  author  = {McCall, Brian P.},
  title   = {Unemployment Insurance Rules, Joblessness, and Part-Time Work},
  journal = {Econometrica},
  volume  = {64},
  number  = {3},
  pages   = {647--682},
  year    = {1996}
}

@article{drysdale2022,
  title={{SurvSet}: An open-source time-to-event dataset repository},
  author={Drysdale, Erik},
  journal={arXiv preprint arXiv:2203.03094},
  year={2022}
}

@article{sa2003survival,
  title={Survival of hemodialysis patients: modeling differences in risk of dialysis centers},
  author={S{\'A} Carvalho, Marilia and Henderson, Robin and Shimakura, Silvia and Sousa, In{\^e}s Pereira Silva Cunha},
  journal={International Journal for Quality in Health Care},
  volume={15},
  number={3},
  pages={189--196},
  year={2003},
  publisher={Oxford University Press}
}

@article{mahmood2014framingham,
  title={The Framingham Heart Study and the epidemiology of cardiovascular disease: a historical perspective},
  author={Mahmood, Syed S and Levy, Daniel and Vasan, Ramachandran S and Wang, Thomas J},
  journal={The lancet},
  volume={383},
  number={9921},
  pages={999--1008},
  year={2014},
  publisher={Elsevier}
}

@techreport{ripley1994note,
  author      = {Ripley, Brian D. and Solomon, P. J.},
  title       = {A Note on Australian AIDS Survival},
  institution = {Department of Statistics, University of Adelaide},
  year        = {1994},
  type        = {Research Report},
}

@article{uno2011c,
  title={On the C-statistics for evaluating overall adequacy of risk prediction procedures with censored survival data},
  author={Uno, Hajime and Cai, Tianxi and Pencina, Michael J and D'Agostino, Ralph B and Wei, Lee-Jen},
  journal={Statistics in medicine},
  volume={30},
  number={10},
  pages={1105--1117},
  year={2011},
  publisher={Wiley Online Library}
}

@article{harrell1982evaluating,
  title={Evaluating the yield of medical tests},
  author={Harrell, Frank E and Califf, Robert M and Pryor, David B and Lee, Kerry L and Rosati, Robert A},
  journal={Jama},
  volume={247},
  number={18},
  pages={2543--2546},
  year={1982},
  publisher={American Medical Association}
}

@article{graf1999assessment,
  title={Assessment and comparison of prognostic classification schemes for survival data},
  author={Graf, Erika and Schmoor, Claudia and Sauerbrei, Willi and Schumacher, Martin},
  journal={Statistics in medicine},
  volume={18},
  number={17-18},
  pages={2529--2545},
  year={1999},
  publisher={Wiley Online Library}
}

@article{consolo2025,
  title={Soft regression trees: a model variant and a decomposition training algorithm},
  author={Consolo, Antonio and Amaldi, Edoardo and Manno, Andrea},
  journal={arXiv preprint 	arXiv:2501.05942},
  year={2025}
}

@article{lambert2016summary,
  title={Summary measure of discrimination in survival models based on cumulative/dynamic time-dependent ROC curves},
  author={Lambert, J{\'e}r{\^o}me and Chevret, Sylvie},
  journal={Statistical methods in medical research},
  volume={25},
  number={5},
  pages={2088--2102},
  year={2016},
  publisher={SAGE Publications Sage UK: London, England}
}

@article{kiossou2025generic,
  title={A Generic Complete Anytime Beam Search for Optimal Decision Tree},
  author={Kiossou, Harold Silv{\`e}re and Nijssen, Siegfried and Schaus, Pierre},
  journal={arXiv preprint arXiv:2508.06064},
  year={2025}
}

@inproceedings{brita2025optimal,
  title={Optimal Classification Trees for Continuous Feature Data Using Dynamic Programming with Branch-and-Bound},
  author={Brița, C{\u{a}}t{\u{a}}lin E and van der Linden, Jacobus GM and Demirovi{\'c}, Emir},
  booktitle={Proceedings of the AAAI Conference on Artificial Intelligence},
  volume={39},
  number={11},
  pages={11131--11139},
  year={2025}
}

@article{ENGUR2024127354,
title = {A linear multivariate decision tree with branch-and-bound components},
journal = {Neurocomputing},
volume = {576},
pages = {127354},
year = {2024},
issn = {0925-2312},
author = {Enver Engür and Banu Soylu},
}

@article{lin2024improved,
  title={An improved decision tree algorithm based on boundary mixed attribute dependency},
  author={Lin, Bowen and Liu, Caihui and Miao, Duoqian},
  journal={Applied Intelligence},
  volume={54},
  number={2},
  pages={2136--2153},
  year={2024},
  publisher={Springer Nature BV}
}

@article{yan2008investigating,
  title={Investigating the effects of ties on measures of concordance},
  author={Yan, Guofen and Greene, Tom},
  journal={Statistics in medicine},
  volume={27},
  number={21},
  pages={4190--4206},
  year={2008},
  publisher={Wiley Online Library}
}

\newpage

%% The Appendices part is started with the command \appendix;
%% appendix sections are then done as normal sections
\appendix

\section{Experimental setup}

\subsection{Datasets details}\label{sec:dataset_details}

This section provides details on the event type and the sources of the datasets used in this study:

\begin{itemize}
 \setlength{\itemindent}{-20pt}
\item \textbf{Aids}: AIDS Clinical Trial \citep{lemeshow2008applied}. The event refers to the onset of AIDS.
\item \textbf{Aids$\_$death}: AIDS Clinical Trial \citep{lemeshow2008applied}. The event refers to the death.
\item \textbf{Aids2}: Patients diagnosed with AIDS in Australia before 1 July 1991 \citep{ripley1994note}. The event refers to death.
\item \textbf{Churn}:
The task involves predicting when customers of a software-as-a-service (SaaS) company are likely to cancel their monthly subscription. The event refers to the end of the subscription. This dataset is from pysurvival \citep{fotso2019pysurvival}.
\item \textbf{Credit}: Lenders aim to predict when a borrower will repay their loan. The event refers to loan repayment. This dataset comes from PySurvival \citep{fotso2019pysurvival} and has been adapted from the UCI Machine Learning Repository. %\citep{asuncion2007uci}.
\item \textbf{Dialysis}: \citep{sa2003survival} . the survival of dialysis patients to assess the quality of renal replacement therapy at dialysis centers in Rio de Janeiro \citep{sa2003survival}. The event refers to death. This
dataset is from \citep{drysdale2022}.
\item \textbf{Employee}: The task involves predicting when an employee will resign. The event refers to an employee's departure. This
dataset is from pysurvival \citep{fotso2019pysurvival}.
\item \textbf{Flchain}: Nonclonal serum immunoglobulin free light chains are used to predict overall survival in the population \citep{dispenzieri2012use}. The event refers to death.
\item \textbf{Framingham}: Survival after the onset of congestive heart failure in Framingham Heart Study \citep{mahmood2014framingham}. The event refers to the onset of heart failure. This
dataset is from \citep{drysdale2022}.
\item \textbf{Gbsg2}: 
The German Breast Cancer Study Group 2 \citep{schumacher1994randomized} focuses on recurrence-free survival. The event refers to cancer recurrence-free survival.
\item \textbf{Maintenance}: The task involves predicting when equipment failure will occur. The event refers to machine failure. This dataset is from PySurvival \citep{fotso2019pysurvival}.
\item \textbf{Uissurv}: The UMASS AIDS Research Unit IMPACT Study \citep{lemeshow2008applied} focuses on the drug use. 
The event refers to the resumption of drug use.
\item \textbf{Undempdur}: The impact of changes in unemployment insurance disregards on job search behavior \citep{mccall1996unemployment}. This
dataset is from \citep{drysdale2022}.
The event refers to reemployment. 
\item \textbf{Unemployement}: Unemployment duration and job search theory, using revised Current Population Survey data from September–December 1993. This
dataset is from \citep{drysdale2022}.
The event refers to reemployment. \citep{romeo1999conducting}. 
\item \textbf{Veterans}: Veterans’ Administration Lung Cancer Trial \citep{kalbfleisch2002statistical}. The event refers to death.
\item \textbf{Whas500}: Worcester Heart Attack Study \citep{lemeshow2008applied}. The event refers to death.

\end{itemize}

\subsection{Software packages}\label{sec:soft_pack}

The Sksurv version used in this work is the \texttt{SurvivalTree} function of the Scikit-Survival package version 0.23.0 available at \url{https://scikit-survival.readthedocs.io/en/stable/index.html}. For CTree and RPART, we use the CRAN package pec (version 2023.4.12), accessible at \url{https://cran.r-project.org/web/packages/pec/index.html}. This package serves as a wrapper for the ctree model from the CRAN package party (version 1.2.17, \url{https://cran.r-project.org/web/packages/party/index.html}) and the CRAN package rpart (version 4.1.23, \url{https://cran.r-project.org/web/packages/rpart/index.html}).
\section{Survival tree performance measures}\label{sec:performance}

In this appendix, we briefly recall the accuracy measures considered in this work to evaluate the accuracy of our soft survival trees which are commonly used to assess the performance of survival models. Performance measures for survival models generally focus on their discrimination and calibration capabilities. Discrimination refers to a model's ability to distinguish between subjects who experience an event and those who do not, assigning higher predicted probabilities to the former. Calibration assesses how closely the predicted probabilities match the observed outcomes. The reader is referred to \citep{steyerberg2009clinical} for a detailed description.

\subsubsection{Concordance Indices}

The Concordance Statistic, a well-established measure of discrimination, has been extended from logistic regression to survival models \cite{harrell1982evaluating}. Harrell’s C-index is a concordance statistic based on the principle that a good survival model should assign lower survival probabilities to data points with shorter times-to-event.

A pair of data points $(i,j)$ is considered \textit{comparable} if it is certain that the event of $i$ occurred before the one of $j$, or vice versa. This occurs either when the event times are known for both data points or when one data point's event is observed prior to the other being censored. A pair is defined as \textit{concordant} if the predicted survival probability (i.e., $\hat{S}_{\mathbf{x}_i}(t_i) < \hat{S}_{\mathbf{x}_j}(t_j),\, t_i < t_j$) is lower for the data point with the earlier event. Given that the predicted survival function depends on time, we use the earlier observed time $\text{min} \{t_i,t_j\}$ to assess whether a comparable pair is concordant. Formally, Harrell's C-index is defined as:

$$C_H = \frac{\sum\limits_{i} \sum\limits_{j} c_i \cdot \mathds{1}_{t_i < t_j} \cdot \left( \mathds{1}_{\hat{S}_{\mathbf{x}_i}(t_i) < \hat{S}_{\mathbf{x}_j}(t_i)} + 0.5 \cdot \mathds{1}_{\hat{S}_{\mathbf{x}_i}(t_i) = \hat{S}_{\mathbf{x}_j}(t_i)} \right)}{\sum\limits_{i} \sum\limits_{j} c_i \cdot \mathds{1}_{t_i < t_j}}$$

Harrell’s C statistic ranges from 0 to 1, where higher values indicate better model fit, and randomly assigned predictions yield an expected score of 0.5. 

A main drawback of Harrell’s C is its bias in datasets with a high proportion of censored data. To mitigate this limitation, \cite{uno2011c} enhanced Harrell’s C by applying Inverse Probability of Censoring Weights (IPCW), making it more robust to extensive censoring. The Uno’s C-index is given by:

$$C_U = \frac{\sum\limits_{i} \sum\limits_{j} c_i \cdot \hat{G}^{-2}(t_i) \cdot \mathds{1}_{t_i < t_j} \cdot 
\left( \mathds{1}_{\hat{S}_{\mathbf{x}_i}(t_i) < \hat{S}_{\mathbf{x}_j}(t_i)} + 0.5 \cdot \mathds{1}_{\hat{S}_{\mathbf{x}_i}(t_i) = \hat{S}_{\mathbf{x}_j}(t_i)} \right)}{\sum\limits_{i} \sum\limits_{j} c_i \cdot \hat{G}^{-2}(t_i) \cdot \mathds{1}_{t_i < t_j}}
$$

\noindent where $\hat{G}$ represents the Kaplan–Meier estimate of the censoring distribution $\mathbf{c}$, assuming independence from the features.

In general concordance indices overlook incomparable observation pairs, which can be problematic under heavy censoring, and their binary definition fails to capture the magnitude of risk differences (i.e. survival probabilities) between comparable data points, reducing their informativeness in datasets with substantial risk variation.

\subsubsection{Cumulative Dynamic AUC}

In classical binary classification the Receiver Operating Characteristic (ROC) curve plots the true positive rate against the false positive rate, with the Area Under the Curve (AUC) representing the probability that the model ranks a randomly chosen positive data point higher than a negative one. 

In survival analysis, for a given time $t$, the concept is adapted: true positives refer to data points  that experienced the event (e.g., death) at or before $t$ ($c_i = 1,\, t_i < t $), while true negatives are those that remain event-free beyond $t$ ($t_i > t$). 
As the time $t$ of the predicted survival function varies, the true positive and false positive rates change, resulting in a time-dependent ROC curve. The corresponding time-dependent AUC serves as a valuable tool for evaluating how well a survival models discriminates the event occurring within a time interval up to $t$, rather than at a single specific time point. The AUC at a specific time $t$ is expressed as follows:

$$\hat{\text{AUC}}(t) = \frac{\sum\limits_{i} \sum\limits_{j} c_i \cdot \hat{G}^{-1}(t_i) \cdot \mathds{1}_{t_i \leq t} \cdot \mathds{1}_{t_j > t} \cdot \mathds{1}_{\hat{S}_{\mathbf{x}_i}(t) < \hat{S}_{\mathbf{x}_j}(t)}}{\left(\sum\limits_{i} c_i \cdot \hat{G}^{-1}(t_i) \cdot \mathds{1}_{t_i \leq t}\right) \cdot \left(\sum\limits_{j} \mathds{1}_{t_j > t}\right)}$$

\noindent where $\hat{G}(\cdot)$ is the usual IPCW computed using the Kaplan-Meier estimator.

\cite{lambert2016summary} proposed the Cumulative Dynamic AUC (CD-AUC), which restricts the time-dependent weighted AUC estimator to a specific time interval, providing a summary measure of the mean AUC over the selected range:

$$\text{CD-AUC} = \frac{1}{\hat{S}(t_{\text{min}}) - \hat{S}(t_{\text{max}})} \int_{t_{\text{min}}}^{t_{\text{max}}} \hat{\text{AUC}}(t) \, d\hat{S}(t)$$

\noindent where $\hat{S}(\cdot)$ is the Kaplan-Meier estimator of the survival function and the time range is derived from the dataset as $t_{\text{min}} = \min\{t_i\}_{i=1}^N$ and $t_{\text{max}} = \max\{t_i\}_{i=1}^N$.

%$$\hat{S}_e(\tau, t) = \frac{\sum\limits_{i} c_i \cdot \hat{G}^{-1}(t_i) \cdot \mathds{1}_{t_i \leq t} \cdot \mathds{1}_{\hat{S}(t|\mathbf{x}_i) < \tau}}{\sum\limits_{i} c_i \cdot \hat{G}^{-1}(t_i) \cdot \mathds{1}_{t_i \leq t}}$$

\subsubsection{Integrated \textcolor{black}{Brier} score}

The Brier Score (BS) is a widely used measure for assessing the accuracy of probabilistic classification models. \cite{graf1999assessment} introduced an adapted version of the \textcolor{black}{BS} to handle survival datasets with censored outputs. In survival analysis, BS is commonly used to evaluate the calibration of survival models. At a specific time $t$, $BS(t)$ represents the mean \textcolor{black}{squared} error between the observed data and the predicted survival function $\hat{S}_{\mathbf{x}_i}(t)$, adjusted using IPCW:

\begin{equation}
BS(t) = \frac{1}{N} \sum_{i=1}^N \left( 
\frac{(\hat{S}_{\mathbf{x}_i}(t) - 0)^2}{\hat{G}(t_i)} \cdot \mathds{1}_{t_i \leq t, c_i = 1} 
+ \frac{(\hat{S}_{\mathbf{x}_i}(t) - 1)^2}{\hat{G}(t)} \cdot \mathds{1}_{t_i > t} 
\right). 
\end{equation}

Similar to the CD-AUC, the Integrated Brier Score (IBS) offers a comprehensive assessment of model performance across a specified time interval:

\begin{equation}
\text{IBS} = \frac{1}{t_{\text{max}}} \int_{0}^{t_{\text{max}}} BS(t) \, dt 
\end{equation}

\noindent where $ t_{\text{max}} = \max\{t_i\}_{i=1}^N $ is the latest time point of all observed \textcolor{black}{data points}.
\section{Initialization and branch node update in NODEC-DR-SST}\label{sec:dec_details}

\textcolor{black}{In this appendix, we describe two important components of NODEC-DR-SST which are just mentioned or summarized in Section \ref{sec:decomposition}, namely, the procedure to generate initial solutions and the procedure to update the branch nodes variables.}

\textcolor{black}{\subsubsection{Initialization procedure}}

\textcolor{black}{Since the NLO training formulation \eqref{eq:obj} is nonconvex, appropriate initial solutions can improve the training phase and enhance testing accuracy. To find suitable initial solutions, we adapt the initialization procedure proposed in \citep{consolo2025} for soft regression trees to survival analysis. Starting from the root node with all the input vectors  $\mathbf{x}_i$ of the training set $I$, we proceed level by level towards the leaf nodes and we apply at each branch node $n \in \tau_B$ a binary clustering algorithm in order to partition the input vectors assigned to node $n$ into two subsets of input vectors to be assigned to the two child nodes of $n$, denoted by $X^{right}_n$ and  $X^{left}_n$. Once each input vector has been assigned to a single root-to-leaf-node path, we determine for each branch node $n$ the corresponding values of the variables $\bm{\w}^0_n$ by solving a binary Logistic Regression problem where we try to route all input vectors in $X^{right}_n$ ($X^{left}_n$) along the right (left) branch of node $n$. Similarly, the corresponding variables $\bm{\beta}^0_n$ characterizing the survival curves associated to each leaf node $n$ are determined by solving a maximum likelihood problem where only the input vectors falling into the leaf node $n$ are considered. It is worth pointing out that this initialization procedure is applicable to any type of (parametric or semiparametric) survival curves which can be estimated via a maximum likelihood approach.}

\textcolor{black}{This sequence of binary clustering problems from the root node to the bottom level leaf nodes is repeated several times. At the end, the set of clusters associated to the leaf nodes that maximizes an appropriate clustering quality metric is used to generate an initial solution $(\bm{\w}^{0},\bm{\beta}^{0})$. The quality of the clusters at the leaf nodes is assessed using silhouette scores, which indicate the average similarity between each cluster and its most similar one \citep{dudek2020silhouette}. As in \cite{consolo2025}, we adopt the Davies-Bouldin index as the silhouette score \citep{davies1979cluster}, which represents the average similarity of each cluster to its most similar cluster. Once the best set of clusters of input vectors associated to the leaf nodes is selected based on the silhouette score, the variables $\bm{\w}^0$ and $\bm{\beta}^{0}$ for all the nodes are determined as above. The reader is referred to \citep{consolo2025} for details on the initialization procedure.}

\textcolor{black}{\subsubsection{Branch node update procedure}}
 
\textcolor{black}{As described in Section \ref{sec:decomposition}, at each inner iteration $k$ of NODEC-DR-SST, the working set is divided into a branch node working set, $W^k_B\subseteq \tau_B$, and a leaf node working set, $W^k_L \subseteq \tau_L$. NODEC-DR-SST first optimizes \eqref{eq:obj} over the branch node variables associated with $W^k_B$ ({\bf BN Step}) and then over the leaf node variables corresponding to $W^k_L$ ({\bf LN Step}).}

\textcolor{black}{To improve efficiency and reduce computational time in the \textbf{BN Step} and \textbf{LN Step}, we solve a surrogate subproblem. Specifically, for each selected branch node $s \in \tau_B$, we restrict the optimization to the subtree rooted at $s$, neglecting the variables $\bm{\w}$ and $\bm{\beta}$ (and the probabilities $p_{in}$) associated to the other nodes of the tree. Furthermore, we consider only the set of data points $(\textbf{x}_i,t_i,c_i)$ whose HBP paths include $s$ (namely $I_s$), as they are the most relevant ones for optimizing the variables in $W_B^k$ and $W_L^k$.}

\textcolor{black}{Since the {\bf BN Step} problem is nonconvex, standard optimization methods often yield to solutions such that the HBP paths of a large proportion of data points end at a few leaf nodes. To mitigate this imbalance, the \textsc{UpdateBranchNode} procedure includes a reassignment heuristic which reroutes the corresponding input vectors $\textbf{x}_i$ by partially modifying their HBP paths.  To detect and correct imbalance along the tree branches, three positive thresholds, $\varepsilon_1$,  $\varepsilon_2$, and $\varepsilon_3$, are introduced.}

\textcolor{black}{To determine the level of imbalance at a branch node $s$, we define $L_s$ as the ratio of the number of input vectors $\mathbf{x}_i$ deterministically routed along the left branch of node $s$ to the cardinality of the restricted training set $I_s$. If the data point routing at branch node $s$ is sufficiently balanced, the update
$\bm{\w}_{W^k_{B}}^{k+1}$ is obtained by minimizing the error function in \eqref{eq:obj} associated with the restricted subtree rooted at $s$ and the restricted training set $I_s$, with respect to the variables $\bm{\w}_n$ for $n \in W_B^k$.} 

\textcolor{black}{If $L_s$ exceeds $\varepsilon_1$ but is below $\varepsilon_2$ (moderate imbalance), one tries to improve the routing balance at branch node $s$ by solving a surrogate subproblem that updates $\bm{\w}_s$ while accounting for the relative weights of the data points routed along its left and right branches. Specifically, $\bm{\w}_{W^k_{B}}^{k+1}$ is determined by solving the following two-class Weighted Logistic Regression (WLR) problem with respect to the variables $\bm{\w}_s$ and the restricted training set $I_s$, while keeping fixed all $\bm{\w}_{l}$ for $l \in \tau_B \setminus \{s\}$: 
\begin{equation}
\label{eq:reshuffle}
 \min -\frac{1}{|I_s|} \sum_{i \in I_s} w_ir_iln(p_{{\bf x}_i s}) + w_i(1 - r_i)ln(1-p_{{\bf x}_is}),  
\end{equation}
where if the input vector $\mathbf{x}_i$ is routed along the left branch of node $s$ we set $w_i=\frac{1}{2 L_s}$ and the routing class parameter as $r_i = 1$, and if ${\bf x}_i$ is routed along the right branch %of node $s$ then 
we set $w_i = \frac{1}{2 (1-L_s)}$ and %the routing class parameter as 
$r_i = 0$. The values of the weights are defined so that $w_i$ is larger for input vectors $\mathbf{x}_i$ that are routed towards the child node of $s$ where the smallest number of input vectors $\mathbf{x}_i$ in $I_s$ fall into it.
}

\textcolor{black}{In the case of high level imbalance, that is, when $L_s$ exceeds $\varepsilon_2$, $\bm{\w}^{k+1}_{W_B^k}$ is obtained, as above, 
by minimizing a WLR function with respect to only the $\bm{\w}_s$ variables of branch node $s$. However, to mitigate the high imbalance, a fraction $\varepsilon_3$ of the input vectors $\mathbf{x}_i$ in $I_s$, namely, those with a larger term in the negative log-likelihood, are reassigned to the opposite branch of node $s$ by swapping their routing class parameter $r_i$ in the corresponding WLR problem. Notice that the solution of the WLR problem is not guaranteed to reroute all such input vectors at branch node $s$ according to the updated routing class parameters.}

\renewcommand{\thealgorithm}{Update Branch Node} %RIMUOVE IL NUMERO
\begin{algorithm}[h!]
\floatname{algorithm}{} %PER RIMUOVERE ALGORITHM
\caption{- Procedure with data points reassignment heuristic}
\label{alg:updatebranching}
$\textbf{Input:} \ \text{training set} \ I; \ W^k_B\, \text{working} \, \text{set};\, \text{previous}\;\text{update}\, \bm{\w}^{k}_{W^k_B};\,\varepsilon_{1},\,\varepsilon_{2},\,\varepsilon_{3} \in (0,1),\, \varepsilon_1 > \varepsilon_2\\$ 
$\textbf{Output:}\,\text{update}\, \tilde{\bm{\w}}_{W^k_B}$
  % t is the ancestor of all the other nodes in Wkb 
  {\fontsize{9}{10}\selectfont
  \begin{algorithmic}[1]
    
    \Procedure{UpdateBranchNode}{$I$,$W^k_B,\bm{\w}^{k}_{W^k_B},\varepsilon_{1},\varepsilon_{2},\varepsilon_{3}$}
        \State $s\gets$  $\min \{ {\hat s}\ | \;{\hat s}\in W^k_B\}$ \Comment{s is the ancestor of all the other nodes in $W^k_B$}
        \State $I_{s}\gets$  $\{ (\textbf{x}_i,t_i,c_i) \in I |$ \text{HBP path of} $(\textbf{x}_i,t_i,c_i)$ \text{contains} $s \}$ \Comment{data points of $I_{s}$ %deterministically  
        falling into $s$}
        
        %\ifthenelse{}{}{}
        \If{($ L_{s}\leq \varepsilon_{1}\; \text{or}\; L_{s} \geq 1 - \varepsilon_{1}) \;\text{and}\;({\varepsilon_1}N \geq 1)$ }\Comment{imbalanced data points routing}
        \State $\tilde{\bm{\w}}_{W^k_B \setminus t} \gets \bm{\w}^{k}_{W^k_B \setminus t}$
        \State \textbf{for} $(\textbf{x}_i,t_i,c_i) \in I_s$  \textbf{do} 
        \State $\;\quad$\textbf{if} $\mathbf{x}_i$ is routed towards the left child node of $t$ \textbf{then} 
        \State $\;\quad\quad$ set $r_i=1$ in \eqref{eq:reshuffle} \textbf{else} set $r_i=0$ in \eqref{eq:reshuffle}
            
            \If{$L_{s} \leq \varepsilon_{2}\;\text{or}\;L_{s} \geq 1 - { \varepsilon_{2}}$}\Comment{high imbalanced routing}
        \State   $\bullet$  let $d_{max}\; \text{be the index of the child node of $s$ with maximum number of $\mathbf{x}_i$ routed to it}$
        
        %$\bullet$  set $d_{max} = argmax \{N_{left_{t}},N_{right_{t}}\}$\Comment{child node of $t$ with max number of $\mathbf{x}_i$ routed to it}
        \State $\bullet$ let $\hat{I}_{s}$ as the $\varepsilon_{3}\%$ of data points in $I_{s}$ with largest negative log-likelihood value routed to $d_{max}$ 
        \State $\bullet$ for every $(\textbf{x}_i,t_i,c_i) \in \hat{I}_{s}$ set  $r_i = 1-r_i$ in \eqref{eq:reshuffle} \Comment{aims at routing $\textbf{x}_i$ along the other branch}
        \State
        $\bullet$ determine $\tilde{\bm{\w}}_{s}$ by minimizing the WLR function \eqref{eq:reshuffle} restricted to $I_s$
        
            \Else \Comment{moderate imbalanced routing}
            \State $\bullet$ determine $\tilde{\bm{\w}}_{t}$ by minimizing the WLR function \eqref{eq:reshuffle} restricted to $I_s$
            \EndIf
            \State $\tilde{\bm{\w}}_{W^k_B} \gets \{\tilde{\bm{\w}}_{s},\tilde{\bm{\w}}_{W^k_B \setminus s} \}$ \Comment{$\tilde{\bm{\w}}_{s}$ are the only updated variables}
        \Else
            \State determine $\tilde{\bm{\w}}_{W^k_B}$ by solving \eqref{eq:obj} with respect to variables ${\bm{\w}}_{W^k_B}$  %\gets OptmizeBranchNode(W^k_B,\bm{\w}^{k-1}_{W^k_B})$
            \Comment{minimize with any %nonlinear unconstrained 
            algorithm} 
        \EndIf
        
        \State \textbf{return} $\tilde{\bm{\w}}_{W^k_B}$
 \EndProcedure
  \end{algorithmic}
  }
\end{algorithm}

\textcolor{black}{To do so, we denote as $d_{max}$ the index of the child node of node $s$ where the largest number of input vectors $\mathbf{x}_i$ from $I_s$ fall deterministically. Given $\varepsilon_3 \in (0,1)$, we define $\hat{I}_s$ as the subset of  $I_s$ containing the $\varepsilon_3\%$ of data points with the largest negative log-likelihood among those routed towards the child node indexed by $d_{max}$. Since a high negative log-likelihood may indicate that these points may be better suited for a different routing, we attempt to reassign each data point $(\textbf{x}_i,t_i,c_i) \in \hat{I}_s$ to the other child node of node $s$ by setting $r_i = 1 - r_i$ and then optimizing the WLR function \eqref{eq:reshuffle} using an unconstrained nonlinear optimization method.
%\footnote{Clearly, the solution obtained for the surrogate subproblem may not reroute all the data points in $\hat{I}_s$ according to the selected values of the parameter $r_i$.}.
}

\textcolor{black}{Note that the data points to be reassigned to different child nodes are those corresponding to larger terms in the error function \eqref{eq:obj} associated to the restricted subtree rooted at $s$ and the restricted dataset $I_s$. Specifically,  the error term for each data point $(\mathbf{x}_i,t_i,c_i)$ is given by:
\begin{equation}
   \label{eq:partial_objective}
    \sum_{n\in {\cal D }_L(s)}P_{in}L^-_n(\mathbf{x}_i,t_i,c_i;\bm{\beta}_n),
\end{equation}
\noindent where ${\cal D }_L(s)$ denotes the subset of leaf nodes that are descendants of node $s$.
}

\textcolor{black}{In general, given the form of the negative log-likelihood, it is necessary to ensure that, for any leaf node $n$ and any data point $(\mathbf{x}_i,t_i,c_i)$ for which the event has been observed (i.e., $c_i=1$), the argument of the logarithm is always non-negative. \textcolor{black}{In SSTs with spline-based semiparametric survival function estimation,} this implies ensuring that the term associated with the spline formulation, $\frac{\text{d}s(y;\bm{\eta}_n)}{\text{d}y} $, is non-negative. After a few macro iterations, in the \textbf{BN Step} it may occur that for some data points $(\mathbf{x}_i,t_i,c_i) \in I_s$ the values of $\frac{\text{d}s(y;\bm{\eta}_n)}{\text{d}y}$ corresponding to the leaf nodes $n \in {\cal D}_L(s)$ are negative. This tends to happen when the probabilities $P_{in}$ of falling into the leaf nodes $n$ are very low, and is due to the fact that in the \textbf{LN Step} the variables $\bm{\beta}$ are updated using only a subset of the data points. In practice, when $\frac{\text{d}s(y;\bm{\eta_}n)}{\text{d}y}$ is negative for a leaf node $n \in \tau_L$, we exclude the corresponding term in the objective function $P_{in}L^-_n(\mathbf{x}_i,t_i,c_i;\bm{\beta}_n)$. Similarly, in high imbalance cases it may occur that some of the data points in $I_s$ with the highest impact on the negative log-likelihood (more likely to be selected among the ${\varepsilon_3}\%$ data points to be rerouted) have a negative $\frac{\text{d}s(y;\bm{\eta}_n)}{\text{d}y}$ for the alternative leaf node $n$ towards which we aim at rerouting them. Then such data points are also excluded and not considered for rerouting.
}

\section{Experimental results}\label{sec:tables_sd}

In this appendix, we present additional experimental results. In paritcular, we provide a comparison of SST at depths 1 and 2 with the benchmark models across all four discrimination and calibration measures ($C_H,\, C_U$, CD-AUC, and IBS). Moreover, we report the training and testing results for the four discrimination and calibration measures applied to the three benchmark models (CTree, RPART, and SkSurv) considering depths from $D=2$ to $D=5$. Lastly, for the three models discussed in Section \ref{sec:results_exp}, we include the corresponding boxplots illustrating the testing $C_U$, $C_H$, CD-AUC, and IBS.

\subsection{Tables reporting the average and standard deviation of training and testing measures for each method}

\begin{landscape}
\noindent \textbf{Depth $D=1$}
\vspace{-20pt}
\begin{table}[H]
\centering
\resizebox{0.85\textwidth}{!}{
\begin{tabular}{lccccccccc}
\cline{2-10}
             & \multicolumn{9}{c}{Testing CD-AUC} \\
\cline{2-10}
             & \multicolumn{4}{c}{D=1} & \multicolumn{3}{c}{D=5} & \multicolumn{2}{c}{D=1} \\\hline
Dataset      & \multicolumn{1}{l}{\textbf{Llog}} & \multicolumn{1}{l}{\textbf{Llog-init}} & \multicolumn{1}{l}{\textbf{Exp}} & \multicolumn{1}{l}{\textbf{W}} & \multicolumn{1}{l}{\textbf{SkSurv}} & \multicolumn{1}{l}{\textbf{RPART}} & \multicolumn{1}{l}{\textbf{CTree}} & \multicolumn{1}{l}{\textbf{PO}} & \multicolumn{1}{l}{\textbf{PH}} \\\hline
Aids         & 0.748 (7.89) & 0.749 (6.78) & 0.751 (8.3) & 0.748 (7.7) & 0.642 (8.15) & 0.743 (9.25) & 0.724 (10.14) & 0.737 (5.55) & 0.743 (5.58) \\
Aids\_death  & 0.807 (7.82) & 0.843 (4.58) & 0.797 (7.33) & 0.808 (7.72) & 0.569 (12.65) & 0.637 (11.95) & 0.651 (9.78) & 0.737 (13.4) & 0.735 (12.94) \\
Aids2        & 0.562 (1.79) & 0.561 (1.91) & 0.558 (1.48) & 0.560 (1.62) & 0.543 (0.89) & 0.519 (1.69) & 0.542 (0.9) & 0.557 (1.93) & 0.545 (2.62) \\
Churn        & 0.924 (0.79) & 0.925 (0.95) & 0.915 (0.78) & 0.914 (0.63) & 0.851 (2.22) & 0.850 (1.21) & 0.856 (1.22) & 0.923 (0.94) & 0.917 (0.93) \\
Credit\_risk & 0.856 (2.37) & 0.855 (2.42) & 0.808 (2.89) & 0.850 (2.51) & 0.813 (1.74) & 0.824 (0.9) & 0.816 (1.57) & 0.855 (2.67) & 0.850 (2.77) \\
Dialysis     & 0.794 (1.21) & 0.786 (2.54) & 0.794 (1.22) & 0.795 (1.23) & 0.671 (2.52) & 0.630 (1.6) & 0.672 (1.33) & 0.768 (1.39) & 0.772 (1.5) \\
Employee     & 0.785 (2.19) & 0.783 (2.53) & 0.775 (1.99) & 0.793 (2.44) & 0.946 (0.56) & 0.942 (0.53) & 0.934 (0.51) & 0.771 (2.83) & 0.782 (3.06) \\
Flchain      & 0.955 (0.51) & 0.955 (0.54) & 0.953 (0.48) & 0.954 (0.52) & 0.875 (1.53) & 0.901 (0.62) & 0.906 (0.36) & 0.954 (0.27) & 0.871 (17.38) \\
Framingham   & 0.773 (1.36) & 0.773 (1.31) & 0.771 (1.43) & 0.772 (1.39) & 0.722 (1.24) & 0.713 (1.41) & 0.733 (1.44) & 0.767 (1.46) & 0.771 (1.31) \\
Gbsg2        & 0.711 (4.11) & 0.700 (4.98) & 0.703 (4.49) & 0.705 (4.68) & 0.659 (2.27) & 0.696 (1.76) & 0.693 (1.85) & 0.699 (3.28) & 0.694 (2.84) \\
Maintenance  & 1.000 (0.0) & 0.976 (2.87) & 0.942 (1.69) & 1.000 (0.0) & 0.992 (0.47) & 0.989 (0.29) & 0.982 (0.47) & 1.000 (0.0) & 1.000 (0.01) \\
Uissurv      & 0.845 (2.16) & 0.838 (2.56) & 0.810 (1.85) & 0.816 (1.92) & 0.822 (2.77) & 0.824 (2.18) & 0.796 (2.9) & 0.843 (1.97) & 0.828 (1.93) \\
Unempdur     & 0.727 (1.19) & 0.731 (1.36) & 0.728 (1.38) & 0.726 (1.44) & 0.703 (1.2) & 0.699 (1.2) & 0.723 (1.24) & 0.722 (1.12) & 0.722 (1.21) \\
Veterans     & 0.742 (6.41) & 0.758 (7.55) & 0.744 (7.78) & 0.714 (9.09) & 0.613 (4.17) & 0.684 (4.1) & 0.654 (5.59) & 0.739 (5.9) & 0.735 (8.16) \\
Whas500      & 0.800 (5.02) & 0.796 (5.17) & 0.793 (5.47) & 0.788 (5.95) & 0.745 (3.71) & 0.748 (4.02) & 0.741 (6.36) & 0.773 (4.71) & 0.779 (5.0) \\\hline
AVERAGE & \textbf{0.802} & \textbf{0.802} & \textbf{0.789} & \textbf{0.796} & \textbf{0.744} & \textbf{0.76} & \textbf{0.762} & \textbf{0.79} & \textbf{0.783}
\\\hline
\end{tabular}
}
\end{table}
\vspace{-15pt}

\begin{table}[H]
\centering
\resizebox{0.85\textwidth}{!}{
\begin{tabular}{lccccccccc}
\cline{2-10}
             & \multicolumn{9}{c}{Testing $C_H$} \\
\cline{2-10}
             & \multicolumn{4}{c}{D=1} & \multicolumn{3}{c}{D=5} & \multicolumn{2}{c}{D=1} \\\hline
Dataset      & \multicolumn{1}{l}{\textbf{Llog}} & \multicolumn{1}{l}{\textbf{Llog-init}} & \multicolumn{1}{l}{\textbf{Exp}} & \multicolumn{1}{l}{\textbf{W}} & \multicolumn{1}{l}{\textbf{SkSurv}} & \multicolumn{1}{l}{\textbf{RPART}} & \multicolumn{1}{l}{\textbf{CTree}} & \multicolumn{1}{l}{\textbf{PO}} & \multicolumn{1}{l}{\textbf{PH}} \\\hline
Aids         & 0.757 (4.9) & 0.760 (4.08) & 0.757 (5.3) & 0.756 (4.76) & 0.621 (9.16) & 0.727 (9.52) & 0.706 (9.41) & 0.733 (2.49) & 0.736 (2.52) \\
Aids\_death  & 0.787 (6.7) & 0.822 (4.39) & 0.771 (6.98) & 0.789 (6.82) & 0.545 (10.9) & 0.624 (14.6) & 0.645 (8.4) & 0.702 (12.12) & 0.708 (11.88) \\
Aids2        & 0.556 (0.92) & 0.559 (0.95) & 0.556 (0.52) & 0.554 (0.99) & 0.540 (1.42) & 0.518 (1.64) & 0.537 (1.01) & 0.557 (1.51) & 0.547 (1.88) \\
Churn        & 0.841 (0.69) & 0.841 (0.98) & 0.832 (0.73) & 0.831 (0.61) & 0.778 (1.77) & 0.779 (0.88) & 0.781 (0.69) & 0.840 (0.93) & 0.835 (0.97) \\
Credit\_risk & 0.758 (2.2) & 0.758 (2.36) & 0.721 (2.16) & 0.749 (2.22) & 0.718 (2.84) & 0.723 (0.67) & 0.726 (0.77) & 0.755 (2.26) & 0.748 (2.31) \\
Dialysis     & 0.738 (1.41) & 0.732 (2.82) & 0.736 (1.41) & 0.737 (1.44) & 0.642 (2.37) & 0.609 (0.92) & 0.642 (1.36) & 0.723 (1.3) & 0.727 (1.37) \\
Employee     & 0.822 (0.83) & 0.825 (0.77) & 0.777 (1.35) & 0.806 (1.8) & 0.912 (0.88) & 0.912 (0.92) & 0.903 (0.64) & 0.824 (1.31) & 0.822 (1.13) \\
Flchain      & 0.934 (0.36) & 0.934 (0.36) & 0.932 (0.33) & 0.933 (0.36) & 0.848 (1.62) & 0.876 (0.2) & 0.880 (0.54) & 0.933 (0.28) & 0.854 (16.62) \\
Framingham   & 0.711 (1.36) & 0.711 (1.33) & 0.710 (1.41) & 0.710 (1.42) & 0.666 (1.68) & 0.658 (1.67) & 0.676 (1.26) & 0.706 (1.75) & 0.709 (1.63) \\
Gbsg2        & 0.667 (1.9) & 0.665 (2.28) & 0.659 (2.11) & 0.662 (2.13) & 0.626 (3.14) & 0.643 (1.99) & 0.633 (2.63) & 0.659 (2.12) & 0.657 (2.05) \\
Maintenance  & 0.944 (1.0) & 0.906 (4.02) & 0.883 (1.01) & 0.945 (1.03) & 0.934 (1.57) & 0.922 (1.37) & 0.901 (1.34) & 0.944 (0.98) & 0.944 (0.97) \\
Uissurv      & 0.743 (1.85) & 0.738 (2.34) & 0.715 (1.63) & 0.719 (1.81) & 0.717 (2.73) & 0.721 (2.04) & 0.700 (2.41) & 0.740 (1.65) & 0.728 (1.73) \\
Unempdur     & 0.678 (1.46) & 0.683 (1.47) & 0.678 (1.56) & 0.677 (1.62) & 0.667 (1.11) & 0.659 (1.43) & 0.676 (1.31) & 0.679 (1.3) & 0.680 (1.36) \\
Veterans     & 0.610 (8.02) & 0.620 (7.68) & 0.626 (8.29) & 0.608 (9.32) & 0.549 (2.88) & 0.575 (5.57) & 0.575 (6.01) & 0.654 (3.77) & 0.659 (6.24) \\
Whas500      & 0.735 (2.75) & 0.733 (2.68) & 0.731 (2.81) & 0.725 (3.44) & 0.674 (3.21) & 0.705 (2.36) & 0.688 (3.41) & 0.721 (3.27) & 0.730 (3.54) \\\hline
AVERAGE      & \textbf{0.752} & \textbf{0.753} & \textbf{0.739} & \textbf{0.747} & \textbf{0.696} & \textbf{0.710} & \textbf{0.711} & \textbf{0.745} & \textbf{0.739} \\\hline

\end{tabular}
}
\caption{Testing results using the CD-AUC and $C_H$ measures. The comparison includes SST models with depth $D=1$ with parametric distributions (Exp, W, Llog) and spline-based semiparametric \textcolor{black}{survival functions} (PO and PH), as well as the three benchmark survival tree models (SkSurv, CTree, and RPART) with depth $D=5$. Llog-init refers to Llog with the clustering-based initialization procedure. In brackets the standard deviation divided by a factor of $1e^{-2}$ for visualization purposes.}
\end{table}
\end{landscape}

\begin{landscape}
\noindent \textbf{Depth $D=1$}
\vspace{-20pt}
\begin{table}[H]
\centering
\resizebox{0.85\textwidth}{!}{
\begin{tabular}{lccccccccc}
\cline{2-10}
             & \multicolumn{9}{c}{Testing IBS} \\
\cline{2-10}
             & \multicolumn{4}{c}{D=1} & \multicolumn{3}{c}{D=5} & \multicolumn{2}{c}{D=1} \\\hline
Dataset      & \multicolumn{1}{l}{\textbf{Llog}} & \multicolumn{1}{l}{\textbf{Llog-init}} & \multicolumn{1}{l}{\textbf{Exp}} & \multicolumn{1}{l}{\textbf{W}} & \multicolumn{1}{l}{\textbf{SkSurv}} & \multicolumn{1}{l}{\textbf{RPART}} & \multicolumn{1}{l}{\textbf{CTree}} & \multicolumn{1}{l}{\textbf{PO}} & \multicolumn{1}{l}{\textbf{PH}} \\\hline
Aids    & 0.057 (1.86) & 0.056 (1.81) & 0.057 (1.9) & 0.057 (1.86) & 0.069 (1.79) & 0.068 (2.69) & 0.059 (1.94) & 0.028 (0.78) & 0.028 (0.8) \\
Aids\_death  & 0.016 (0.44) & 0.016 (0.44) & 0.016 (0.43) & 0.016 (0.44) & 0.020 (0.56) & 0.020 (0.7) & 0.016 (0.43) & 0.007 (0.2) & 0.007 (0.17) \\
Aids2        & 0.158 (1.57) & 0.160 (1.68) & 0.141 (2.29) & 0.145 (1.97) & 0.141 (2.26) & 0.141 (2.39) & 0.140 (2.27) & 0.033 (0.25) & 0.033 (0.3) \\
Churn    & 0.083 (1.12) & 0.065 (1.58) & 0.101 (1.31) & 0.097 (1.25) & 0.103 (1.69) & 0.101 (2.06) & 0.101 (2.05) & 0.009 (0.09) & 0.009 (0.1) \\
Credit\_risk  & 0.106 (1.52) & 0.106 (1.38) & 0.127 (1.34) & 0.106 (1.39) & 0.122 (0.6) & 0.113 (0.81) & 0.114 (1.03) & 0.103 (1.26) & 0.105 (1.23) \\
Dialysis     & 0.145 (0.77) & 0.149 (1.05) & 0.145 (0.79) & 0.144 (0.79) & 0.175 (0.84) & 0.183 (0.93) & 0.176 (0.77) & 0.111 (0.57) & 0.110 (0.58) \\
Employee & 0.160 (1.6) & 0.156 (1.72) & 0.234 (1.86) & 0.165 (1.4) & 0.055 (0.31) & 0.057 (0.18) & 0.065 (0.36) & 0.143 (0.88) & 0.145 (1.21) \\
Flchain & 0.044 (0.15) & 0.044 (0.15) & 0.046 (0.15) & 0.043 (0.15) & 0.062 (0.28) & 0.064 (0.51) & 0.060 (0.17) & 0.012 (0.73) & 0.097 (22.89) \\
Framingham & 0.111 (0.31) & 0.111 (0.29) & 0.113 (0.28) & 0.111 (0.31) & 0.119 (0.46) & 0.118 (0.32) & 0.117 (0.33) & 0.045 (0.79) & 0.045 (0.79) \\
Gbsg2    & 0.169 (1.54) & 0.170 (1.46) & 0.174 (1.41) & 0.171 (1.49) & 0.189 (2.24) & 0.180 (1.1) & 0.173 (1.25) & 0.085 (1.88) & 0.086 (1.89) \\
Maintenance  & 0.017 (0.27) & 0.022 (0.34) & 0.097 (0.68) & 0.017 (0.24) & 0.001 (0.06) & 0.003 (0.17) & 0.007 (0.13) & 0.001 (0.02) & 0.001 (0.02) \\
Uissurv     & 0.142 (1.69) & 0.144 (1.61) & 0.147 (1.82) & 0.146 (1.81) & 0.146 (2.41) & 0.144 (1.67) & 0.146 (1.71) & 0.095 (0.65) & 0.100 (0.64) \\
Unempdur     & 0.163 (1.28) & 0.163 (1.27) & 0.164 (1.22) & 0.164 (1.26) & 0.170 (1.1) & 0.164 (1.21) & 0.161 (1.07) & 0.168 (0.8) & 0.169 (0.77) \\
Veterans & 0.119 (3.48) & 0.119 (3.62) & 0.120 (3.89) & 0.122 (4.27) & 0.160 (5.39) & 0.134 (4.33) & 0.134 (3.39) & 0.121 (0.98) & 0.120 (0.98) \\
Whas500      & 0.168 (2.06) & 0.169 (2.17) & 0.171 (2.23) & 0.172 (2.37) & 0.206 (2.7) & 0.190 (2.52) & 0.177 (1.77) & 0.109 (0.73) & 0.108 (0.73) \\\hline
AVERAGE      & \textbf{0.111} & \textbf{0.110} & \textbf{0.124} & \textbf{0.112} & \textbf{0.116} & \textbf{0.112} & \textbf{0.110} & \textbf{0.071} & \textbf{0.078} \\\hline
\end{tabular}
}
\end{table}
\vspace{-15pt}

\begin{table}[H]
\centering
\resizebox{0.85\textwidth}{!}{
\begin{tabular}{lccccccccc}
\cline{2-10}
             & \multicolumn{9}{c}{Testing $C_U$}                                                                                                                                                                                                                                                                                                            \\
\cline{2-10}
             & \multicolumn{4}{c}{D=1} & \multicolumn{3}{c}{D=5} & \multicolumn{2}{c}{D=1} \\\hline 
Dataset      & \multicolumn{1}{l}{\textbf{Llog}} & \multicolumn{1}{l}{\textbf{Llog-init}} & \multicolumn{1}{l}{\textbf{Exp}} & \multicolumn{1}{l}{\textbf{W}} & \multicolumn{1}{l}{\textbf{SkSurv}} & \multicolumn{1}{l}{\textbf{RPART}} & \multicolumn{1}{l}{\textbf{CTree}} & \multicolumn{1}{l}{\textbf{PO}} & \multicolumn{1}{l}{\textbf{PH}} \\\hline
Aids         & 0.755 (4.62) & 0.758 (4.09) & 0.753 (4.9) & 0.754 (4.52) & 0.624 (8.55) & 0.719 (9.57) & 0.697 (9.47) & 0.724 (2.23) & 0.726 (2.28) \\
Aids\_death  & 0.774 (7.69) & 0.811 (5.8) & 0.756 (7.39) & 0.775 (7.76) & 0.520 (9.01) & 0.609 (13.65) & 0.632 (7.65) & 0.668 (12.7) & 0.677 (12.74) \\
Aids2        & 0.541 (0.96) & 0.544 (1.04) & 0.543 (0.63) & 0.541 (0.92) & 0.528 (1.14) & 0.511 (1.01) & 0.528 (0.65) & 0.542 (1.09) & 0.534 (1.72) \\
Churn        & 0.742 (14.15) & 0.742 (15.76) & 0.731 (12.82) & 0.737 (13.02) & 0.703 (5.43) & 0.707 (5.55) & 0.692 (6.3) & 0.737 (16.04) & 0.713 (14.73) \\
Credit\_risk & 0.736 (2.06) & 0.738 (2.06) & 0.709 (1.8) & 0.731 (2.03) & 0.699 (3.01) & 0.702 (1.12) & 0.704 (1.92) & 0.736 (2.11) & 0.731 (2.06) \\
Dialysis     & 0.720 (1.79) & 0.715 (2.19) & 0.718 (1.79) & 0.719 (1.61) & 0.651 (2.44) & 0.634 (2.44) & 0.656 (1.98) & 0.701 (1.63) & 0.706 (1.61) \\
Employee     & 0.712 (2.29) & 0.710 (2.5) & 0.714 (2.44) & 0.726 (2.6) & 0.896 (0.54) & 0.894 (0.56) & 0.886 (0.61) & 0.692 (2.68) & 0.708 (3.31) \\
Flchain      & 0.941 (0.37) & 0.941 (0.37) & 0.939 (0.34) & 0.940 (0.36) & 0.852 (1.96) & 0.880 (0.3) & 0.884 (0.47) & 0.940 (0.29) & 0.860 (16.91) \\
Framingham   & 0.697 (1.73) & 0.696 (1.68) & 0.696 (1.74) & 0.696 (1.76) & 0.655 (1.93) & 0.643 (1.75) & 0.664 (1.62) & 0.691 (1.81) & 0.694 (1.69) \\
Gbsg2        & 0.648 (4.13) & 0.644 (4.41) & 0.645 (4.21) & 0.646 (4.31) & 0.614 (3.51) & 0.631 (1.44) & 0.619 (4.08) & 0.642 (2.66) & 0.640 (2.38) \\
Maintenance  & 0.935 (1.05) & 0.889 (4.92) & 0.851 (1.44) & 0.936 (1.08) & 0.924 (1.47) & 0.909 (1.39) & 0.893 (1.28) & 0.935 (1.01) & 0.935 (1.01) \\
Uissurv      & 0.743 (1.83) & 0.738 (2.32) & 0.714 (1.69) & 0.718 (1.8) & 0.718 (2.68) & 0.721 (1.97) & 0.700 (2.37) & 0.740 (1.6) & 0.727 (1.66) \\
Unempdur     & 0.646 (1.57) & 0.650 (1.56) & 0.647 (1.68) & 0.645 (1.67) & 0.628 (1.42) & 0.624 (1.66) & 0.639 (1.52) & 0.648 (1.74) & 0.645 (1.7) \\
Veterans     & 0.596 (7.87) & 0.607 (7.54) & 0.615 (8.11) & 0.597 (8.91) & 0.540 (2.65) & 0.566 (5.36) & 0.563 (5.69) & 0.642 (3.66) & 0.648 (6.18) \\
Whas500      & 0.727 (3.38) & 0.726 (3.17) & 0.724 (3.43) & 0.718 (4.1) & 0.678 (2.89) & 0.707 (2.76) & 0.692 (3.1) & 0.703 (4.14) & 0.710 (4.25) \\\hline
AVERAGE      & \textbf{0.727} & \textbf{0.727} & \textbf{0.717} & \textbf{0.725} & \textbf{0.682} & \textbf{0.697} & \textbf{0.697} & \textbf{0.716} & \textbf{0.710} \\\hline
\end{tabular}
}
\caption{Testing results using the IBS and $C_U$ measures. The comparison includes SST models with depth $D=1$ with parametric distributions (Exp, W, Llog) and spline-based semiparametric \textcolor{black}{survival functions} (PO and PH), as well as the three benchmark survival tree models (SkSurv, CTree, and RPART) with depth $D=5$. Llog-init refers to Llog with the clustering-based initialization procedure. In brackets the standard deviation divided by a factor of $1e^{-2}$ for visualization purposes.}
\end{table}
\end{landscape}

\vspace{-100pt}
\begin{landscape}
\noindent \textbf{Depth $D=1$}
\vspace{-25pt}
\begin{table}[H]
\centering
\resizebox{0.85\textwidth}{!}{
\begin{tabular}{lccccccccc}
\cline{2-10}
             & \multicolumn{9}{c}{Training CD-AUC}                                                                                                                                                                                                                                                                                                            \\
\cline{2-10}
             & \multicolumn{4}{c}{D=1} & \multicolumn{3}{c}{D=5} & \multicolumn{2}{c}{D=1} \\\hline 
Dataset      & \multicolumn{1}{l}{\textbf{Llog}} & \multicolumn{1}{l}{\textbf{Llog-init}} & \multicolumn{1}{l}{\textbf{Exp}} & \multicolumn{1}{l}{\textbf{W}} & \multicolumn{1}{l}{\textbf{SkSurv}} & \multicolumn{1}{l}{\textbf{RPART}} & \multicolumn{1}{l}{\textbf{CTree}} & \multicolumn{1}{l}{\textbf{PO}} & \multicolumn{1}{l}{\textbf{PH}} \\\hline
Aids         & 0.772 (2.16) & 0.765 (1.65) & 0.768 (2.31) & 0.772 (2.13) & 0.894 (2.02) & 0.820 (1.19) & 0.757 (1.98) & 0.788 (1.24) & 0.788 (1.28) \\
Aids\_death  & 0.895 (1.23) & 0.889 (0.94) & 0.869 (1.32) & 0.894 (1.24) & 0.934 (7.71) & 0.933 (2.21) & 0.738 (8.4) & 0.860 (2.74) & 0.850 (2.83) \\
Aids2        & 0.580 (0.78) & 0.577 (1.36) & 0.581 (0.66) & 0.581 (0.68) & 0.585 (0.62) & 0.525 (2.07) & 0.549 (0.32) & 0.570 (0.74) & 0.549 (2.77) \\
Churn        & 0.929 (0.28) & 0.930 (0.36) & 0.920 (0.45) & 0.919 (0.26) & 0.891 (0.38) & 0.865 (0.46) & 0.878 (0.29) & 0.928 (0.29) & 0.922 (0.47) \\
Credit\_risk & 0.894 (0.59) & 0.893 (0.69) & 0.853 (0.67) & 0.894 (0.45) & 0.887 (1.05) & 0.857 (1.42) & 0.850 (0.64) & 0.892 (0.41) & 0.890 (0.36) \\
Dialysis     & 0.814 (0.46) & 0.806 (2.56) & 0.811 (0.46) & 0.814 (0.45) & 0.701 (0.8) & 0.638 (1.24) & 0.697 (0.98) & 0.795 (0.26) & 0.796 (0.26) \\
Employee     & 0.792 (1.14) & 0.787 (0.93) & 0.779 (1.61) & 0.800 (1.85) & 0.956 (0.11) & 0.947 (0.28) & 0.940 (0.15) & 0.776 (0.73) & 0.787 (1.56) \\
Flchain      & 0.956 (0.14) & 0.956 (0.13) & 0.954 (0.15) & 0.955 (0.15) & 0.889 (0.98) & 0.905 (0.35) & 0.915 (0.11) & 0.956 (0.09) & 0.872 (17.44) \\
Framingham   & 0.780 (0.31) & 0.779 (0.29) & 0.778 (0.29) & 0.779 (0.3) & 0.773 (0.41) & 0.727 (0.92) & 0.765 (0.36) & 0.771 (0.32) & 0.775 (0.22) \\
Gbsg2        & 0.762 (0.66) & 0.759 (0.87) & 0.746 (0.59) & 0.754 (0.66) & 0.810 (0.7) & 0.759 (1.34) & 0.714 (1.56) & 0.737 (1.09) & 0.732 (0.64) \\
Maintenance  & 1.000 (0.0) & 0.980 (2.01) & 0.943 (0.88) & 1.000 (0.0) & 1.000 (0.01) & 0.991 (0.17) & 0.983 (0.29) & 1.000 (0.01) & 1.000 (0.01) \\
Uissurv      & 0.855 (0.45) & 0.853 (1.12) & 0.833 (0.94) & 0.837 (0.89) & 0.868 (0.5) & 0.847 (0.44) & 0.804 (1.0) & 0.855 (0.49) & 0.845 (0.59) \\
Unempdur     & 0.729 (0.4) & 0.731 (0.31) & 0.730 (0.38) & 0.729 (0.4) & 0.758 (0.87) & 0.699 (0.29) & 0.726 (0.37) & 0.725 (0.34) & 0.727 (0.54) \\
Veterans     & 0.842 (2.27) & 0.829 (2.12) & 0.813 (2.18) & 0.831 (2.75) & 0.901 (1.52) & 0.857 (1.92) & 0.729 (5.66) & 0.855 (1.96) & 0.838 (2.07) \\
Whas500      & 0.849 (1.43) & 0.850 (1.38) & 0.854 (1.38) & 0.835 (2.02) & 0.869 (3.63) & 0.865 (2.19) & 0.820 (1.99) & 0.832 (2.35) & 0.840 (2.25) \\\hline
AVERAGE      & \textbf{0.830} & \textbf{0.826} & \textbf{0.815} & \textbf{0.826} & \textbf{0.848} & \textbf{0.816} & \textbf{0.791} & \textbf{0.823} & \textbf{0.814} \\\hline
\end{tabular}
}
\end{table}
\vspace{-20pt}

\begin{table}[H]
\centering
\resizebox{0.85\textwidth}{!}{
\begin{tabular}{lccccccccc}
\cline{2-10}
             & \multicolumn{9}{c}{Training $C_H$}                                                                                                                                                                                                                                                                                                            \\
\cline{2-10}
             & \multicolumn{4}{c}{D=1} & \multicolumn{3}{c}{D=5} & \multicolumn{2}{c}{D=1} \\\hline 
Dataset      & \multicolumn{1}{l}{\textbf{Llog}} & \multicolumn{1}{l}{\textbf{Llog-init}} & \multicolumn{1}{l}{\textbf{Exp}} & \multicolumn{1}{l}{\textbf{W}} & \multicolumn{1}{l}{\textbf{SkSurv}} & \multicolumn{1}{l}{\textbf{RPART}} & \multicolumn{1}{l}{\textbf{CTree}} & \multicolumn{1}{l}{\textbf{PO}} & \multicolumn{1}{l}{\textbf{PH}} \\\hline
Aids         & 0.780 (1.03) & 0.773 (0.73) & 0.777 (1.13) & 0.780 (1.01) & 0.888 (2.04) & 0.812 (0.98) & 0.751 (2.22) & 0.787 (0.96) & 0.787 (1.03) \\
Aids\_death  & 0.879 (1.42) & 0.874 (1.21) & 0.850 (1.44) & 0.879 (1.45) & 0.944 (6.54) & 0.937 (1.74) & 0.776 (8.29) & 0.847 (2.64) & 0.836 (2.65) \\
Aids2        & 0.564 (0.52) & 0.567 (1.2) & 0.566 (0.32) & 0.562 (0.6) & 0.568 (0.75) & 0.521 (1.71) & 0.540 (0.19) & 0.569 (0.64) & 0.550 (2.01) \\
Churn        & 0.848 (0.19) & 0.851 (0.32) & 0.839 (0.43) & 0.838 (0.23) & 0.823 (0.35) & 0.791 (0.47) & 0.805 (0.29) & 0.848 (0.27) & 0.842 (0.5) \\
Credit\_risk & 0.802 (0.61) & 0.801 (0.84) & 0.764 (0.55) & 0.794 (0.55) & 0.794 (1.71) & 0.759 (1.14) & 0.759 (0.97) & 0.794 (0.51) & 0.787 (0.43) \\
Dialysis     & 0.757 (0.4) & 0.752 (2.61) & 0.752 (0.37) & 0.755 (0.38) & 0.678 (0.63) & 0.619 (0.99) & 0.671 (0.75) & 0.750 (0.28) & 0.751 (0.28) \\
Employee     & 0.828 (0.66) & 0.829 (0.44) & 0.781 (1.3) & 0.812 (1.63) & 0.930 (0.15) & 0.920 (0.32) & 0.917 (0.1) & 0.827 (0.39) & 0.825 (0.71) \\
Flchain      & 0.000 (0.1) & 0.000 (0.09) & 0.000 (0.12) & 0.000 (0.12) & 0.000 (0.97) & 0.000 (0.46) & 0.000 (0.05) & 0.000 (0.06) & 0.000 (18.83) \\
Framingham   & 0.718 (0.33) & 0.717 (0.31) & 0.717 (0.31) & 0.716 (0.32) & 0.720 (0.35) & 0.674 (0.73) & 0.712 (0.4) & 0.710 (0.32) & 0.713 (0.3) \\
Gbsg2        & 0.695 (0.44) & 0.695 (0.42) & 0.683 (0.53) & 0.688 (0.53) & 0.765 (0.73) & 0.707 (1.31) & 0.665 (1.33) & 0.677 (0.85) & 0.674 (0.69) \\
Maintenance  & 0.953 (0.27) & 0.919 (3.31) & 0.891 (0.77) & 0.953 (0.26) & 0.952 (0.2) & 0.933 (0.55) & 0.910 (0.46) & 0.953 (0.26) & 0.953 (0.26) \\
Uissurv      & 0.757 (0.5) & 0.754 (1.23) & 0.736 (0.74) & 0.737 (0.8) & 0.771 (0.7) & 0.744 (0.57) & 0.713 (0.84) & 0.755 (0.53) & 0.744 (0.71) \\
Unempdur     & 0.682 (0.42) & 0.687 (0.37) & 0.682 (0.38) & 0.681 (0.41) & 0.708 (0.5) & 0.661 (0.35) & 0.683 (0.46) & 0.681 (0.34) & 0.683 (0.62) \\
Veterans     & 0.720 (2.11) & 0.715 (2.18) & 0.709 (1.93) & 0.711 (2.26) & 0.788 (3.37) & 0.750 (3.5) & 0.647 (4.35) & 0.764 (1.59) & 0.750 (1.92) \\
Whas500      & 0.741 (0.78) & 0.741 (0.8) & 0.744 (0.76) & 0.736 (0.83) & 0.801 (1.52) & 0.772 (0.84) & 0.735 (0.98) & 0.764 (0.93) & 0.765 (1.04) \\\hline
AVERAGE      & \textbf{0.715} & \textbf{0.712} & \textbf{0.699} & \textbf{0.710} & \textbf{0.742} & \textbf{0.707} & \textbf{0.686} & \textbf{0.715} & \textbf{0.711} \\\hline
\end{tabular}
}
\caption{Training results using the CD-AUC and $C_H$ measures. The comparison includes SST models with depth $D=1$ with parametric distributions (Exp, W, Llog) and spline-based semiparametric \textcolor{black}{survival functions} (PO and PH), as well as the three benchmark survival tree models (SkSurv, CTree, and RPART) with depth $D=5$. Llog-init refers to Llog with the clustering-based initialization procedure. In brackets the standard deviation divided by a factor of $1e^{-2}$ for visualization purposes.}
\end{table}

\end{landscape}

\begin{landscape}
\noindent \textbf{Depth $D=1$}
\vspace{-20pt}
\begin{table}[H]
\centering
\resizebox{0.85\textwidth}{!}{
\begin{tabular}{lccccccccc}
\cline{2-10}
             & \multicolumn{9}{c}{Training $C_U$}                                                                                                                                                                                                                                                                                                            \\
\cline{2-10}
             & \multicolumn{4}{c}{D=1} & \multicolumn{3}{c}{D=5} & \multicolumn{2}{c}{D=1} \\\hline 
Dataset      & \multicolumn{1}{l}{\textbf{Llog}} & \multicolumn{1}{l}{\textbf{Llog-init}} & \multicolumn{1}{l}{\textbf{Exp}} & \multicolumn{1}{l}{\textbf{W}} & \multicolumn{1}{l}{\textbf{SkSurv}} & \multicolumn{1}{l}{\textbf{RPART}} & \multicolumn{1}{l}{\textbf{CTree}} & \multicolumn{1}{l}{\textbf{PO}} & \multicolumn{1}{l}{\textbf{PH}} \\\hline
Aids         & 0.775 (1.06) & 0.769 (1.04) & 0.772 (1.1) & 0.775 (1.11) & 0.874 (2.56) & 0.797 (0.76) & 0.739 (2.76) & 0.779 (1.33) & 0.779 (1.42) \\
Aids\_death  & 0.878 (1.99) & 0.871 (1.96) & 0.848 (2.03) & 0.879 (2.02) & 0.937 (6.86) & 0.935 (1.45) & 0.767 (8.43) & 0.841 (2.86) & 0.830 (2.91) \\
Aids2        & 0.547 (0.5) & 0.549 (0.86) & 0.553 (0.36) & 0.548 (0.59) & 0.558 (0.8) & 0.516 (1.28) & 0.531 (0.2) & 0.550 (0.44) & 0.535 (1.79) \\
Churn        & 0.792 (1.89) & 0.790 (1.16) & 0.793 (1.79) & 0.791 (2.14) & 0.742 (1.37) & 0.731 (1.86) & 0.744 (2.59) & 0.783 (1.65) & 0.784 (1.48) \\
Credit\_risk & 0.782 (0.55) & 0.781 (0.81) & 0.752 (0.51) & 0.778 (0.47) & 0.778 (1.27) & 0.739 (1.0) & 0.742 (0.67) & 0.776 (0.51) & 0.772 (0.41) \\
Dialysis     & 0.740 (0.37) & 0.736 (1.66) & 0.739 (0.4) & 0.741 (0.34) & 0.679 (1.2) & 0.636 (1.39) & 0.677 (1.0) & 0.729 (0.4) & 0.734 (0.4) \\
Employee     & 0.717 (1.3) & 0.714 (1.07) & 0.719 (1.84) & 0.732 (2.17) & 0.909 (0.23) & 0.901 (0.32) & 0.895 (0.21) & 0.698 (0.79) & 0.712 (2.08) \\
Flchain      & 0.000 (0.11) & 0.000 (0.11) & 0.000 (0.12) & 0.000 (0.13) & 0.000 (0.91) & 0.000 (0.44) & 0.000 (0.07) & 0.000 (0.07) & 0.000 (19.15) \\
Framingham   & 0.704 (0.43) & 0.703 (0.4) & 0.703 (0.4) & 0.702 (0.41) & 0.705 (0.33) & 0.660 (0.82) & 0.699 (0.53) & 0.695 (0.31) & 0.698 (0.31) \\
Gbsg2        & 0.682 (0.58) & 0.679 (0.66) & 0.673 (0.65) & 0.678 (0.68) & 0.745 (1.61) & 0.695 (1.12) & 0.647 (1.97) & 0.661 (0.88) & 0.657 (0.63) \\
Maintenance  & 0.945 (0.26) & 0.903 (4.17) & 0.857 (0.96) & 0.945 (0.26) & 0.944 (0.21) & 0.921 (0.51) & 0.903 (0.53) & 0.945 (0.24) & 0.945 (0.24) \\
Uissurv      & 0.756 (0.53) & 0.753 (1.25) & 0.735 (0.76) & 0.737 (0.83) & 0.770 (0.68) & 0.743 (0.51) & 0.712 (0.82) & 0.754 (0.55) & 0.743 (0.74) \\
Unempdur     & 0.649 (0.6) & 0.654 (0.47) & 0.650 (0.58) & 0.649 (0.59) & 0.679 (0.61) & 0.626 (0.55) & 0.649 (0.63) & 0.650 (0.61) & 0.651 (0.59) \\
Veterans     & 0.714 (1.88) & 0.709 (1.98) & 0.704 (1.73) & 0.707 (2.07) & 0.784 (3.41) & 0.745 (3.27) & 0.642 (4.18) & 0.756 (1.57) & 0.743 (1.82) \\
Whas500      & 0.745 (0.64) & 0.745 (0.51) & 0.748 (0.56) & 0.740 (0.87) & 0.774 (2.41) & 0.746 (2.33) & 0.731 (0.98) & 0.730 (2.58) & 0.740 (2.27) \\\hline
AVERAGE      & \textbf{0.695} & \textbf{0.690} & \textbf{0.683} & \textbf{0.694} & \textbf{0.725} & \textbf{0.693} & \textbf{0.672} & \textbf{0.690} & \textbf{0.688} \\\hline
\end{tabular}
}
%\caption{Training results using the $C_U$ measure. The comparison includes SST models with depth $D=1$ with parametric distributions (Exp, W, Llog) and splines-based semiparametric approaches (PO and PH), as well as the three benchmark survival tree models (SkSurv, CTree, and RPART) with depth $D=5$. Llog clust refers to Llog with the clustering-based initialization procedure. In brackets the standard deviation divided by a factor of $1e^{-2}$ for visualization purposes.}
\end{table}
\vspace{-15pt}

\noindent \textbf{Depth $D=2$}
\vspace{-20pt}
\begin{table}[H]
\centering
\resizebox{0.85\textwidth}{!}{
\begin{tabular}{lccccccccc}
\cline{2-10}
             & \multicolumn{9}{c}{Training $C_U$} \\
\cline{2-10}
             & \multicolumn{4}{c}{D=2} & \multicolumn{3}{c}{D=5} & \multicolumn{2}{c}{D=2} \\\hline
Dataset      & \multicolumn{1}{l}{\textbf{Llog}} & \multicolumn{1}{l}{\textbf{Llog-init}} & \multicolumn{1}{l}{\textbf{Exp}} & \multicolumn{1}{l}{\textbf{W}} & \multicolumn{1}{l}{\textbf{SkSurv}} & \multicolumn{1}{l}{\textbf{RPART}} & \multicolumn{1}{l}{\textbf{CTree}} & \multicolumn{1}{l}{\textbf{PO}} & \multicolumn{1}{l}{\textbf{PH}} \\\hline
Aids         & 0.786 (1.26) & 0.775 (1.11) & 0.776 (1.22) & 0.783 (1.29) & 0.874 (2.56) & 0.797 (0.76) & 0.739 (2.76) & 0.765 (8.54) & 0.771 (6.76) \\
Aids\_death  & 0.885 (2.19) & 0.875 (2.06) & 0.843 (2.18) & 0.884 (2.13) & 0.937 (6.86) & 0.935 (1.45) & 0.767 (8.43) & 0.823 (9.52) & 0.814 (9.71) \\
Aids2        & 0.547 (0.55) & 0.549 (0.4) & 0.555 (0.39) & 0.551 (0.54) & 0.558 (0.8) & 0.516 (1.28) & 0.531 (0.2) & 0.554 (0.36) & 0.54 (1.75) \\
Churn        & 0.798 (1.71) & 0.795 (1.43) & 0.793 (1.66) & 0.792 (1.85) & 0.742 (1.37) & 0.731 (1.86) & 0.744 (2.59) & 0.794 (2.16) & 0.788 (1.89) \\
Credit\_risk & 0.793 (0.85) & 0.798 (0.91) & 0.758 (0.63) & 0.789 (0.74) & 0.778 (1.27) & 0.739 (1.0) & 0.742 (0.67) & 0.787 (1.58) & 0.787 (0.79) \\
Dialysis     & 0.744 (0.48) & 0.745 (0.51) & 0.741 (0.44) & 0.745 (0.43) & 0.679 (1.2) & 0.636 (1.39) & 0.677 (1.0) & 0.719 (4.0) & 0.731 (0.52) \\
Employee     & 0.736 (1.98) & 0.845 (4.11) & 0.746 (2.75) & 0.756 (3.77) & 0.909 (0.23) & 0.901 (0.32) & 0.895 (0.21) & 0.722 (2.58) & 0.828 (6.88) \\
Flchain      & 0.942 (0.11) & 0.942 (0.11) & 0.941 (0.13) & 0.942 (0.14) & 0.871 (0.91) & 0.886 (0.44) & 0.894 (0.07) & 0.869 (16.71) & 0.83 (19.11) \\
Framingham   & 0.705 (0.44) & 0.704 (0.4) & 0.704 (0.39) & 0.703 (0.43) & 0.705 (0.33) & 0.66 (0.82) & 0.699 (0.53) & 0.69 (1.25) & 0.695 (0.87) \\
Gbsg2        & 0.683 (0.7) & 0.685 (0.75) & 0.671 (0.81) & 0.677 (0.76) & 0.745 (1.61) & 0.695 (1.12) & 0.647 (1.97) & 0.683 (1.1) & 0.669 (1.21) \\
Maintenance  & 0.942 (1.22) & 0.937 (2.2) & 0.846 (1.87) & 0.943 (0.31) & 0.944 (0.21) & 0.921 (0.51) & 0.903 (0.53) & 0.945 (0.24) & 0.945 (0.23) \\
Uissurv      & 0.757 (0.67) & 0.758 (0.87) & 0.734 (0.8) & 0.743 (0.88) & 0.77 (0.68) & 0.743 (0.51) & 0.712 (0.82) & 0.762 (0.6) & 0.744 (0.84) \\
Unempdur     & 0.654 (0.64) & 0.658 (0.54) & 0.654 (0.55) & 0.654 (0.59) & 0.679 (0.61) & 0.626 (0.55) & 0.649 (0.63) & 0.65 (0.59) & 0.651 (0.59) \\
Veterans     & 0.72 (2.25) & 0.723 (1.92) & 0.71 (1.78) & 0.721 (2.02) & 0.784 (3.41) & 0.745 (3.27) & 0.642 (4.18) & 0.777 (1.97) & 0.767 (1.94) \\
Whas500      & 0.744 (1.57) & 0.745 (0.66) & 0.751 (1.24) & 0.734 (2.33) & 0.774 (2.41) & 0.746 (2.33) & 0.731 (0.98) & 0.751 (4.37) & 0.745 (3.11) \\\hline
AVERAGE      & \textbf{0.762} & \textbf{0.769} & \textbf{0.748} & \textbf{0.761} & \textbf{0.783} & \textbf{0.752} & \textbf{0.731} & \textbf{0.753} & \textbf{0.754} \\\hline
\end{tabular}
}
\caption{Training results using the $C_U$ measure. The comparison includes SST models with depth $D=1$ and $D=2$ with parametric distributions (Exp, W, Llog) and spline-based semiparametric \textcolor{black}{survival functions} (PO and PH), as well as the three benchmark survival tree models (SkSurv, CTree, and RPART) with depth $D=5$. Llog-init refers to Llog with the clustering-based initialization procedure. In brackets the standard deviation divided by a factor of $1e^{-2}$ for visualization purposes.}
\end{table}
\end{landscape}

%\noindent \textbf{Depth $D=2$}

\begin{landscape}
\noindent \textbf{Depth $D=2$}
\vspace{-5pt}
\begin{table}[H]
\centering
\resizebox{0.85\textwidth}{!}{
\begin{tabular}{lccccccccc}
\cline{2-10}
             & \multicolumn{9}{c}{Training CD-AUC} \\
\cline{2-10}
             & \multicolumn{4}{c}{D=2} & \multicolumn{3}{c}{D=5} & \multicolumn{2}{c}{D=2} \\\hline
Dataset      & \multicolumn{1}{l}{\textbf{Llog}} & \multicolumn{1}{l}{\textbf{Llog-init}} & \multicolumn{1}{l}{\textbf{Exp}} & \multicolumn{1}{l}{\textbf{W}} & \multicolumn{1}{l}{\textbf{SkSurv}} & \multicolumn{1}{l}{\textbf{RPART}} & \multicolumn{1}{l}{\textbf{CTree}} & \multicolumn{1}{l}{\textbf{PO}} & \multicolumn{1}{l}{\textbf{PH}} \\\hline
Aids         & 0.787 (2.22) & 0.772 (1.99) & 0.775 (2.41) & 0.783 (2.18) & 0.894 (2.02) & 0.82 (1.19) & 0.757 (1.98) & 0.772 (9.17) & 0.777 (7.51) \\
Aids\_death  & 0.897 (1.4) & 0.891 (1.08) & 0.85 (2.46) & 0.894 (1.51) & 0.934 (7.71) & 0.933 (2.21) & 0.738 (8.4) & 0.838 (8.94) & 0.829 (9.2) \\
Aids2        & 0.582 (0.77) & 0.582 (0.79) & 0.585 (0.76) & 0.585 (0.82) & 0.585 (0.62) & 0.525 (2.07) & 0.549 (0.32) & 0.58 (0.77) & 0.558 (2.68) \\
Churn        & 0.931 (0.44) & 0.933 (0.45) & 0.92 (0.57) & 0.923 (0.6) & 0.891 (0.38) & 0.865 (0.46) & 0.878 (0.29) & 0.929 (0.43) & 0.925 (0.62) \\
Credit\_risk  & 0.903 (0.76) & 0.908 (0.8) & 0.857 (0.82) & 0.902 (0.65) & 0.887 (1.05) & 0.857 (1.42) & 0.85 (0.64) & 0.901 (1.6) & 0.901 (0.7) \\
Dialysis     & 0.82 (0.69) & 0.82 (0.62) & 0.815 (0.46) & 0.819 (0.6) & 0.701 (0.8) & 0.638 (1.24) & 0.697 (0.98) & 0.784 (4.12) & 0.794 (0.35) \\
Employee     & 0.809 (1.78) & 0.903 (3.47) & 0.804 (2.55) & 0.822 (3.27) & 0.956 (0.11) & 0.947 (0.28) & 0.94 (0.16) & 0.797 (2.17) & 0.885 (8.04) \\
Flchain      & 0.956 (0.14) & 0.956 (0.14) & 0.954 (0.15) & 0.955 (0.17) & 0.889 (0.98) & 0.905 (0.35) & 0.915 (0.11) & 0.89 (16.59) & 0.872 (17.44) \\
Framingham   & 0.781 (0.33) & 0.78 (0.29) & 0.778 (0.31) & 0.779 (0.33) & 0.773 (0.41) & 0.727 (0.92) & 0.765 (0.36) & 0.765 (1.53) & 0.77 (1.04) \\
Gbsg2        & 0.763 (0.83) & 0.765 (0.94) & 0.744 (0.91) & 0.754 (0.89) & 0.81 (0.7) & 0.759 (1.34) & 0.714 (1.56) & 0.762 (1.39) & 0.743 (1.39) \\
Maintenance  & 0.999 (0.58) & 0.997 (1.08) & 0.937 (1.4) & 0.999 (0.07) & 1.0 (0.01) & 0.991 (0.17) & 0.983 (0.29) & 1.0 (0.02) & 1.0 (0.02) \\
Uissurv      & 0.857 (0.6) & 0.856 (0.79) & 0.832 (1.03) & 0.846 (0.88) & 0.868 (0.5) & 0.847 (0.44) & 0.804 (1.0) & 0.863 (0.56) & 0.847 (0.84) \\
Unempdur     & 0.732 (0.51) & 0.735 (0.46) & 0.733 (0.39) & 0.734 (0.46) & 0.758 (0.88) & 0.699 (0.29) & 0.726 (0.37) & 0.725 (0.51) & 0.727 (0.51) \\
Veterans     & 0.856 (2.15) & 0.856 (1.9) & 0.82 (2.5) & 0.854 (2.54) & 0.901 (1.52) & 0.857 (1.92) & 0.729 (5.66) & 0.877 (2.21) & 0.866 (2.17) \\
Whas500      & 0.837 (2.7) & 0.845 (2.09) & 0.851 (1.69) & 0.822 (3.23) & 0.869 (3.63) & 0.865 (2.19) & 0.82 (1.99) & 0.837 (6.19) & 0.826 (3.95) \\\hline
AVERAGE      & \textbf{0.834} & \textbf{0.84} & \textbf{0.817} & \textbf{0.832} & \textbf{0.848} & \textbf{0.816} & \textbf{0.791} & \textbf{0.821} & \textbf{0.821} \\\hline
\end{tabular}
}
\end{table}
\vspace{-15pt}
\begin{table}[H]
\centering
\resizebox{0.85\textwidth}{!}{
\begin{tabular}{lccccccccc}
\cline{2-10}
             & \multicolumn{9}{c}{Training $C_H$} \\
\cline{2-10}
             & \multicolumn{4}{c}{D=2} & \multicolumn{3}{c}{D=5} & \multicolumn{2}{c}{D=2} \\\hline
Dataset      & \multicolumn{1}{l}{\textbf{Llog}} & \multicolumn{1}{l}{\textbf{Llog-init}} & \multicolumn{1}{l}{\textbf{Exp}} & \multicolumn{1}{l}{\textbf{W}} & \multicolumn{1}{l}{\textbf{SkSurv}} & \multicolumn{1}{l}{\textbf{RPART}} & \multicolumn{1}{l}{\textbf{CTree}} & \multicolumn{1}{l}{\textbf{PO}} & \multicolumn{1}{l}{\textbf{PH}} \\\hline
Aids         & 0.791 (1.21) & 0.779 (1.03) & 0.782 (1.17) & 0.788 (1.18) & 0.888 (2.04) & 0.812 (0.98) & 0.751 (2.22) & 0.773 (8.39) & 0.779 (6.71) \\
Aids\_death  & 0.884 (1.7) & 0.879 (1.43) & 0.84 (1.66) & 0.881 (1.63) & 0.944 (6.54) & 0.937 (1.74) & 0.776 (8.29) & 0.83 (9.26) & 0.819 (9.52) \\
Aids2        & 0.563 (0.64) & 0.566 (0.42) & 0.569 (0.35) & 0.564 (0.54) & 0.568 (0.75) & 0.521 (1.71) & 0.54 (0.19) & 0.572 (0.45) & 0.552 (2.2) \\
Churn        & 0.852 (0.47) & 0.854 (0.53) & 0.84 (0.57) & 0.842 (0.67) & 0.823 (0.35) & 0.791 (0.47) & 0.805 (0.29) & 0.85 (0.42) & 0.845 (0.65) \\
Credit\_risk & 0.813 (0.85) & 0.818 (0.92) & 0.769 (0.71) & 0.804 (0.76) & 0.794 (1.71) & 0.759 (1.14) & 0.759 (0.97) & 0.806 (1.46) & 0.801 (0.75) \\
Dialysis     & 0.763 (0.6) & 0.763 (0.55) & 0.755 (0.38) & 0.76 (0.51) & 0.678 (0.63) & 0.619 (0.99) & 0.671 (0.75) & 0.741 (3.64) & 0.749 (0.3) \\
Employee     & 0.836 (0.89) & 0.888 (1.94) & 0.801 (1.91) & 0.825 (2.47) & 0.93 (0.15) & 0.92 (0.32) & 0.917 (0.1) & 0.835 (1.03) & 0.869 (7.34) \\
Flchain      & 0.935 (0.1) & 0.935 (0.1) & 0.933 (0.12) & 0.934 (0.14) & 0.869 (0.97) & 0.882 (0.46) & 0.892 (0.05) & 0.861 (16.55) & 0.825 (18.78) \\
Framingham   & 0.719 (0.36) & 0.718 (0.31) & 0.717 (0.31) & 0.717 (0.35) & 0.72 (0.35) & 0.674 (0.73) & 0.712 (0.4) & 0.704 (1.33) & 0.709 (0.93) \\
Gbsg2        & 0.698 (0.6) & 0.701 (0.62) & 0.682 (0.72) & 0.689 (0.72) & 0.765 (0.73) & 0.707 (1.31) & 0.665 (1.33) & 0.698 (1.02) & 0.684 (1.14) \\
Maintenance  & 0.95 (0.98) & 0.946 (1.8) & 0.879 (1.61) & 0.952 (0.31) & 0.952 (0.2) & 0.933 (0.55) & 0.91 (0.46) & 0.953 (0.26) & 0.953 (0.25) \\
Uissurv      & 0.758 (0.65) & 0.759 (0.84) & 0.735 (0.78) & 0.743 (0.87) & 0.771 (0.7) & 0.744 (0.57) & 0.713 (0.84) & 0.762 (0.57) & 0.745 (0.78) \\
Unempdur     & 0.687 (0.55) & 0.693 (0.42) & 0.687 (0.36) & 0.687 (0.41) & 0.708 (0.5) & 0.661 (0.35) & 0.683 (0.46) & 0.683 (0.67) & 0.684 (0.62) \\
Veterans     & 0.726 (2.43) & 0.729 (2.15) & 0.714 (1.95) & 0.725 (2.2) & 0.788 (3.37) & 0.75 (3.5) & 0.647 (4.35) & 0.785 (2.01) & 0.773 (2.0) \\
Whas500      & 0.745 (0.97) & 0.741 (0.87) & 0.748 (0.75) & 0.739 (0.94) & 0.801 (1.52) & 0.772 (0.84) & 0.735 (0.98) & 0.782 (3.98) & 0.778 (1.48) \\\hline
AVERAGE      & \textbf{0.781} & \textbf{0.785} & \textbf{0.763} & \textbf{0.777} & \textbf{0.8} & \textbf{0.765} & \textbf{0.745} & \textbf{0.776} & \textbf{0.771} \\\hline
\end{tabular}
}
\caption{Training results using the CD-AUC $C_H$ measures. The comparison includes SST models with depth $D=2$ with parametric distributions (Exp, W, Llog) and spline-based semiparametric \textcolor{black}{survival functions} (PO and PH), as well as the three benchmark survival tree models (SkSurv, CTree, and RPART) with depth $D=5$. Llog-init refers to Llog with the clustering-based initialization procedure. In brackets the standard deviation divided by a factor of $1e^{-2}$ for visualization purposes.}
\end{table}
\end{landscape}

\begin{landscape}
\noindent \textbf{Depth $D=2$}
\vspace{-20pt}
\begin{table}[H]
\centering
\resizebox{0.85\textwidth}{!}{
\begin{tabular}{lccccccccc}
\cline{2-10}
             & \multicolumn{9}{c}{Testing $C_U$} \\
\cline{2-10}
             & \multicolumn{4}{c}{D=2} & \multicolumn{3}{c}{D=5} & \multicolumn{2}{c}{D=2} \\\hline
Dataset      & \multicolumn{1}{l}{\textbf{Llog}} & \multicolumn{1}{l}{\textbf{Llog-init}} & \multicolumn{1}{l}{\textbf{Exp}} & \multicolumn{1}{l}{\textbf{W}} & \multicolumn{1}{l}{\textbf{SkSurv}} & \multicolumn{1}{l}{\textbf{RPART}} & \multicolumn{1}{l}{\textbf{CTree}} & \multicolumn{1}{l}{\textbf{PO}} & \multicolumn{1}{l}{\textbf{PH}} \\\hline
Aids         & 0.740 (4.97) & 0.752 (4.34) & 0.745 (5.13) & 0.742 (4.91) & 0.624 (8.55) & 0.719 (9.57) & 0.697 (9.47) & 0.700 (5.35) & 0.708 (5.09) \\
Aids\_death  & 0.743 (8.47) & 0.799 (5.92) & 0.716 (9.51) & 0.742 (9.22) & 0.520 (9.01) & 0.609 (13.65) & 0.632 (7.65) & 0.672 (15.57) & 0.659 (15.13) \\
Aids2        & 0.539 (1.06) & 0.543 (0.92) & 0.544 (0.78) & 0.542 (1.02) & 0.528 (1.14) & 0.511 (1.01) & 0.528 (0.65) & 0.545 (1.17) & 0.536 (1.97) \\
Churn  & 0.737 (13.19) & 0.745 (12.96) & 0.732 (12.54) & 0.729 (12.78) & 0.703 (5.43) & 0.707 (5.55) & 0.692 (6.3) & 0.745 (12.74) & 0.741 (11.86) \\
Credit\_risk & 0.731 (2.14) & 0.728 (2.26) & 0.698 (1.9) & 0.723 (2.1) & 0.699 (3.01) & 0.702 (1.12) & 0.704 (1.92) & 0.727 (2.24) & 0.724 (2.04) \\
Dialysis     & 0.715 (1.64) & 0.716 (1.51) & 0.714 (1.63) & 0.715 (1.5) & 0.651 (2.44) & 0.634 (2.44) & 0.656 (1.98) & 0.696 (4.01) & 0.707 (1.82) \\
Employee& 0.726 (2.7) & 0.840 (4.21) & 0.740 (3.31) & 0.746 (4.16) & 0.896 (0.54) & 0.894 (0.56) & 0.886 (0.61) & 0.714 (3.81) & 0.820 (7.04) \\
Flchain & 0.940 (0.38) & 0.941 (0.36) & 0.939 (0.38) & 0.940 (0.39) & 0.852 (1.96) & 0.880 (0.3) & 0.884 (0.47) & 0.880 (15.26) & 0.863 (16.17) \\
Framingham & 0.695 (1.73) & 0.695 (1.68) & 0.694 (1.69) & 0.693 (1.71) & 0.655 (1.93) & 0.643 (1.75) & 0.664 (1.62) & 0.684 (1.81) & 0.691 (1.71) \\
Gbsg2   & 0.644 (3.97) & 0.643 (4.34) & 0.629 (4.14) & 0.638 (4.03) & 0.614 (3.51) & 0.631 (1.44) & 0.619 (4.08) & 0.641 (3.42) & 0.639 (2.93) \\
Maintenance & 0.932 (1.55) & 0.929 (1.92) & 0.837 (2.64) & 0.934 (1.04) & 0.924 (1.47) & 0.909 (1.39) & 0.893 (1.28) & 0.935 (1.0) & 0.935 (1.01) \\
Uissurv      & 0.736 (1.87) & 0.726 (1.65) & 0.710 (1.79) & 0.712 (2.17) & 0.718 (2.68) & 0.721 (1.97) & 0.700 (2.37) & 0.736 (1.84) & 0.723 (2.0) \\
Unempdur     & 0.647 (1.73) & 0.650 (1.76) & 0.648 (1.89) & 0.648 (1.86) & 0.628 (1.42) & 0.624 (1.66) & 0.639 (1.52) & 0.647 (1.67) & 0.645 (1.7) \\
Veterans     & 0.581 (7.41) & 0.591 (7.63) & 0.598 (6.82) & 0.579 (6.24) & 0.540 (2.65) & 0.566 (5.36) & 0.563 (5.69) & 0.624 (4.51) & 0.622 (5.3) \\
Whas500      & 0.708 (3.85) & 0.722 (3.78) & 0.714 (3.47) & 0.705 (4.06) & 0.678 (2.89) & 0.707 (2.76) & 0.692 (3.1) & 0.689 (4.34) & 0.703 (3.27) \\\hline
AVERAGE      & \textbf{0.720} & \textbf{0.733} & \textbf{0.708} & \textbf{0.718} & \textbf{0.686} & \textbf{0.696} & \textbf{0.697} & \textbf{0.710} & \textbf{0.714} \\\hline
\end{tabular}
}
\end{table}

\vspace{-15pt}
\begin{table}[H]
\centering
\resizebox{0.85\textwidth}{!}{
\begin{tabular}{lccccccccc}
\cline{2-10}
             & \multicolumn{9}{c}{Testing CD-AUC} \\
\cline{2-10}
             & \multicolumn{4}{c}{D=2} & \multicolumn{3}{c}{D=5} & \multicolumn{2}{c}{D=2} \\\hline
Dataset      & \multicolumn{1}{l}{\textbf{Llog}} & \multicolumn{1}{l}{\textbf{Llog-init}} & \multicolumn{1}{l}{\textbf{Exp}} & \multicolumn{1}{l}{\textbf{W}} & \multicolumn{1}{l}{\textbf{SkSurv}} & \multicolumn{1}{l}{\textbf{RPART}} & \multicolumn{1}{l}{\textbf{CTree}} & \multicolumn{1}{l}{\textbf{PO}} & \multicolumn{1}{l}{\textbf{PH}} \\\hline
Aids         & 0.732 (7.94) & 0.745 (7.0) & 0.744 (8.4) & 0.736 (7.77) & 0.642 (8.15) & 0.743 (9.25) & 0.724 (10.14) & 0.706 (8.78) & 0.714 (8.64) \\
Aids\_death  & 0.758 (10.92) & 0.824 (7.04) & 0.747 (11.36) & 0.760 (10.72) & 0.569 (12.65) & 0.637 (11.95) & 0.651 (9.78) & 0.727 (14.49) & 0.724 (14.69) \\
Aids2        & 0.564 (1.76) & 0.567 (1.21) & 0.560 (1.62) & 0.564 (1.62) & 0.543 (0.89) & 0.519 (1.69) & 0.542 (0.9) & 0.568 (1.97) & 0.548 (3.01) \\
Churn        & 0.920 (0.91) & 0.921 (1.04) & 0.912 (1.03) & 0.915 (0.97) & 0.851 (2.22) & 0.850 (1.21) & 0.856 (1.22) & 0.918 (1.04) & 0.916 (1.07) \\
Credit\_risk & 0.848 (2.49) & 0.845 (2.54) & 0.791 (3.1) & 0.837 (2.77) & 0.813 (1.74) & 0.824 (0.9) & 0.816 (1.57) & 0.844 (2.86) & 0.841 (2.72) \\
Dialysis     & 0.794 (1.2) & 0.791 (1.17) & 0.792 (1.19) & 0.794 (1.23) & 0.671 (2.52) & 0.630 (1.6) & 0.672 (1.33) & 0.763 (4.01) & 0.772 (1.53) \\
Employee     & 0.798 (2.45) & 0.897 (3.68) & 0.798 (2.96) & 0.811 (3.69) & 0.946 (0.56) & 0.942 (0.53) & 0.934 (0.51) & 0.788 (3.26) & 0.875 (8.38) \\
Flchain      & 0.955 (0.53) & 0.955 (0.52) & 0.953 (0.53) & 0.954 (0.56) & 0.875 (1.53) & 0.901 (0.62) & 0.906 (0.36) & 0.889 (16.28) & 0.875 (16.6) \\
Framingham   & 0.771 (1.42) & 0.771 (1.37) & 0.769 (1.39) & 0.769 (1.39) & 0.722 (1.24) & 0.713 (1.41) & 0.733 (1.44) & 0.758 (1.69) & 0.766 (1.45) \\
Gbsg2        & 0.700 (4.74) & 0.701 (4.42) & 0.681 (5.6) & 0.691 (4.89) & 0.659 (2.27) & 0.696 (1.76) & 0.693 (1.85) & 0.701 (4.28) & 0.690 (3.68) \\
Maintenance  & 0.999 (0.64) & 0.998 (0.69) & 0.930 (3.06) & 0.999 (0.11) & 0.992 (0.47) & 0.989 (0.29) & 0.982 (0.47) & 1.000 (0.05) & 1.000 (0.02) \\
Uissurv      & 0.837 (2.25) & 0.824 (1.86) & 0.805 (2.19) & 0.810 (2.68) & 0.822 (2.77) & 0.824 (2.18) & 0.796 (2.9) & 0.834 (2.13) & 0.822 (2.44) \\
Unempdur     & 0.729 (1.19) & 0.732 (1.34) & 0.730 (1.42) & 0.730 (1.45) & 0.703 (1.2) & 0.699 (1.2) & 0.723 (1.24) & 0.723 (0.98) & 0.723 (1.11) \\
Veterans     & 0.721 (7.46) & 0.723 (8.93) & 0.715 (7.46) & 0.694 (6.26) & 0.613 (4.17) & 0.684 (4.1) & 0.654 (5.59) & 0.705 (7.54) & 0.703 (7.84) \\
Whas500      & 0.773 (5.35) & 0.792 (5.77) & 0.774 (5.6) & 0.769 (5.99) & 0.745 (3.71) & 0.748 (4.02) & 0.741 (6.36) & 0.751 (5.88) & 0.771 (3.66) \\\hline
AVERAGE      & \textbf{0.793} & \textbf{0.806} & \textbf{0.780} & \textbf{0.789} & \textbf{0.744} & \textbf{0.760} & \textbf{0.761} & \textbf{0.778} & \textbf{0.783} \\\hline
\end{tabular}
}
\caption{Testing results using the $C_U$ and CD-AUC measures. The comparison includes SST models with depth $D=2$ with parametric distributions (Exp, W, Llog) and spline-based semiparametric \textcolor{black}{survival functions} (PO and PH), as well as the three benchmark survival tree models (SkSurv, CTree, and RPART) with depth $D=5$. Llog-init refers to Llog with the clustering-based initialization procedure. In brackets the standard deviation divided by a factor of $1e^{-2}$ for visualization purposes.}\label{tab:auc_test}
\end{table}
\end{landscape}

\subsection{Experiments at different depths for RPART, CTree and SkSurv}

\noindent \textbf{$C_H$}

\begin{table}[H]
\centering
\resizebox{0.6\textwidth}{!}{
\begin{tabular}{lcccccccc}
\cline{2-9}
            & \multicolumn{8}{c}{CTree $C_H$}                                                                               \\ \cline{2-9} 
            & \multicolumn{2}{c}{$D=2$} & \multicolumn{2}{c}{$D=3$} & \multicolumn{2}{c}{$D=4$} & \multicolumn{2}{c}{$D=5$} \\ \cline{2-9} 
Dataset     & Train       & Test        & Train       & Test        & Train       & Test        & Train       & Test        \\ \hline
Aids        & 0.745       & 0.704       & 0.751       & 0.706       & 0.751       & 0.706       & 0.751       & 0.706       \\
Aids\_death & 0.765       & 0.649       & 0.776       & 0.645       & 0.776       & 0.645       & 0.776       & 0.645       \\
Aids2       & 0.539       & 0.537       & 0.54        & 0.537       & 0.54        & 0.537       & 0.54        & 0.537       \\
Churn       & 0.73        & 0.727       & 0.763       & 0.744       & 0.79        & 0.769       & 0.805       & 0.781       \\
Credit risk & 0.676       & 0.67        & 0.725       & 0.7         & 0.756       & 0.727       & 0.759       & 0.726       \\
Dialysis    & 0.632       & 0.624       & 0.65        & 0.634       & 0.661       & 0.639       & 0.671       & 0.642       \\
Employee    & 0.863       & 0.855       & 0.892       & 0.882       & 0.908       & 0.894       & 0.917       & 0.903       \\
Flchain     & 0.777           & 0.775       & 0.844           & 0.839       & 0.876           & 0.867       & 0.891           & 0.88        \\
Framingham  & 0.651       & 0.637       & 0.689       & 0.666       & 0.708       & 0.677       & 0.712       & 0.676       \\
Gbsg2       & 0.661       & 0.637       & 0.665       & 0.633       & 0.665       & 0.633       & 0.665       & 0.633       \\
Maintenance & 0.814       & 0.808       & 0.907       & 0.898       & 0.91        & 0.901       & 0.91        & 0.901       \\
Uissurv         & 0.713       & 0.7         & 0.713       & 0.7         & 0.713       & 0.7         & 0.713       & 0.7         \\
Unempdur    & 0.681       & 0.675       & 0.683       & 0.676       & 0.683       & 0.676       & 0.683       & 0.676       \\
Veterans    & 0.647       & 0.575       & 0.647       & 0.575       & 0.647       & 0.575       & 0.647       & 0.575       \\
Whas500     & 0.719       & 0.678       & 0.735       & 0.688       & 0.735       & 0.688       & 0.735       & 0.688       \\ \hline
AVERAGE     & \textbf{0.656} & \textbf{0.683} & \textbf{0.676} & \textbf{0.702} & \textbf{0.683} & \textbf{0.709} & \textbf{0.686} & \textbf{0.711} \\ \hline
\end{tabular}
}
\end{table}

\begin{table}[H]
\centering
\resizebox{0.6\textwidth}{!}{
\begin{tabular}{lcccccccc}
\cline{2-9}
            & \multicolumn{8}{c}{RPART $C_H$}                                                                               \\ \cline{2-9} 
            & \multicolumn{2}{c}{$D=2$} & \multicolumn{2}{c}{$D=3$} & \multicolumn{2}{c}{$D=4$} & \multicolumn{2}{c}{$D=5$} \\ \cline{2-9} 
Dataset     & Train       & Test        & Train       & Test        & Train       & Test        & Train       & Test        \\ \hline
Aids        & 0.756       & 0.745       & 0.787       & 0.744       & 0.797       & 0.727       & 0.812       & 0.727       \\
Aids\_death      & 0.829       & 0.668       & 0.901       & 0.634       & 0.927       & 0.614       & 0.937       & 0.624       \\
Aids2       & 0.521       & 0.518       & 0.521       & 0.518       & 0.521       & 0.518       & 0.521       & 0.518       \\
Churn       & 0.726       & 0.723       & 0.759       & 0.749       & 0.78        & 0.765       & 0.791       & 0.779       \\
Credit\_risk     & 0.701       & 0.692       & 0.741       & 0.711       & 0.752       & 0.716       & 0.759       & 0.723       \\
Dialysis   & 0.619       & 0.609       & 0.619       & 0.609       & 0.619       & 0.609       & 0.619       & 0.609       \\
Employee\_attrition  & 0.883       & 0.877       & 0.907       & 0.9         & 0.915       & 0.906       & 0.92        & 0.912       \\
Flchain     & 0.77           & 0.763       & 0.842           & 0.838       & 0.869           & 0.866       & 0.882           & 0.876       \\
Framingham    & 0.653       & 0.641       & 0.674       & 0.658       & 0.674       & 0.658       & 0.674       & 0.658       \\
Gbsg2      & 0.67        & 0.631       & 0.694       & 0.641       & 0.704       & 0.644       & 0.707       & 0.643       \\
Maintenance  & 0.832       & 0.824       & 0.911       & 0.901       & 0.933       & 0.922       & 0.933       & 0.922       \\
Uissurv      & 0.727       & 0.718       & 0.732       & 0.716       & 0.738       & 0.717       & 0.744       & 0.721       \\
Unempdur   & 0.661       & 0.659       & 0.661       & 0.659       & 0.661       & 0.659       & 0.661       & 0.659       \\
Veterans & 0.666       & 0.579       & 0.715       & 0.581       & 0.732       & 0.568       & 0.75        & 0.575       \\
Whas500    & 0.716       & 0.677       & 0.754       & 0.688       & 0.761       & 0.706       & 0.772       & 0.705       \\ \hline
AVERAGE    & \textbf{0.664} & \textbf{0.688} & \textbf{0.692} & \textbf{0.703} & \textbf{0.701} & \textbf{0.706} & \textbf{0.707} & \textbf{0.71} \\ \hline
\end{tabular}
}
\end{table}

\begin{table}[H]
\centering
\resizebox{0.6\textwidth}{!}{
\begin{tabular}{lcccccccc}
\cline{2-9}
            & \multicolumn{8}{c}{SkSurv $C_H$}                                                                               \\ \cline{2-9} 
            & \multicolumn{2}{c}{$D=2$} & \multicolumn{2}{c}{$D=3$} & \multicolumn{2}{c}{$D=4$} & \multicolumn{2}{c}{$D=5$} \\ \cline{2-9} 
Dataset     & Train       & Test        & Train       & Test        & Train       & Test        & Train       & Test        \\ \hline
Aids        & 0.758       & 0.71        & 0.799       & 0.74        & 0.844       & 0.705       & 0.888       & 0.621       \\
Aids\_death & 0.797       & 0.642       & 0.869       & 0.634       & 0.914       & 0.589       & 0.944       & 0.545       \\
Aids2       & 0.531       & 0.527       & 0.55        & 0.533       & 0.557       & 0.538       & 0.568       & 0.54        \\
Churn       & 0.731       & 0.72        & 0.761       & 0.745       & 0.799       & 0.771       & 0.823       & 0.778       \\
Credit\_risk & 0.705       & 0.682       & 0.737       & 0.7         & 0.765       & 0.705       & 0.794       & 0.718       \\
Dialysis    & 0.62        & 0.611       & 0.638       & 0.625       & 0.662       & 0.639       & 0.678       & 0.642       \\
Employee    & 0.883       & 0.877       & 0.908       & 0.896       & 0.923       & 0.912       & 0.93        & 0.912       \\
Flchain     & 0.761           & 0.759       & 0.811           & 0.807       & 0.842           & 0.829       & 0.869           & 0.848       \\
Framingham  & 0.627       & 0.609       & 0.669       & 0.643       & 0.697       & 0.66        & 0.72        & 0.666       \\
Gbsg2       & 0.644       & 0.596       & 0.695       & 0.65        & 0.726       & 0.637       & 0.765       & 0.626       \\
Maintenance & 0.832       & 0.824       & 0.914       & 0.901       & 0.945       & 0.928       & 0.952       & 0.934       \\
Uissurv         & 0.675       & 0.669       & 0.732       & 0.714       & 0.752       & 0.713       & 0.771       & 0.717       \\
Unempdur    & 0.682       & 0.674       & 0.69        & 0.671       & 0.699       & 0.672       & 0.708       & 0.667       \\
Veterans    & 0.651       & 0.556       & 0.702       & 0.564       & 0.752       & 0.551       & 0.788       & 0.549       \\
Whas500     & 0.701       & 0.684       & 0.732       & 0.69        & 0.766       & 0.68        & 0.801       & 0.674       \\ \hline
AVERAGE     & \textbf{0.656} & \textbf{0.676} & \textbf{0.693} & \textbf{0.701} & \textbf{0.72} & \textbf{0.702} & \textbf{0.742} & \textbf{0.696} \\ \hline
\end{tabular}
}
\end{table}

\noindent \textbf{$C_U$}

\begin{table}[H]
\centering
\resizebox{0.6\textwidth}{!}{
\begin{tabular}{lcccccccc}
\cline{2-9}
            & \multicolumn{8}{c}{CTree $C_U$}                                                                               \\ \cline{2-9} 
            & \multicolumn{2}{c}{$D=2$} & \multicolumn{2}{c}{$D=3$} & \multicolumn{2}{c}{$D=4$} & \multicolumn{2}{c}{$D=5$} \\ \cline{2-9} 
Dataset     & Train       & Test        & Train       & Test        & Train       & Test        & Train       & Test        \\ \hline
Aids        & 0.735       & 0.697       & 0.739       & 0.697       & 0.739       & 0.697       & 0.739       & 0.697       \\
Aids\_death & 0.754       & 0.635       & 0.767       & 0.632       & 0.767       & 0.632       & 0.767       & 0.632       \\
Aids2       & 0.53        & 0.527       & 0.531       & 0.528       & 0.531       & 0.528       & 0.531       & 0.528       \\
Churn       & 0.667       & 0.66        & 0.69        & 0.673       & 0.715       & 0.693       & 0.744       & 0.692       \\
Credit\_risk & 0.673       & 0.663       & 0.713       & 0.682       & 0.739       & 0.705       & 0.742       & 0.704       \\
Dialysis    & 0.643       & 0.64        & 0.66        & 0.652       & 0.67        & 0.654       & 0.677       & 0.656       \\
Employee    & 0.766       & 0.763       & 0.818       & 0.815       & 0.874       & 0.866       & 0.895       & 0.886       \\
Flchain     & 0.769           & 0.769       & 0.844           & 0.838       & 0.878           & 0.87        & 0.894           & 0.884       \\
Framingham  & 0.638       & 0.624       & 0.675       & 0.656       & 0.695       & 0.666       & 0.699       & 0.664       \\
Gbsg2       & 0.643       & 0.623       & 0.647       & 0.619       & 0.647       & 0.619       & 0.647       & 0.619       \\
Maintenance & 0.822       & 0.813       & 0.9         & 0.89        & 0.903       & 0.893       & 0.903       & 0.893       \\
Uissurv     & 0.712       & 0.7         & 0.712       & 0.7         & 0.712       & 0.7         & 0.712       & 0.7         \\
Unempdur    & 0.648       & 0.639       & 0.649       & 0.639       & 0.649       & 0.639       & 0.649       & 0.639       \\
Veterans    & 0.642       & 0.563       & 0.642       & 0.563       & 0.642       & 0.563       & 0.642       & 0.563       \\
Whas500     & 0.717       & 0.683       & 0.731       & 0.692       & 0.731       & 0.692       & 0.731       & 0.692       \\ \hline
AVERAGE     & \textbf{0.639} & \textbf{0.667} & \textbf{0.658} & \textbf{0.685} & \textbf{0.668} & \textbf{0.694} & \textbf{0.672} & \textbf{0.697} \\ \hline
\end{tabular}
}
\end{table}

\begin{table}[H]
\centering
\resizebox{0.6\textwidth}{!}{
\begin{tabular}{lcccccccc}
\cline{2-9}
            & \multicolumn{8}{c}{RPART $C_U$}                                                                               \\ \cline{2-9} 
            & \multicolumn{2}{c}{$D=2$} & \multicolumn{2}{c}{$D=3$} & \multicolumn{2}{c}{$D=4$} & \multicolumn{2}{c}{$D=5$} \\ \cline{2-9} 
Dataset     & Train       & Test        & Train       & Test        & Train       & Test        & Train       & Test        \\\hline
Aids        & 0.742       & 0.73        & 0.77        & 0.732       & 0.781       & 0.719       & 0.797       & 0.719       \\
Aids\_death & 0.825       & 0.638       & 0.896       & 0.611       & 0.922       & 0.593       & 0.935       & 0.609       \\
Aids2       & 0.516       & 0.511       & 0.516       & 0.511       & 0.516       & 0.511       & 0.516       & 0.511       \\
Churn       & 0.666       & 0.668       & 0.698       & 0.691       & 0.715       & 0.699       & 0.731       & 0.707       \\
Credit\_risk & 0.69        & 0.675       & 0.724       & 0.691       & 0.733       & 0.696       & 0.739       & 0.702       \\
Dialysis    & 0.636       & 0.634       & 0.636       & 0.634       & 0.636       & 0.634       & 0.636       & 0.634       \\
Employee    & 0.798       & 0.798       & 0.87        & 0.866       & 0.892       & 0.884       & 0.901       & 0.894       \\
Flchain     & 0.768           & 0.756       & 0.842           & 0.836       & 0.871           & 0.868       & 0.886           & 0.88        \\
Framingham  & 0.64        & 0.627       & 0.66        & 0.643       & 0.66        & 0.643       & 0.66        & 0.643       \\
Gbsg2       & 0.655       & 0.62        & 0.682       & 0.631       & 0.69        & 0.63        & 0.695       & 0.631       \\
Maintenance & 0.797       & 0.79        & 0.895       & 0.883       & 0.921       & 0.909       & 0.921       & 0.909       \\
Uissurv     & 0.725       & 0.718       & 0.73        & 0.716       & 0.737       & 0.717       & 0.743       & 0.721       \\
Unempdur    & 0.626       & 0.624       & 0.626       & 0.624       & 0.626       & 0.624       & 0.626       & 0.624       \\
Veterans    & 0.662       & 0.573       & 0.71        & 0.572       & 0.727       & 0.559       & 0.745       & 0.566       \\
Whas500     & 0.701       & 0.681       & 0.731       & 0.692       & 0.737       & 0.705       & 0.746       & 0.707       \\\hline
AVERAGE     & \textbf{0.645} & \textbf{0.67} & \textbf{0.676} & \textbf{0.689} & \textbf{0.686} & \textbf{0.693} & \textbf{0.693} & \textbf{0.697} \\\hline
\end{tabular}
}
\end{table}

\begin{table}[H]
\centering
\resizebox{0.6\textwidth}{!}{
\begin{tabular}{lcccccccc}
\cline{2-9}
            & \multicolumn{8}{c}{SkSurv $C_U$}                                                                               \\ \cline{2-9} 
            & \multicolumn{2}{c}{$D=2$} & \multicolumn{2}{c}{$D=3$} & \multicolumn{2}{c}{$D=4$} & \multicolumn{2}{c}{$D=5$} \\ \cline{2-9} 
Dataset     & Train       & Test        & Train       & Test        & Train       & Test        & Train       & Test        \\ \hline
Aids        & 0.748       & 0.702       & 0.785       & 0.722       & 0.83        & 0.704       & 0.874       & 0.624       \\
Aids\_death & 0.784       & 0.637       & 0.853       & 0.586       & 0.897       & 0.567       & 0.937       & 0.52        \\
Aids2       & 0.523       & 0.519       & 0.54        & 0.523       & 0.547       & 0.527       & 0.558       & 0.528       \\
Churn       & 0.667       & 0.656       & 0.687       & 0.673       & 0.719       & 0.695       & 0.742       & 0.703       \\
Credit risk & 0.691       & 0.67        & 0.72        & 0.686       & 0.747       & 0.689       & 0.778       & 0.699       \\
Dialysis    & 0.626       & 0.617       & 0.64        & 0.632       & 0.666       & 0.648       & 0.679       & 0.651       \\
Employee    & 0.79        & 0.787       & 0.865       & 0.857       & 0.897       & 0.89        & 0.909       & 0.896       \\
Flchain     & 0.762           & 0.759       & 0.813           & 0.807       & 0.844           & 0.829       & 0.871           & 0.852       \\
Framingham  & 0.615       & 0.6         & 0.656       & 0.63        & 0.682       & 0.649       & 0.705       & 0.655       \\
Gbsg2       & 0.622       & 0.583       & 0.671       & 0.637       & 0.703       & 0.633       & 0.745       & 0.614       \\
Maintenance & 0.797       & 0.79        & 0.897       & 0.884       & 0.936       & 0.919       & 0.944       & 0.924       \\
Uissurv         & 0.672       & 0.666       & 0.731       & 0.714       & 0.751       & 0.713       & 0.77        & 0.718       \\
Unempdur    & 0.649       & 0.639       & 0.657       & 0.636       & 0.668       & 0.634       & 0.679       & 0.628       \\
Veterans    & 0.645       & 0.55        & 0.697       & 0.558       & 0.747       & 0.542       & 0.784       & 0.54        \\
Whas500     & 0.685       & 0.677       & 0.71        & 0.688       & 0.736       & 0.677       & 0.774       & 0.678       \\ \hline
AVERAGE     & \textbf{0.634} & \textbf{0.657} & \textbf{0.674} & \textbf{0.682} & \textbf{0.702} & \textbf{0.688} & \textbf{0.725} & \textbf{0.682} \\ \hline
\end{tabular}
}
\end{table}

\noindent \textbf{AUC}

\begin{table}[H]
\centering
\resizebox{0.6\textwidth}{!}{
\begin{tabular}{lcccccccc}
\cline{2-9}
            & \multicolumn{8}{c}{CTree   AUC}                                                                               \\ \cline{2-9} 
            & \multicolumn{2}{c}{$D=2$} & \multicolumn{2}{c}{$D=3$} & \multicolumn{2}{c}{$D=4$} & \multicolumn{2}{c}{$D=5$} \\ \cline{2-9} 
Dataset     & Train       & Test        & Train       & Test        & Train       & Test        & Train       & Test        \\ \hline
Aids        & 0.753       & 0.723       & 0.757       & 0.724       & 0.757       & 0.724       & 0.757       & 0.724       \\
Aids\_death & 0.725       & 0.657       & 0.738       & 0.651       & 0.738       & 0.651       & 0.738       & 0.651       \\
Aids2       & 0.548       & 0.542       & 0.549       & 0.542       & 0.549       & 0.542       & 0.549       & 0.542       \\
Churn       & 0.789       & 0.787       & 0.826       & 0.813       & 0.858       & 0.841       & 0.878       & 0.856       \\
Credit risk & 0.782       & 0.771       & 0.827       & 0.794       & 0.849       & 0.817       & 0.85        & 0.816       \\
Dialysis    & 0.652       & 0.645       & 0.672       & 0.659       & 0.685       & 0.666       & 0.697       & 0.672       \\
Employee    & 0.833       & 0.831       & 0.87        & 0.869       & 0.921       & 0.917       & 0.94        & 0.934       \\
Flchain     & 0.807       & 0.806       & 0.876       & 0.872       & 0.903       & 0.896       & 0.915       & 0.906       \\
Framingham  & 0.702       & 0.69        & 0.741       & 0.721       & 0.76        & 0.732       & 0.765       & 0.733       \\
Gbsg2       & 0.708       & 0.698       & 0.714       & 0.693       & 0.714       & 0.693       & 0.714       & 0.693       \\
Maintenance & 0.919       & 0.92        & 0.982       & 0.981       & 0.983       & 0.982       & 0.983       & 0.982       \\
Uissurv     & 0.804       & 0.796       & 0.804       & 0.796       & 0.804       & 0.796       & 0.804       & 0.796       \\
Unempdur    & 0.723       & 0.721       & 0.726       & 0.722       & 0.726       & 0.723       & 0.726       & 0.723       \\
Veterans    & 0.729       & 0.654       & 0.729       & 0.654       & 0.729       & 0.654       & 0.729       & 0.654       \\
Whas500     & 0.795       & 0.733       & 0.82        & 0.741       & 0.82        & 0.741       & 0.82        & 0.741       \\ \hline
AVERAGE     & \textbf{0.751} & \textbf{0.731} & \textbf{0.775} & \textbf{0.749} & \textbf{0.787} & \textbf{0.758} & \textbf{0.791} & \textbf{0.761}   \\ \hline
\end{tabular}
}
\end{table}

\begin{table}[H]
\centering
\resizebox{0.6\textwidth}{!}{
\begin{tabular}{lcccccccc}
\cline{2-9}
            & \multicolumn{8}{c}{RPART AUC}                                                                               \\ \cline{2-9} 
            & \multicolumn{2}{c}{$D=2$} & \multicolumn{2}{c}{$D=3$} & \multicolumn{2}{c}{$D=4$} & \multicolumn{2}{c}{$D=5$} \\ \cline{2-9} 
Dataset     & Train       & Test        & Train       & Test        & Train       & Test        & Train       & Test        \\ \hline
Aids        & 0.757       & 0.757       & 0.793       & 0.757       & 0.805       & 0.736       & 0.82        & 0.743       \\
Aids\_death & 0.813       & 0.683       & 0.895       & 0.652       & 0.92        & 0.625       & 0.933       & 0.637       \\
Aids2       & 0.525       & 0.519       & 0.525       & 0.519       & 0.525       & 0.519       & 0.525       & 0.519       \\
Churn       & 0.787       & 0.782       & 0.828       & 0.817       & 0.853       & 0.838       & 0.865       & 0.85        \\
Credit risk & 0.814       & 0.799       & 0.847       & 0.819       & 0.854       & 0.822       & 0.857       & 0.824       \\
Dialysis    & 0.638       & 0.63        & 0.638       & 0.63        & 0.638       & 0.63        & 0.638       & 0.63        \\
Employee    & 0.85        & 0.851       & 0.922       & 0.92        & 0.94        & 0.934       & 0.947       & 0.942       \\
Flchain     & 0.798       & 0.791       & 0.872       & 0.869       & 0.894       & 0.892       & 0.905       & 0.901       \\
Framingham  & 0.704       & 0.695       & 0.727       & 0.713       & 0.727       & 0.713       & 0.727       & 0.713       \\
Gbsg2       & 0.716       & 0.68        & 0.745       & 0.702       & 0.756       & 0.696       & 0.759       & 0.696       \\
Maintenance & 0.922       & 0.918       & 0.983       & 0.982       & 0.991       & 0.989       & 0.991       & 0.989       \\
Uissurv     & 0.828       & 0.826       & 0.834       & 0.827       & 0.841       & 0.825       & 0.847       & 0.824       \\
Unempdur    & 0.699       & 0.699       & 0.699       & 0.699       & 0.699       & 0.699       & 0.699       & 0.699       \\
Veterans    & 0.758       & 0.658       & 0.817       & 0.674       & 0.839       & 0.665       & 0.857       & 0.684       \\
Whas500     & 0.803       & 0.727       & 0.84        & 0.731       & 0.855       & 0.747       & 0.865       & 0.748       \\ \hline
AVERAGE     & \textbf{0.761} & \textbf{0.734} & \textbf{0.798} & \textbf{0.754} & \textbf{0.809} & \textbf{0.755} & \textbf{0.816} & \textbf{0.76} \\ \hline
\end{tabular}
}
\end{table}

\begin{table}[H]
\centering
\resizebox{0.6\textwidth}{!}{
\begin{tabular}{lcccccccc}
\cline{2-9}
            & \multicolumn{8}{c}{SkSurv AUC}                                                                               \\ \cline{2-9} 
            & \multicolumn{2}{c}{$D=2$} & \multicolumn{2}{c}{$D=3$} & \multicolumn{2}{c}{$D=4$} & \multicolumn{2}{c}{$D=5$} \\ \cline{2-9} 
Dataset     & Train       & Test        & Train       & Test        & Train       & Test        & Train       & Test        \\ \hline
Aids        & 0.761       & 0.722       & 0.803       & 0.751       & 0.852       & 0.72        & 0.894       & 0.642       \\
Aids\_death & 0.777       & 0.651       & 0.856       & 0.653       & 0.909       & 0.625       & 0.934       & 0.569       \\
Aids2       & 0.537       & 0.53        & 0.563       & 0.535       & 0.571       & 0.537       & 0.585       & 0.543       \\
Churn       & 0.79        & 0.782       & 0.822       & 0.806       & 0.866       & 0.843       & 0.891       & 0.851       \\
Credit risk & 0.806       & 0.792       & 0.845       & 0.808       & 0.862       & 0.806       & 0.887       & 0.813       \\
Dialysis    & 0.638       & 0.629       & 0.659       & 0.646       & 0.685       & 0.665       & 0.701       & 0.671       \\
Employee    & 0.855       & 0.854       & 0.918       & 0.913       & 0.944       & 0.938       & 0.956       & 0.946       \\
Flchain     & 0.789       & 0.788       & 0.837       & 0.834       & 0.868       & 0.858       & 0.889       & 0.875       \\
Framingham  & 0.675       & 0.655       & 0.725       & 0.699       & 0.755       & 0.718       & 0.773       & 0.722       \\
Gbsg2       & 0.682       & 0.653       & 0.741       & 0.707       & 0.773       & 0.691       & 0.81        & 0.659       \\
Maintenance & 0.922       & 0.918       & 0.984       & 0.982       & 0.998       & 0.993       & 1.0         & 0.992       \\
Uissurv     & 0.782       & 0.774       & 0.834       & 0.825       & 0.853       & 0.823       & 0.868       & 0.822       \\
Unempdur    & 0.727       & 0.719       & 0.738       & 0.715       & 0.748       & 0.712       & 0.758       & 0.703       \\
Veterans    & 0.732       & 0.612       & 0.8         & 0.639       & 0.862       & 0.616       & 0.901       & 0.613       \\
Whas500     & 0.744       & 0.738       & 0.783       & 0.743       & 0.814       & 0.739       & 0.869       & 0.745       \\ \hline
AVERAGE     & \textbf{0.748} & \textbf{0.721} & \textbf{0.794} & \textbf{0.75} & \textbf{0.824} & \textbf{0.752} & \textbf{0.848} & \textbf{0.744} \\ \hline
\end{tabular}
}
\end{table}

\noindent \textbf{IBS}

\begin{table}[h]
    \centering
    \begin{minipage}{0.32\textwidth} % Regola la larghezza in base alle necessità
        \centering
        \resizebox{\textwidth}{!}{
        \begin{tabular}{lcccc}
\cline{2-5}
            & \multicolumn{4}{c}{CTree Testing IBS}                                                     \\ \hline
Dataset     & $D=2$                & $D=3$                & $D=4$                & $D=5$                \\ \hline
Aids        & 0.059                & 0.059                & 0.059                & 0.059                \\
Aids\_death & 0.016                & 0.016                & 0.016                & 0.016                \\
Aids2       & 0.140                & 0.140                & 0.140                & 0.140                \\
Churn       & 0.132                & 0.123                & 0.108                & 0.101                \\
Credit risk & 0.116                & 0.116                & 0.113                & 0.114                \\
Dialysis    & 0.180                & 0.178                & 0.177                & 0.176                \\
Employee    & 0.145                & 0.117                & 0.083                & 0.065                \\
Flchain     & 0.087                & 0.072                & 0.064                & 0.060                \\
Framingham  & 0.121                & 0.118                & 0.117                & 0.117                \\
Gbsg2       & 0.173                & 0.173                & 0.173                & 0.173                \\
Maintenance & 0.029                & 0.008                & 0.007                & 0.007                \\
Uissurv     & 0.146                & 0.146                & 0.146                & 0.146                \\
Unempdur    & 0.162                & 0.162                & 0.161                & 0.161                \\
Veterans    & 0.134                & 0.134                & 0.134                & 0.134                \\
Whas500     & 0.182                & 0.177                & 0.177                & 0.177                \\ \hline
AVERAGE     & \textbf{0.122}       & \textbf{0.116}       & \textbf{0.112}       & \textbf{0.110}       \\ \hline
\end{tabular}
        }
        %\caption{Tabella 1}
    \end{minipage}
    \hfill
    \begin{minipage}{0.32\textwidth}
        \centering
        \resizebox{\textwidth}{!}{
        \begin{tabular}{lcccc}
\cline{2-5}
            & \multicolumn{4}{c}{RPART Testing IBS}                                                     \\ \hline
Dataset     & $D=2$                & $D=3$                & $D=4$                & $D=5$                \\ \hline
Aids        & 0.066                & 0.067                & 0.067                & 0.068                \\
Aids\_death & 0.017                & 0.019                & 0.020                & 0.020                \\
Aids2       & 0.141                & 0.141                & 0.141                & 0.141                \\
Churn       & 0.136                & 0.122                & 0.109                & 0.101                \\
Credit risk & 0.114                & 0.113                & 0.114                & 0.113                \\
Dialysis    & 0.183                & 0.183                & 0.183                & 0.183                \\
Employee    & 0.123                & 0.081                & 0.063                & 0.057                \\
Flchain     & 0.087                & 0.075                & 0.067                & 0.064                \\
Framingham  & 0.121                & 0.118                & 0.118                & 0.118                \\
Gbsg2       & 0.178                & 0.176                & 0.179                & 0.180                \\
Maintenance & 0.019                & 0.009                & 0.003                & 0.003                \\
Uissurv     & 0.141                & 0.141                & 0.143                & 0.144                \\
Unempdur    & 0.164                & 0.164                & 0.164                & 0.164                \\
Veterans    & 0.135                & 0.136                & 0.136                & 0.134                \\
Whas500     & 0.187                & 0.197                & 0.190                & 0.190                \\ \hline
AVERAGE     & \textbf{0.121}       & \textbf{0.116}       & \textbf{0.113}       & \textbf{0.112}       \\ \hline
\end{tabular}
        }
        %\caption{Tabella 2}
    \end{minipage}
    \hfill
    \begin{minipage}{0.32\textwidth}
        \centering
        \resizebox{\textwidth}{!}{
\begin{tabular}{lcccc}
\cline{2-5}
            & \multicolumn{4}{c}{SkSurv Testing IBS}                                                   \\ \hline
Dataset     & $D=2$                & $D=3$                & $D=4$                & $D=5$                \\ \hline
Aids        & 0.059                & 0.058                & 0.060                & 0.069                \\
Aids\_death & 0.016                & 0.017                & 0.019                & 0.020                \\
Aids2       & 0.141                & 0.141                & 0.141                & 0.141                \\
Churn       & 0.133                & 0.125                & 0.109                & 0.103                \\
Credit risk & 0.113                & 0.113                & 0.118                & 0.122                \\
Dialysis    & 0.181                & 0.177                & 0.175                & 0.175                \\
Employee    & 0.125                & 0.086                & 0.066                & 0.055                \\
Flchain     & 0.083                & 0.072                & 0.067                & 0.062                \\
Framingham  & 0.123                & 0.120                & 0.119                & 0.119                \\
Gbsg2       & 0.178                & 0.173                & 0.180                & 0.189                \\
Maintenance & 0.019                & 0.009                & 0.001                & 0.001                \\
Uissurv     & 0.148                & 0.142                & 0.144                & 0.146                \\
Unempdur    & 0.162                & 0.164                & 0.166                & 0.170                \\
Veterans    & 0.141                & 0.139                & 0.153                & 0.160                \\
Whas500     & 0.192                & 0.185                & 0.200                & 0.206                \\ \hline
AVERAGE     & \textbf{0.121}       & \textbf{0.115}       & \textbf{0.115}       & \textbf{0.116}       \\ \hline
\end{tabular}
        }
        %\caption{Tabella 3}
    \end{minipage}
\end{table}

\subsection{Box plots for Llog, CTree, and PO}

\begin{figure}[H]
    \centering
    \includegraphics[width=0.7\linewidth]{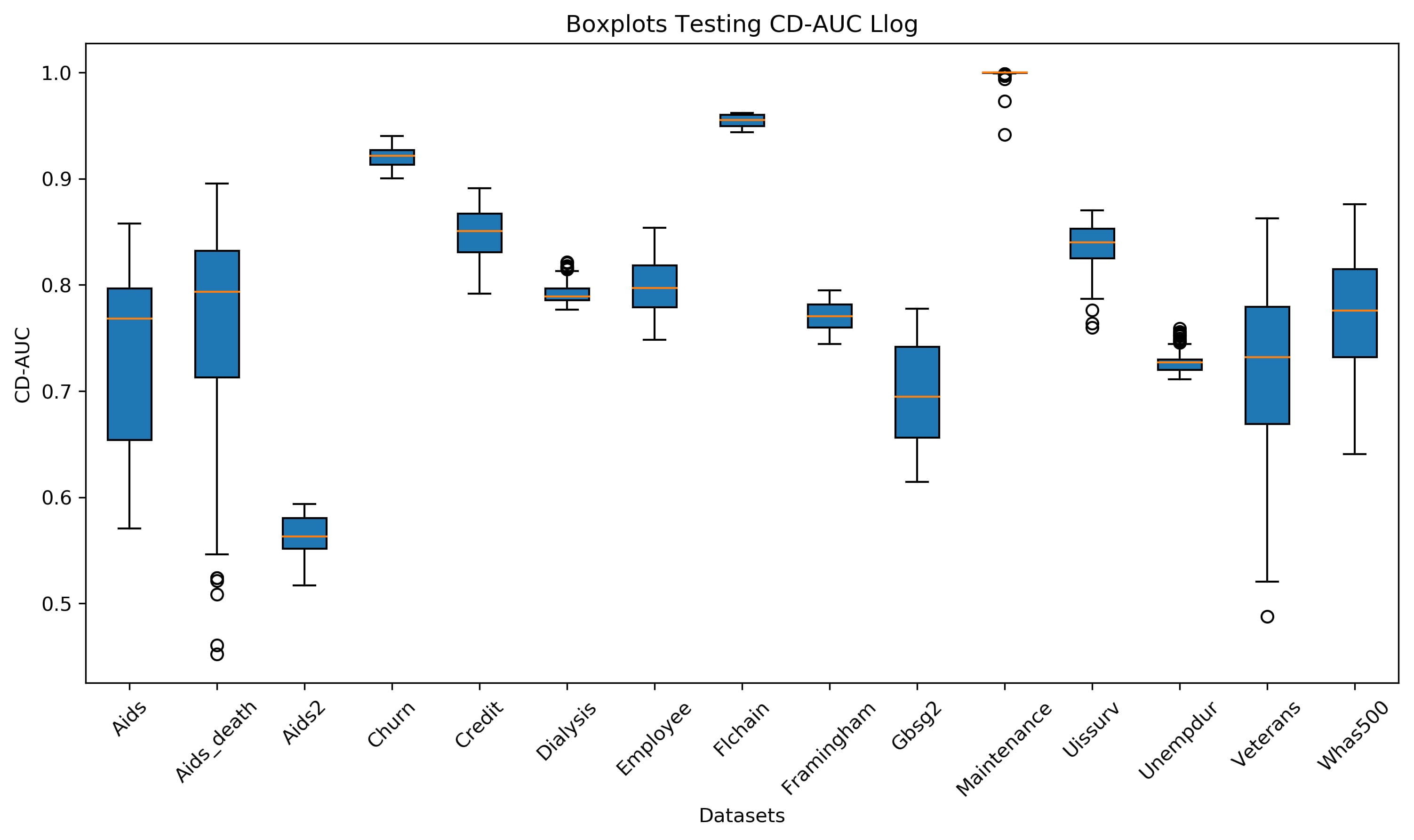}

    \label{fig:enter-label}
\end{figure}

\begin{figure}[H]
    \centering
    \includegraphics[width=0.7\linewidth]{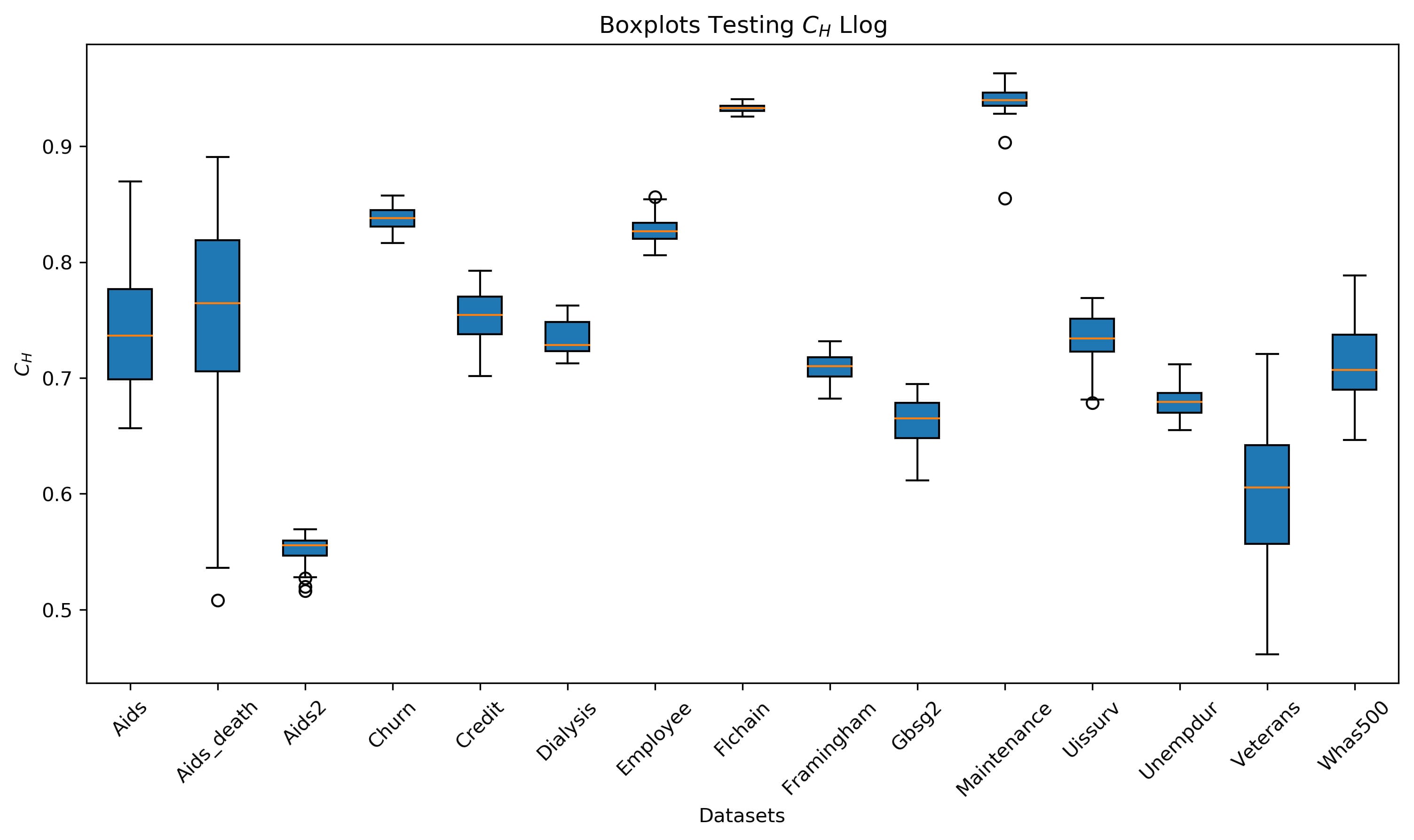}

    \label{fig:enter-label}
\end{figure}

\begin{figure}[H]
    \centering
    \includegraphics[width=0.7\linewidth]{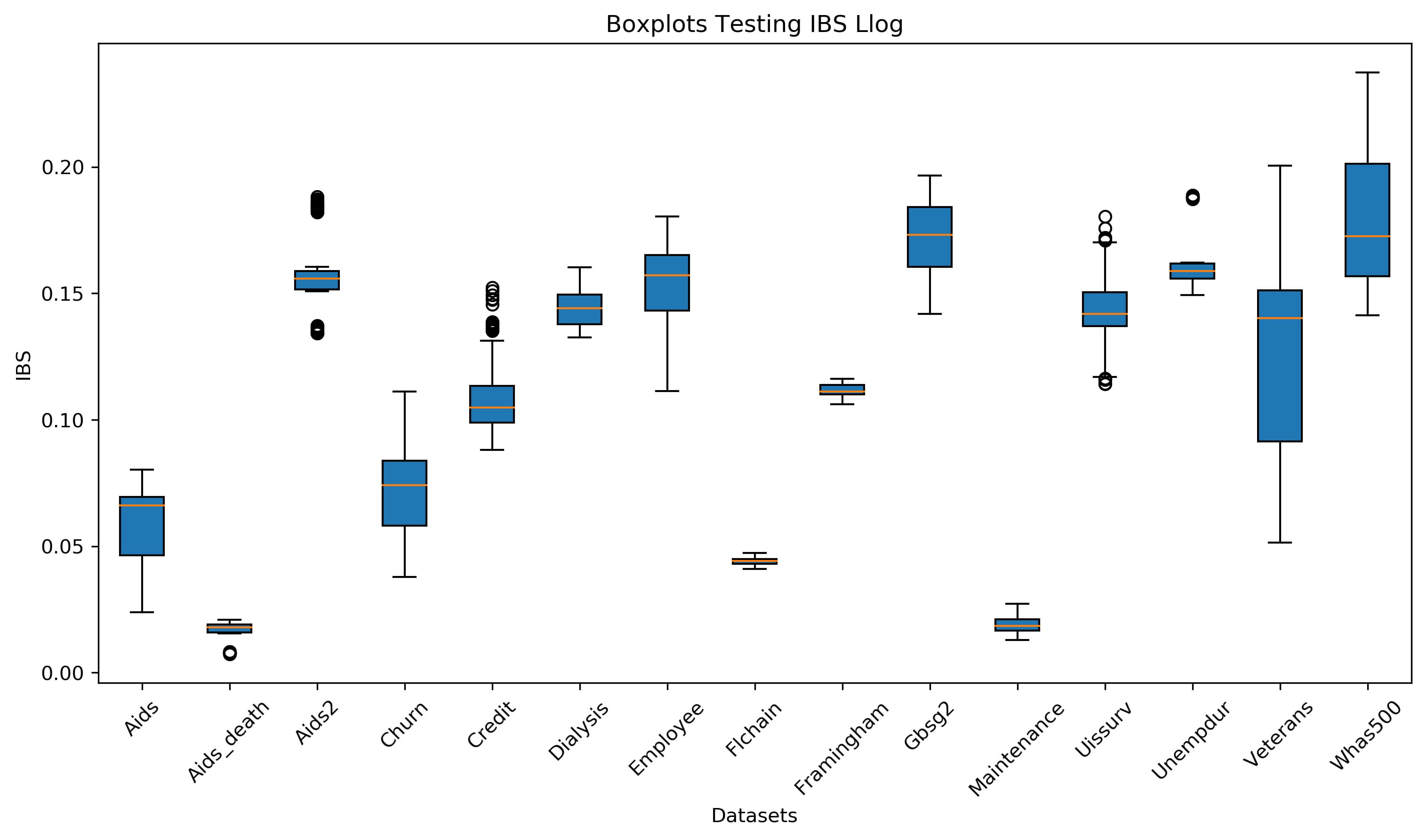}

    \label{fig:enter-label}
\end{figure}

\begin{figure}[H]
    \centering
    \includegraphics[width=0.7\linewidth]{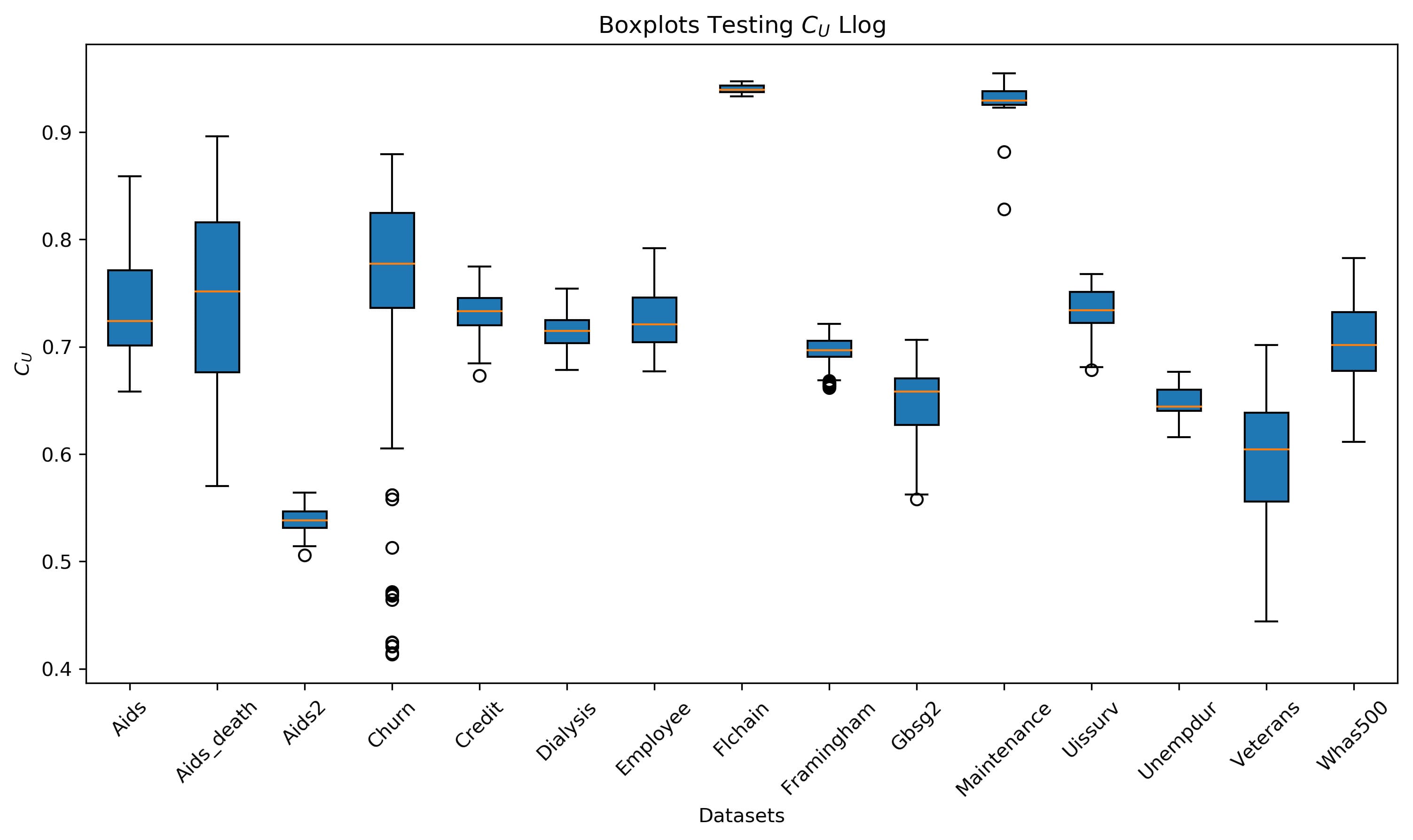}

    \label{fig:enter-label}
\end{figure}

\begin{figure}[H]
    \centering
    \includegraphics[width=0.7\linewidth]{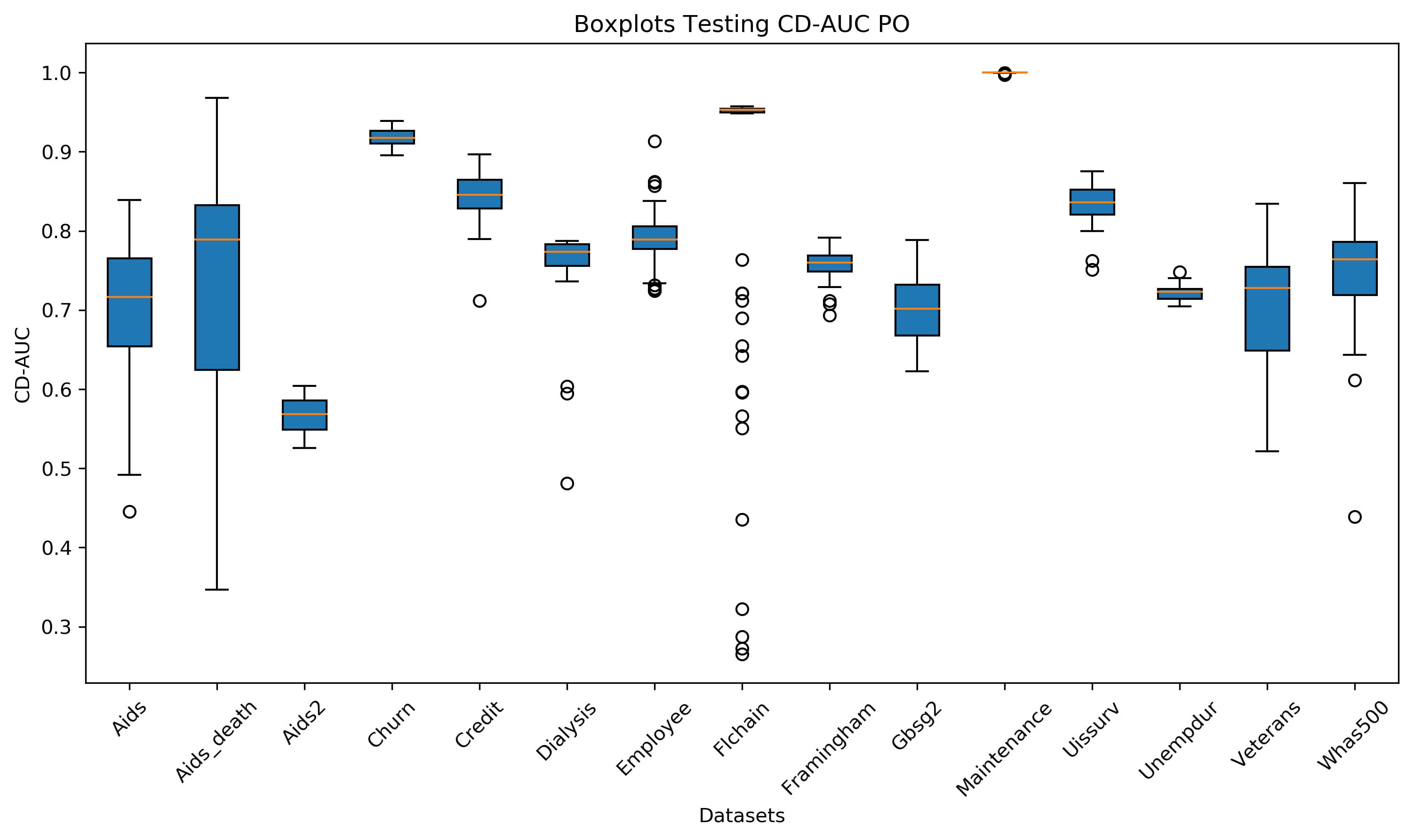}

    \label{fig:enter-label}
\end{figure}

\begin{figure}[H]
    \centering
    \includegraphics[width=0.7\linewidth]{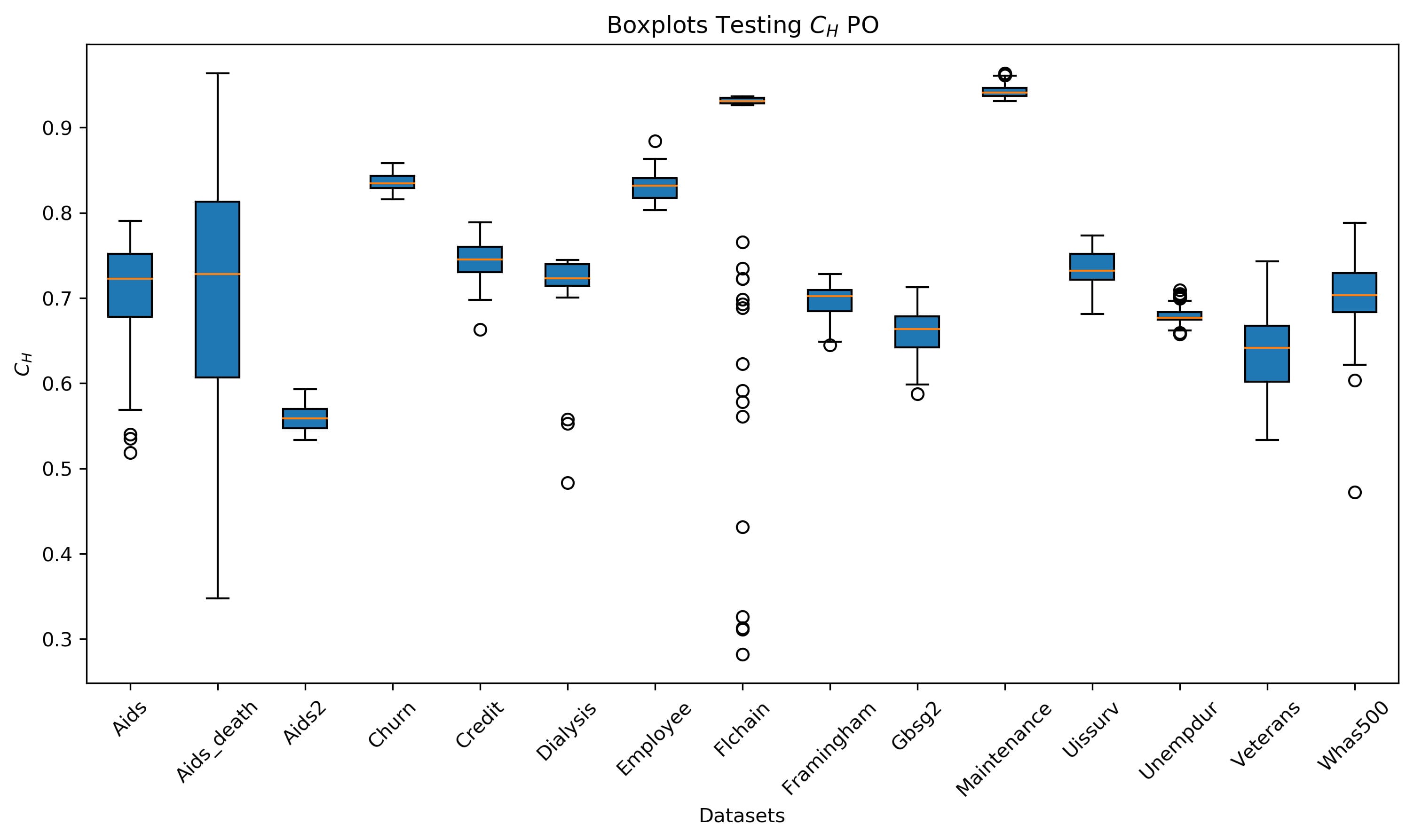}

    \label{fig:enter-label}
\end{figure}

\begin{figure}[H]
    \centering
    \includegraphics[width=0.7\linewidth]{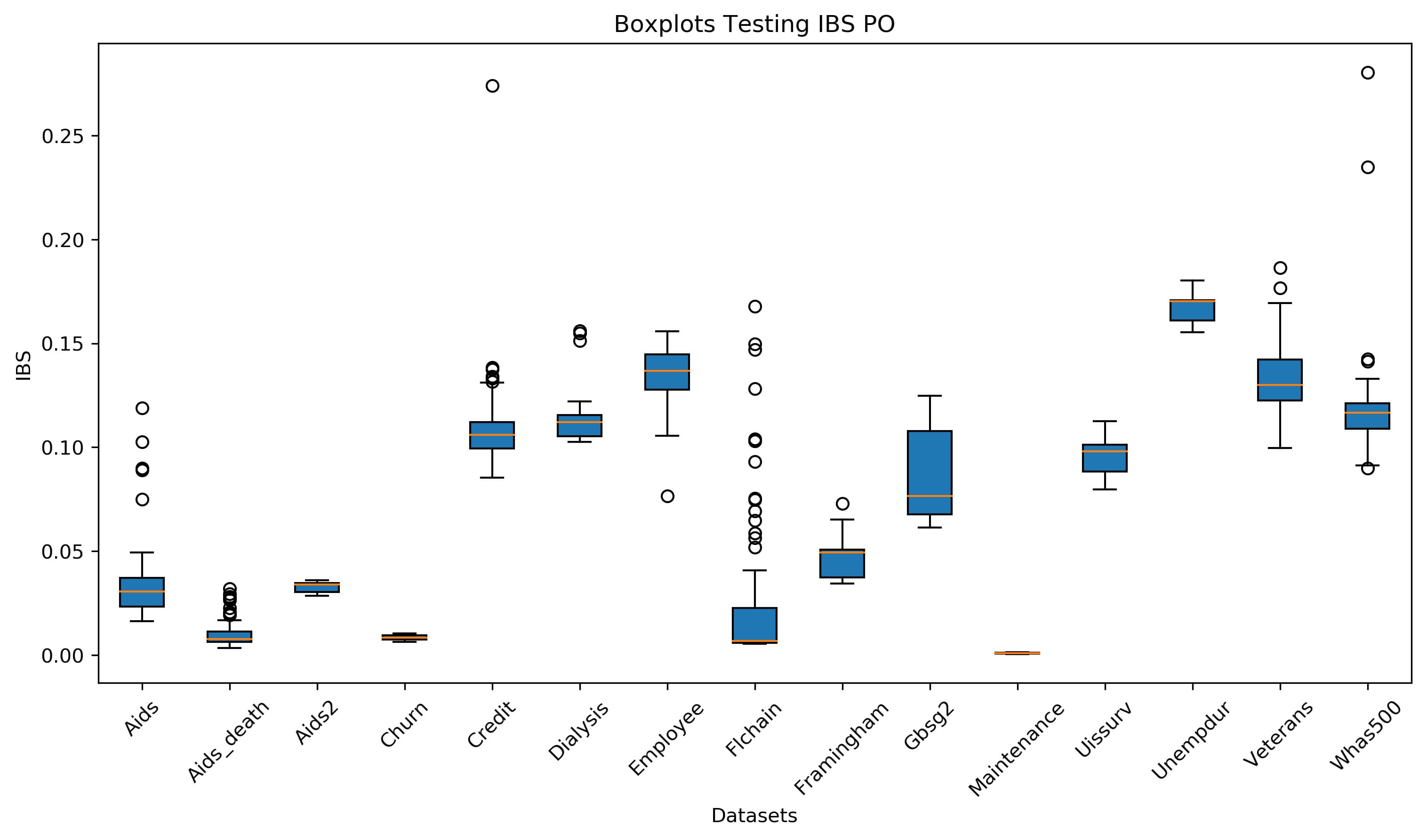}

    \label{fig:enter-label}
\end{figure}

\begin{figure}[H]
    \centering
    \includegraphics[width=0.7\linewidth]{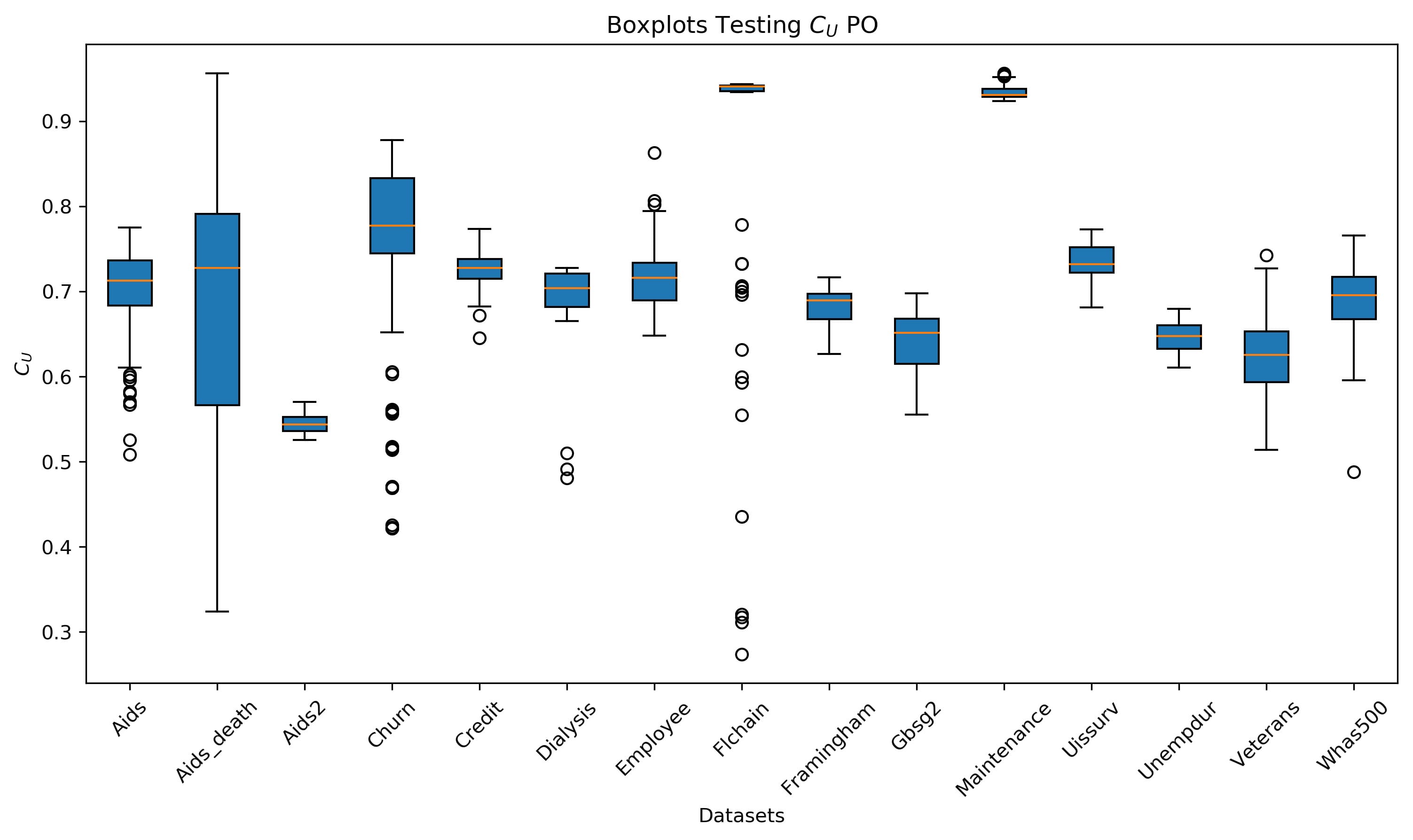}

    \label{fig:enter-label}
\end{figure}

\begin{figure}[H]
    \centering
    \includegraphics[width=0.7\linewidth]{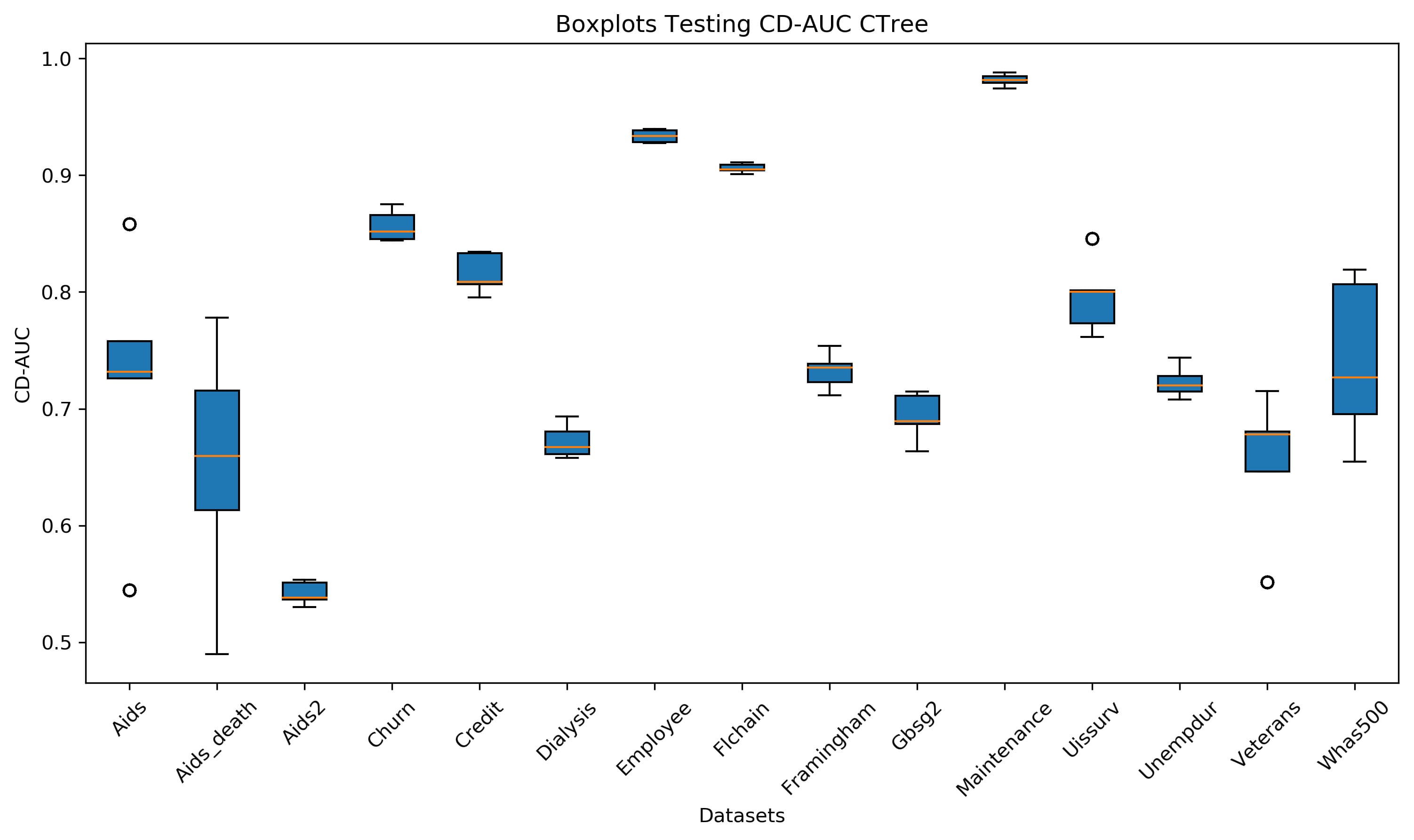}

    \label{fig:enter-label}
\end{figure}

\begin{figure}[H]
    \centering
    \includegraphics[width=0.7\linewidth]{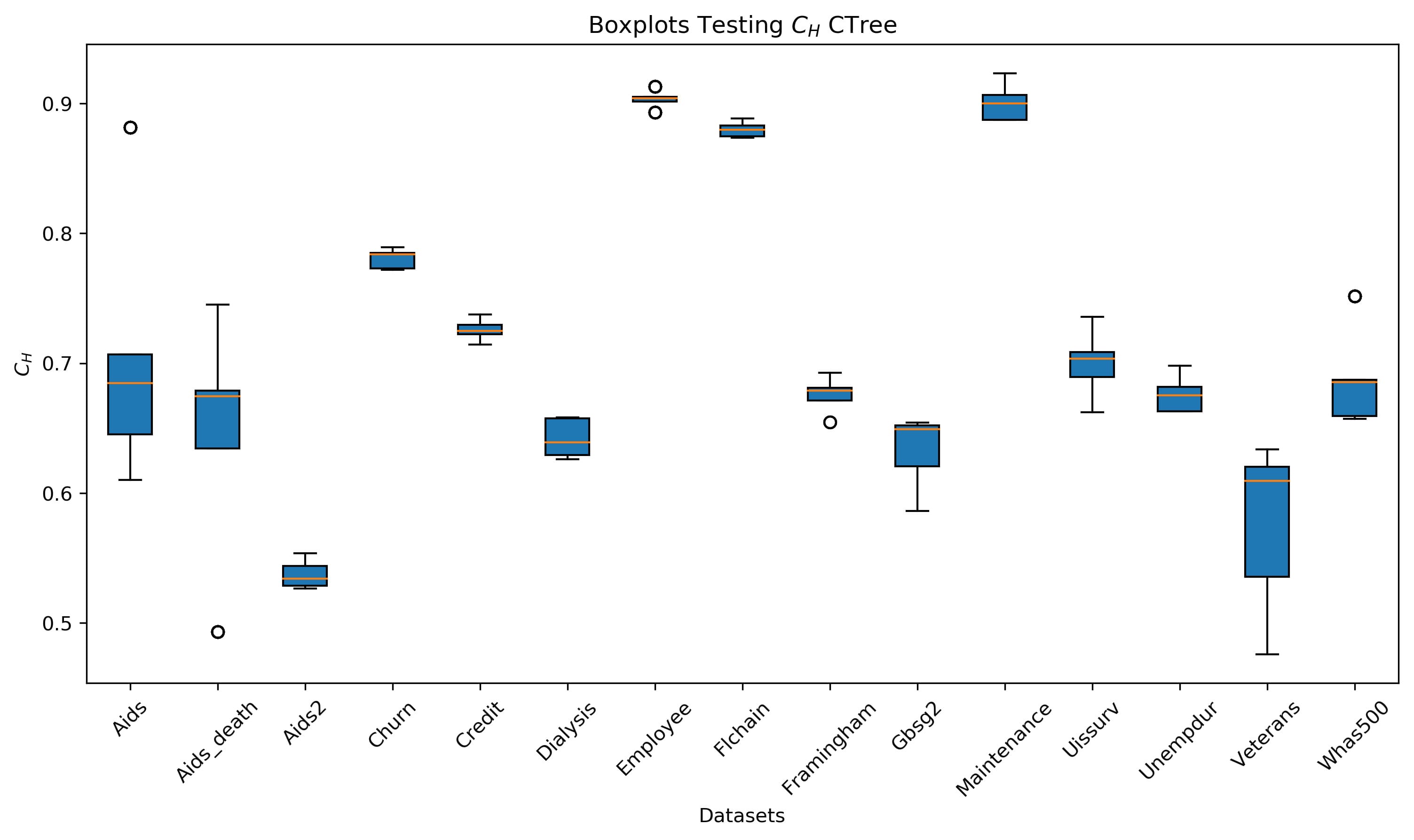}

    \label{fig:enter-label}
\end{figure}

\begin{figure}[H]
    \centering
    \includegraphics[width=0.7\linewidth]{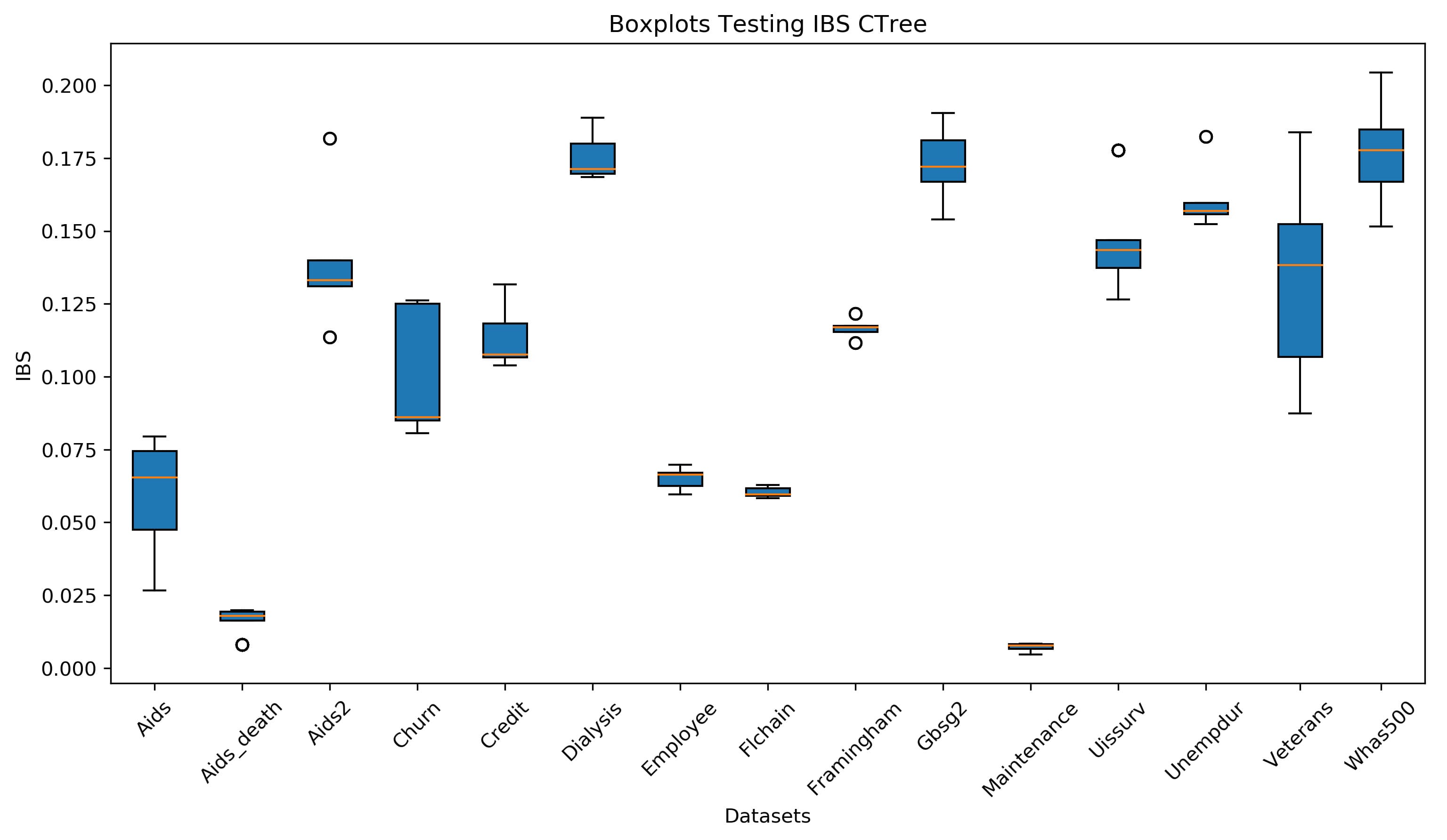}
    
    \label{fig:enter-label}
\end{figure}

\begin{figure}[H]
    \centering
    \includegraphics[width=0.7\linewidth]{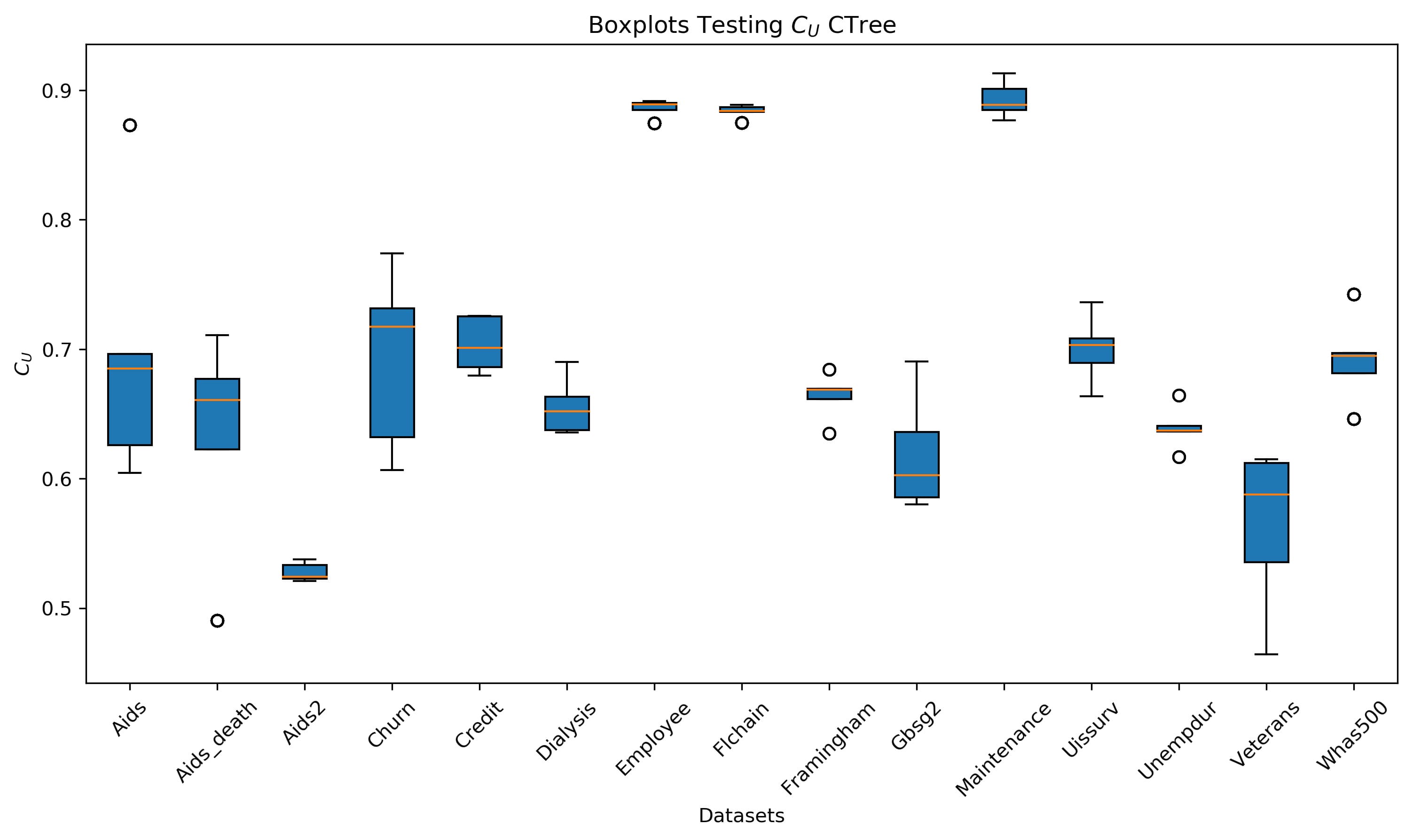}

    \label{fig:enter-label}
\end{figure}
\section{Interpretability results}\label{sec:additional_explain}

This appendix provides illustrative examples to highlight the capability of SSTs in enhancing interpretability by exploring differences in the survival functions of individual data points, both between and within leaf node clusters. 

Figure \ref{fig:explain_app} presents the survival functions distribution from a single run (a single fold and initial solution) on the Aids$\_$death, Aids, Churn, Credit, Gbsg2, Uissurv, and Whas500 datasets. The results for depth $D=1$ are shown in the first column, while those for depth $D=2$ are displayed in the second column. Colors are used to identify specific leaf nodes, with varying shades within each node illustrating differences in the survival functions among data points.

As discussed in Section \ref{sec:interpretability}, it is clear that in many other datasets, survival functions are readily distinguishable at both depths, not just between leaf nodes but also within them, where distinct patterns can be observed among individual data points. For instance, the Churn, Credit, Gbsg2, Veterans, and Uissurv datasets highlight how leaf nodes generate survival functions with varying shapes and scales.

\begin{figure}[htbp]
    \centering
    % Titoli sopra le colonne
    \makebox[0.43\textwidth]{\textbf{D=1}}
    \hfill
    \makebox[0.53\textwidth]{\textbf{D=2}}
    \begin{minipage}[b]{0.48\textwidth}
        \centering
        \includegraphics[width=\textwidth]{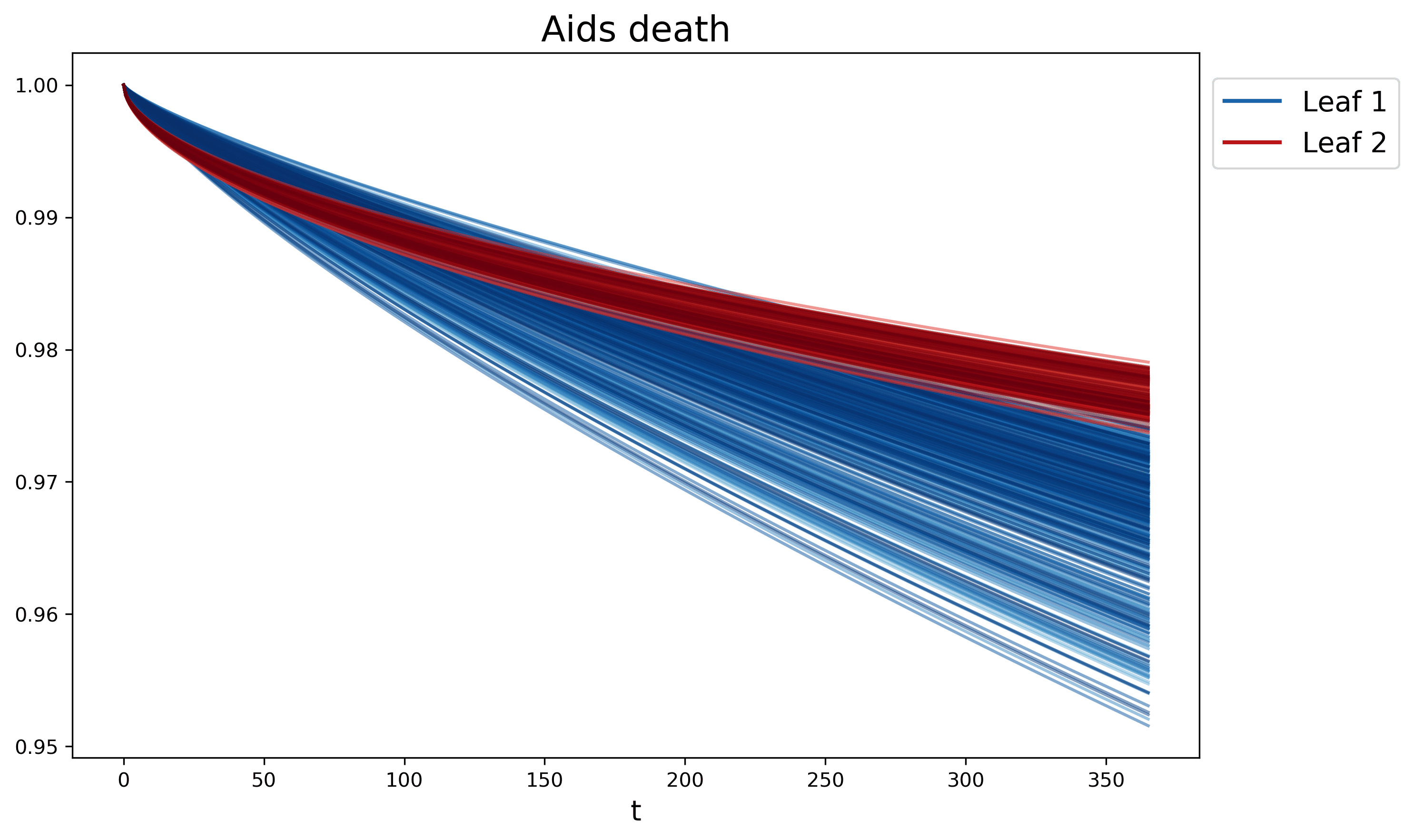}
    \end{minipage}
    \hfill
    \begin{minipage}[b]{0.48\textwidth}
        \centering
        \includegraphics[width=\textwidth]{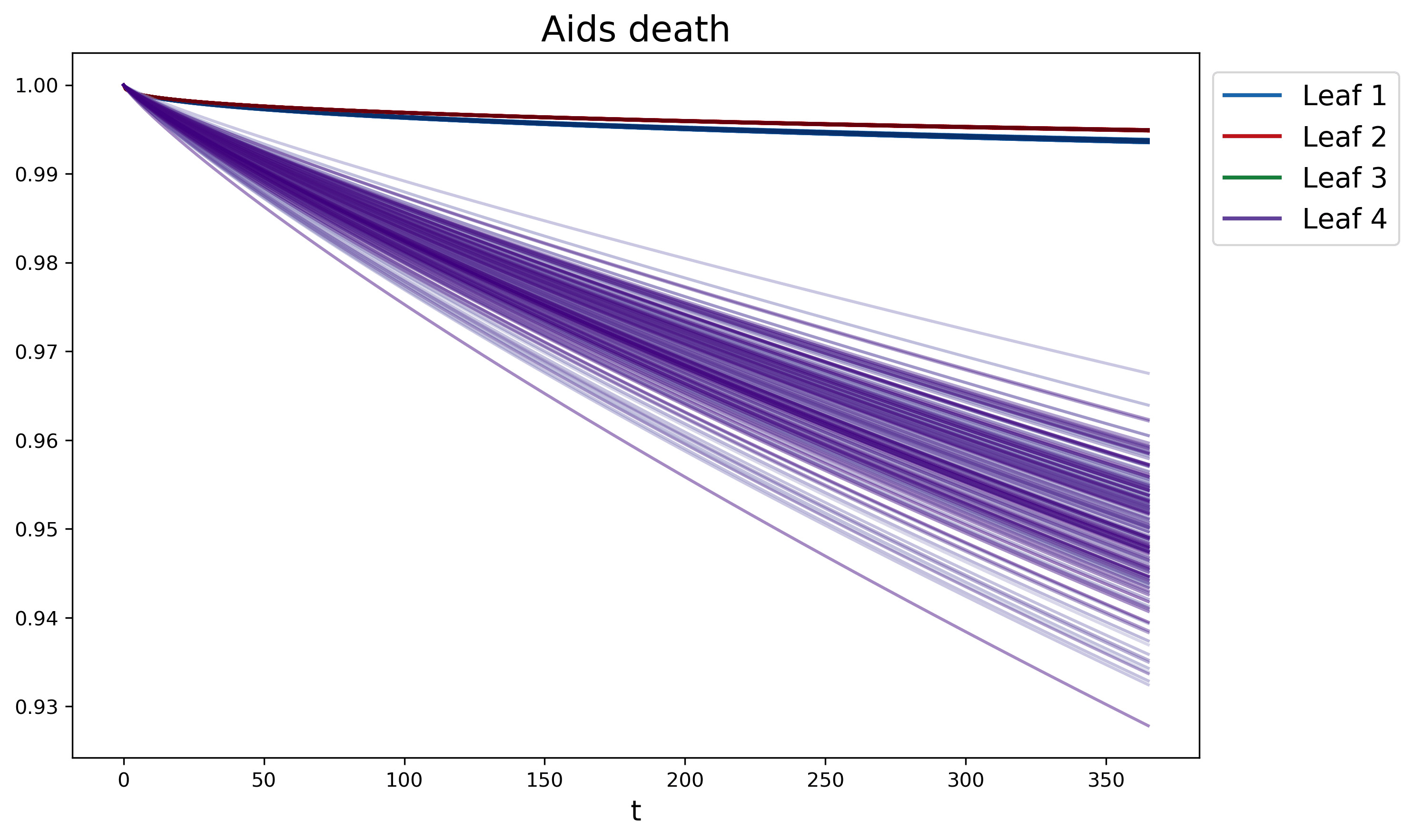}
    \end{minipage}
    \vspace{1em}
    \begin{minipage}[b]{0.48\textwidth}
        \centering
        \includegraphics[width=\textwidth]{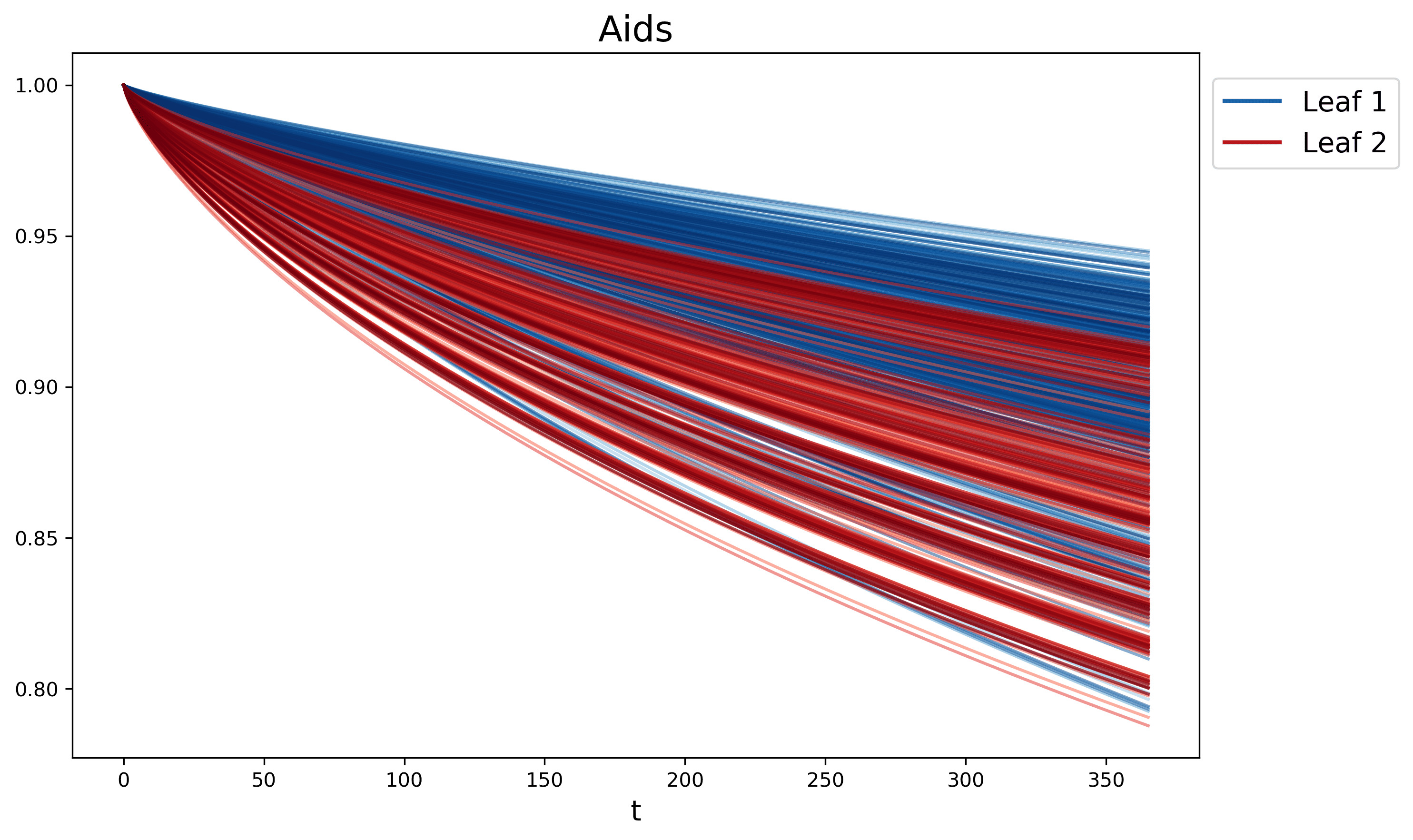}
    \end{minipage}
    \hfill
    \begin{minipage}[b]{0.48\textwidth}
        \centering
        \includegraphics[width=\textwidth]{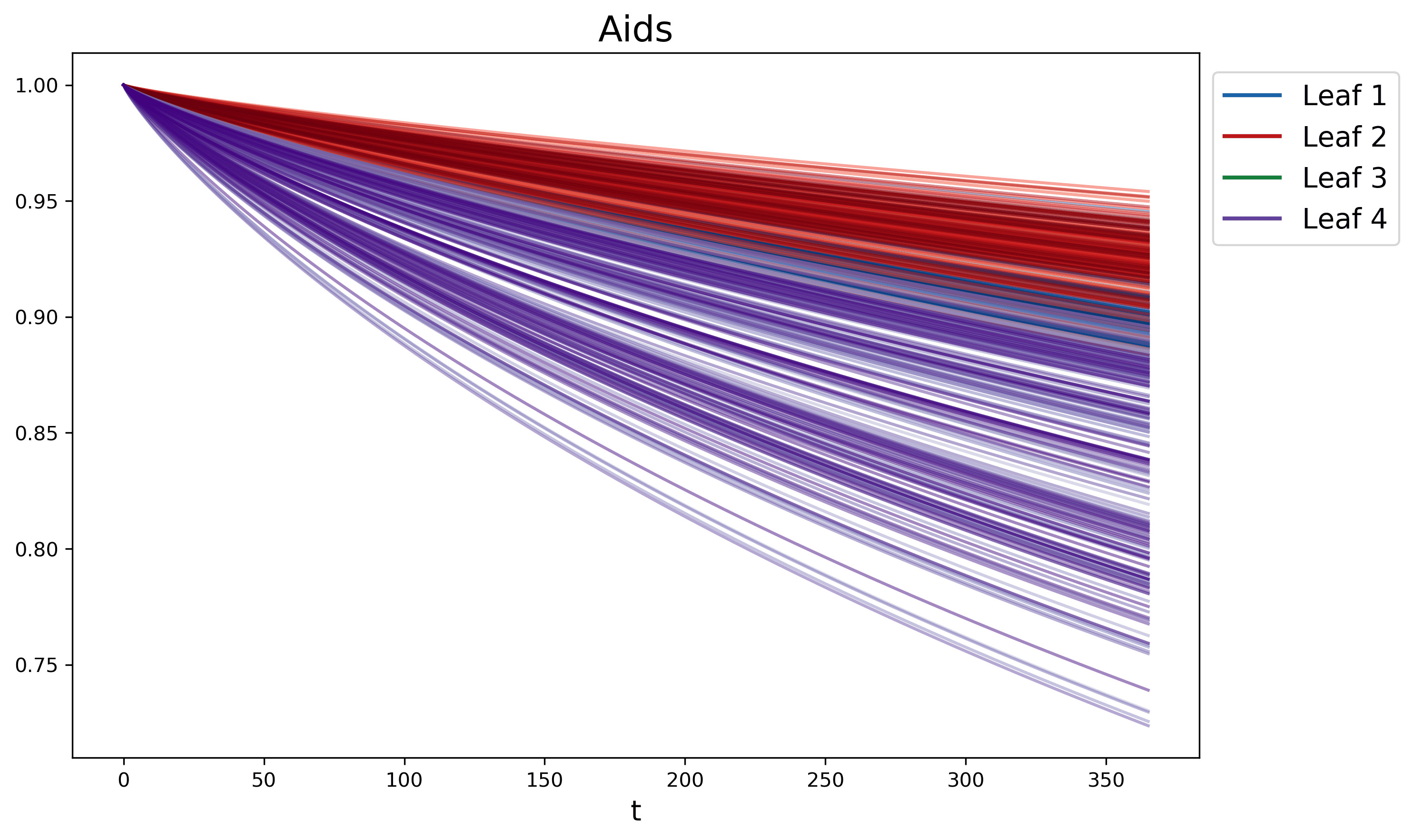}
    \end{minipage}
        \vspace{1em}
    \iffalse
    \begin{minipage}[b]{0.48\textwidth}
        \centering
        \includegraphics[width=\textwidth]{images_explain/churn_risk_dataset_d1_nuovo.jpg}
    \end{minipage}
    \hfill
    \begin{minipage}[b]{0.48\textwidth}
        \centering
        \includegraphics[width=\textwidth]{images_explain/churn_dataset_d2_attempt_1_nuovo.jpg}
    \end{minipage}
    \vspace{1em}
    \fi
    \begin{minipage}[b]{0.48\textwidth}
        \centering
        \includegraphics[width=\textwidth]{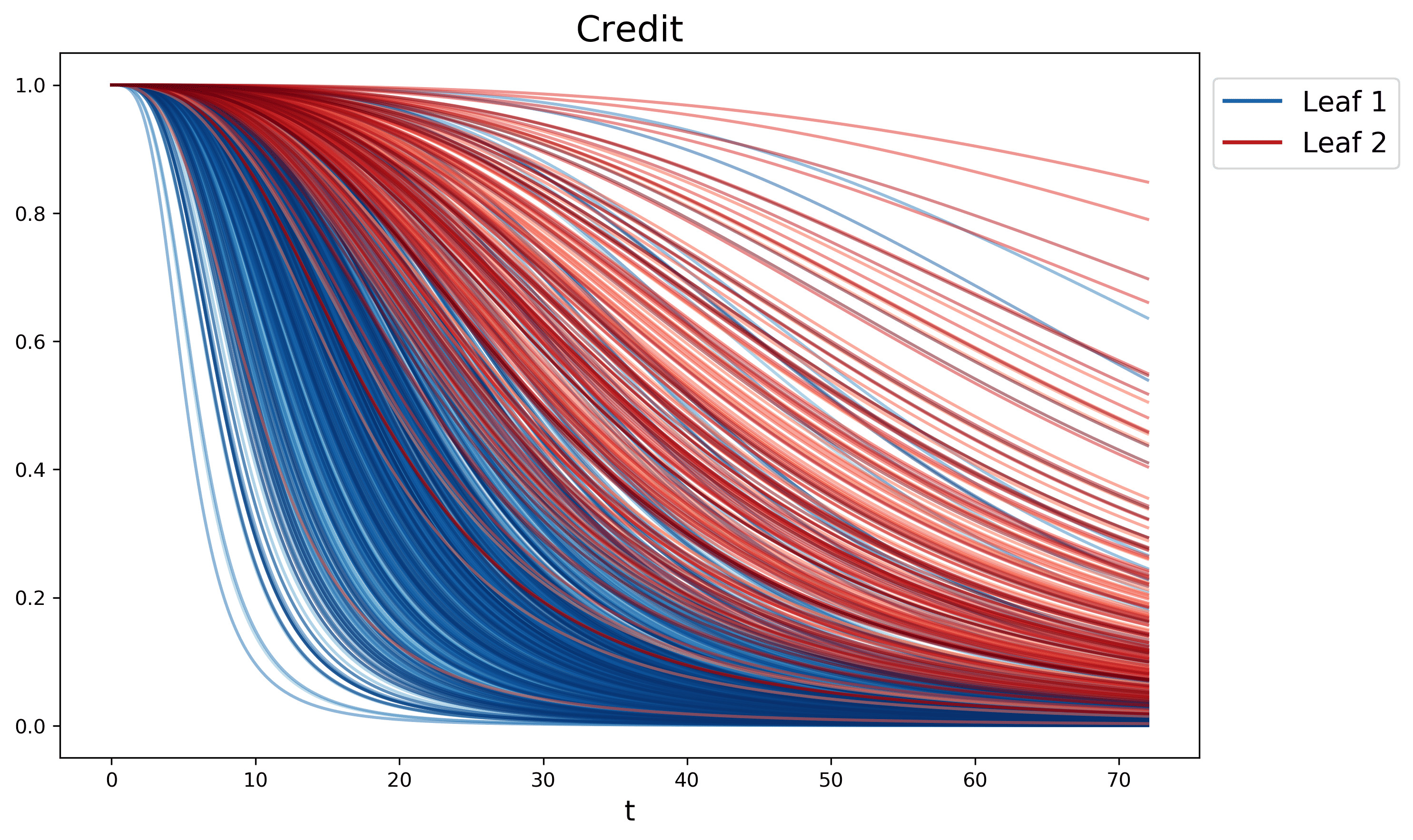}
    \end{minipage}
    \hfill
    \begin{minipage}[b]{0.48\textwidth}
        \centering
        \includegraphics[width=\textwidth]{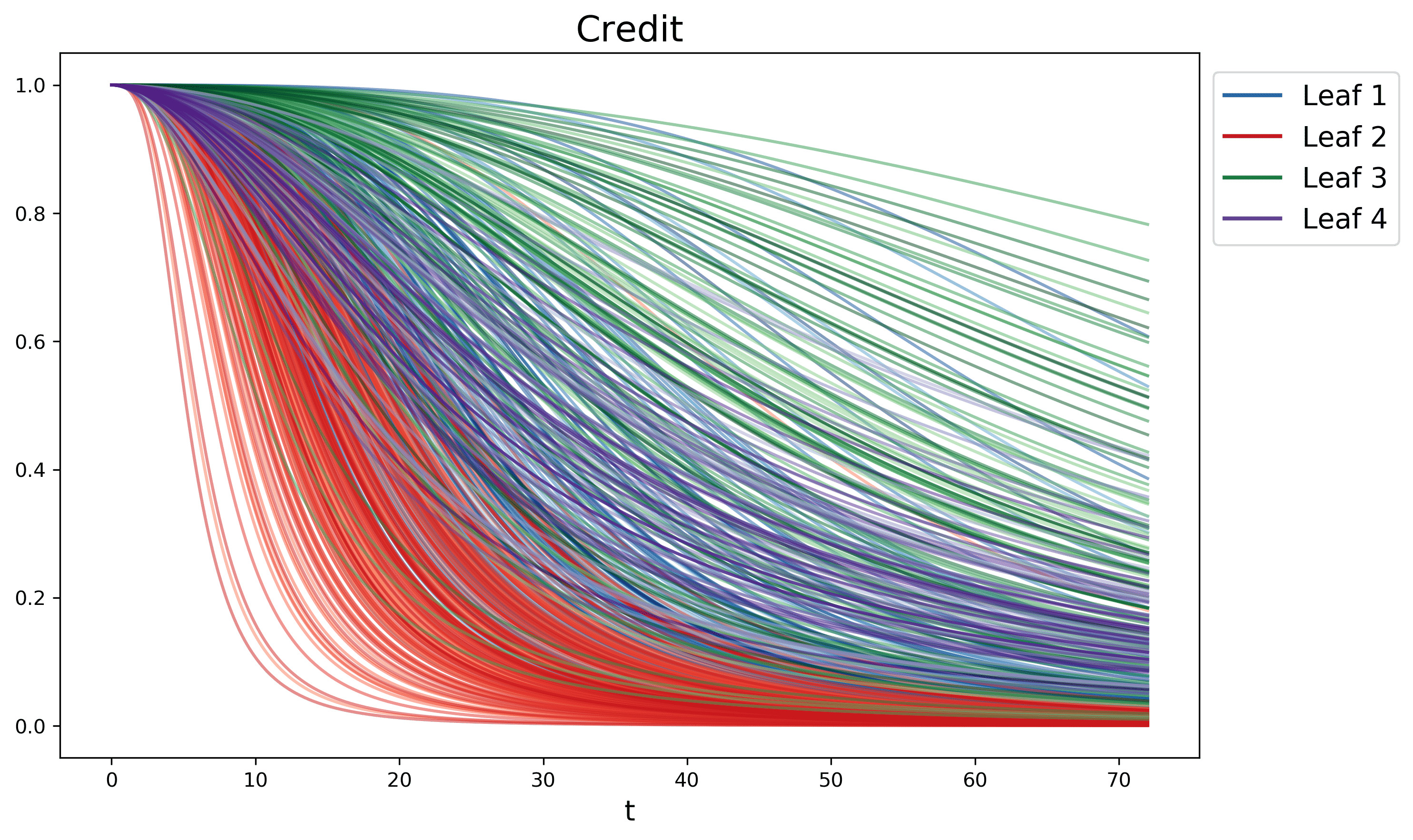}
    \end{minipage}
    \vspace{1em}
    \begin{minipage}[b]{0.48\textwidth}
        \centering
        \includegraphics[width=\textwidth]{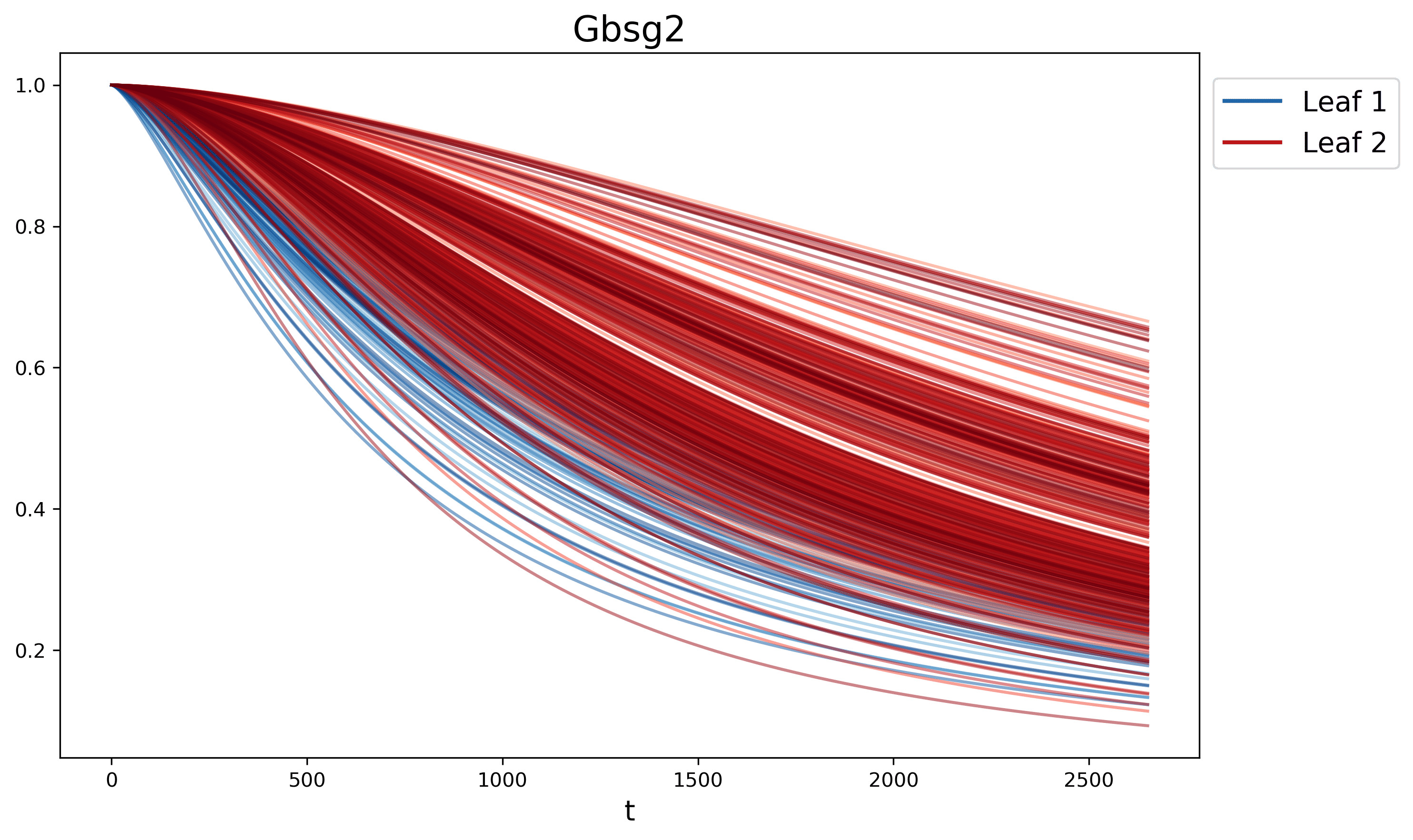}
    \end{minipage}
    \hfill
    \begin{minipage}[b]{0.48\textwidth}
        \centering
        \includegraphics[width=\textwidth]{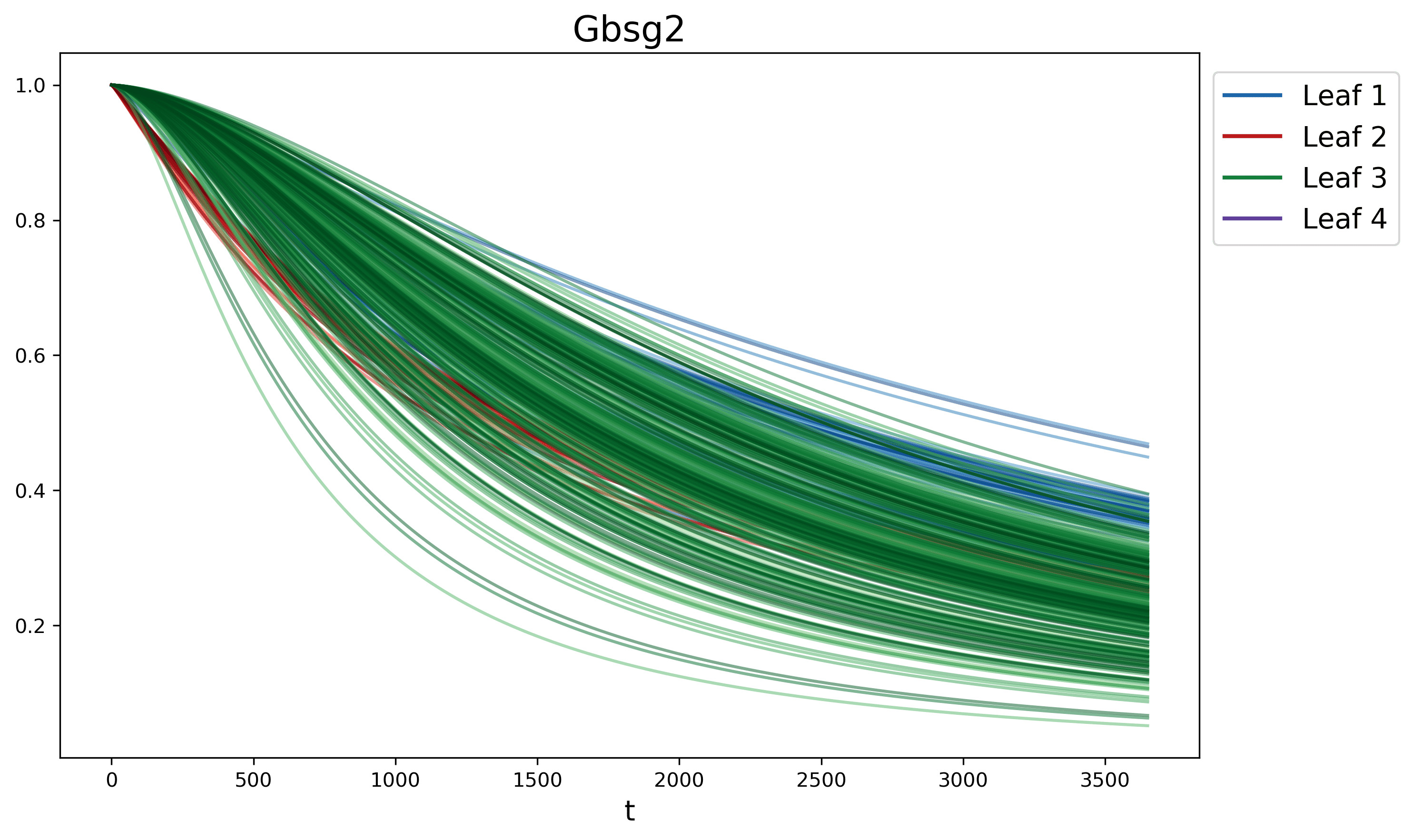}
    \end{minipage}
    \vspace{1em}
    \begin{minipage}[b]{0.48\textwidth}
        \centering
        \includegraphics[width=\textwidth]{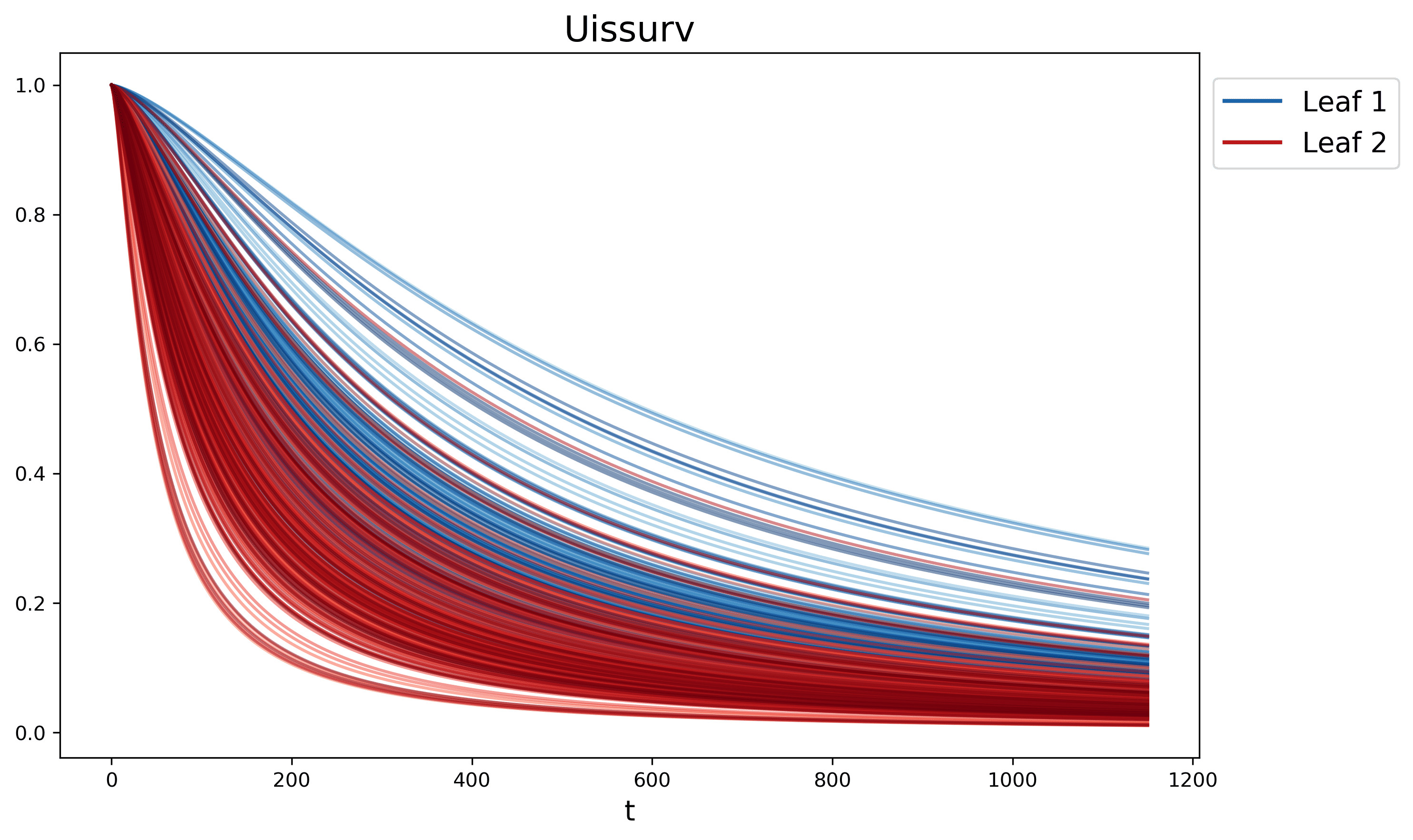}
    \end{minipage}
    \hfill
    \begin{minipage}[b]{0.48\textwidth}
        \centering
        \includegraphics[width=\textwidth]{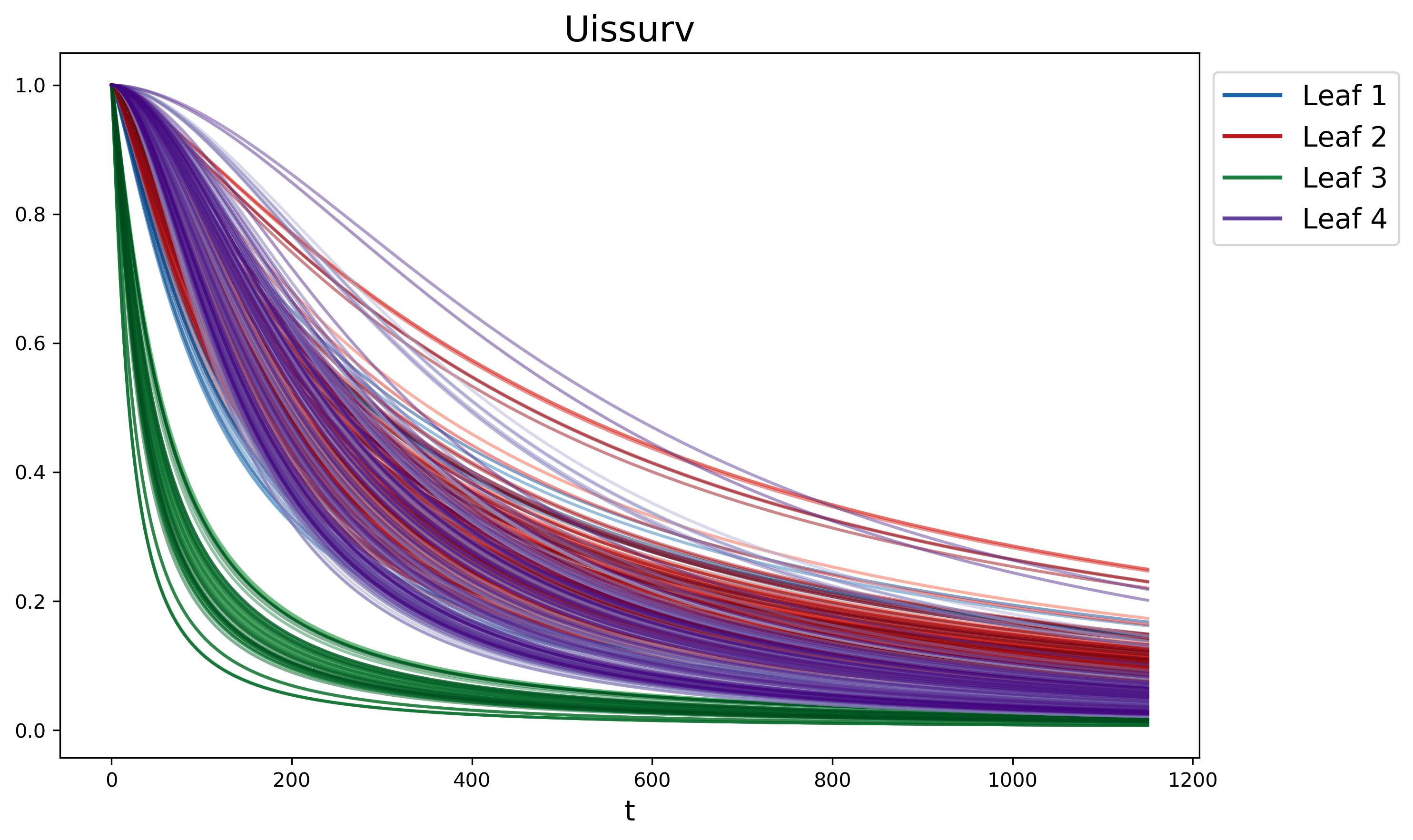}
    \end{minipage}
    \end{figure}
    \newpage
    \begin{figure}
        \centering
        % Titoli sopra le colonne
    \makebox[0.43\textwidth]{\textbf{D=1}}
    \hfill
    \makebox[0.53\textwidth]{\textbf{D=2}}
    %%%%%%%%%%%%%%%%%%%%%%%%%%%%%%%
    \iffalse
    \begin{minipage}[b]{0.48\textwidth}
        \centering
        \includegraphics[width=\textwidth]{images_explain/maintenance_dataset_d1_nuovo.jpg}
    \end{minipage}
    \hfill
    \begin{minipage}[b]{0.48\textwidth}
        \centering
        \includegraphics[width=\textwidth]{images_explain/maintenance_dataset_d2_attempt_1_nuovo.jpg}
    \end{minipage}
    \vspace{1em}
    \begin{minipage}[b]{0.48\textwidth}
        \centering
        \includegraphics[width=\textwidth]{images_explain/lung_cancer_dataset_d1_nuovo.jpg}
    \end{minipage}
    \hfill
    \begin{minipage}[b]{0.48\textwidth}
        \centering
        \includegraphics[width=\textwidth]{images_explain/lung_cancer_dataset_d2_attempt_1_nuovo.jpg}
    \end{minipage}
    \vspace{1em}
    \fi
    %%%%%%%%%%%%%%%%%%%%%%%%%%%%%%%%
    \iffalse
    \begin{minipage}[b]{0.48\textwidth}
        \centering
        \includegraphics[width=\textwidth]{images_explain/uis_dataset_d1_nuovo.jpg}
    \end{minipage}
    \hfill
    \begin{minipage}[b]{0.48\textwidth}
        \centering
        \includegraphics[width=\textwidth]{images_explain/uis_dataset_d2_attempt_1_nuovo.jpg}
    \end{minipage}
    \vspace{1em}
    \fi
    \begin{minipage}[b]{0.48\textwidth}
        \centering
        \includegraphics[width=\textwidth]{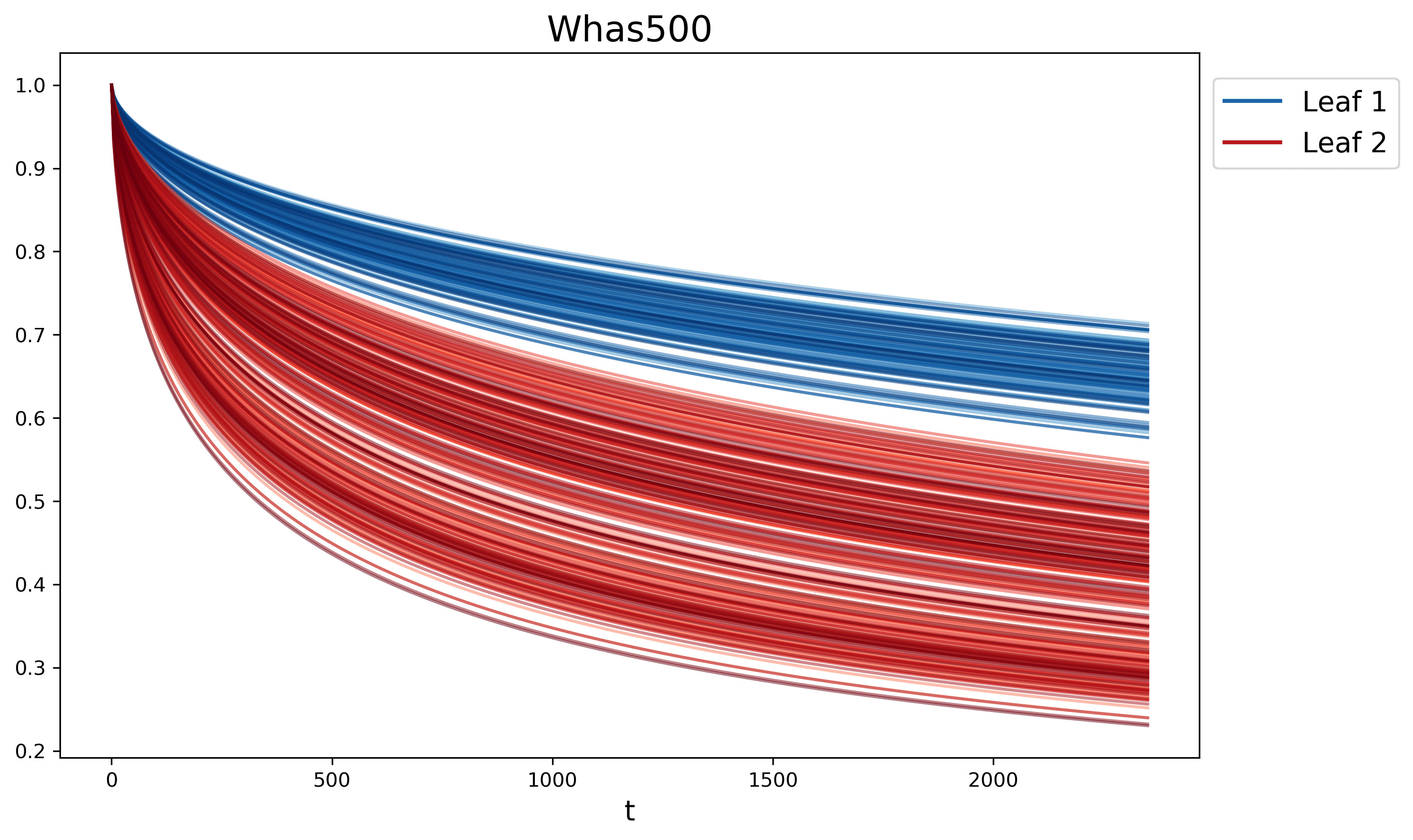}
    \end{minipage}
    \hfill
    \begin{minipage}[b]{0.48\textwidth}
        \centering
        \includegraphics[width=\textwidth]{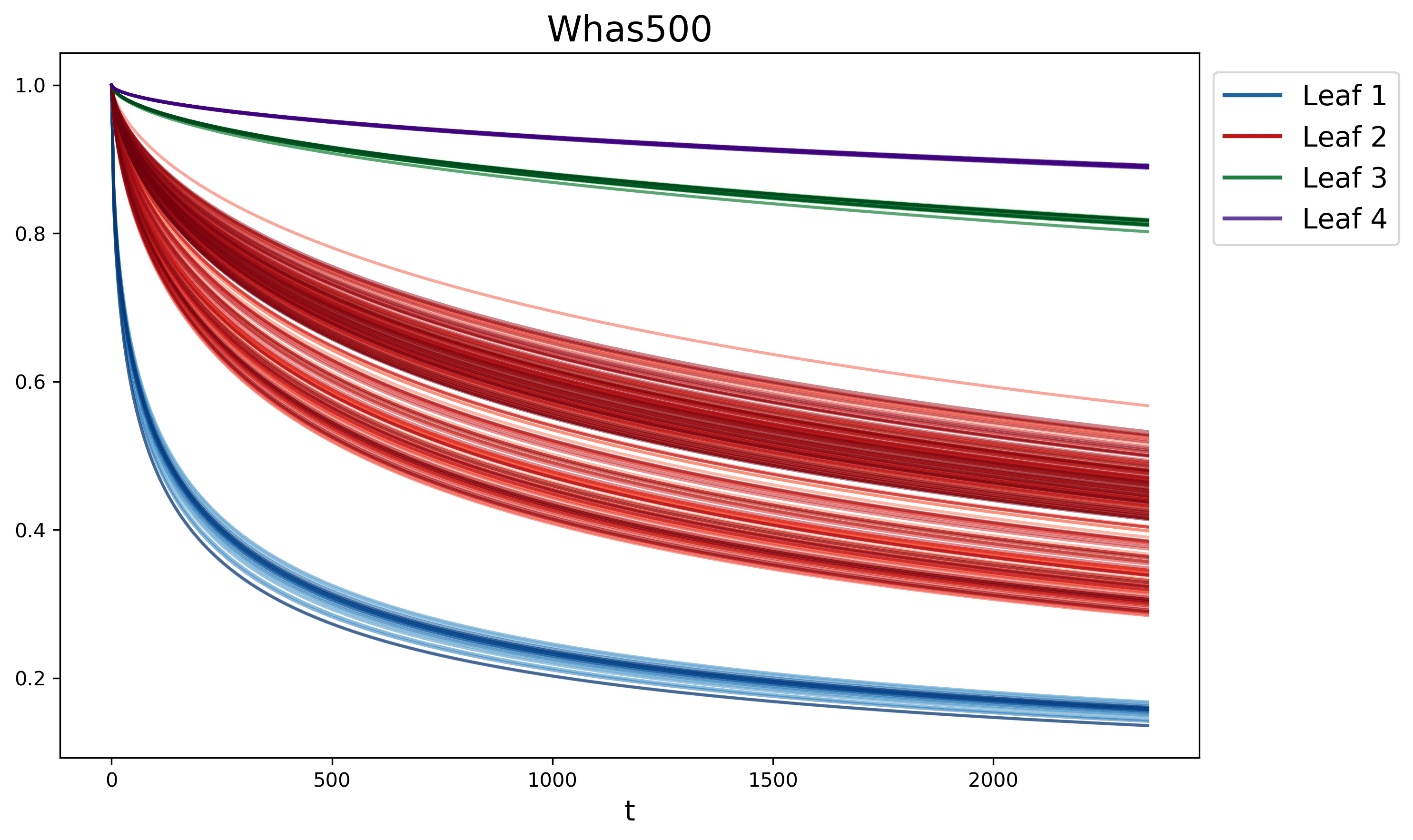}
    \end{minipage}
    
    \caption{The survival function distributions for a single run across the Aids$\_$death, Aids, Churn, Credit, Gbsg2, and Whas500 datasets. The results for depth $D=1$ are shown on the left, while those for $D=2$ are on the right. Each color corresponds to a specific leaf node, with variations in shading within each node illustrating the differences in survival functions among individual data points.}\label{fig:explain_app}
\end{figure}

\section{Fairness results}\label{sec:additional_fair}
In this appendix we evaluate the performance of our fairness-promoting formulation \eqref{eq:formulation_fair} optimized using the node-based decomposition training algorithm on a socio-economic dataset \citep{romeo1999conducting} where a sensitive group is identifiable. Moreover, we report some plots to illustrate the impact on the survival function of the penalty term on the individual survival functions as $\rho$ grows. 

The $\rho$ hyperparameter begins at $0$ and increases to $\frac{20}{N_M N_F}$ in five steps ($\frac{1}{N_M N_F} [0,1,5,10,15,20]$), where $N_M$ and $N_F$ denote the number of males and females in the training dataset. 

The four plots in Figure \ref{fig:fair_measures} display the training and testing values of $C_H$, $C_U$, $\text{CD-AUC}$, and $\text{IBS}$, as well as the value of the fairness penalty term \eqref{eq:fair_term}\footnote{The calculation of the integral of the difference between the survival functions in \eqref{eq:fair_term} was carried out by summing the differences at each observed time in the training set}. In particular, the x-axis represents the value
of $\rho$, the left y-axis the performance measure values while the right y-axis
describes the training and testing fairness penalty value (i.e., the \textcolor{black}{squared} difference between survival functions belonging to the sensitive and complementary groups). The blue and green lines represent the performance measures for training and testing, respectively. The red and purple lines, on the other hand, depict the fairness penalty values for training and testing, respectively.

\begin{figure}[htbp]
    \centering
    \begin{minipage}[b]{0.48\textwidth}
        \centering
        \includegraphics[width=\textwidth]{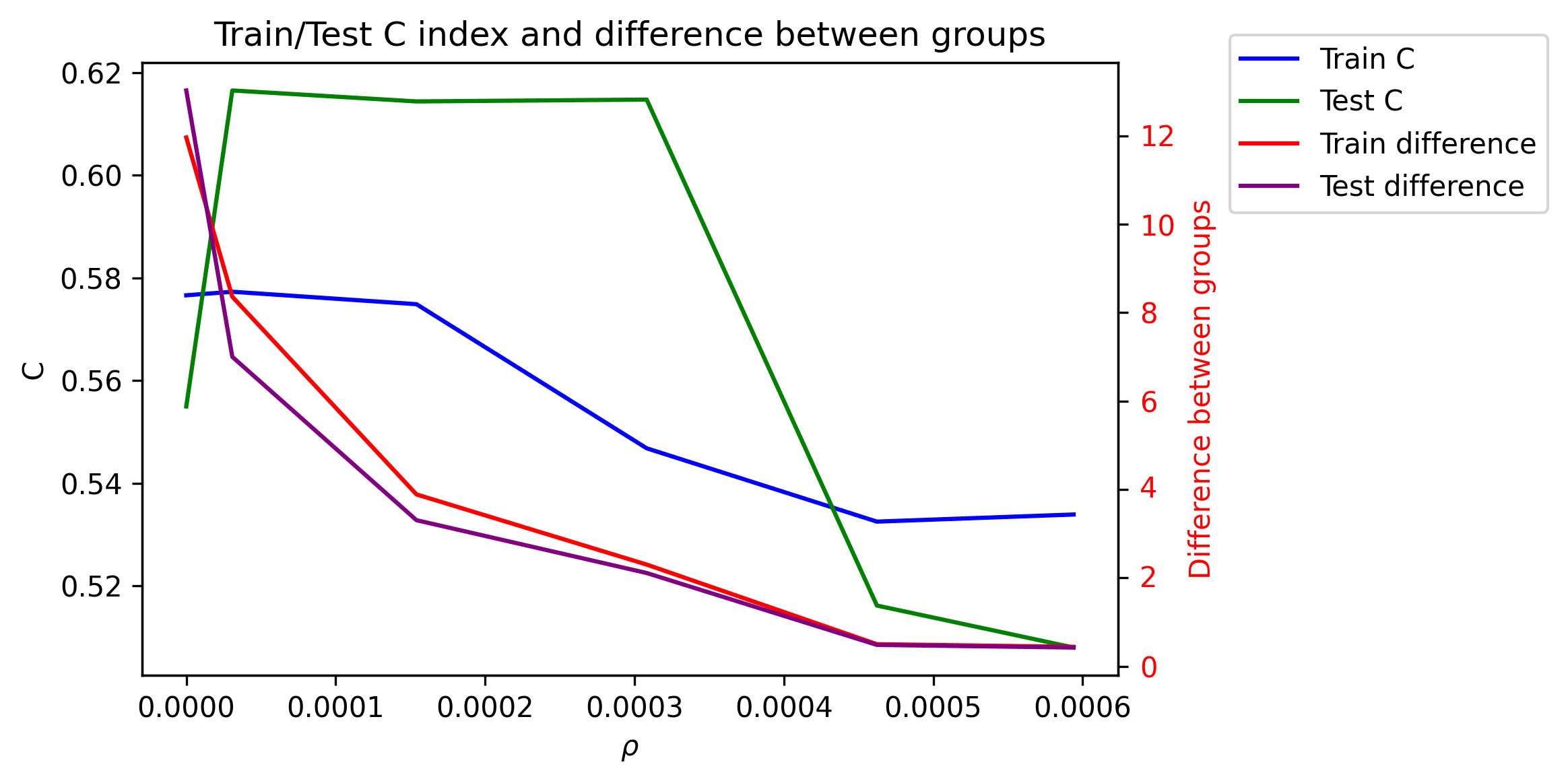}
    \end{minipage}
    \hfill
    \begin{minipage}[b]{0.48\textwidth}
        \centering
        \includegraphics[width=\textwidth]{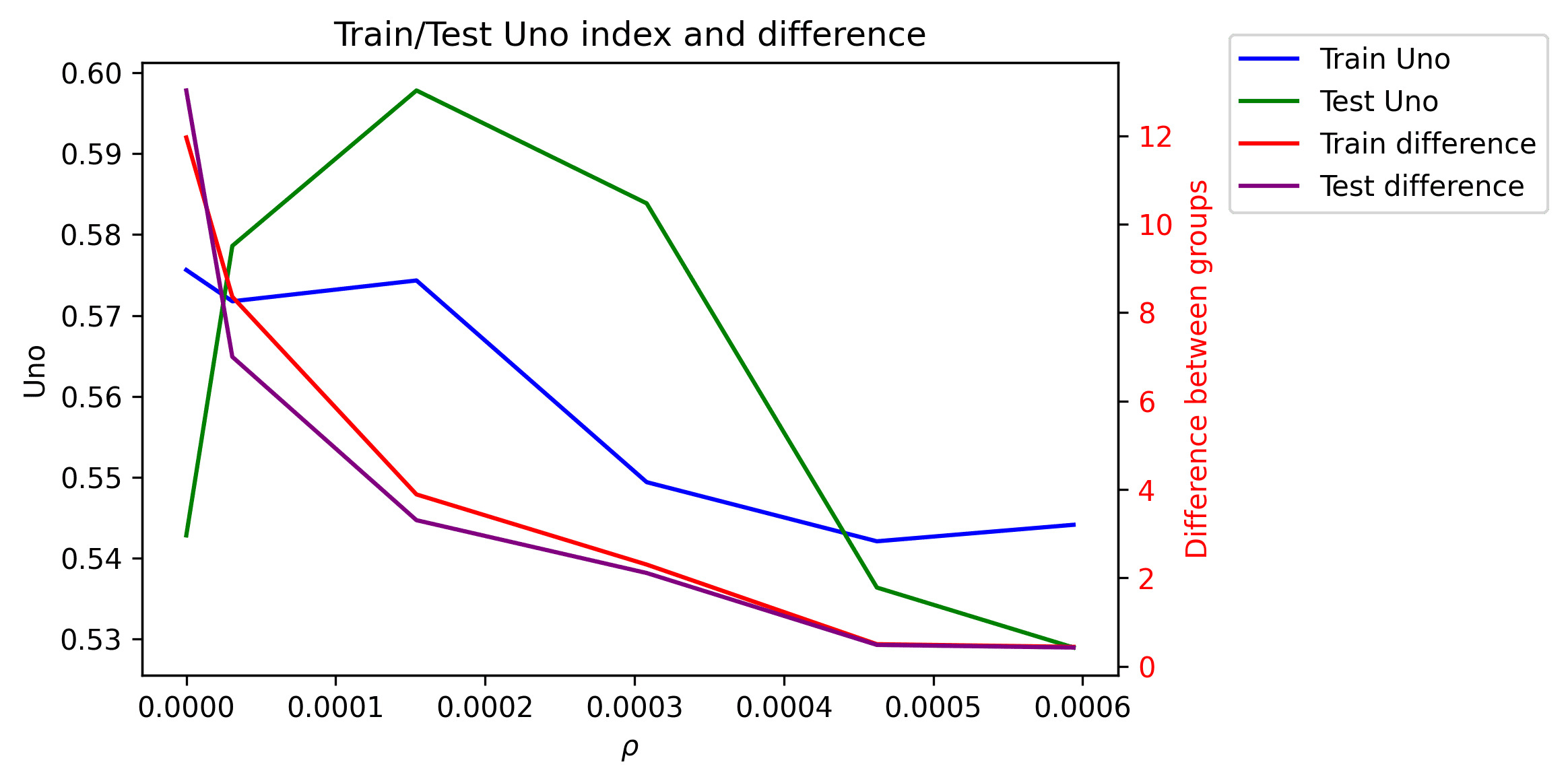}
    \end{minipage}
    \vspace{1em}
    \begin{minipage}[b]{0.48\textwidth}
        \centering
        \includegraphics[width=\textwidth]{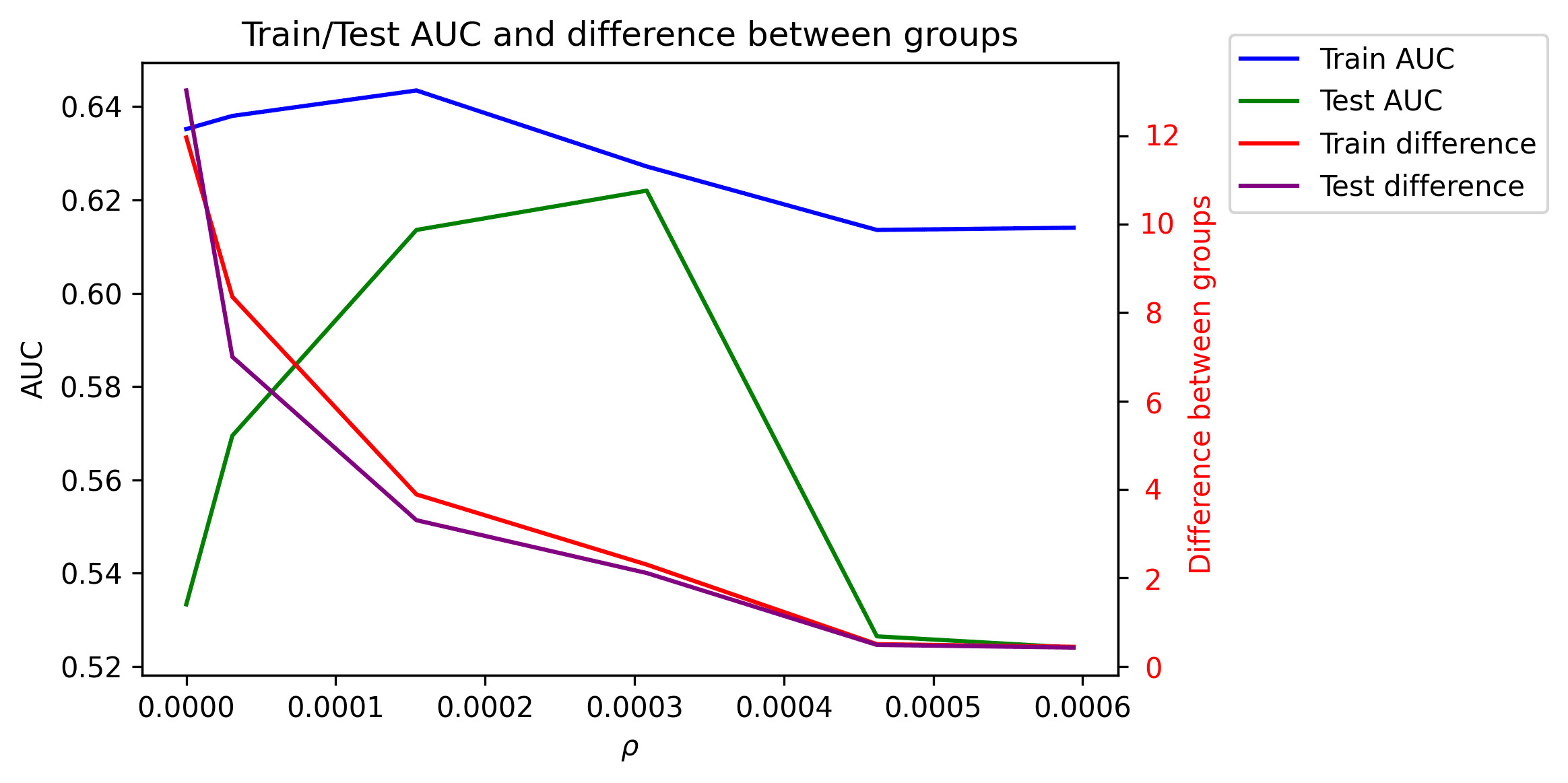}
    \end{minipage}
    \hfill
    \begin{minipage}[b]{0.48\textwidth}
        \centering
        \includegraphics[width=\textwidth]{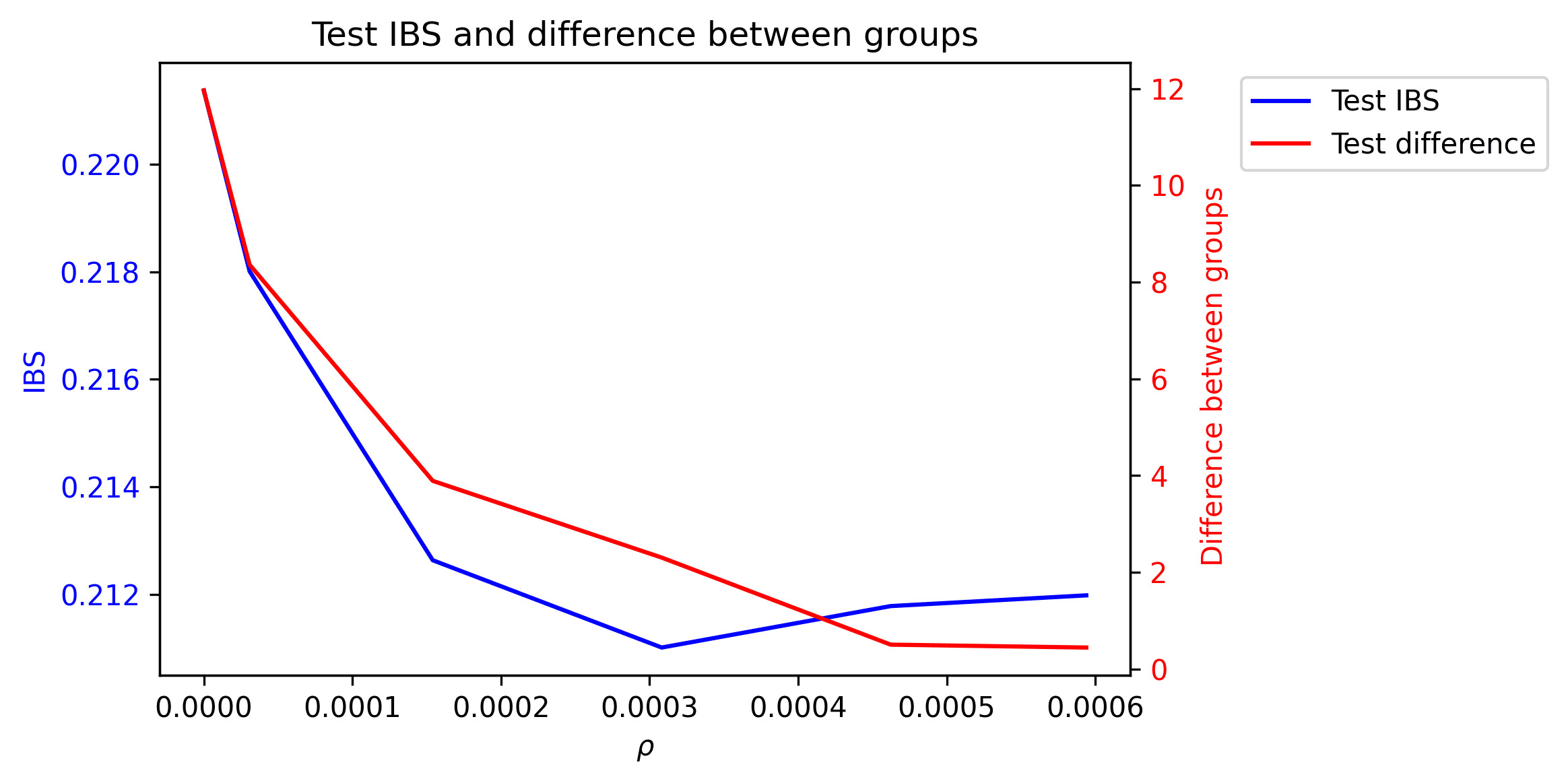}
    \end{minipage}
    \caption{Trends in the training (blue) and testing (green) accuracy measures $C_H$, $C_U$, $\text{CD-AUC}$, and IBS as the fairness penalty $\rho$ increases. The red and purple functions represent the differences between the survival functions for the two groups \eqref{eq:fair_term} on the training and testing sets, respectively.}\label{fig:fair_measures}
\end{figure}

In general, we see that the introduction of the penalty term has a positive impact on both training and testing group fairness by clearly reducing the distances between survival functions of the two groups. For the calibration measures $C_H$, $C_U$, and $\text{CD-AUC}$, it is clear that as $\rho$ increases, the training performance decreases. In contrast, the testing performance seems to improve for small values of $\rho$, before decreasing again.  It is noteworthy that for the green training function, there exists a value of $\rho > 0$ that results in a slight improvement. This apparently counterintuitive observation can be explained by the non-convexity of the problem, where the optimization process, starting from the same initial solution, may lead to different solutions depending on the value of $\rho$. Furthermore, the three calibration measures are proxies for the original maximum likelihood problem. Concerning IBS, the introduction of the penalty term leads to improvements in IBS testing for low values of $\rho$, followed by a rise in the value and subsequent decline in performance.

Figure \ref{fig:fair_appendix_complete} presents the survival functions for individual data points, where different leaf nodes are distinguished by varying colors. The left column depicts the training set, while the right column represents the testing set. From top to bottom, the plots illustrate how survival functions change with increasing values of $\rho$.

\begin{figure}[H]
    \centering
    % Prima riga
        % Titoli sopra le colonne
    \makebox[0.5\textwidth]{\textbf{Train}}
    \hfill
    \makebox[0.48\textwidth]{\textbf{Test}\textcolor{white}{aaaaaaaa}}
    \begin{subfigure}{0.45\textwidth}
        \centering
        \includegraphics[width=\linewidth]{immagini_nuove/unemployement_seed1039_rho0_reg_drop_train.jpg}
        %\caption{Figura 1}
    \end{subfigure}
    \begin{subfigure}{0.45\textwidth}
        \centering
        \includegraphics[width=\linewidth]{immagini_nuove/unemployement_seed1039_rho0_reg_drop_test.jpg}
        %\caption{Figura 2}
    \end{subfigure}
    
    % Seconda riga
    \begin{subfigure}{0.45\textwidth}
        \centering
        \includegraphics[width=\linewidth]{immagini_nuove/unemployement_seed1039_rho1_reg_drop_train.jpg}
        %\caption{Figura 3}
    \end{subfigure}
    \begin{subfigure}{0.45\textwidth}
        \centering
        \includegraphics[width=\linewidth]{immagini_nuove/unemployement_seed1039_rho1_reg_drop_test.jpg}
        %\caption{Figura 4}
    \end{subfigure}
    
    % Terza riga
    \begin{subfigure}{0.45\textwidth}
        \centering
        \includegraphics[width=\linewidth]{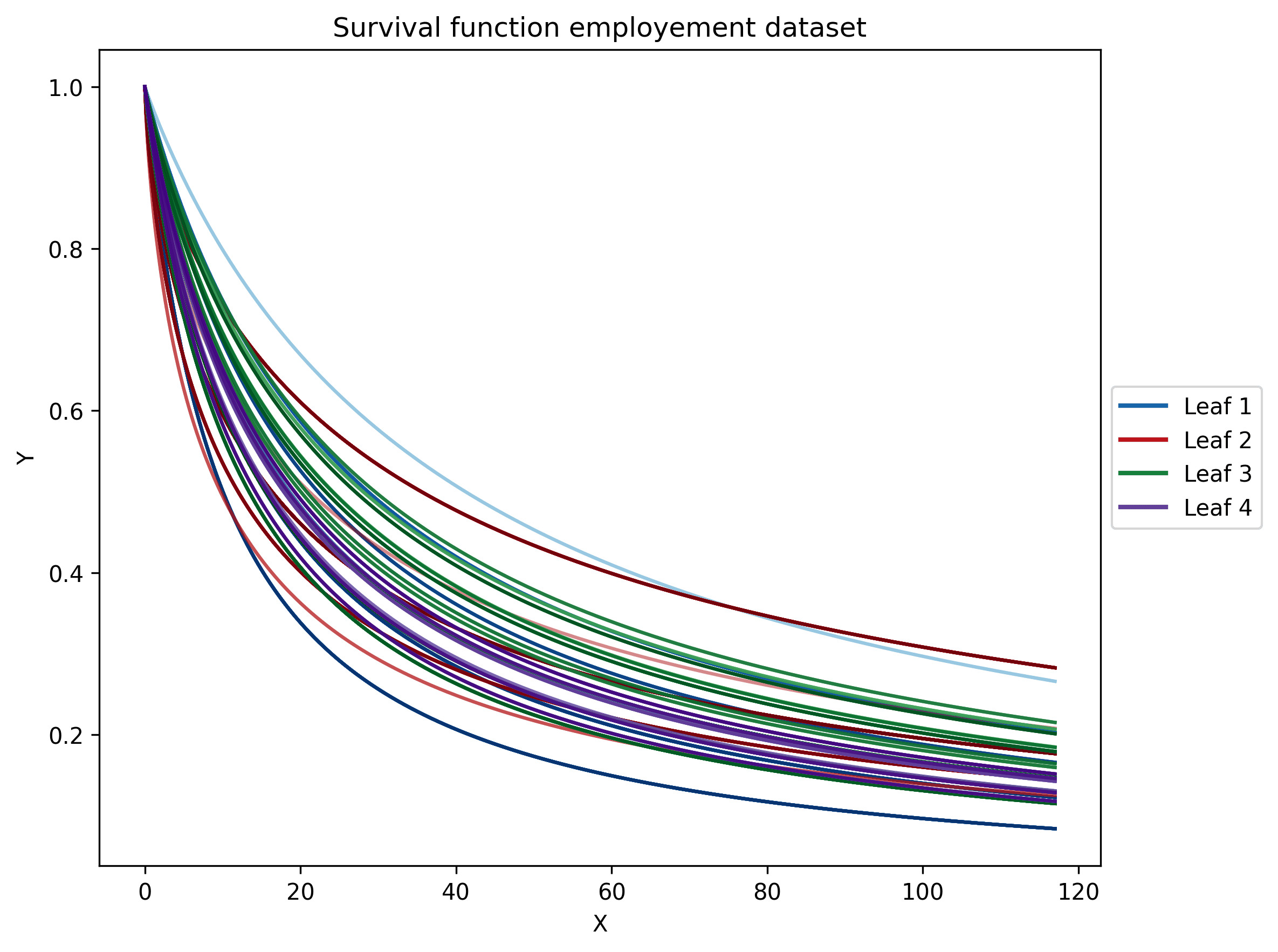}
        %\caption{Figura 5}
    \end{subfigure}
    \begin{subfigure}{0.45\textwidth}
        \centering
        \includegraphics[width=\linewidth]{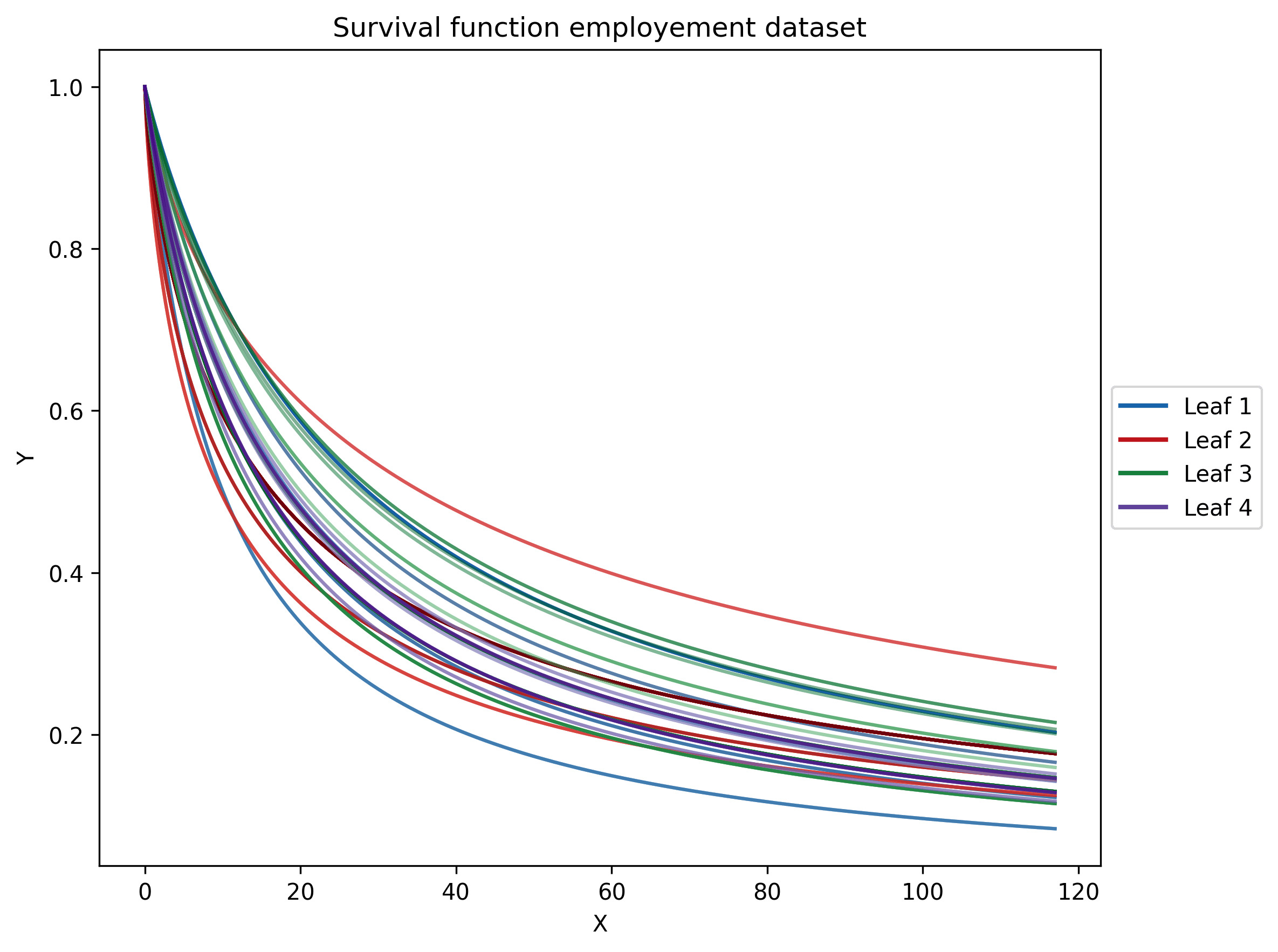}
        %\caption{Figura 6}
    \end{subfigure}
    
    % Quarta riga
    \begin{subfigure}{0.45\textwidth}
        \centering
        \includegraphics[width=\linewidth]{immagini_nuove/unemployement_seed1039_rho3_reg_drop_train.jpg}
        %\caption{Figura 7}
    \end{subfigure}
    \begin{subfigure}{0.45\textwidth}
        \centering
        \includegraphics[width=\linewidth]{immagini_nuove/unemployement_seed1039_rho3_reg_drop_test.jpg}
        %\caption{Figura 8}
    \end{subfigure}

\end{figure}
\newpage
\begin{figure}[H]
\centering
    % Titoli sopra le colonne
    \makebox[0.5\textwidth]{\textbf{Train}}
    \hfill
    \makebox[0.48\textwidth]{\textbf{Test}\textcolor{white}{aaaaaaaa}}
        \begin{subfigure}{0.45\textwidth}
        \centering
        \includegraphics[width=\linewidth]{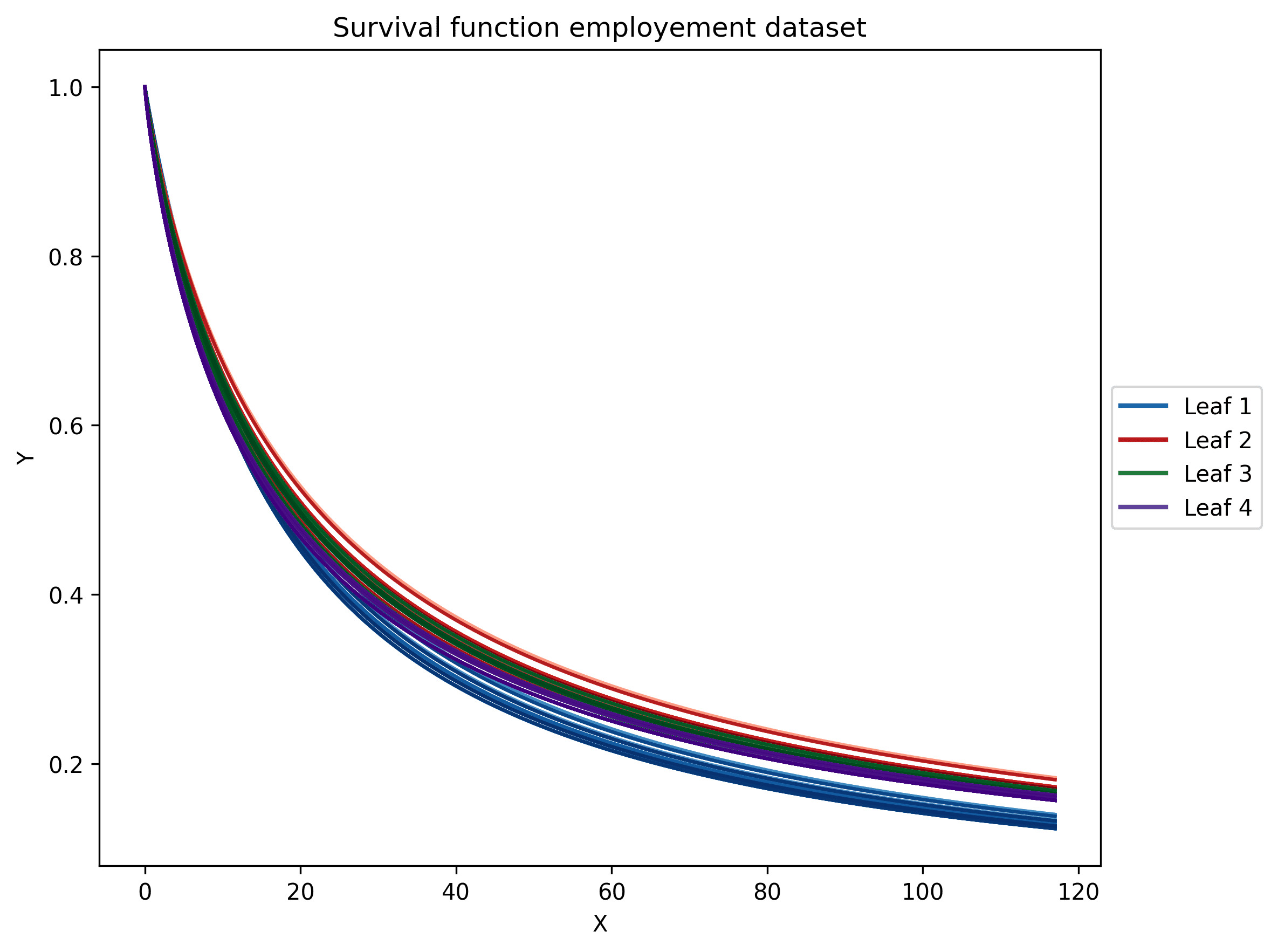}
        %\caption{Figura 7}
    \end{subfigure}
    \begin{subfigure}{0.45\textwidth}
        \centering
        \includegraphics[width=\linewidth]{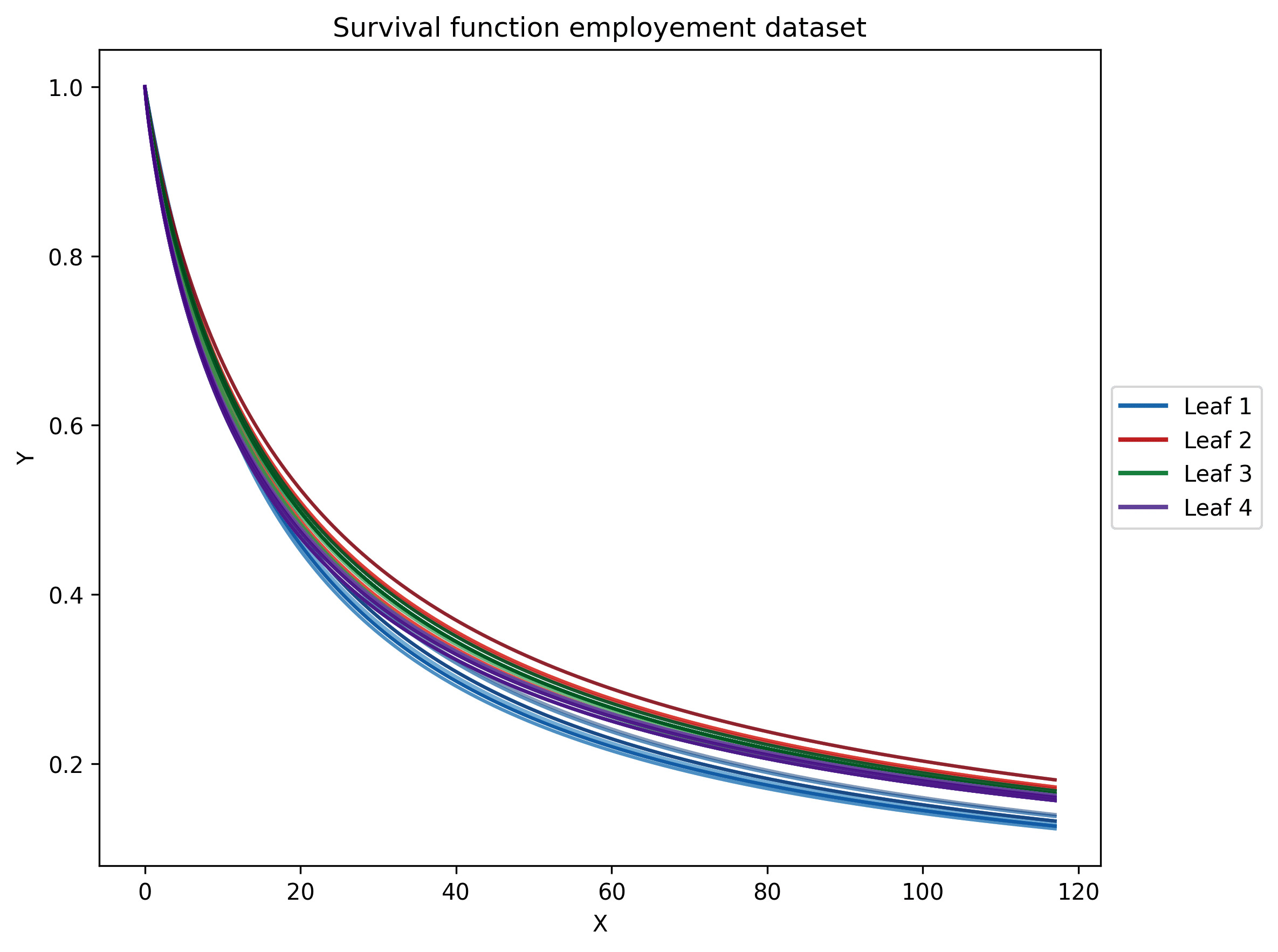}
        %\caption{Figura 8}
    \end{subfigure}

        \begin{subfigure}{0.45\textwidth}
        \centering
        \includegraphics[width=\linewidth]{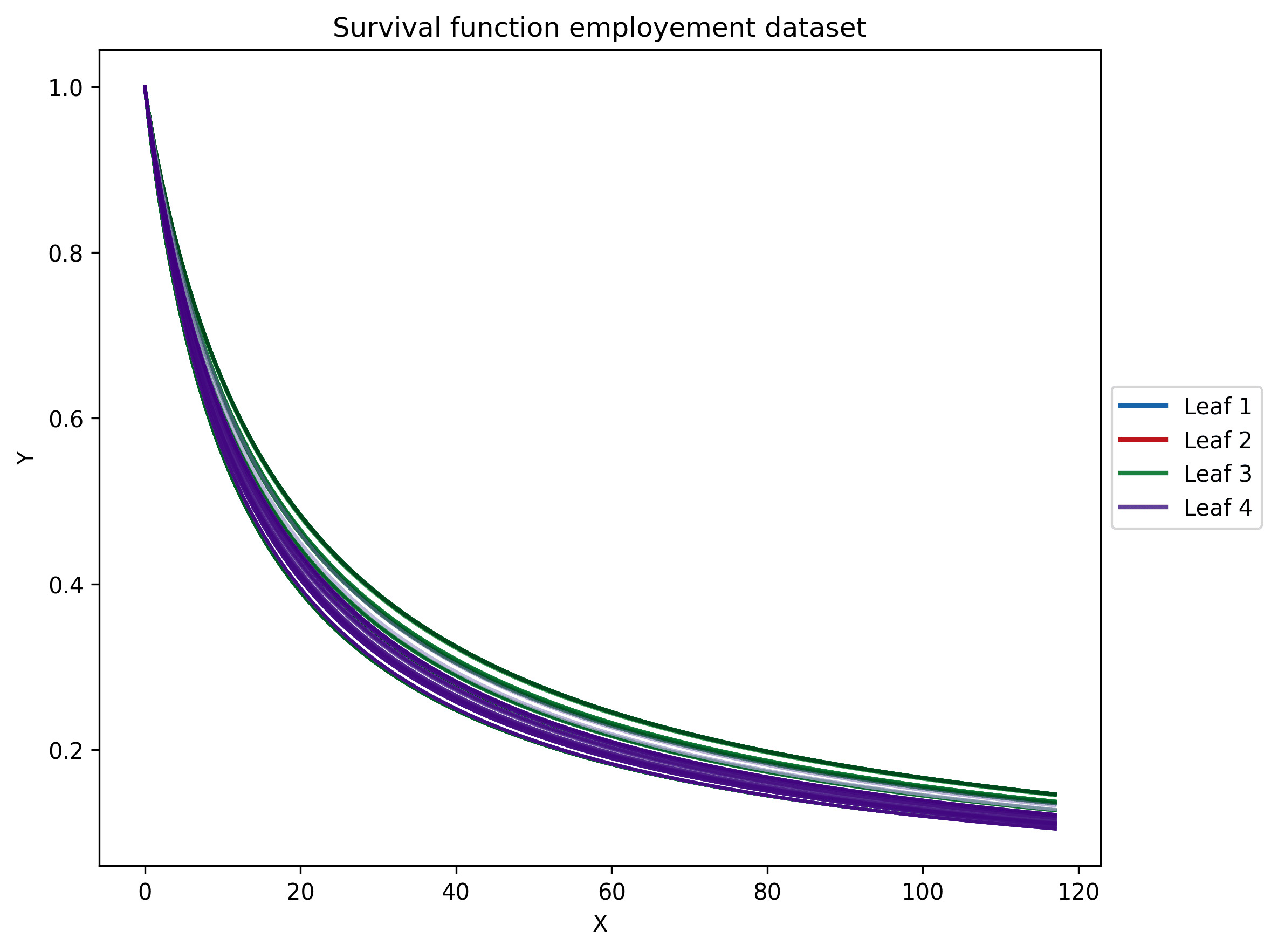}
        %\caption{Figura 7}
    \end{subfigure}
    \begin{subfigure}{0.45\textwidth}
        \centering
        \includegraphics[width=\linewidth]{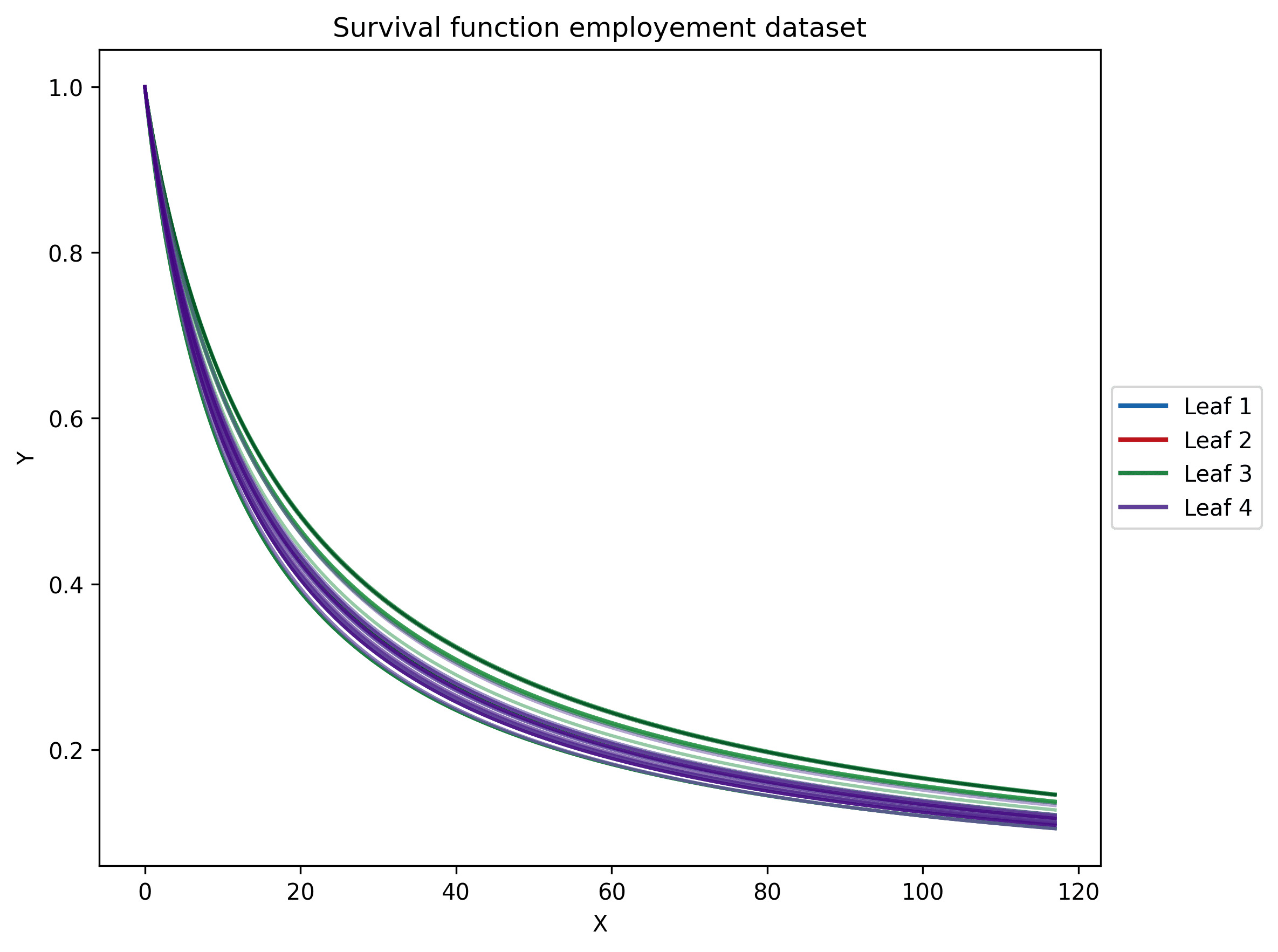}
        %\caption{Figura 8}
    \end{subfigure}
    \caption{\textcolor{black}{Plot of survival functions where each row represents the results for a different value of fairness penalty $\rho$. The red functions of the the first row (i.e. penalty equal to 0) represents a group composed mainly by women. Note that only in the first row red denotes the group of women; when $\rho > 0$, data points are redistributed among the leaf nodes, with colors indicating leaf membership only.
}}\label{fig:fair_appendix_complete}
\end{figure}

%\newpage

\end{document}